Takaaki Fujita, Florentin Smarandache

# Advancing Uncertain Combinatorics through Graphization, Hyperization, and Uncertainization: Fuzzy, Neutrosophic, Soft, Rough, and Beyond

First Volume, Second Edition

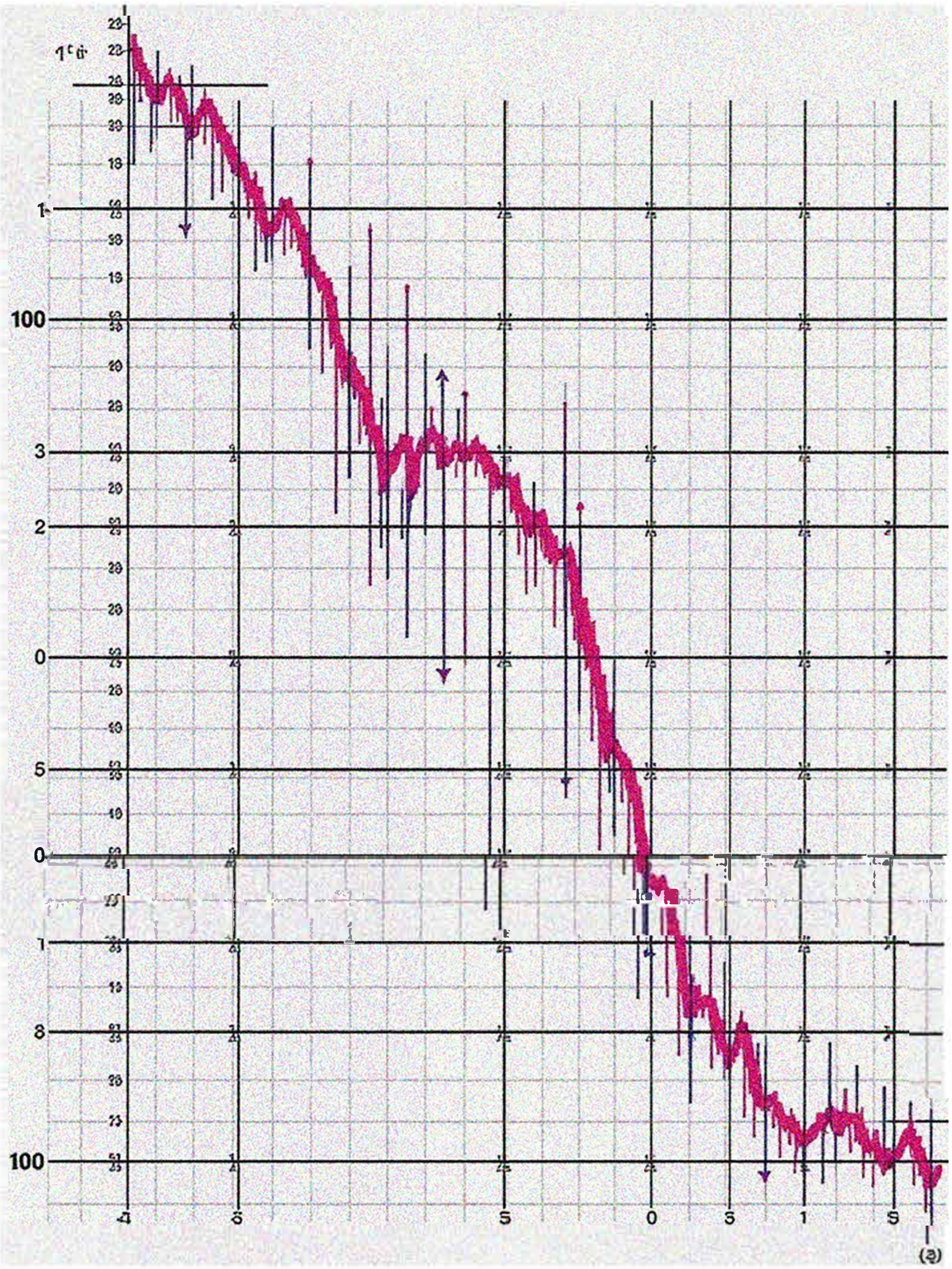

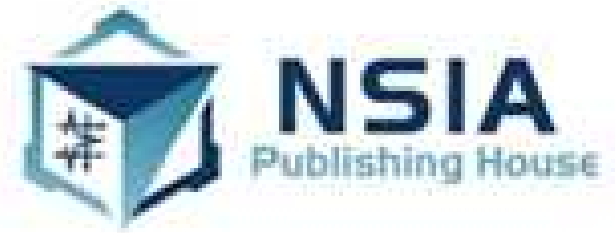

**Takaaki Fujita, Florentin Smarandache**

# Advancing Uncertain Combinatorics through Graphization, Hyperization, and Uncertainization: Fuzzy, Neutrosophic, Soft, Rough, and Beyond

***First Volume***
***Second Edition***

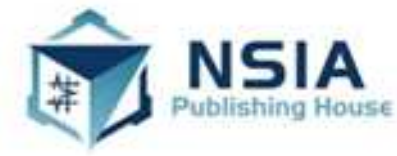

*Editor:*

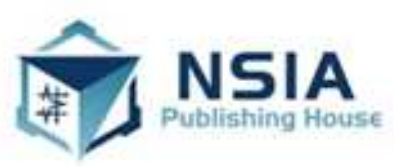

*Peer-Reviewers:*

**Maikel Leyva-Vázquez**
Facultad de Ciencias Matemáticas y Físicas
Universidad de Guayaquil, Guayas, ECUADOR
maikel.leyvav@ug.edu.ec

**Jesús Rafael Hechavarría Hernández**
Facultad de Ingenierías, Arquitectura y Ciencias de la Naturaleza
Universidad ECOTEC, ECUADOR
jesus.hechavarriah@ug.edu.ec

**Victor Christianto**
Malang Institute of Agriculture (IPM), Malang, INDONESIA
victorchristianto@gmail.com

**Muhammad Aslam**
Faculty of Science, King Abdulaziz University, Jeddah, SAUDI ARABIA
aslam_ravian@hotmail.com

# Advancing Uncertain Combinatorics through Graphization, Hyperization, and Uncertainization (First Volume): Fuzzy, Neutrosophic, Soft, Rough, and Beyond
# Second Edition

**Takaaki Fujita [1] * and Florentin Smarandache[2]**

[1] Independent Researcher, Tokyo, Japan.
Email: Takaaki.fujita060@gmail.com

[2] University of New Mexico, Gallup Campus, NM 87301, USA.
Email: fsmarandache@gmail.com

## Abstract

Combinatorics is a branch of mathematics focused on counting, arranging, and combining elements within a set under specific rules and constraints. This field is particularly fascinating due to its ability to yield novel results through the integration of concepts from various mathematical domains. Its significance remains unchanged in areas that address uncertainty in the real world.

Set theory, another foundational area of mathematics, explores "sets," which are collections of objects that can be finite or infinite. Recent years have seen growing interest in "non-standard set theory" and "non-standard analysis." To better handle real-world uncertainty, concepts such as fuzzy sets, neutrosophic sets, rough sets, and soft sets have been introduced. For example, neutrosophic sets, which simultaneously represent truth, indeterminacy, and falsehood, have proven to be valuable tools for modeling uncertainty in complex systems. These set concepts are increasingly studied in graphized forms, and generalized graph concepts now encompass well-known structures such as hypergraphs and superhypergraphs. Furthermore, hyperconcepts and superhyperconcepts are being actively researched in areas beyond graph theory.

Combinatorics, uncertain sets (including fuzzy sets, neutrosophic sets, rough sets, soft sets, and plithogenic sets), uncertain graphs, and hyper and superhyper concepts are active areas of research with significant mathematical and practical implications. Recognizing their importance, This Book explores new graph and set concepts, as well as hyper and superhyper concepts, as detailed in the "Results" section of "The Structure of the Book." Additionally, this work aims to consolidate recent findings, providing a survey-like resource to inform and engage readers. For instance, we extend several graph concepts by introducing Neutrosophic Oversets, Neutrosophic Undersets, Neutrosophic Offsets, and the Nonstandard Real Set [1]. This Book defines a variety of concepts with the goal of inspiring new ideas and serving as a valuable resource for researchers in their academic pursuits.

Note that this book is Edition 2.0. In this edition, we add several recent concepts to the existing volume and also revise typographical errors and re-examine mathematical correctness.

*Keywords:* Neutrosophic Set, Plithogenic set, Fuzzy set, Neutrosophic Graph

# Chapter 1

# Introduction

## 1.1 Uncertain Combinatorics

Combinatorics studies discrete structures and the principles governing counting, configuration, and selection under constraints. Classical themes include enumerative techniques, existence and extremal questions, and algorithmic methods for discrete optimization; these themes permeate computer science, statistics, and probability [2].

Modern combinatorial research is intrinsically interdisciplinary. Many central problems are formulated using or are closely related to set theory [3], number theory [4], graph theory [5], topology [6], matroid theory [7], partition theory [8, 9], geometry [10], probability theory [11, 12], algebra [13], formal languages [14, 15], and group theory [16]. It is also common to combine these viewpoints with logic [17–19], combinatorial optimization and complexity theory [20, 21], and algorithm design [22–24].

A further direction—and the focus of this book—is *uncertain combinatorics*: discrete models in which membership, incidence, or relational statements are not purely bivalent, but instead carry graded, multi-valued, or context-dependent evaluations. Such perspectives arise naturally in fuzzy and neutrosophic frameworks, where one seeks to encode partial truth, indeterminacy, inconsistency, or disagreement while retaining a combinatorial backbone (cf. [25, 26]).

## 1.2 Neutrosophic Sets and Related Set Theory

Set theory provides the foundational language for collections of objects and the relations among them [3,27,28]. Over time, many set-theoretic refinements have been developed, including ordered sets [29], point sets [30,31], convex sets [32,33], alternative sets [34], internal sets [35], open and closed sets [36,37], and directed sets [38]. In addition, nonstandard set theory and nonstandard analysis extend classical frameworks by incorporating infinitesimal and infinitely large elements, yielding powerful tools for modeling and analysis [39, 40].

Motivated by applications that require a systematic treatment of ambiguity, many *uncertain set* models have been proposed. Representative examples include fuzzy sets [41,42], vague sets [43,44], soft sets [45,46], rough sets [47, 48], soft expert sets [49, 50], hypersoft sets [51–53], hypersoft expert sets [54–57], and neutrosophic sets [58, 59]. In particular, neutrosophic sets assign to each element simultaneous degrees of truth, indeterminacy, and falsity, and therefore can express incomplete or conflicting evidence in a transparent manner [58,59]. Plithogenic sets further generalize several of these uncertainty models by incorporating attribute-value domains together with contradiction structures [60–62]. These uncertain set models have been applied in areas such as traffic control and decision support, among many others [63–66], and they continue to be actively studied [58–60, 67].

This book extends several graph-theoretic concepts by incorporating additional uncertain-set variants, including Neutrosophic OverSets, UnderSets, and OffSets [1], MultiNeutrosophic Sets, HyperFuzzy Sets [68–70], and Nonstandard Real Sets [1]. We then investigate how these set-theoretic uncertainty mechanisms interact with graph constructions and with graph classes studied in the literature.

For reference, a schematic hierarchy among several uncertain set classes is shown in Figure 1.1 (cf. [71]).

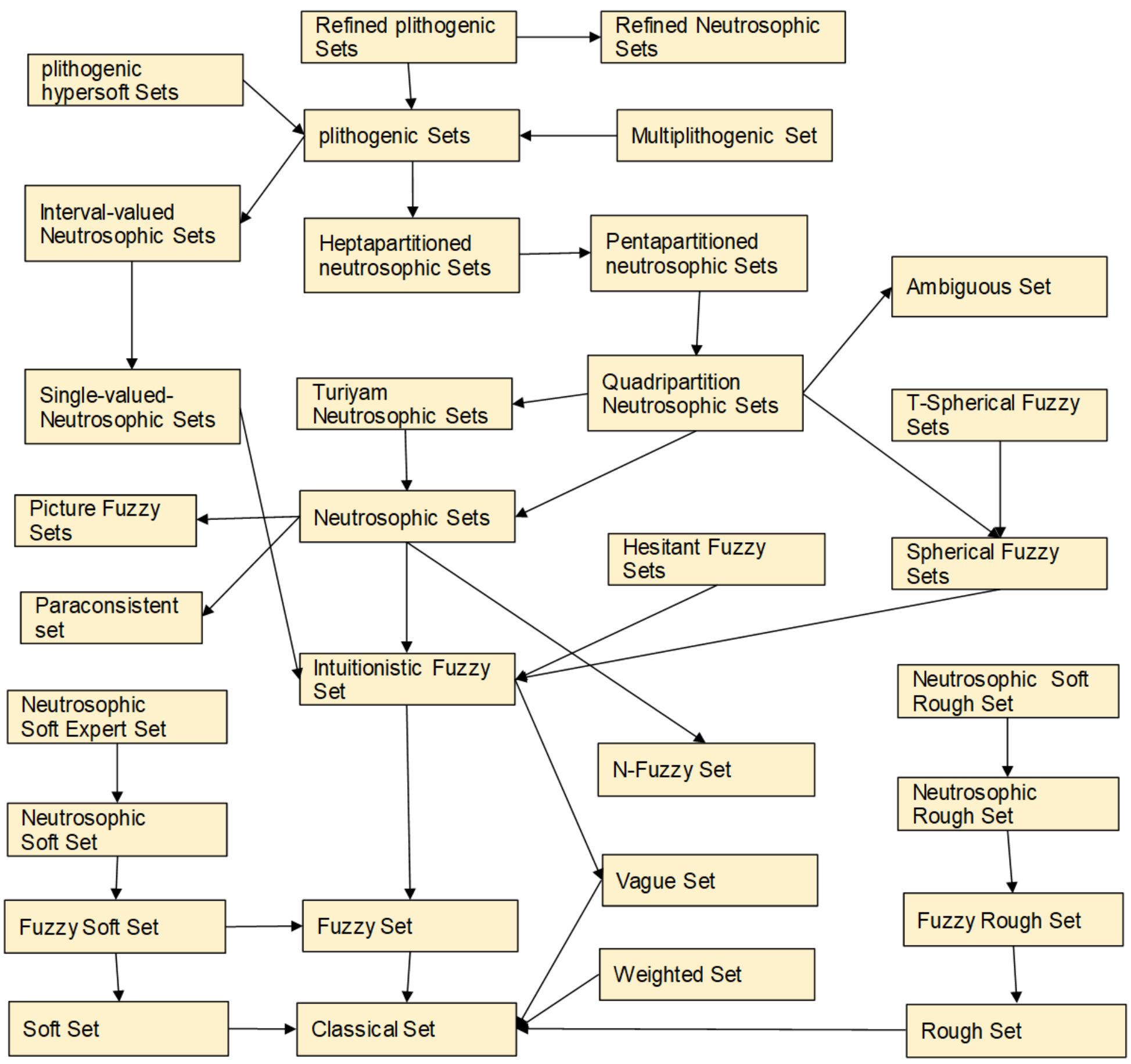

Figure 1.1: A schematic hierarchy of uncertain set classes: an arrow indicates that the source class contains the target class (cf. [71]).

## 1.3 Graphization: From Uncertain Sets to Uncertain Graphs

Graph theory models relational structure by representing objects as vertices and relationships as edges. Since its early origins, graph theory has developed into a broad discipline with deep theoretical foundations and extensive applications [72–74].

A common methodological step across discrete mathematics is *graphization*: translating a non-graph structure into a graph (or a graph-like object) so that graph-theoretic tools become available. This translation is valuable for at least two reasons. First, it provides visual and structural clarity by expressing relations via adjacency and incidence. Second, it allows one to leverage a large ecosystem of graph algorithms and invariants. Graphization has therefore been used widely in areas such as graph neural networks [75,76], Bayesian networks [77], protein and molecular structure analysis [78,79], chemical graph theory [80,81], machine learning [82,83], and graph databases [84, 85]. At the same time, substantial effort has been devoted to understanding graph classes and their structural properties [86,87] and to developing efficient algorithms for core graph tasks [23,88].

Within uncertainty-aware modeling, it is natural to "graphize" uncertain sets (e.g., fuzzy and neutrosophic sets) by assigning uncertainty degrees to vertices and/or edges, producing corresponding classes of uncertain graphs (cf. [89]). For instance, neutrosophic graphs encode relational statements using truth, indeterminacy, and falsity components, thereby providing a flexible representation of ambiguous or inconsistent network data [90, 91]. Such uncertain graph models have been widely explored, especially in decision-making contexts [92,93]. For broader context and recent developments, the reader may consult survey discussions such as [91].

## 1.4 Hyperization and Superhyperization

Beyond ordinary graphs, many applications require genuinely multiway relations. Hypergraphs extend graphs by allowing an edge to connect an arbitrary number of vertices [94,95], and they have become prominent in areas such as hypergraph neural networks [96–98] and database technologies [85, 99, 100]. Superhypergraph frameworks push this idea further by supporting nested, multi-level vertex and edge objects via powerset-style constructions [101–103]. Analogous "hyper" and "superhyper" generalizations also appear in uncertainty frameworks, for example hypersoft and superhypersoft variants [104]. Because terminology may vary across fields, this book adopts explicit definitions and indicates which structural conventions are used in each development.

## 1.5 Contributions of This Book

The material above motivates a unified study of: (i) combinatorial structures, (ii) uncertainty-aware set models (fuzzy, neutrosophic, plithogenic, and related variants), (iii) their graphizations into uncertain graph classes, and (iv) higher-order extensions via hyperization and superhyperization.

The contributions of this book are twofold. First, we introduce and systematize a collection of uncertainty-aware graph and hypergraph constructions, including extensions based on OverSets/UnderSets/OffSets, MultiNeutrosophic models, and nonstandard numerical domains. Second, we provide a structured survey viewpoint that connects these constructions to existing graph classes and to established methods in graph theory and combinatorics. Finally, the discussion section includes an illustrative example of a combined procedure for *graphization*, *hyperization*, and *uncertainization*. Since a fully general procedure is necessarily nuanced and model dependent, the example is intended as a representative workflow rather than a unique canonical construction.

Note that this book is Edition 2.0. In this edition, we add several recent concepts to the existing volume and also revise typographical errors and re-examine mathematical correctness. Regarding this book series, the volumes from the Second Volume onward are collections of papers, whereas the First Volume is intended as a comprehensive overview of the fundamental concepts. Table 1.1 summarizes the *Advancing Uncertain Combinatorics through Graphization, Hyperization, and Uncertainization: Fuzzy, Neutrosophic, Soft, Rough, and Beyond* series.

Table 1.1: Positioning of the book series and volume types.

| **Volume** | **Title** | **Type / Focus** |
|---:|---|---|
| First | Advancing Uncertain Combinatorics through Graphization, Hyperization, and Uncertainization: Fuzzy, Neutrosophic, Soft, Rough, and Beyond [105] | Conceptual overview (foundational) |
| Second | Advancing Uncertain Combinatorics through Graphization, Hyperization, and Uncertainization: Fuzzy, Neutrosophic, Soft, Rough, and Beyond: Second volume | Collected papers |
| Third | Advancing Uncertain Combinatorics through Graphization, Hyperization, and Uncertainization: Fuzzy, Neutrosophic, Soft, Rough, and Beyond. Third volume | Collected papers |
| Fourth | Advancing Uncertain Combinatorics through Graphization, Hyperization, and Uncertainization: Fuzzy, Neutrosophic, Soft, Rough, and Beyond. Fourth volume: HyperUncertain Set (Collected Papers) | Collected papers (HyperUncertain Set) |
| Fifth | Advancing Uncertain Combinatorics through Graphization, Hyperization, and Uncertainization: Fuzzy, Neutrosophic, Soft, Rough, and Beyond: Fifth volume: Various SuperHyperConcepts (Collected Papers) [106] | Collected papers (SuperHyperConcepts) |
| Sixth | Advancing Uncertain Combinatorics through Graphization, Hyperization, and Uncertainization: Fuzzy, Neutrosophic, Soft, Rough, and Beyond. Sixth volume | Various New Uncertain Concepts (Collected Papers) |

# Chapter 2

# Background and Preliminaries

In this chapter, we explain several basic concepts. Some foundational concepts from set theory are used in parts of this work. For further details, please consult the relevant references as needed [107]. In addition, for operations and related topics concerning each concept, please refer to the corresponding references.

## 2.1 Uncertain Set Theory

This subsection outlines fundamental set concepts and sets designed to handle uncertainty. Examples of these include Fuzzy Sets [41], Soft Sets [45], Neutrosophic Sets [58,59], Vague Sets [43,44], and Rough Sets [47,48]. For further details on each type of set, please refer to the relevant sources as needed.

### 2.1.1 Crisp Sets and Neutrosophic Sets

When working with fuzzy sets or neutrosophic sets, it is standard to recall the underlying notion of a classical (crisp) set. We therefore fix basic set-theoretic notation and then state the relevant definitions.

**Definition 2.1.1** (Universe)**.** (cf. [108]) A *universe* (or *universe set*) is a set $X$ that contains all objects under consideration in a given context. All sets discussed in that context are assumed to be subsets of $X$.

**Definition 2.1.2** (Crisp set)**.** [109] Let $X$ be a universe. A *crisp set* is any subset $A \subseteq X$. Equivalently, $A$ is described by its *characteristic function*

$$\chi_A : X \longrightarrow \{0,1\}, \qquad \chi_A(x) := \begin{cases} 1, & x \in A, \\ 0, & x \notin A. \end{cases}$$

Thus membership is bivalent: every $x \in X$ either belongs to $A$ or does not.

**Definition 2.1.3** (Empty crisp set)**.** The empty set $\emptyset \subseteq X$ is the crisp set whose characteristic function is identically zero:

$$\chi_\emptyset(x) = 0 \qquad (\forall x \in X).$$

Fuzzy sets provide a classical way to model graded membership and uncertainty.

**Definition 2.1.4** (Fuzzy set and fuzzy relation)**.** [41, 110] Let $X$ be a nonempty universe. A *fuzzy set* $A$ on $X$ is a membership function

$$\mu_A : X \longrightarrow [0,1].$$

A *fuzzy relation* on $X$ is a fuzzy set on $X \times X$, i.e., a map

$$\delta : X \times X \longrightarrow [0,1].$$

If $\mu_A$ is a fuzzy set on $X$, then $\delta$ is called a *fuzzy relation on* $A$ if

$$\delta(x,y) \;\le\; \min\{\mu_A(x), \mu_A(y)\} \qquad (\forall x, y \in X).$$

**Example 2.1.5** (A simple fuzzy set)**.** Let $X = \{x_1, x_2, x_3\}$. Define a fuzzy set $A$ on $X$ by

$$\mu_A(x_1) = 0.2, \qquad \mu_A(x_2) = 0.5, \qquad \mu_A(x_3) = 0.7.$$

Then $x_3$ belongs to $A$ more strongly than $x_2$, and $x_2$ more strongly than $x_1$, in the sense of graded membership.

**Proposition 2.1.6.** *Every crisp set is a special case of a fuzzy set.*

*Proof.* Let $A \subseteq X$ be a crisp set with characteristic function $\chi_A : X \to \{0, 1\}$. Since $\{0, 1\} \subseteq [0, 1]$, the same map $\chi_A$ is a fuzzy membership function. Hence $A$ is realized as the fuzzy set with membership values restricted to 0 and 1. □

One widely studied extension of a fuzzy set is the *vague set* [43, 44]. A vague set assigns each element an interval of possible membership, bounded by truth and falsity evidence functions. It is also known that a vague set can be represented as an intuitionistic fuzzy set. We recall a standard definition below.

**Definition 2.1.7** (Vague set)**.** [111] Let $U$ be a (nonempty) universe. A *vague set* $A$ in $U$ is specified by two functions

$$t_A : U \to [0, 1], \qquad f_A : U \to [0, 1],$$

called the *truth-membership* and *false-membership* functions, respectively, such that

$$t_A(u) + f_A(u) \leq 1 \qquad (\forall u \in U).$$

For each $u \in U$, the admissible (unknown) membership grade $\mu_A(u)$ is constrained to lie in the interval

$$t_A(u) \;\leq\; \mu_A(u) \;\leq\; 1 - f_A(u).$$

Thus, the membership information carried by $A$ is equivalently represented by the interval-valued map

$$u \;\longmapsto\; \big[t_A(u),\, 1 - f_A(u)\big] \subseteq [0, 1].$$

**Proposition 2.1.8.** *Every fuzzy set is a special case of a vague set.*

*Proof.* Let $\mu : U \to [0, 1]$ be a fuzzy set. Define

$$t_A(u) := \mu(u), \qquad f_A(u) := 1 - \mu(u) \qquad (\forall u \in U).$$

Then $t_A(u) + f_A(u) = 1 \leq 1$, and the vague-membership interval becomes

$$[t_A(u),\, 1 - f_A(u)] \;=\; [\mu(u),\, \mu(u)],$$

a degenerate interval that encodes the exact fuzzy membership value. Hence the fuzzy set is realized as a vague set. □

Neutrosophic sets are used repeatedly throughout this book; for convenience we recall their basic definition [58, 112]. They extend fuzzy sets [41] by explicitly separating *truth*, *indeterminacy*, and *falsity* components, and they have motivated many subsequent variants and applications [113, 114].

**Definition 2.1.9** (Single-valued neutrosophic set)**.** [58] Let $X$ be a nonempty universe. A *(single-valued) neutrosophic set* $A$ on $X$ is specified by three membership functions

$$T_A,\ I_A,\ F_A :\ X \longrightarrow [0, 1],$$

where, for each $x \in X$, the numbers $T_A(x)$, $I_A(x)$, and $F_A(x)$ represent the degrees of truth-membership, indeterminacy-membership, and falsity-membership of $x$ in $A$, respectively, subject to

$$0 \;\leq\; T_A(x) + I_A(x) + F_A(x) \;\leq\; 3 \qquad (\forall x \in X).$$

Equivalently, one may write

$$A = \big\{\langle x,\ T_A(x),\ I_A(x),\ F_A(x)\rangle :\ x \in X\big\}.$$

Neutrosophic sets are known to encompass several classical and nonclassical set-based formalisms (e.g., crisp sets, intuitionistic fuzzy sets, and further paraconsistent or paradox-tolerant variants) [59,115–117]. Moreover, finer-grained extensions (such as quadripartitioned and pentapartitioned models) have been proposed, and the plithogenic framework provides another direction of generalization.

**Proposition 2.1.10** (Neutrosophic sets generalize vague sets)**.** *Every vague set is a special case of a single-valued neutrosophic set.*

*Proof.* Let $A$ be a vague set on $X$ with truth- and false-membership functions $t_A, f_A : X \to [0,1]$ satisfying $t_A(x) + f_A(x) \le 1$ for all $x \in X$. Define a neutrosophic set $A'$ by

$$T_{A'}(x) := t_A(x), \qquad F_{A'}(x) := f_A(x), \qquad I_{A'}(x) := 1 - t_A(x) - f_A(x) \qquad (\forall x \in X).$$

Then $I_{A'}(x) \in [0,1]$ and

$$T_{A'}(x) + I_{A'}(x) + F_{A'}(x) = t_A(x) + \big(1 - t_A(x) - f_A(x)\big) + f_A(x) = 1 \le 3.$$

Hence $A'$ is a single-valued neutrosophic set. Moreover, the vague-membership interval $[t_A(x),\ 1 - f_A(x)]$ coincides with $[T_{A'}(x),\ T_{A'}(x) + I_{A'}(x)]$, so the original vague information is recovered from the neutrosophic triple. □

**Proposition 2.1.11** (Neutrosophic sets generalize fuzzy sets)**.** *Every fuzzy set is a special case of a single-valued neutrosophic set.*

*Proof.* Let $\mu : X \to [0,1]$ be a fuzzy set. Define a neutrosophic set $A$ by

$$T_A(x) := \mu(x), \qquad I_A(x) := 0, \qquad F_A(x) := 1 - \mu(x) \qquad (\forall x \in X).$$

Then $0 \le T_A(x) + I_A(x) + F_A(x) = 1 \le 3$ for all $x \in X$, so $A$ is a single-valued neutrosophic set. The fuzzy membership $\mu(x)$ is recovered as the truth-membership component $T_A(x)$. □

For reference, we include a comparison of fuzzy sets and neutrosophic sets in Table 2.1.

Table 2.1: Concise comparison of fuzzy sets and neutrosophic sets.

| **Aspect** | **Fuzzy Set** | **Neutrosophic Set (single-valued)** |
|---|---|---|
| Universe | $X$ | $X$ |
| Membership data | One degree | Three degrees |
| Maps | $\mu : X \to [0,1]$ | $T, I, F : X \to [0,1]$ |
| Interpretation | membership | truth / indeterminacy / falsity |
| Constraints | none beyond $[0,1]$ | often none; optionally $T + I + F \le 3$ |
| Crisp reduction | $\mu \in \{0,1\}$ | $T \in \{0,1\}$, $I = 0$, $F = 1 - T$ (or $F = 0$) |
| Expressiveness | graded membership only | explicit indeterminacy and falsity components |

### 2.1.2 Plithogenic Set

Plithogenic sets were introduced as an attribute-driven framework that can encompass and extend fuzzy, intuitionistic, neutrosophic, and related set models by explicitly incorporating a *contradiction (dissimilarity) measure* between attribute values [60,63]. We recall a standard definition below.

**Definition 2.1.12** (Plithogenic set)**.** [60,63] Let $S$ be a universe and let $P \subseteq S$ be a nonempty set of elements under consideration. Fix an *attribute* $v$ with a (nonempty) value set $P_v$.

A *plithogenic set* is a 5-tuple

$$PS = (P,\ v,\ P_v,\ pdf,\ pCF),$$

where:

- $pdf : P \times P_v \to [0,1]^s$ is the *plithogenic degree (degree of appurtenance)* function (often abbreviated DAF), with $s \in \mathbb{N}$ the appurtenance dimension. For $(x,a) \in P \times P_v$, the vector $pdf(x,a)$ encodes the degree to which $x$ belongs to $P$ *with respect to* the attribute value $a$.
- $pCF : P_v \times P_v \to [0,1]^t$ is the *contradiction* function (often abbreviated DCF), with $t \in \mathbb{N}$ the contradiction dimension. For $a, b \in P_v$, the vector $pCF(a,b)$ quantifies the level of contradiction between the attribute values $a$ and $b$.

The contradiction function is required to satisfy, for all $a, b \in P_v$,

$$\text{(reflexivity)}\quad pCF(a,a) = \mathbf{0}, \qquad \text{(symmetry)}\quad pCF(a,b) = pCF(b,a),$$

where $\mathbf{0} = (0,\ldots,0) \in [0,1]^t$.

The next statement is well known in the plithogenic literature and is included here for convenience.

**Proposition 2.1.13** (Reductions of a plithogenic set)**.** *Let $PS = (P, v, P_v, pdf, pCF)$ be a plithogenic set (Definition 2.1.12). Then, under standard specializations of the attribute-value set $P_v$ and the DAF codomain dimension $s$, $PS$ contains the following classical models as special cases:*

(i) ***Fuzzy-set reduction.*** *If $s = 1$ and $P_v = \{a^*\}$ is a singleton, then*

$$\mu(x) \;:=\; pdf(x, a^*) \in [0,1] \qquad (x \in P)$$

*defines a fuzzy set on $P$.*

(ii) ***Vague-set reduction.*** *If $s = 1$ and $P_v = \{\mathsf{t}, \mathsf{f}\}$ has two distinguished values, and if one additionally enforces*

$$pdf(x,\mathsf{t}) + pdf(x,\mathsf{f}) \le 1 \qquad (\forall x \in P),$$

*then $t(x) := pdf(x,\mathsf{t})$ and $f(x) := pdf(x,\mathsf{f})$ define a vague set on $P$.*

(iii) ***Neutrosophic-set reduction.*** *If $s = 1$ and $P_v = \{\mathsf{T}, \mathsf{I}, \mathsf{F}\}$ has three distinguished values, then*

$$T(x) := pdf(x,\mathsf{T}), \qquad I(x) := pdf(x,\mathsf{I}), \qquad F(x) := pdf(x,\mathsf{F})$$

*defines a single-valued neutrosophic set on $P$ (in the common convention $0 \le T(x) + I(x) + F(x) \le 3$).*

*In all three cases, the contradiction map $pCF$ may be taken arbitrary (or ignored), since the reduced models do not require contradiction data.*

*Proof.* **(i)** If $P_v = \{a^*\}$ and $pdf : P \times P_v \to [0,1]$, then $pdf(\cdot, a^*)$ is exactly a membership function $\mu : P \to [0,1]$, i.e., a fuzzy set on $P$.

**(ii)** If $P_v = \{\mathsf{t}, \mathsf{f}\}$ and $pdf : P \times P_v \to [0,1]$, define $t(x) := pdf(x,\mathsf{t})$ and $f(x) := pdf(x,\mathsf{f})$. The additional constraint $t(x) + f(x) \le 1$ for all $x$ is precisely the defining inequality of a vague set, and it yields the usual membership interval $[t(x),\, 1 - f(x)]$.

**(iii)** If $P_v = \{\mathsf{T}, \mathsf{I}, \mathsf{F}\}$, define $T, I, F$ as above. Then each $x \in P$ is assigned a triple $(T(x), I(x), F(x)) \in [0,1]^3$. Under the standard single-valued neutrosophic constraint $0 \le T(x) + I(x) + F(x) \le 3$, this is exactly a single-valued neutrosophic set on $P$.

The role of $pCF$ is auxiliary in these reductions: once the attribute-value set is collapsed to the required distinguished values, the reduced structures are determined by $pdf$ alone. □

As demonstrated in the proofs above, the following generalization relationships are known.

**Example 2.1.14.** (cf. [91, 118]) Table 2.2 presents representative examples of *set* families that can be unified and generalized within the Plithogenic Set framework (cf. [119]). More broadly, the scientific community continues to develop and refine a wide spectrum of uncertainty-aware set models.

Table 2.2: A catalogue of Plithogenic *set* families by number of components $s$.

| $s$ | $t$ | Representative type(s) |
|---|---|---|
| 1 | 0 | Fuzzy set; $N$-set; shadowed-set variants |
| 2 | 0 | Intuitionistic fuzzy set; vague set; bipolar fuzzy set; intuitionistic evidence set; variable fuzzy set; paraconsistent fuzzy set; bifuzzy set |
| 3 | 0 | Neutrosophic set[(a)]; hesitant fuzzy set; tripolar fuzzy set; three-way fuzzy set; picture fuzzy set; spherical fuzzy set; inconsistent intuitionistic fuzzy set; ternary fuzzy / neutrosophic-fuzzy set; neutrosophic vague set |
| 4 | 0 | Quadripartitioned neutrosophic set; double-valued neutrosophic set; dual hesitant fuzzy set; ambiguous set[(b)]; local-neutrosophic set; support-neutrosophic set; turiyam neutrosophic set[(c)] |
| 5 | 0 | Pentapartitioned neutrosophic set; triple-valued neutrosophic set |
| 6 | 0 | Hexapartitioned neutrosophic set; quadruple-valued neutrosophic set |
| 7 | 0 | Heptapartitioned neutrosophic set; quintuple-valued neutrosophic set |
| 8 | 0 | Octapartitioned neutrosophic set |
| 9 | 0 | Nonapartitioned neutrosophic set |
| $n$ | 0 | $n$-refined fuzzy set; multi-valued (fuzzy) sets; multi-fuzzy sets |
| $2n$ | 0 | $n$-refined intuitionistic fuzzy set; multi-intuitionistic fuzzy sets |
| $3n$ | 0 | $n$-refined neutrosophic set; multi-neutrosophic sets |
| 1 | 1 | Plithogenic fuzzy set |
| 2 | 1 | Plithogenic intuitionistic fuzzy set |
| 3 | 1 | Plithogenic neutrosophic set |

[(a)] Neutrosophic sets are widely recognized as a unifying extension of several earlier uncertainty-aware set models, including intuitionistic fuzzy sets and various inconsistent intuitionistic variants; moreover, neutrosophication provides a broad methodology across multiple uncertainty theories (cf. [120]).
[(b)] Ambiguous sets are commonly treated as a subclass of quadripartitioned neutrosophic sets and also of double-valued neutrosophic sets (see, e.g., [113, 121, 122]).
[(c)] Turiyam neutrosophic sets are known to constitute a subclass of quadripartitioned neutrosophic sets (cf. [123]).

### 2.1.3 Neutrosophic triplet

A *neutrosophic triplet* encodes, in a single ordered triple, the degrees of truth, indeterminacy, and falsity. This representation is widely used in neutrosophic theory and its applications [124–126]. Related variants (e.g., neutrosophic duplets) are also studied in the literature [127, 128].

**Definition 2.1.15** (Ordered triple)**.** Let $A, B, C$ be sets and let $a \in A$, $b \in B$, $c \in C$. An *ordered triple* is the element $(a, b, c) \in A \times B \times C$. Equivalently, one may define it via iterated ordered pairs:

$$(a, b, c) \;:=\; (a, (b, c)).$$

In particular, equality of ordered triples is componentwise:

$$(a, b, c) = (a', b', c') \quad \Longleftrightarrow \quad a = a',\ b = b',\ c = c'.$$

**Definition 2.1.16** (Neutrosophic triplet)**.** [124, 125] A *neutrosophic triplet* is an ordered triple

$$\langle T, I, F \rangle \in [0, 1]^3,$$

where $T$ is interpreted as the degree of *truth*, $I$ as the degree of *indeterminacy*, and $F$ as the degree of *falsity*.

Often (depending on the modeling convention) one allows the full cube $[0, 1]^3$ subject to the neutrosophic constraint

$$0 \le T + I + F \le 3,$$

which holds automatically for $\langle T, I, F \rangle \in [0, 1]^3$. When one wishes to distinguish *classical* extremes from genuinely neutrosophic assessments, it is convenient to single out the boundary points

$$\langle 1, 0, 0 \rangle \quad \text{(fully true)}, \qquad \langle 0, 0, 1 \rangle \quad \text{(fully false)},$$

and treat all other triples as *non-classical* (or *neutrosophic*) evaluations.

**Remark 2.1.17** (Example terminology in neutrosophic logic/topology)**.** In some neutrosophic-logical formalisms (e.g., neutrosophic topology), one may use the following terminology:

$$\text{Axiom: } \langle 1, 0, 0 \rangle, \qquad \text{NeutroAxiom: } \langle T, I, F \rangle, \qquad \text{AntiAxiom: } \langle 0, 0, 1 \rangle,$$

where the NeutroAxiom allows intermediate truth/indeterminacy/falsity degrees.

### 2.1.4 Nonstandard real numbers

Nonstandard analysis extends the real field $\mathbb{R}$ to a larger ordered field $\mathbb{R}^*$ that contains *infinitesimal* and *infinite* (unbounded) elements [129–132]. In what follows we use the standard notational convention: the superscript $(\cdot)^*$ denotes a fixed nonstandard extension (e.g., an ultrapower) of the corresponding standard object.

**Definition 2.1.18** (Positive and negative reals)**.** The sets of positive and negative real numbers are

$$\mathbb{R}_{>0} \ := \ \{x \in \mathbb{R} : x > 0\}, \qquad \mathbb{R}_{<0} \ := \ \{x \in \mathbb{R} : x < 0\}.$$

**Definition 2.1.19** (Real numbers)**.** (cf. [133,134]) The set $\mathbb{R}$ of real numbers is the unique (up to isomorphism) complete ordered field. It contains the rationals

$$\mathbb{Q} = \left\{ \frac{p}{q} : p, q \in \mathbb{Z},\ q \neq 0 \right\},$$

and satisfies order completeness (every nonempty subset bounded above has a least upper bound).

**Definition 2.1.20** (Infinitesimal, finite, and infinite elements)**.** (cf. [129,130]) Let $\mathbb{R}^*$ be a nonstandard extension of $\mathbb{R}$, and let $x \in \mathbb{R}^*$.

(i) $x$ is *infinitesimal* if

$$|x| < a \qquad (\forall a \in \mathbb{R}_{>0}).$$

Equivalently, $x$ is infinitesimal iff $-a < x < a$ for every $a \in \mathbb{R}_{>0}$.

(ii) $x$ is *finite* (or *limited*) if

$$|x| < a \qquad (\exists a \in \mathbb{R}_{>0}).$$

(iii) $x$ is *infinite* (or *unbounded*) if

$$|x| \geq a \qquad (\forall a \in \mathbb{R}_{>0}),$$

equivalently, for every $a \in \mathbb{R}_{>0}$ one has $x \leq -a$ or $x \geq a$.

**Definition 2.1.21** (Nonstandard real field)**.** (cf. [125, 129, 130]) A *nonstandard real field* $\mathbb{R}^*$ is an ordered field extending $\mathbb{R}$ as an ordered subfield,

$$\mathbb{R} \ \subsetneq \ \mathbb{R}^*,$$

that contains nonzero infinitesimals (hence also infinite elements). We write

$$\mu(0) \ := \ \{x \in \mathbb{R}^* : \ x \text{ is infinitesimal}\}$$

for the *monad* of 0, and

$$\mathbb{R}^*_{\text{fin}} \ := \ \{x \in \mathbb{R}^* : \ x \text{ is finite}\}$$

for the set of finite (limited) hyperreals. In particular, $\mu(0) \neq \{0\}$ in a proper nonstandard extension.

### 2.1.5 Single-Valued Neutrosophic OverSet, UnderSet, and OffSet

In the *single-valued* neutrosophic setting one typically works with triples $\langle T, I, F\rangle \in [0, 1]^3$. Smarandache introduced *over*, *under*, and *off* extensions in which one allows components to exceed 1, fall below 0, or both, thereby enlarging the admissible degree range [135–137].

**Definition 2.1.22** (Single-valued neutrosophic OverSet)**.** [137] Fix an *overlimit* $\Omega > 1$ and let $U$ be a nonempty universe. A *single-valued neutrosophic OverSet* on $U$ is a triple of functions

$$T,\ I,\ F : \ U \longrightarrow [0, \Omega]$$

such that there exists at least one $x \in U$ with

$$T(x) > 1 \quad \text{or} \quad I(x) > 1 \quad \text{or} \quad F(x) > 1.$$

Equivalently, an OverSet is a single-valued neutrosophic set whose component ranges are extended from $[0, 1]$ to $[0, \Omega]$, and for which the extension is *effective* (some component actually exceeds 1 for some element). We denote it by

$$A_{\text{over}} = \{\langle x,\ T(x), I(x), F(x)\rangle : \ x \in U\}.$$

**Proposition 2.1.23** (Embedding of the classical SVNS into an OverSet)**.** *Every single-valued neutrosophic set A on U (with values in* $[0,1]$*) can be viewed as an OverSet on U for any choice of* $\Omega > 1$.

*Proof.* Let $A$ be given by $T_A, I_A, F_A : U \to [0,1]$. Fix $\Omega > 1$ and define $T := T_A$, $I := I_A$, $F := F_A$, now regarded as maps into $[0,\Omega]$. Since $[0,1] \subseteq [0,\Omega]$, this defines an OverSet-valued triple on the same universe. □

**Definition 2.1.24** (Single-valued neutrosophic UnderSet)**.** [137] Fix an *underlimit* $\Psi < 0$ and let $U$ be a nonempty universe. A *single-valued neutrosophic UnderSet* on $U$ is a triple of functions

$$T,\ I,\ F :\ U \longrightarrow [\Psi, 1]$$

such that there exists at least one $x \in U$ with

$$T(x) < 0 \quad \text{or} \quad I(x) < 0 \quad \text{or} \quad F(x) < 0.$$

We denote it by

$$A_{\text{under}} = \{\langle x,\ T(x), I(x), F(x)\rangle :\ x \in U\}.$$

**Proposition 2.1.25** (Embedding of the classical SVNS into an UnderSet)**.** *Every single-valued neutrosophic set A on U (with values in* $[0,1]$*) can be viewed as an UnderSet on U for any choice of* $\Psi < 0$.

*Proof.* Let $A$ be given by $T_A, I_A, F_A : U \to [0,1]$. Fix $\Psi < 0$ and define $T := T_A$, $I := I_A$, $F := F_A$, now regarded as maps into $[\Psi,1]$. Since $[0,1] \subseteq [\Psi,1]$, this defines an UnderSet-valued triple on the same universe. □

**Definition 2.1.26** (Single-valued neutrosophic OffSet)**.** [137] Fix parameters $\Psi < 0 < 1 < \Omega$ and let $U$ be a nonempty universe. A *single-valued neutrosophic OffSet* on $U$ is a triple of functions

$$T,\ I,\ F :\ U \longrightarrow [\Psi, \Omega]$$

such that there exists at least one $x \in U$ with a component outside the classical range $[0,1]$, i.e.,

$$T(x) \notin [0,1] \quad \text{or} \quad I(x) \notin [0,1] \quad \text{or} \quad F(x) \notin [0,1].$$

We denote it by

$$A_{\text{off}} = \{\langle x,\ T(x), I(x), F(x)\rangle :\ x \in U\}.$$

**Proposition 2.1.27** (OffSet restrictions yield OverSets and UnderSets)**.** *Let* $A_{\text{off}} = \{\langle x, T(x), I(x), F(x)\rangle : x \in U\}$ *be a single-valued neutrosophic OffSet with* $\Psi < 0 < 1 < \Omega$*. Define the subsets*

$$U_{\text{over}} := \{x \in U :\ T(x) > 1 \text{ or } I(x) > 1 \text{ or } F(x) > 1\},$$

$$U_{\text{under}} := \{x \in U :\ T(x) < 0 \text{ or } I(x) < 0 \text{ or } F(x) < 0\}.$$

*Then the restrictions of* $T, I, F$ *to* $U_{\text{over}}$ *define a single-valued neutrosophic OverSet on* $U_{\text{over}}$*, and the restrictions to* $U_{\text{under}}$ *define a single-valued neutrosophic UnderSet on* $U_{\text{under}}$.

*Proof.* By Definition 2.1.26, the maps $T, I, F$ take values in $[\Psi,\Omega]$. Restricting the domain to $U_{\text{over}}$ preserves the codomain inclusion in $[0,\Omega]$ (since $[\Psi,\Omega] \subseteq (-\infty,\Omega]$), and by construction at least one component exceeds 1 for each $x \in U_{\text{over}}$, so the restriction satisfies Definition 2.1.22. Similarly, restricting to $U_{\text{under}}$ preserves the codomain inclusion in $[\Psi,1]$, and by construction at least one component is negative for each $x \in U_{\text{under}}$, so the restriction satisfies Definition 2.1.24. □

For reference, a brief overview of the single-valued neutrosophic set versus the single-valued neutrosophic OffSet is provided in Table 2.3.

Table 2.3: Concise comparison: single-valued neutrosophic set vs. single-valued neutrosophic OffSet.

| **Aspect** | **SVNS** | **SVNS OffSet** |
|---|---|---|
| Universe | $U$ | $U$ |
| Membership maps | $T, I, F : U \to [0, 1]$ | $T, I, F : U \to [\Psi, \Omega]$ |
| Range parameters | none | $\Psi < 0 < 1 < \Omega$ |
| Out-of-$[0, 1]$ values | not allowed | allowed (and typically required) |
| Pointwise sum bound | automatic: $0 \le T + I + F \le 3$ | not automatic; often use $T + I + F \le 3\Omega$ |
| Classical reduction | — | set $\Psi = 0, \Omega = 1$ (or restrict to $[0, 1]$) |

### 2.1.6 Fuzzy and Intuitionistic Fuzzy OverSets, UnderSets, and OffSets

Following the nonstandard extension philosophy in Smarandache's work (cf. [138]), one may enlarge the admissible range of membership degrees beyond the classical interval $[0, 1]$. This leads to *over*-variants (some degrees exceed 1), *under*-variants (some degrees are below 0), and *off*-variants (both phenomena may occur).

**Definition 2.1.28** (Fuzzy OverSet)**.** (cf. [138]) Let $X$ be a nonempty universe and fix an *overlimit* $\Omega > 1$. A *fuzzy OverSet* on $X$ is a membership function

$$\mu : X \longrightarrow [0, \Omega]$$

such that $\mu(x) > 1$ for at least one $x \in X$. We denote it by

$$\widetilde{A}_{\text{over}} = \{(x, \mu(x)) : x \in X\}.$$

**Definition 2.1.29** (Fuzzy UnderSet)**.** (cf. [138]) Let $X$ be a nonempty universe and fix an *underlimit* $\Psi < 0$. A *fuzzy UnderSet* on $X$ is a membership function

$$\mu : X \longrightarrow [\Psi, 1]$$

such that $\mu(x) < 0$ for at least one $x \in X$. We denote it by

$$\widetilde{A}_{\text{under}} = \{(x, \mu(x)) : x \in X\}.$$

**Definition 2.1.30** (Fuzzy OffSet)**.** (cf. [138]) Let $X$ be a nonempty universe and fix parameters $\Psi < 0 < 1 < \Omega$. A *fuzzy OffSet* on $X$ is a membership function

$$\mu : X \longrightarrow [\Psi, \Omega]$$

such that there exist (not necessarily distinct) elements $x, y \in X$ with

$$\mu(x) > 1 \qquad \text{and} \qquad \mu(y) < 0.$$

We denote it by

$$\widetilde{A}_{\text{off}} = \{(x, \mu(x)) : x \in X\}.$$

**Remark 2.1.31.** From a fuzzy OffSet $\mu : X \to [\Psi, \Omega]$, one obtains a fuzzy OverSet by restricting the domain to $X_{\text{over}} := \{x \in X : \mu(x) > 1\}$, and a fuzzy UnderSet by restricting to $X_{\text{under}} := \{x \in X : \mu(x) < 0\}$, provided these sets are nonempty.

For reference, we include a schematic overview of a fuzzy OffSet in Figure 2.1.

**Corollary 2.1.32** (OffSet restrictions)**.** *Every fuzzy OffSet admits natural restrictions that form a fuzzy OverSet and a fuzzy UnderSet (on the corresponding nonempty subuniverses).*

*Proof.* Let $\mu : X \to [\Psi, \Omega]$ be a fuzzy OffSet. By definition, there exist $x, y \in X$ with $\mu(x) > 1$ and $\mu(y) < 0$, hence $X_{\text{over}} \neq \emptyset$ and $X_{\text{under}} \neq \emptyset$. The restricted maps $\mu|_{X_{\text{over}}} : X_{\text{over}} \to [0, \Omega]$ and $\mu|_{X_{\text{under}}} : X_{\text{under}} \to [\Psi, 1]$ satisfy the defining conditions of Definitions 2.1.28 and 2.1.29, respectively. □

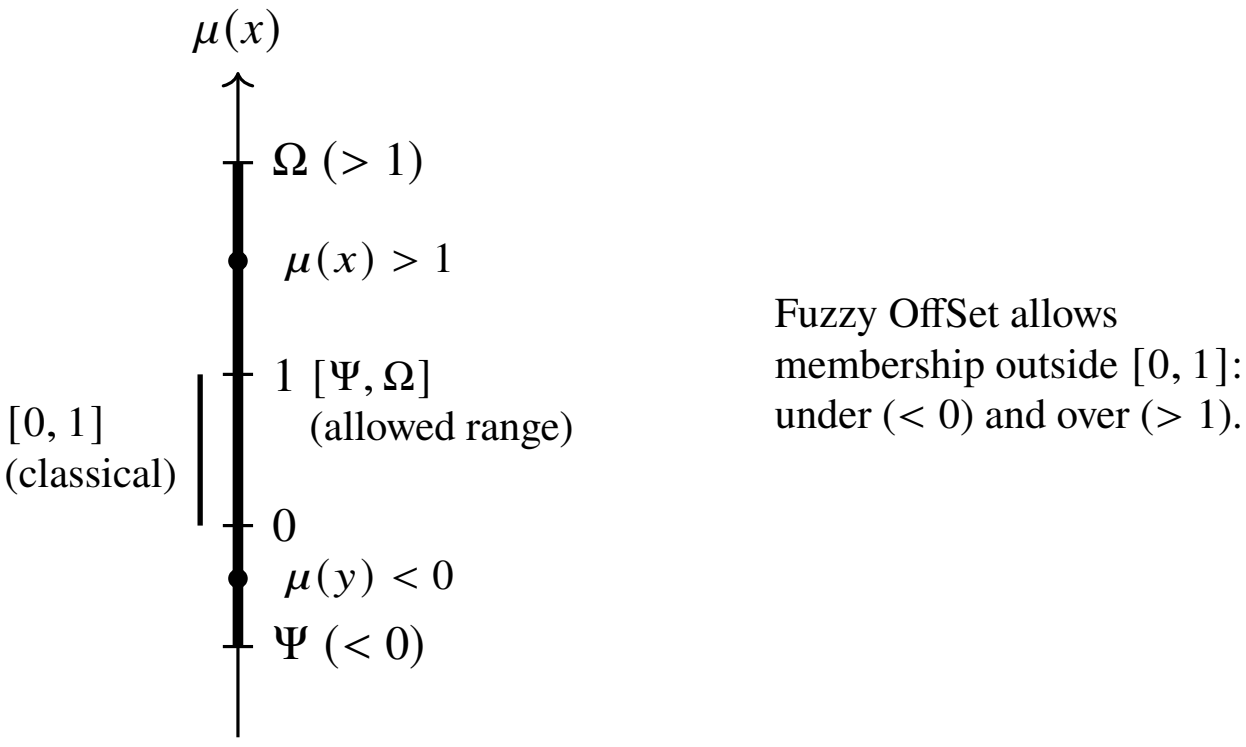

Figure 2.1: Schematic overview of a fuzzy OffSet (Definition 2.1.30).

We employ the standard intuitionistic fuzzy convention: each element $x \in X$ is assigned a pair $(\mu(x), \nu(x))$ (membership / nonmembership). In the classical case one requires $0 \leq \mu(x), \nu(x) \leq 1$ and $\mu(x)+\nu(x) \leq 1$. For the over/under/off variants, we enlarge the codomain while keeping an analogous "Atanassov-type" constraint in the enlarged range.

**Definition 2.1.33** (Intuitionistic Fuzzy OverSet)**.** (cf. [138]) Let $X$ be a nonempty universe and fix an overlimit $\Omega > 1$. An *intuitionistic fuzzy OverSet* on $X$ is a pair of maps

$$\mu, \nu : X \longrightarrow [0, \Omega]$$

such that

$$0 \leq \mu(x) + \nu(x) \leq \Omega \qquad (\forall x \in X),$$

and there exists at least one $x \in X$ with $\mu(x) > 1$ or $\nu(x) > 1$. We denote it by

$$A_{\text{over}} = \{(x, \mu(x), \nu(x)) : \ x \in X\}.$$

**Definition 2.1.34** (Intuitionistic Fuzzy UnderSet)**.** (cf. [138]) Let $X$ be a nonempty universe and fix an underlimit $\Psi < 0$. An *intuitionistic fuzzy UnderSet* on $X$ is a pair of maps

$$\mu, \nu : X \longrightarrow [\Psi, 1]$$

such that

$$\Psi \leq \mu(x) + \nu(x) \leq 1 \qquad (\forall x \in X),$$

and there exists at least one $x \in X$ with $\mu(x) < 0$ or $\nu(x) < 0$. We denote it by

$$A_{\text{under}} = \{(x, \mu(x), \nu(x)) : \ x \in X\}.$$

**Definition 2.1.35** (Intuitionistic Fuzzy OffSet)**.** (cf. [138]) Let $X$ be a nonempty universe and fix $\Psi < 0 < 1 < \Omega$. An *intuitionistic fuzzy OffSet* on $X$ is a pair of maps

$$\mu, \nu : X \longrightarrow [\Psi, \Omega]$$

such that

$$\Psi \leq \mu(x) + \nu(x) \leq \Omega \qquad (\forall x \in X),$$

and there exist elements $x, y \in X$ for which

$$\mu(x) > 1 \text{ or } \nu(x) > 1, \qquad \text{and} \qquad \mu(y) < 0 \text{ or } \nu(y) < 0.$$

We denote it by

$$A_{\text{off}} = \{(x, \mu(x), \nu(x)) : \ x \in X\}.$$

**Corollary 2.1.36** (OffSet restrictions)**.** *Every intuitionistic fuzzy OffSet admits natural restrictions that form an intuitionistic fuzzy OverSet and an intuitionistic fuzzy UnderSet (on the corresponding nonempty subuniverses).*

*Proof.* Let $(\mu, \nu)$ define an intuitionistic fuzzy OffSet on $X$. Set

$$X_{\text{over}} := \{x \in X : \mu(x) > 1 \text{ or } \nu(x) > 1\}, \qquad X_{\text{under}} := \{x \in X : \mu(x) < 0 \text{ or } \nu(x) < 0\}.$$

By Definition 2.1.35, both sets are nonempty. Restricting $(\mu, \nu)$ to $X_{\text{over}}$ yields maps into $[0, \Omega]$ (since $[\Psi, \Omega] \subseteq (-\infty, \Omega]$), and the inequality $\Psi \le \mu + \nu \le \Omega$ implies $0 \le \mu + \nu \le \Omega$ on $X_{\text{over}}$ after truncating the lower bound (because $\mu, \nu \ge 0$ on $[0, \Omega]$). Hence Definition 2.1.33 holds. Similarly, restricting to $X_{\text{under}}$ yields maps into $[\Psi, 1]$ and preserves $\Psi \le \mu + \nu \le 1$, so Definition 2.1.34 holds. □

Single-valued neutrosophic over/under/off sets (Section 2.1.5) allow three independent components $T, I, F$ to leave $[0, 1]$. The following embeddings formalize that the neutrosophic variants subsume the fuzzy and intuitionistic fuzzy variants.

**Proposition 2.1.37** (Neutrosophic OffSet subsumes fuzzy and intuitionistic fuzzy OffSets)**.** *Fix* $\Psi < 0 < 1 < \Omega$.

(i) *Let* $\widetilde{A}_{\text{off}}$ *be a fuzzy OffSet on X with membership* $\mu : X \to [\Psi, \Omega]$. *Define*

$$T(x) := \mu(x), \qquad I(x) := 0, \qquad F(x) := 0 \qquad (x \in X).$$

*Then* $(T, I, F)$ *defines a single-valued neutrosophic OffSet on X.*

(ii) *Let* $A_{\text{off}}$ *be an intuitionistic fuzzy OffSet on X with* $\mu, \nu : X \to [\Psi, \Omega]$. *Define*

$$T(x) := \mu(x), \qquad F(x) := \nu(x), \qquad I(x) := 0 \qquad (x \in X).$$

*Then* $(T, I, F)$ *defines a single-valued neutrosophic OffSet on X.*

*In both cases, the construction is an embedding at the level of membership data.*

*Proof.* In (i), each component lies in $[\Psi, \Omega]$, and since $\widetilde{A}_{\text{off}}$ has points with $\mu > 1$ and $\mu < 0$, the resulting $T$ has values outside $[0, 1]$, so the neutrosophic OffSet condition holds. In (ii), $T, F \in [\Psi, \Omega]$ and $I \equiv 0$, and the OffSet condition holds because there exist points where $\mu$ or $\nu$ exceeds 1 and points where $\mu$ or $\nu$ is negative. Thus $(T, I, F)$ is a neutrosophic OffSet. □

**Proposition 2.1.38** (Neutrosophic OverSet and UnderSet subsume the fuzzy and intuitionistic fuzzy cases)**.** *Fix* $\Omega > 1$ *and* $\Psi < 0$. *The constructions in Proposition 2.1.37 restrict to the over-only and under-only regimes:*

(i) *A fuzzy OverSet (resp. UnderSet) embeds into a neutrosophic OverSet (resp. UnderSet) by* $T = \mu$ *and* $I = F \equiv 0$.

(ii) *An intuitionistic fuzzy OverSet (resp. UnderSet) embeds into a neutrosophic OverSet (resp. UnderSet) by* $T = \mu$, $F = \nu$, *and* $I \equiv 0$.

*Proof.* The verification is identical to Proposition 2.1.37, using the corresponding over/under domain constraints and the existence of at least one element with a component $> 1$ (over) or $< 0$ (under). □

For crisp sets, membership is binary. One may still present a "crisp OffSet" perspective by allowing the two values to be any chosen pair $\Psi < \Omega$ (typically $\Psi = 0, \Omega = 1$).

**Definition 2.1.39** (Crisp set and crisp OffSet representation)**.** Let $X$ be a nonempty universe and fix two scalars $\Psi < \Omega$. A *crisp set* $A \subseteq X$ is equivalently specified by its characteristic function

$$\chi_A : X \longrightarrow \{\Psi, \Omega\}, \qquad \chi_A(x) = \begin{cases} \Omega, & x \in A, \\ \Psi, & x \notin A. \end{cases}$$

When $\Psi = 0$ and $\Omega = 1$, this is the usual crisp membership map $\chi_A : X \to \{0, 1\}$.

**Proposition 2.1.40** (Fuzzy OffSet generalizes the crisp case)**.** *Let* $\Psi < 0 < 1 < \Omega$ *and let* $A \subseteq X$ *be a crisp set with characteristic function* $\chi_A : X \to \{0, 1\}$*. Then* $\chi_A$ *is also a fuzzy OffSet membership function* $\mu : X \to [\Psi, \Omega]$*. In particular, crisp membership is obtained from fuzzy OffSet membership by restricting the codomain to* $\{0, 1\}$*.*

*Proof.* Since $\{0, 1\} \subseteq [\Psi, \Omega]$, the map $\chi_A$ can be viewed as a function into $[\Psi, \Omega]$, hence it is a fuzzy OffSet membership function in the sense of Definition 2.1.30. Conversely, restricting a fuzzy OffSet membership map $\mu$ to the two values $\{0, 1\}$ yields a crisp membership map. □

**Corollary 2.1.41** (Crisp Over/Under specializations)**.** *Crisp membership maps are special cases of fuzzy OverSets and fuzzy UnderSets as well, by the same codomain restriction argument.*

*Proof.* Since $\{0, 1\} \subseteq [0, \Omega]$ and $\{0, 1\} \subseteq [\Psi, 1]$, the same reasoning as in Proposition 2.1.40 applies. □

### 2.1.7 Neutrosophic Triplet Strong Set

In neutrosophic algebra (cf. [125]), one often works with structures in which each element $x$ carries its own *local neutral* (identity-like) element and its own *local opposite* (anti-element), both defined relative to a fixed binary operation.

**Definition 2.1.42** (Neutrosophic Triplet Strong Set)**.** (cf. [125]) A *neutrosophic triplet strong set* is an algebraic structure

$$(N, *),$$

where $N \neq \emptyset$ is a set and $* : N \times N \to N$ is a binary operation, for which there exist two maps

$$\mathrm{neut}, \mathrm{anti} : N \longrightarrow N$$

satisfying the following axioms for every $x \in N$:

(i) **Associativity.**

$$(x * y) * z \;=\; x * (y * z) \qquad (\forall\, x, y, z \in N).$$

(ii) **Local neutral (two-sided).**

$$x * \mathrm{neut}(x) \;=\; x \qquad \text{and} \qquad \mathrm{neut}(x) * x \;=\; x.$$

(iii) **Local opposite (two-sided).**

$$x * \mathrm{anti}(x) \;=\; \mathrm{neut}(x) \qquad \text{and} \qquad \mathrm{anti}(x) * x \;=\; \mathrm{neut}(x).$$

(iv) **Uniqueness (strongness).** For each fixed $x \in N$, the elements $\mathrm{neut}(x)$ and $\mathrm{anti}(x)$ are *unique* in the sense that:

  (a) if $e \in N$ satisfies $x * e = e * x = x$, then $e = \mathrm{neut}(x)$;

  (b) if $a \in N$ satisfies $x * a = a * x = \mathrm{neut}(x)$, then $a = \mathrm{anti}(x)$.

The ordered triple

$$\langle x,\ \mathrm{neut}(x),\ \mathrm{anti}(x) \rangle$$

is called the *neutrosophic triplet* associated with $x$.

**Remark 2.1.43.** (i) Definition 2.1.42 does *not* require a global identity element $e$ satisfying $x * e = e * x = x$ for all $x \in N$. Instead, each element $x$ has its own $\mathrm{neut}(x)$, which may depend on $x$.

(ii) Likewise, $\mathrm{anti}(x)$ is a *local* opposite: it neutralizes $x$ to $\mathrm{neut}(x)$, not necessarily to a global identity.

(iii) The term "strong" refers to the uniqueness clause in (iv). If one drops uniqueness, the structure is often called a *neutrosophic triplet set* (or simply a neutrosophic triplet structure) in the weaker sense.

**Proposition 2.1.44** (Basic consequences)**.** *Let $(N, *)$ be a neutrosophic triplet strong set. Then, for every $x \in N$,*

(i) $\mathrm{neut}(x) = x * \mathrm{anti}(x) = \mathrm{anti}(x) * x$.

(ii) *If* $\mathrm{anti}(x) = \mathrm{anti}(y)$ *and* $\mathrm{neut}(x) = \mathrm{neut}(y)$*, then* $x = y$ *whenever* $x * \mathrm{neut}(y) = x$ *and* $\mathrm{neut}(y) * x = x$. *(In particular, if* $\mathrm{neut}(x) = \mathrm{neut}(y)$ *and* $\mathrm{neut}(y)$ *is also a neutral for* $x$*, then* $\mathrm{neut}(x) = \mathrm{neut}(y)$ *by uniqueness.)*

*Proof.* (i) is exactly axiom (iii). For (ii), note that if $\mathrm{neut}(y)$ is a two-sided neutral for $x$, then by Definition 2.1.42(iv)(a) we must have $\mathrm{neut}(y) = \mathrm{neut}(x)$. Similar uniqueness reasoning applies to $\mathrm{anti}(\cdot)$ using (iv)(b). □

**Example 2.1.45** (A small explicit instance)**.** Let $N := \{0, 1, 2\}$ and define $x * y := x$ (the left-projection operation) for all $x, y \in N$. Then $*$ is associative. For each $x \in N$, set

$$\mathrm{neut}(x) := x, \qquad \mathrm{anti}(x) := 0.$$

We have

$$x * \mathrm{neut}(x) = x * x = x, \qquad \mathrm{neut}(x) * x = x * x = x,$$

and

$$x * \mathrm{anti}(x) = x * 0 = x = \mathrm{neut}(x), \qquad \mathrm{anti}(x) * x = 0 * x = 0.$$

Hence the last equality fails unless $x = 0$. Therefore this structure is *not* a neutrosophic triplet strong set. This illustrates that both-sided conditions in Definition 2.1.42(ii)–(iii) are essential: the operation must allow *both* $x * \mathrm{anti}(x) = \mathrm{neut}(x)$ and $\mathrm{anti}(x) * x = \mathrm{neut}(x)$ simultaneously.

### 2.1.8 MultiNeutrosophic Set

MultiNeutrosophic sets (cf. [139]) extend (single-valued) neutrosophic sets by allowing *multiple* truth-, indeterminacy-, and falsity-evaluations for the same element, typically coming from different sources (experts, sensors, models, or criteria). As a further generalization of the MultiNeutrosophic Set, the concept of a Refined Neutrosophic Set is also known [112].

**Definition 2.1.46** (MultiNeutrosophic Set)**.** (cf. [139, 140]) Let $\mathcal{U}$ be a nonempty universe. Fix integers $p, r, s \geq 0$ and set $n := p + r + s$. Write $\mathcal{P}^*([0,1]) := \mathcal{P}([0,1]) \setminus \{\emptyset\}$ for the family of nonempty subsets of $[0, 1]$.

A *MultiNeutrosophic set $M$* on $\mathcal{U}$ (of type $(p, r, s)$) is specified by three families of maps

$$T_j : \mathcal{U} \to \mathcal{P}^*([0,1]) \quad (j = 1, \ldots, p), \qquad I_k : \mathcal{U} \to \mathcal{P}^*([0,1]) \quad (k = 1, \ldots, r), \qquad F_\ell : \mathcal{U} \to \mathcal{P}^*([0,1]) \quad (\ell = 1, \ldots, s),$$

and is denoted by

$$M = \Big\{ \big\langle x;\ T_1(x), \ldots, T_p(x);\ I_1(x), \ldots, I_r(x);\ F_1(x), \ldots, F_s(x) \big\rangle \ \Big|\ x \in \mathcal{U} \Big\}.$$

Here $T_1(x), \ldots, T_p(x)$ are the *multi-truth (multi-membership) components*, $I_1(x), \ldots, I_r(x)$ are the *multi-indeterminacy components*, and $F_1(x), \ldots, F_s(x)$ are the *multi-falsity (multi-nonmembership) components*.

**Admissibility bounds.** For each $x \in \mathcal{U}$, assume the aggregate bounds

$$0 \leq \sum_{j=1}^{p} \inf T_j(x) + \sum_{k=1}^{r} \inf I_k(x) + \sum_{\ell=1}^{s} \inf F_\ell(x) \leq \sum_{j=1}^{p} \sup T_j(x) + \sum_{k=1}^{r} \sup I_k(x) + \sum_{\ell=1}^{s} \sup F_\ell(x) \leq n,$$

where $\inf(\cdot)$ and $\sup(\cdot)$ are taken in $[0, 1]$ (well-defined since each component is a nonempty bounded subset). No further relations among the components are imposed unless specified by the application.

**Remark 2.1.47** (Interpretation and common special cases)**.** (cf. [139])

(i) The intended meaning is that each element $x$ receives several evaluations of truth/indeterminacy/falsity, often from different information sources. Set-valued components $T_j(x) \subseteq [0,1]$ allow one to encode ranges or uncertainty about the $j$-th source itself.

(ii) **Single-Valued MultiNeutrosophic Set (SVMNS).** If each component is a singleton,

$$T_j(x) = \{t_j(x)\}, \quad I_k(x) = \{i_k(x)\}, \quad F_\ell(x) = \{f_\ell(x)\} \qquad (x \in \mathcal{U}),$$

then the MultiNeutrosophic data become numerical tuples

$$\big(t_1(x), \dots, t_p(x);\ i_1(x), \dots, i_r(x);\ f_1(x), \dots, f_s(x)\big) \in [0,1]^n,$$

subject to the corresponding summed bounds.

(iii) **Interval-Valued MultiNeutrosophic Set (IVMNS).** If each component is an interval of $[0,1]$, i.e., $T_j(x) = [\underline{t}_j(x), \overline{t}_j(x)]$ etc., then the admissibility inequalities in Definition 4.1.3 become bounds on sums of interval endpoints.

**Proposition 2.1.48** (Reduction to an ordinary (single-valued) neutrosophic set)**.** *A MultiNeutrosophic set generalizes the usual (single-valued) neutrosophic set: it is obtained as the special case $p = r = s = 1$ with singleton-valued components.*

*Proof.* Take $p = r = s = 1$. Assume that for each $x \in \mathcal{U}$,

$$T_1(x) = \{T(x)\}, \qquad I_1(x) = \{I(x)\}, \qquad F_1(x) = \{F(x)\}$$

for some functions $T, I, F : \mathcal{U} \to [0,1]$. Then the MultiNeutrosophic representation in Definition 4.1.3 reduces to the assignment

$$x \longmapsto (T(x), I(x), F(x)) \in [0,1]^3,$$

which is precisely the usual single-valued neutrosophic membership triple. Moreover, the admissibility bound becomes

$$0 \le T(x) + I(x) + F(x) \le 3,$$

so the ordinary neutrosophic constraint is recovered. □

### 2.1.9 MultiCrisp Set and its relation to MultiNeutrosophic Sets

**Definition 2.1.49** (MultiCrisp Set)**.** Let $\mathcal{U}$ be a nonempty universe and fix an integer $k \ge 1$. A *MultiCrisp set* (with $k$ crisp evaluations) is specified by $k$ characteristic functions

$$\mu_1, \dots, \mu_k : \mathcal{U} \to \{0,1\},$$

and can be represented as

$$M_{\mathrm{cr}} = \Big\{ (x, \mu_1(x), \dots, \mu_k(x)) \ \Big|\ x \in \mathcal{U} \Big\}.$$

When $k = 1$, this reduces to an ordinary crisp set via its single characteristic function. When $k \ge 2$, the $k$-tuple records multiple crisp inclusion/exclusion decisions (e.g., by different rules or agents).

**Proposition 2.1.50** (MultiCrisp sets generalize crisp sets)**.** *An ordinary crisp set is the special case of a MultiCrisp set with $k = 1$.*

*Proof.* Let $C \subseteq \mathcal{U}$ be a crisp set with characteristic function $\chi_C : \mathcal{U} \to \{0,1\}$. Taking $k = 1$ and $\mu_1 = \chi_C$ in Definition 2.1.49 yields

$$M_{\mathrm{cr}} = \{(x, \chi_C(x)) \mid x \in \mathcal{U}\},$$

which is exactly the usual representation of $C$ by a single membership indicator. □

**Theorem 2.1.51** (MultiNeutrosophic sets subsume MultiCrisp sets)**.** *Every MultiCrisp set with $k \ge 2$ can be embedded as a MultiNeutrosophic set of type $(p, r, s) = (k, 0, 0)$ by taking the truth-components as singleton $\{0,1\}$-valued sets.*

*Proof.* Let $M_{\mathrm{cr}}$ be a MultiCrisp set given by $\mu_1, \ldots, \mu_k : \mathcal{U} \to \{0, 1\}$ with $k \geq 2$. Define a MultiNeutrosophic set $M$ of type $(p, r, s) = (k, 0, 0)$ by setting, for each $j = 1, \ldots, k$,

$$T_j(x) := \{\mu_j(x)\} \subseteq [0, 1] \qquad (x \in \mathcal{U}),$$

and taking no indeterminacy or falsity components (since $r = s = 0$).

Then $T_j : \mathcal{U} \to \mathcal{P}^*([0, 1])$ is well-defined and singleton-valued. Moreover, for each $x$,

$$\sum_{j=1}^{k} \inf T_j(x) = \sum_{j=1}^{k} \mu_j(x) = \sum_{j=1}^{k} \sup T_j(x) \in \{0, 1, \ldots, k\},$$

so the admissibility inequalities in Definition 4.1.3 hold with $n = p = k$. Hence $M$ is a valid MultiNeutrosophic set, and its stored data $\langle x; T_1(x), \ldots, T_k(x)\rangle$ recovers exactly the MultiCrisp tuple $(x, \mu_1(x), \ldots, \mu_k(x))$. □

### 2.1.10 Single-Valued MultiNeutrosophic OffSet

The notion of a *multi*-neutrosophic OffSet combines two extensions of the classical single-valued neutrosophic setting: (i) *multi-evaluation* (several assessments of truth/indeterminacy/falsity for the same element) and (ii) the *OffSet* philosophy (allowing degrees to go below 0 and/or above 1). Using the same pattern, one may analogously introduce MultiNeutrosophic OverSets and UnderSets; we omit them here.

**Definition 2.1.52** (Single-Valued MultiNeutrosophic OffSet)**.** Fix real bounds $\Psi < 0 < 1 < \Omega$ and let $\mathcal{U}$ be a universe. A *Single-Valued MultiNeutrosophic OffSet* (abbrev. *SVMNO*) on $\mathcal{U}$ is a specification of three finite index sets

$$[p] := \{1, \ldots, p\}, \qquad [r] := \{1, \ldots, r\}, \qquad [s] := \{1, \ldots, s\}, \quad \text{with } p, r, s \in \mathbb{N},$$

together with families of functions

$$T_i,\ I_j,\ F_k : \mathcal{U} \longrightarrow [\Psi, \Omega] \qquad (i \in [p],\ j \in [r],\ k \in [s]),$$

such that for every $x \in \mathcal{U}$ one has the *bounded-sum (multi-OffSet) constraint*

$$\Psi \ \leq\ \sum_{i=1}^{p} T_i(x) \ +\ \sum_{j=1}^{r} I_j(x) \ +\ \sum_{k=1}^{s} F_k(x) \ \leq\ \Omega. \tag{2.1}$$

We write the resulting SVMNO as

$$\mathcal{A} = \Big\{\Big\langle x;\ (T_i(x))_{i=1}^{p},\ (I_j(x))_{j=1}^{r},\ (F_k(x))_{k=1}^{s}\Big\rangle :\ x \in \mathcal{U}\Big\}.$$

Here $T_i(x)$ are the $p$ truth-evaluations, $I_j(x)$ are the $r$ indeterminacy-evaluations, and $F_k(x)$ are the $s$ falsity-evaluations of $x$, each permitted to lie outside $[0, 1]$ due to $[\Psi, \Omega]$.

**OffSet condition.** We call $\mathcal{A}$ *properly OffSet* if at least one element exhibits an out-of-range component, i.e., there exists $x \in \mathcal{U}$ such that

$$\max\Big( \max_i T_i(x),\ \max_j I_j(x),\ \max_k F_k(x) \Big) > 1 \quad \text{or} \quad \min\Big( \min_i T_i(x),\ \min_j I_j(x),\ \min_k F_k(x) \Big) < 0.$$

(If this does not occur, then $\mathcal{A}$ is simply a multi-neutrosophic set with all values in $[0, 1]$.)

**Remark 2.1.53** (Single-valued neutrosophic OffSet as a special case)**.** If $p = r = s = 1$, then Definition 2.1.52 reduces to a (single-valued) neutrosophic OffSet with one triple $(T_1(x), I_1(x), F_1(x)) \in [\Psi, \Omega]^3$ per element $x \in \mathcal{U}$, subject to $\Psi \leq T_1(x) + I_1(x) + F_1(x) \leq \Omega$.

**Theorem 2.1.54** (SVMNO generalizes the single-valued neutrosophic OffSet)**.** *Let $\Psi < 0 < 1 < \Omega$ and let SVNO be any single-valued neutrosophic OffSet on $\mathcal{U}$, given by three functions $T, I, F : \mathcal{U} \to [\Psi, \Omega]$ satisfying*

$$\Psi \leq T(x) + I(x) + F(x) \leq \Omega \qquad (\forall x \in \mathcal{U}).$$

*Then SVNO can be embedded into a Single-Valued MultiNeutrosophic OffSet (SVMNO).*

*Proof.* Take $p = r = s = 1$ in Definition 2.1.52 and set

$$T_1 := T, \qquad I_1 := I, \qquad F_1 := F.$$

Then $T_1, I_1, F_1 : \mathcal{U} \to [\Psi, \Omega]$ and the bounded-sum constraint (2.1) becomes exactly $\Psi \leq T(x) + I(x) + F(x) \leq \Omega$ for all $x \in \mathcal{U}$. Hence the SVMNO determined by $(T_1, I_1, F_1)$ coincides with the given $SVNO$. Therefore every single-valued neutrosophic OffSet is obtained as the special case $p = r = s = 1$ of an SVMNO, proving that SVMNOs generalize $SVNO$s. □

### 2.1.11 Soft Sets and Soft Expert Sets

Soft set theory (Molodtsov) provides a parameterized way to describe collections of objects without requiring additional structure such as membership grades or probability measures. A soft expert set extends a soft set by indexing the approximations not only by parameters but also by *experts* and their *opinions*, which is useful when multiple assessments must be represented explicitly (e.g., [49, 50, 141]).

**Definition 2.1.55** (Soft set)**.** [45] Let $U$ be a nonempty universe and let $E$ be a set of parameters. A *soft set* over $U$ (with respect to $E$) is a pair $(F, E)$, where

$$F : E \longrightarrow \mathcal{P}(U)$$

is a mapping from parameters to subsets of $U$. For each $e \in E$, the set $F(e) \subseteq U$ is interpreted as the collection of objects of $U$ that are *approximately* compatible with the parameter $e$.

**Remark 2.1.56** (Crisp sets as a special case)**.** Every crisp set $A \subseteq U$ is obtained from a soft set by taking a singleton parameter set $E = \{e\}$ and defining $F(e) := A$. Conversely, a soft set $(F, \{e\})$ on a singleton parameter set corresponds uniquely to the crisp set $F(e) \subseteq U$.

**Proposition 2.1.57** (Soft sets generalize crisp sets)**.** *Every crisp subset $A \subseteq U$ can be represented as a soft set over $U$.*

*Proof.* Fix a crisp set $A \subseteq U$. Let $E := \{e\}$ and define $F : E \to \mathcal{P}(U)$ by $F(e) := A$. Then $(F, E)$ is a soft set over $U$ (Definition 3.9.3), and its unique approximation set equals $A$. Thus crisp sets embed into soft sets as the singleton-parameter case. □

**Definition 2.1.58** (Soft subset)**.** [45] Let $(F, A)$ and $(G, B)$ be soft sets over the same universe $U$, where $A, B \subseteq E$. We say that $(F, A)$ is a *soft subset* of $(G, B)$, written $(F, A) \mathrel{\widetilde{\subseteq}} (G, B)$, if

$$A \subseteq B \qquad \text{and} \qquad F(e) \subseteq G(e) \quad (\forall e \in A).$$

**Definition 2.1.59** (Null soft set)**.** [45] Let $U$ be a universe and let $A \subseteq E$ be a nonempty parameter set. A soft set $(F, A)$ over $U$ is called a *null soft set* (denoted $\Phi_A$) if

$$F(e) = \emptyset \qquad (\forall e \in A).$$

**Definition 2.1.60** (Full soft set)**.** [142] Let $(F, A)$ be a soft set over $U$. It is called *full* if

$$\bigcup_{a \in A} F(a) \;=\; U.$$

Equivalently, every element of $U$ is contained in at least one approximation set $F(a)$.

We now incorporate *experts* and their *opinions*.

**Definition 2.1.61** (Soft expert set)**.** [49] Let $U$ be a nonempty universe, $E$ a set of parameters, $X$ a set of experts, and $O$ a set of opinions. (For example, one may take $O = \{1, 0\}$ to encode "agree" vs. "disagree", but $O$ is arbitrary.) Put

$$Z \;:=\; E \times X \times O, \qquad A \subseteq Z.$$

A *soft expert set* over $U$ (with respect to the index set $A$) is a pair $(F, A)$, where

$$F : A \longrightarrow \mathcal{P}(U).$$

For $\alpha = (e, x, o) \in A$, the set $F(\alpha) = F(e, x, o) \subseteq U$ is interpreted as the collection of objects of $U$ selected (or judged acceptable) under the parameter $e$, by the expert $x$, with opinion $o$.

**Remark 2.1.62** (Reduction to a soft set)**.** If $|X| = |O| = 1$, say $X = \{x_0\}$ and $O = \{o_0\}$, then $A \subseteq E \times \{x_0\} \times \{o_0\}$ can be identified with a subset $A_E \subseteq E$, and a soft expert set $(F, A)$ becomes a soft set $(F', A_E)$ by $F'(e) := F(e, x_0, o_0)$. Hence soft expert sets extend soft sets by adding expert/opinion indexing.

### 2.1.12 Neutrosophic Axial Sets and Partner Multineutrosophic Sets

Recently, the notions of a *neutrosophic axial set* [143] and a *partner multineutrosophic set* [140, 144] have been proposed as set-theoretic constructions related to neutrosophic modeling. For completeness, we restate clean definitions below and fix notation.

**Definition 2.1.63** (Discrete subset)**.** Let $(X, \tau)$ be a topological space. A subset $D \subseteq X$ is called *discrete* (in $X$) if every point of $D$ is isolated relative to $D$; equivalently,

$$\forall x \in D\ \exists U \in \tau \text{ such that } U \cap D = \{x\}.$$

**Definition 2.1.64** (Neutrosophic axial set)**.** [143] Let $X$ be a nonempty set. A *neutrosophic axial set* (NAS) on $X$ is an ordered triple

$$\mathsf{NA} = \langle A,\ A_1,\ A_2 \rangle,$$

where $A, A_1, A_2 \subseteq X$ satisfy the disjointness constraints

$$A \cap A_1 = \emptyset \qquad \text{and} \qquad A \cap A_2 = \emptyset.$$

We call $A$ the *axis* of $\mathsf{NA}$, and $A_1, A_2$ its *axial parts*. The class of all neutrosophic axial sets on $X$ is denoted by

$$\mathfrak{NA}(X) := \big\{ \langle A, A_1, A_2 \rangle :\ A, A_1, A_2 \subseteq X,\ A \cap A_i = \emptyset\ (i = 1, 2) \big\}.$$

**Remark 2.1.65** (Optional topological specialization)**.** In some applications one takes $X$ to be a topological space and imposes extra conditions on the parts, e.g., requiring $A_1$ and/or $A_2$ to be discrete subsets of $X \setminus A$ (Definition 2.1.63). These requirements are *additional structure* and are not necessary for Definition 2.1.64.

**Example 2.1.66** (A concrete NAS in $\mathbb{R}$)**.** Let $X = \mathbb{R}$ with its usual topology and let $A = (1, 2)$. Choose

$$A_1 = \{0\}, \qquad A_2 = \{3, 4, 5, \dots\}.$$

Then $A \cap A_1 = \emptyset$ and $A \cap A_2 = \emptyset$, so $\langle (1, 2), A_1, A_2 \rangle \in \mathfrak{NA}(\mathbb{R})$. Moreover, both $A_1$ and $A_2$ are discrete subsets of $\mathbb{R} \setminus (1, 2)$.

**Definition 2.1.67** (Partner multineutrosophic set)**.** [140, 144] Let $M_n$ be a multineutrosophic set on $X$ as in Definition **??**, with $n = r + s + t$. Define the *partner (averaged) membership* function

$$f_{M_n} : X \longrightarrow [0, 1], \qquad f_{M_n}(x) := \frac{1}{n} \left( \sum_{i=1}^{r} T_i(x) + \sum_{j=1}^{s} I_j(x) + \sum_{k=1}^{t} F_k(x) \right).$$

The associated fuzzy set

$$M_n^p := \big\{ \langle x,\ f_{M_n}(x) \rangle :\ x \in X \big\}$$

is called the *partner set* of $M_n$, and we refer to the pair $(M_n, M_n^p)$ as a *partner multineutrosophic set* structure.

**Remark 2.1.68** (Well-definedness)**.** Since each component satisfies $T_i(x), I_j(x), F_k(x) \in [0, 1]$, we have $0 \leq \sum T_i(x) + \sum I_j(x) + \sum F_k(x) \leq n$, hence $f_{M_n}(x) \in [0, 1]$ for all $x \in X$. Therefore $M_n^p$ is indeed a (classical) fuzzy set on $X$.

**Example 2.1.69** (Partner set computation)**.** Let $X = \{x_1, x_2\}$ and take $r = s = t = 1$ (so $n = 3$). Define

$$T_1(x_1) = 0.9,\ I_1(x_1) = 0.2,\ F_1(x_1) = 0.1, \qquad T_1(x_2) = 0.3,\ I_1(x_2) = 0.6,\ F_1(x_2) = 0.4.$$

Then the partner membership values are

$$f_{M_3}(x_1) = \frac{0.9 + 0.2 + 0.1}{3} = 0.4, \qquad f_{M_3}(x_2) = \frac{0.3 + 0.6 + 0.4}{3} = \frac{1.3}{3} \approx 0.4333,$$

and hence

$$M_3^p = \{ \langle x_1, 0.4 \rangle,\ \langle x_2, 0.4333 \rangle \}.$$

### 2.1.13 Meta Sets

Meta sets extend the fuzzy-set paradigm by allowing an element to carry *multiple* membership grades organized along a (typically infinite) binary-tree index, thereby supporting hierarchical and multi-resolution membership modeling [145–147].

**Definition 2.1.70** (Infinite full binary tree)**.** [148, 149] Let $\{0,1\}^*$ denote the set of all finite binary words (including the empty word $\epsilon$). The *infinite full binary tree* is the graph

$$\mathcal{T}_\infty = (V,E), \qquad V := \{0,1\}^*, \qquad E := \big\{\{w,w0\},\{w,w1\} : w \in \{0,1\}^*\big\}.$$

The root is $\epsilon$. Each node $w \in V$ has exactly two children $w0$ and $w1$, and (if $w \neq \epsilon$) a unique parent obtained by deleting the last bit of $w$.

**Remark 2.1.71** (Depth and prefix order)**.** For $w \in \{0,1\}^*$, its *depth* is $|w|$ (word length). We write $u \preceq w$ if $u$ is a prefix of $w$. The set of prefixes of $w$ forms the unique root-to-$w$ path.

**Definition 2.1.72** (Meta set)**.** [145–147] Let $X$ be a nonempty universe and let $\mathcal{T}_\infty = (V,E)$ be the infinite full binary tree (Definition 2.1.70). A *meta set* on $X$ is a pair

$$\rho = (X,\mu_\rho), \qquad \mu_\rho : X \times V \longrightarrow [0,1],$$

where $\mu_\rho(x,w)$ is interpreted as the membership grade of $x \in X$ *at the tree node* $w \in V$.

For each $x \in X$, the map

$$\mu_\rho(x,\cdot) : V \to [0,1]$$

is called the *membership profile* of $x$ in $\rho$.

**Remark 2.1.73** (Finite-support meta sets (common restriction))**.** In many applications one additionally assumes that, for each fixed $x \in X$, the set

$$\mathrm{Supp}_\rho(x) := \{\, w \in V : \mu_\rho(x,w) > 0 \,\}$$

is finite (or contained in the nodes up to some bounded depth). This is optional and not required for Definition 4.7.2.

**Proposition 2.1.74** (Meta sets generalize fuzzy sets)**.** *Every fuzzy set on $X$ can be realized as a meta set on $X$.*

*Proof.* Let $A$ be a fuzzy set on $X$, i.e., $A = (X,\mu_A)$ with $\mu_A : X \to [0,1]$. Let $\mathcal{T}_\infty = (V,E)$ be as in Definition 2.1.70 and let $\epsilon \in V$ be its root.

Define $\mu_\rho : X \times V \to [0,1]$ by

$$\mu_\rho(x,w) := \begin{cases} \mu_A(x), & w = \epsilon, \\ 0, & w \neq \epsilon. \end{cases}$$

Then $\rho := (X,\mu_\rho)$ is a meta set (Definition 4.7.2). Moreover, the entire membership information of $A$ is recovered at the single node $\epsilon$, since $\mu_\rho(x,\epsilon) = \mu_A(x)$ for all $x \in X$. Hence every fuzzy set is a special case of a meta set, so meta sets generalize fuzzy sets. □

**Remark 2.1.75** (Interpretation)**.** The meta-set framework permits an element $x \in X$ to have different membership grades across different nodes $w$, which can be viewed as membership information at multiple *levels*, *contexts*, or *resolutions*. The tree index provides a canonical hierarchical address system for these grades.

### 2.1.14 Binary Fuzzy Set and Binary Neutrosophic Set

Binary-valued (crisp) sets provide the classical baseline in which membership is bivalent. Binary fuzzy sets extend fuzzy sets by attaching *two* membership grades to each element, typically interpreted as two viewpoints, dimensions, or evaluation channels [150]. Binary neutrosophic sets further encode truth/indeterminacy/falsity information on a *pair* of universes [151]. We recall clean formulations below.

**Definition 2.1.76** (Binary (crisp) set)**.** [150, 152] Let $X$ be a nonempty universe. A *binary set* (i.e., a crisp subset) $A \subseteq X$ is equivalently specified by its characteristic function

$$\chi_A : X \longrightarrow \{0,1\}, \qquad \chi_A(x) = \begin{cases} 1, & x \in A, \\ 0, & x \notin A. \end{cases}$$

**Definition 2.1.77** (Binary fuzzy set)**.** [150,152] Let $X$ be a nonempty universe. A *binary fuzzy set* on $X$ is an ordered triple

$$\mathcal{B} = (X, \mu_\rho, \mu_\nu),$$

where

$$\mu_\rho, \mu_\nu : X \longrightarrow [0,1]$$

are two membership functions. For each $x \in X$, the pair

$$\big(\mu_\rho(x),\ \mu_\nu(x)\big) \in [0,1]^2$$

records two (possibly distinct) membership grades of $x$ in $\mathcal{B}$, interpreted as two evaluation perspectives or dimensions (e.g., two experts, two criteria, or two sensors).

**Remark 2.1.78.** If $\mu_\rho = \mu_\nu$, then a binary fuzzy set reduces to an ordinary fuzzy set $\mu : X \to [0,1]$. If, moreover, $\mu_\rho(x), \mu_\nu(x) \in \{0,1\}$ for all $x$, then it reduces to a binary (crisp) set.

**Definition 2.1.79** (Binary neutrosophic set)**.** [151] Let $X$ and $Y$ be nonempty sets. A *binary neutrosophic set* (BNCS) on the fixed space $(X, Y)$ is an ordered pair

$$C = \Big( (C_{11}, C_{12}, C_{13}),\ (C_{21}, C_{22}, C_{23}) \Big),$$

where

$$C_{11}, C_{12}, C_{13} \subseteq X, \qquad C_{21}, C_{22}, C_{23} \subseteq Y.$$

Informally, for the $X$-side, $C_{11}$, $C_{12}$, and $C_{13}$ encode the truth-, indeterminacy-, and falsity-related subsets, respectively; similarly, $C_{21}$, $C_{22}$, and $C_{23}$ encode the corresponding three components on $Y$.

**Definition 2.1.80** (Types of BNCS)**.** [151] Let $C$ be a BNCS on $(X, Y)$ as in Definition 2.1.79.

(i) **Type 1.** $C$ is a *BNCS-Type1* if the three components are pairwise disjoint on each side:

$$C_{11} \cap C_{12} = \emptyset, \quad C_{11} \cap C_{13} = \emptyset, \quad C_{12} \cap C_{13} = \emptyset,$$

and

$$C_{21} \cap C_{22} = \emptyset, \quad C_{21} \cap C_{23} = \emptyset, \quad C_{22} \cap C_{23} = \emptyset.$$

(ii) **Type 2.** $C$ is a *BNCS-Type2* if it is Type 1 and, in addition, each side is covered by the union of its three components:

$$C_{11} \cup C_{12} \cup C_{13} = X, \qquad C_{21} \cup C_{22} \cup C_{23} = Y.$$

(iii) **Type 3.** $C$ is a *BNCS-Type3* if the triple intersection is empty on each side and each side is covered by the union:

$$C_{11} \cap C_{12} \cap C_{13} = \emptyset, \qquad C_{21} \cap C_{22} \cap C_{23} = \emptyset,$$

and

$$C_{11} \cup C_{12} \cup C_{13} = X, \qquad C_{21} \cup C_{22} \cup C_{23} = Y.$$

**Remark 2.1.81.** Type 2 implies Type 1 by definition. Type 3 imposes a weaker disjointness condition (only the triple intersection is forced to be empty), but still requires full coverage on each side.

**Example 2.1.82** (A BNCS-Type2 on a small fixed space)**.** Let

$$X = \{x_1, x_2, x_3\}, \qquad Y = \{y_1, y_2, y_3, y_4\}.$$

Define subsets of $X$ by

$$C_{11} = \{x_1\}, \qquad C_{12} = \{x_2\}, \qquad C_{13} = \{x_3\},$$

and subsets of $Y$ by

$$C_{21} = \{y_1, y_2\}, \qquad C_{22} = \{y_3\}, \qquad C_{23} = \{y_4\}.$$

Set

$$C = \Big( (C_{11}, C_{12}, C_{13}),\ (C_{21}, C_{22}, C_{23}) \Big).$$

Then $C$ is a binary neutrosophic set on $(X, Y)$ in the sense of Definition 2.1.79. Moreover, $C$ is BNCS-Type2 (Definition 2.1.80), since on each side the three components are pairwise disjoint and their union covers the entire universe:

$$C_{11} \cap C_{12} = C_{11} \cap C_{13} = C_{12} \cap C_{13} = \emptyset, \qquad C_{11} \cup C_{12} \cup C_{13} = X,$$

$$C_{21} \cap C_{22} = C_{21} \cap C_{23} = C_{22} \cap C_{23} = \emptyset, \qquad C_{21} \cup C_{22} \cup C_{23} = Y.$$

### 2.1.15 Cohesive Fuzzy Set

Cohesive fuzzy sets (CHFS) were introduced as a framework that unifies (i) *complex-valued* membership representations (as in complex fuzzy sets) and (ii) *hesitation* (as in hesitant fuzzy sets) by allowing each element to have a *set* of admissible complex membership values [153–156].

**Definition 2.1.83** (Cohesive fuzzy set (CHFS))**.** [156] Let $S$ be a nonempty universe of discourse, and let

$$\mathbb{D} \ := \ \{z \in \mathbb{C} : |z| \leq 1\}$$

denote the closed unit disk in the complex plane. A *cohesive fuzzy set* on $S$ is a mapping

$$h : S \longrightarrow \mathcal{P}(\mathbb{D}) \setminus \{\emptyset\},$$

assigning to each $x \in S$ a nonempty set $h(x)$ of admissible complex membership values. Equivalently, each $z \in h(x)$ can be written in polar form as

$$z \ = \ r\, e^{i\theta}, \qquad r \in [0, 1],\ \theta \in \mathbb{R},$$

where $r$ is the *amplitude* (membership magnitude) and $\theta$ is the *phase*. We denote the resulting CHFS by

$$\mathcal{H} \ = \ \big\{ \langle x,\ h(x) \rangle : x \in S \big\}.$$

**Remark 2.1.84** (Two standard special cases)**.** [153, 156, 157] Let $\mathcal{H} = \{\langle x, h(x)\rangle : x \in S\}$ be a CHFS.

(i) If $|h(x)| = 1$ for every $x \in S$, then $\mathcal{H}$ reduces to a *complex fuzzy set* (each $x$ has a single complex membership value).

(ii) If $h(x) \subseteq [0, 1] \subseteq \mathbb{C}$ for every $x \in S$ (i.e., all phases are 0), then $\mathcal{H}$ reduces to a *hesitant fuzzy set* (each $x$ has a set of real membership candidates).

**Example 2.1.85** (A cohesive fuzzy set with phase uncertainty)**.** Let $S = \{s_1, s_2, s_3\}$ and let $\mathbb{D} = \{z \in \mathbb{C} : |z| \leq 1\}$. Define $h : S \to \mathcal{P}(\mathbb{D}) \setminus \{\emptyset\}$ by

$$h(s_1) = \Big\{ 0.8\, e^{i0},\ 0.8\, e^{i\pi/3} \Big\}, \qquad h(s_2) = \Big\{ 0.5\, e^{i\pi/2} \Big\}, \qquad h(s_3) = \Big\{ 0.3\, e^{-i\pi/4},\ 0.6\, e^{-i\pi/4} \Big\}.$$

Then $h(s_j) \subseteq \mathbb{D}$ and $h(s_j) \neq \emptyset$ for $j = 1, 2, 3$, because each listed value has modulus 0.8, 0.5, 0.3, or 0.6, all of which lie in $[0, 1]$. Hence

$$\mathcal{H} = \{\langle x, h(x)\rangle :\ x \in S\}$$

is a cohesive fuzzy set (CHFS) on $S$ in the sense of Definition 2.1.83.

*Interpretation.* For $s_1$, the amplitude is fixed at 0.8 but the phase can be either 0 or $\pi/3$, representing uncertain orientation on the unit disk. For $s_3$, the phase is fixed at $-\pi/4$ but the amplitude may be 0.3 or 0.6, representing uncertain strength with a consistent phase.

### 2.1.16 Ranked Soft Set

Ranked soft sets refine the soft-set viewpoint by replacing each parameter's approximation set with a *ranked partition* that encodes qualitative levels of satisfaction [158].

**Definition 2.1.86** (Ranked partition)**.** [158] Let $U$ be a nonempty set. A *ranked partition* of $U$ is a finite ordered tuple

$$\mathbf{V} = (V_0, V_1, \ldots, V_k) \qquad (k \in \mathbb{N})$$

such that:

(i) $V_i \subseteq U$ for all $i = 0, 1, \ldots, k$;

(ii) $V_i \cap V_j = \emptyset$ for all $i \neq j$;

(iii) $\bigcup_{i=0}^{k} V_i = U$.

The block $V_0$ is allowed to be empty and may be interpreted as the "non-satisfying" class, while $V_1, \ldots, V_k$ represent increasing ranks (higher index = higher rank). Let $\mathcal{R}(U)$ denote the set of all ranked partitions of $U$.

**Definition 2.1.87** (Ranked soft set)**.** [158] Let $U$ be a nonempty universe and let $E$ be a (nonempty) set of parameters. A *ranked soft set* over $U$ is a pair $(R, E)$ where

$$R : E \longrightarrow \mathcal{R}(U), \qquad e \longmapsto (V_0^{(e)}, V_1^{(e)}, \ldots, V_{k(e)}^{(e)}).$$

Thus, for each parameter $e \in E$, the universe $U$ is partitioned into ranked blocks $V_0^{(e)}, V_1^{(e)}, \ldots, V_{k(e)}^{(e)}$, representing progressively stronger satisfaction (or confidence) levels with respect to $e$.

**Remark 2.1.88.** If one takes $k(e) = 1$ for every $e \in E$, then $R(e) = (V_0^{(e)}, V_1^{(e)})$ is just a two-block partition of $U$. In this case, setting $F(e) := V_1^{(e)}$ recovers an ordinary soft set $(F, E)$ in the sense of Molodtsov/Maji [45].

**Example 2.1.89** (A ranked soft set for hotel selection)**.** Let

$$U = \{h_1, h_2, h_3, h_4\}$$

be a set of hotels and let

$$E = \{\mathsf{Clean},\ \mathsf{Cheap}\}$$

be the parameter set. Define $R : E \to \mathcal{R}(U)$ by the ranked partitions

$$R(\mathsf{Clean}) = (V_0^{(\mathsf{Clean})}, V_1^{(\mathsf{Clean})}, V_2^{(\mathsf{Clean})})$$

with

$$V_0^{(\mathsf{Clean})} = \{h_4\}, \qquad V_1^{(\mathsf{Clean})} = \{h_2, h_3\}, \qquad V_2^{(\mathsf{Clean})} = \{h_1\},$$

and

$$R(\mathsf{Cheap}) = (V_0^{(\mathsf{Cheap})}, V_1^{(\mathsf{Cheap})}, V_2^{(\mathsf{Cheap})})$$

with

$$V_0^{(\mathsf{Cheap})} = \{h_1\}, \qquad V_1^{(\mathsf{Cheap})} = \{h_3\}, \qquad V_2^{(\mathsf{Cheap})} = \{h_2, h_4\}.$$

Each triple is a ranked partition of $U$ (pairwise disjoint blocks whose union is $U$), so $(R, E)$ is a ranked soft set over $U$ in the sense of Definition 2.1.87. Here $V_0^{(e)}$ indicates "does not satisfy $e$," $V_1^{(e)}$ indicates "moderately satisfies $e$," and $V_2^{(e)}$ indicates "strongly satisfies $e$."

### 2.1.17 Bijective Soft Set

Bijective soft sets strengthen soft sets by requiring the parameterized approximations to form a *partition* of the universe, so that each object is assigned to exactly one parameter class [159–162].

**Definition 2.1.90** (Bijective function)**.** (cf. [163–166]) Let $A$ and $B$ be sets. A function $f : A \to B$ is *bijective* if it is both injective and surjective, i.e.,

$$\big(\forall x_1, x_2 \in A,\ f(x_1) = f(x_2) \Rightarrow x_1 = x_2\big) \quad \text{and} \quad \big(\forall y \in B,\ \exists x \in A \text{ with } f(x) = y\big).$$

In this case, $f$ establishes a one-to-one correspondence between $A$ and $B$.

**Definition 2.1.91** (Bijective soft set)**.** [159] Let $U$ be a nonempty universe and let $B$ be a nonempty set of parameters. A pair $(F, B)$ is called a *bijective soft set* over $U$ if

$$F : B \longrightarrow \mathcal{P}(U)$$

satisfies:

(i) **Covering (exhaustivity):** $\displaystyle\bigcup_{e \in B} F(e) = U$;

(ii) **Disjointness:** $F(e_1) \cap F(e_2) = \emptyset$ whenever $e_1 \neq e_2$ in $B$.

Equivalently, the family $\{F(e)\}_{e \in B}$ is a partition of $U$ indexed by $B$.

**Proposition 2.1.92** (Equivalent "classification" form)**.** *[159] A pair $(F, B)$ is a bijective soft set over U if and only if there exists a surjective map*

$$g : U \longrightarrow B$$

*such that*

$$F(e) = g^{-1}(e) \qquad (\forall e \in B).$$

*Proof.* ($\Rightarrow$) Assume $(F, B)$ is bijective. For each $u \in U$, the covering property gives at least one $e \in B$ with $u \in F(e)$. By disjointness, this $e$ is unique. Define $g(u) := e$. Then $g$ is surjective (since $F(e) \neq \emptyset$ for each $e$ appearing in the cover), and by construction $u \in F(e)$ iff $g(u) = e$, i.e., $F(e) = g^{-1}(e)$.

($\Leftarrow$) Conversely, assume a surjection $g : U \to B$ is given and define $F(e) := g^{-1}(e)$. Then $\bigcup_{e \in B} F(e) = U$ holds because every $u$ has some value $g(u) \in B$, and disjointness holds because preimages of distinct points under a function are disjoint. Hence $(F, B)$ is a bijective soft set. □

**Remark 2.1.93.** If, in addition, $|F(e)| = 1$ for every $e \in B$, then the map $g : U \to B$ in Proposition 2.1.92 is bijective (hence $|U| = |B|$), and $(F, B)$ becomes a literal one-to-one assignment between objects and parameters.

## 2.2 Uncertain Graph Theory

In this subsection, we provide the necessary definitions and terms essential for the discussion.

### 2.2.1 Fuzzy and Neutrosophic Graph

Since the introduction of fuzzy sets in the 1960s, many graph models have been proposed to encode uncertainty on vertices and/or edges [167]. This section fixes a crisp (classical) graph as the underlying skeleton and then records several widely used uncertainty enrichments, including fuzzy [89], intuitionistic fuzzy [168], neutrosophic [90], hesitant fuzzy, and (partitioned) neutrosophic variants. We also recall a plithogenic graph model based on attribute–value memberships together with contradiction information.

**Definition 2.2.1** (Crisp (simple) graph)**.** A *crisp graph* is a finite simple undirected graph $G = (V, E)$ where:

(i) $V \neq \emptyset$ is a finite set of vertices;

(ii) $E \subseteq \binom{V}{2} = \{\{u, v\} \subseteq V : u \neq v\}$ is the set of edges.

If $E = \emptyset$, then $G$ is an *edgeless graph*.

**Definition 2.2.2** (Unified framework for uncertain graphs)**.** Let $G^* = (V, E)$ be a crisp graph (Definition 2.2.1). An *uncertain graph structure on* $G^*$ consists of assigning to each vertex $v \in V$ and each edge $e \in E$ one or more membership values in a prescribed domain, together with model-specific consistency constraints relating edge degrees to their endpoints.

In the items below, we write $uv$ for the edge $\{u, v\} \in E$.

**Definition 2.2.3** (Fuzzy graph)**.** [89, 169] A *fuzzy graph* on a crisp graph $G^* = (V, E)$ is a triple

$$G = (V, \sigma, \mu),$$

where $\sigma : V \to [0, 1]$ is the *vertex-membership function* and $\mu : E \to [0, 1]$ is the *edge-membership function*, satisfying the admissibility condition

$$\mu(uv) \ \leq \ \min\{\sigma(u), \sigma(v)\} \qquad (\forall\, uv \in E).$$

**Definition 2.2.4** (Intuitionistic fuzzy graph)**.** [170, 171] An *intuitionistic fuzzy graph* on $G^* = (V, E)$ is a sextuple

$$G = (V, E, \mu_V, \nu_V, \mu_E, \nu_E),$$

where $\mu_V, \nu_V : V \to [0, 1]$ and $\mu_E, \nu_E : E \to [0, 1]$ satisfy, for all $v \in V$ and $uv \in E$,

$$0 \leq \mu_V(v) + \nu_V(v) \leq 1, \qquad 0 \leq \mu_E(uv) + \nu_E(uv) \leq 1,$$

together with the standard edge–vertex compatibility constraints

$$\mu_E(uv) \ \leq \ \min\{\mu_V(u), \mu_V(v)\}, \qquad \nu_E(uv) \ \leq \ \max\{\nu_V(u), \nu_V(v)\}.$$

**Definition 2.2.5** (Single-valued neutrosophic graph)**.** [172, 173] A *(single-valued) neutrosophic graph* on $G^* = (V, E)$ is a tuple

$$G = \big(V, E,\ T_V, I_V, F_V,\ T_E, I_E, F_E\big),$$

where $T_V, I_V, F_V : V \to [0, 1]$ and $T_E, I_E, F_E : E \to [0, 1]$ satisfy

$$0 \leq T_V(v) + I_V(v) + F_V(v) \leq 3 \qquad (\forall v \in V),$$

$$0 \leq T_E(uv) + I_E(uv) + F_E(uv) \leq 3 \qquad (\forall uv \in E),$$

and the edge–vertex consistency constraints (componentwise)

$$T_E(uv) \ \leq \ \min\{T_V(u), T_V(v)\}, \qquad I_E(uv) \ \leq \ \min\{I_V(u), I_V(v)\}, \qquad F_E(uv) \ \leq \ \max\{F_V(u), F_V(v)\}.$$

**Example 2.2.6** (A small single-valued neutrosophic graph)**.** Let $G^* = (V, E)$ be the crisp path on three vertices

$$V = \{u, v, w\}, \qquad E = \{uv, vw\}.$$

Define vertex memberships $T_V, I_V, F_V : V \to [0, 1]$ by

| $x$ | $T_V(x)$ | $I_V(x)$ | $F_V(x)$ |
|---|---|---|---|
| $u$ | 0.8 | 0.1 | 0.4 |
| $v$ | 0.6 | 0.2 | 0.5 |
| $w$ | 0.3 | 0.7 | 0.6 |

and edge memberships $T_E, I_E, F_E : E \to [0, 1]$ by

| $e$ | $T_E(e)$ | $I_E(e)$ | $F_E(e)$ |
|---|---|---|---|
| $uv$ | 0.6 | 0.1 | 0.5 |
| $vw$ | 0.3 | 0.2 | 0.6 |

Then for each $x \in V$ one has $0 \leq T_V(x) + I_V(x) + F_V(x) \leq 3$, and similarly for each $e \in E$. Moreover, the edge–vertex consistency constraints of Definition 2.2.5 hold:

$$T_E(uv) = 0.6 \leq \min\{0.8, 0.6\} = 0.6, \quad I_E(uv) = 0.1 \leq \min\{0.1, 0.2\} = 0.1, \quad F_E(uv) = 0.5 \leq \max\{0.4, 0.5\} = 0.5,$$

$$T_E(vw) = 0.3 \leq \min\{0.6, 0.3\} = 0.3, \quad I_E(vw) = 0.2 \leq \min\{0.2, 0.7\} = 0.2, \quad F_E(vw) = 0.6 \leq \max\{0.5, 0.6\} = 0.6.$$

Hence

$$G = \big(V, E,\ T_V, I_V, F_V,\ T_E, I_E, F_E\big)$$

is a (single-valued) neutrosophic graph on $G^*$.

We include a figure illustrating this example in Figure 2.2.

$T_E = 0.6$ $I_E = 0.1$ $F_E = 0.5$
$T_E = 0.3$ $I_E = 0.2$ $F_E = 0.6$
$u$ $v$ $w$
$T_V = 0.8$ $I_V = 0.1$ $F_V = 0.4$
$T_V = 0.6$ $I_V = 0.2$ $F_V = 0.5$
$T_V = 0.3$ $I_V = 0.7$ $F_V = 0.6$

Figure 2.2: A single-valued neutrosophic graph with vertex/edge triples $(T, I, F)$ (Example 2.2.6).

**Definition 2.2.7** (Hesitant fuzzy graph). [155, 174] A *hesitant fuzzy graph* on $G^* = (V, E)$ is a triple

$$G = (V, \Sigma, M),$$

where $\Sigma : V \longrightarrow \mathcal{P}^*([0,1])$ and $M : E \longrightarrow \mathcal{P}^*([0,1])$ assign to each vertex and edge a finite nonempty set of possible membership values (hesitation sets). A standard admissibility requirement is

$$\max M(uv) \ \leq\ \min\{\max \Sigma(u), \max \Sigma(v)\} \qquad (\forall\, uv \in E),$$

with analogous constraints possible under alternative aggregation rules (e.g., using inf or averaging operators) depending on the intended hesitant semantics.

**Definition 2.2.8** (Single-valued quadripartitioned neutrosophic graph). [175, 176] A *single-valued quadripartitioned neutrosophic graph* on $G^* = (V, E)$ is a tuple

$$G = \big(V, E,\ T_V, C_V, U_V, F_V,\ T_E, C_E, U_E, F_E\big),$$

where $T_V, C_V, U_V, F_V : V \to [0,1]$ and $T_E, C_E, U_E, F_E : E \to [0,1]$ satisfy

$$0 \leq T_V(v) + C_V(v) + U_V(v) + F_V(v) \leq 4 \qquad (\forall v \in V),$$

$$0 \leq T_E(uv) + C_E(uv) + U_E(uv) + F_E(uv) \leq 4 \qquad (\forall uv \in E),$$

and componentwise edge–vertex constraints

$$T_E(uv) \leq \min\{T_V(u), T_V(v)\}, \quad C_E(uv) \leq \min\{C_V(u), C_V(v)\},$$

$$U_E(uv) \leq \max\{U_V(u), U_V(v)\}, \quad F_E(uv) \leq \max\{F_V(u), F_V(v)\}.$$

Here $T$ denotes truth, $C$ contradiction, $U$ unknown/ignorance, and $F$ falsity.

**Definition 2.2.9** (Single-valued pentapartitioned neutrosophic graph). [177–179] A *single-valued pentapartitioned neutrosophic graph* on $G^* = (V, E)$ is a tuple

$$G = \big(V, E,\ T_V, C_V, R_V, U_V, F_V,\ T_E, C_E, R_E, U_E, F_E\big),$$

where $T_V, C_V, R_V, U_V, F_V : V \to [0,1]$ and $T_E, C_E, R_E, U_E, F_E : E \to [0,1]$ satisfy

$$0 \leq T_V(v) + C_V(v) + R_V(v) + U_V(v) + F_V(v) \leq 5 \qquad (\forall v \in V),$$

$$0 \leq T_E(uv) + C_E(uv) + R_E(uv) + U_E(uv) + F_E(uv) \leq 5 \qquad (\forall uv \in E),$$

and the componentwise edge–vertex constraints

$$T_E(uv) \leq \min\{T_V(u), T_V(v)\}, \qquad C_E(uv) \leq \min\{C_V(u), C_V(v)\},$$

$$R_E(uv) \leq \max\{R_V(u), R_V(v)\}, \qquad U_E(uv) \leq \max\{U_V(u), U_V(v)\}, \qquad F_E(uv) \leq \max\{F_V(u), F_V(v)\}.$$

The symbols typically represent: $T$ (truth), $C$ (contradiction), $R$ (neutrality/refusal, model-dependent), $U$ (unknown/ignorance), and $F$ (falsity).

**Definition 2.2.10** (Plithogenic graph)**.** [60, 62, 180] Let $G^* = (V, E)$ be a crisp graph. Fix two (possibly distinct) attributes: a vertex-attribute $v$ with value domain $P_v$, and an edge-attribute $w$ with value domain $P_w$. A *plithogenic graph* on $G^*$ is a tuple

$$\mathcal{PG} = \big(G^*;\ pdf_V, pCF_V,\ pdf_E, pCF_E\big),$$

where:

- $pdf_V : V \times P_v \to [0,1]^s$ is the *vertex degree of appurtenance* (DAF) and $pCF_V : P_v \times P_v \to [0,1]^t$ is the *vertex contradiction map* (DCF), with

$$pCF_V(a,a) = 0, \qquad pCF_V(a,b) = pCF_V(b,a) \qquad (\forall a, b \in P_v).$$

- $pdf_E : E \times P_w \to [0,1]^s$ is the *edge degree of appurtenance* (DAF) and $pCF_E : P_w \times P_w \to [0,1]^t$ is the *edge contradiction map* (DCF), with

$$pCF_E(\alpha,\alpha) = 0, \qquad pCF_E(\alpha,\beta) = pCF_E(\beta,\alpha) \qquad (\forall \alpha, \beta \in P_w).$$

A common edge–vertex compatibility requirement is that, for every edge $uv \in E$ and for chosen attribute-values $a \in P_v$, $b \in P_v$, the edge-appurtenance (possibly after applying a model-specific aggregation operator) is bounded by the endpoint appurtenances. In the simplest scalar case $s = 1$, this takes the form

$$pdf_E(uv \mid a, b) \ \le\ \min\{pdf_V(u \mid a),\ pdf_V(v \mid b)\},$$

where $pdf_E(uv \mid a, b)$ denotes an edge appurtenance value derived from $(uv, \cdot)$ and the endpoint attribute choices $(a, b)$ (the precise derivation is application dependent).

**Remark 2.2.11.** Plithogenic graph models admit multiple equivalent presentations in the literature, depending on (i) whether appurtenance is scalar or vector-valued ($s \ge 1$), (ii) whether contradiction is scalar or vector-valued ($t \ge 1$), and (iii) how edge appurtenance combines endpoint attribute-values. Definition 2.2.10 records the common structural ingredients: attribute-conditioned appurtenance together with contradiction maps that quantify opposition among attribute values.

**Example 2.2.12** (A small scalar plithogenic graph with symmetric contradiction maps)**.** Let $G^* = (V, E)$ be the crisp graph with

$$V = \{u, v, w\}, \qquad E = \{uv, vw\}.$$

Fix the scalar case $s = t = 1$. Choose a vertex-attribute $v$ ("status") with value domain

$$P_v = \{H, L\} \quad \text{(High, Low)},$$

and an edge-attribute $w$ ("link quality") with value domain

$$P_w = \{S, W\} \quad \text{(Strong, Weak)}.$$

**Vertex DAF.** Define $pdf_V : V \times P_v \to [0,1]$ by

$$pdf_V(u \mid H) = 0.9, \quad pdf_V(u \mid L) = 0.2,$$

$$pdf_V(v \mid H) = 0.6, \quad pdf_V(v \mid L) = 0.5,$$

$$pdf_V(w \mid H) = 0.3, \quad pdf_V(w \mid L) = 0.8.$$

**Vertex DCF.** Define $pCF_V : P_v \times P_v \to [0,1]$ by

$$pCF_V(H,H) = 0, \quad pCF_V(L,L) = 0, \quad pCF_V(H,L) = pCF_V(L,H) = 0.4.$$

**Edge DAF.** Define $pdf_E : E \times P_w \to [0,1]$ by

$$pdf_E(uv \mid S) = 0.55, \quad pdf_E(uv \mid W) = 0.20, \qquad pdf_E(vw \mid S) = 0.30, \quad pdf_E(vw \mid W) = 0.10.$$

**Edge DCF.** Define $pCF_E : P_w \times P_w \to [0,1]$ by

$$pCF_E(S,S) = 0, \quad pCF_E(W,W) = 0, \quad pCF_E(S,W) = pCF_E(W,S) = 0.7.$$

Then

$$\mathcal{PG} = \left(G^*;\ pdf_V, pCF_V,\ pdf_E, pCF_E\right)$$

is a plithogenic graph in the sense of Definition 2.2.10. Moreover, if we adopt the simplest endpoint-bounding rule that uses the same vertex-value $H$ at both endpoints, then for the edge $uv$ we have

$$pdf_E(uv \mid S) = 0.55 \ \leq\ \min\{pdf_V(u \mid H), pdf_V(v \mid H)\} = \min\{0.9, 0.6\} = 0.6,$$

and for $vw$,

$$pdf_E(vw \mid S) = 0.30 \ \leq\ \min\{pdf_V(v \mid H), pdf_V(w \mid H)\} = \min\{0.6, 0.3\} = 0.3,$$

so the basic edge–vertex compatibility holds for this choice.

**Example 2.2.13.** (cf. [91, 181]) The following examples are provided.

- When $s = t = 1$, $PG$ is called a *Plithogenic Fuzzy Graph*.
- When $s = 2, t = 1$, $PG$ is called a *Plithogenic Intuitionistic Fuzzy Graph*.
- When $s = 3, t = 1$, $PG$ is called a *Plithogenic Neutrosophic Graph*.
- When $s = 4, t = 1$, $PG$ is called a *Plithogenic quadripartitioned Neutrosophic Graph*.
- When $s = 5, t = 1$, $PG$ is called a *Plithogenic pentapartitioned Neutrosophic Graph*.

The plithogenic graph encompasses various graph types that have been actively studied. This graph concept is capable of handling multiple layers of uncertainty while generalizing numerous existing graph concepts. Its flexibility allows for selecting different graph types depending on the research objectives or practical applications, making it mathematically significant and versatile. Furthermore, it is anticipated that practical applications and further studies on its utility will be explored in the future.

**Definition 2.2.14** (Plithogenic Graph Type)**.** (cf. [91, 181]) A plithogenic graph $PG$ is a graph that satisfies one of the following types of plithogenic characteristics (referred to as $PG$ of the $i$-th type) or any combination thereof:

(i) $PG_1 = \{G_1, G_2, G_3, \ldots, G_P\}$ where plithogenic characteristics exist in each graph $G_i$, incorporating different attributes and degrees of appurtenance and contradiction for vertices and edges.

(ii) $PG_2 = \{V, E_P\}$ where the edge set $E_P$ is plithogenic, meaning that each edge is associated with a range of possible attributes and corresponding degrees of appurtenance and contradiction.

(iii) $PG_3 = \{V, E(t_P, h_P)\}$ where both the vertex set $V$ and edge set $E$ are crisp, but the edges have plithogenic heads $h(e_i)$ and plithogenic tails $t(e_i)$ with respect to certain attributes.

(iv) $PG_4 = \{V_P, E\}$ where the vertex set $V_P$ is plithogenic, meaning each vertex has attributes with varying degrees of appurtenance and contradiction.

(v) $PG_5 = \{V, E(w_P)\}$ where both the vertex set $V$ and edge set $E$ are crisp, but the edges have plithogenic weights $w_P$, indicating the attribute-based degrees of appurtenance and contradiction.

The General Plithogenic Graph is a generalization of the Plithogenic Graph (cf. [91, 182, 183]). The General Plithogenic Graph relaxes certain conditions, such as the Edge Appurtenance Constraint.

By incorporating constraints from Pythagorean fuzzy sets [184, 185], spherical fuzzy sets [186–193], (m, n)-fuzzy sets [194, 195], and q-rung orthopair fuzzy sets [196–203], we hope to explore new mathematical characteristics and applications, such as in decision-making and other domains.

**Definition 2.2.15** (General Plithogenic Graph)**.** [91] Let $G = (V, E)$ be a classical graph, where $V$ is a finite set of vertices, and $E \subseteq V \times V$ is a set of edges.

A General Plithogenic Graph $G^{GP} = (PM, PN)$ consists of:

1. *General Plithogenic Vertex Set* $PM$:

$$PM = (M, l, Ml, adf, aCf)$$

where:

- $M \subseteq V$: Set of vertices.
- $l$: Attribute associated with the vertices.
- $Ml$: Range of possible attribute values.
- $adf : M \times Ml \to [0, 1]^s$: Degree of Appurtenance Function (DAF) for vertices.
- $aCf : Ml \times Ml \to [0, 1]^t$: Degree of Contradiction Function (DCF) for vertices.

2. *General Plithogenic Edge Set* $PN$:

$$PN = (N, m, Nm, bdf, bCf)$$

where:

- $N \subseteq E$: Set of edges.
- $m$: Attribute associated with the edges.
- $Nm$: Range of possible attribute values.
- $bdf : N \times Nm \to [0, 1]^s$: Degree of Appurtenance Function (DAF) for edges.
- $bCf : Nm \times Nm \to [0, 1]^t$: Degree of Contradiction Function (DCF) for edges.

The General Plithogenic Graph $G^{GP}$ only needs to satisfy the following *Reflexivity and Symmetry* properties of the Contradiction Functions:

- Reflexivity and Symmetry of Contradiction Functions:

$$\begin{aligned} aCf(a, a) &= 0, & \forall a \in Ml \\ aCf(a, b) &= aCf(b, a), & \forall a, b \in Ml \\ bCf(a, a) &= 0, & \forall a \in Nm \\ bCf(a, b) &= bCf(b, a), & \forall a, b \in Nm \end{aligned}$$

In graphs dealing with uncertainty, the following has been established [91].

**Theorem 2.2.16.** *[91] In each graph class, the following relationships hold.*

- *An empty graph and a null graph can be represented as 2-valued graphs and 3-valued graphs.*
- *Every edge-fuzzy graph can be transformed into a 2-valued graph by thresholding the edge membership values.*
- *Every fuzzy graph can be transformed into a 3-valued graph by mapping the fuzzy membership values of vertices and edges to the values {-1, 0, 1}.*
- *Every Intuitionistic Fuzzy Graph can be transformed into a Fuzzy Graph by restricting the non-membership function* $\nu_A$ *to 0 for all vertices.*
- *Every Neutrosophic Graph can be transformed into an Intuitionistic Fuzzy Graph by setting the indeterminacy value to zero.*

- *Every Extended Turiyam Neutrosophic Graph is a generalization of the Turiyam Neutrosophic Graph.*
- *A plithogenic Graphs generalize Fuzzy Graphs, Intuitionistic Fuzzy Graphs, Neutrosophic Graphs, Turiyam Neutrosophic Graphs, Extended Turiyam Neutrosophic Graphs.*
- *Every general plithogenic Graphs can be transformed into General Turiyam Neutrosophic Graph, General Fuzzy Graph, General Intuitionistic Fuzzy Graph, Four-Valued Fuzzy graph, Ambiguous graph, Picture Fuzzy Graph, Hesitant Fuzzy Graph, Intuitionistic Hesitant Fuzzy Graph, Fuzzy Graphs, Intuitionistic Fuzzy Graphs, Neutrosophic Graphs, Quadripartitioned Neutrosophic graph, Pentapartitioned Neutrosophic graph, Turiyam Neutrosophic Graphs, and Spherical Fuzzy Graphs.*

### 2.2.2 Soft Sets, Soft Graphs, and Multisoft Graphs

Soft set theory provides a parameterized way to represent uncertain or incomplete information. Given a universe $U$ and a parameter set $E$, a soft set selects, for each active parameter $a$, a subset $F(a) \subseteq U$. This flexibility makes soft sets a convenient interface between set-based data and graph-based models [45,46,204–206]. For further details on soft-set operations, see [46].

**Definition 2.2.17** (Soft set)**.** [46] Let $U \neq \emptyset$ be a universe and let $E \neq \emptyset$ be a parameter set. A *soft set* over $U$ is an ordered pair $(F, A)$ where $A \subseteq E$ and

$$F : A \longrightarrow \mathcal{P}(U).$$

Thus, each parameter $a \in A$ is assigned a subset $F(a) \subseteq U$.

**Definition 2.2.18** (Soft subset, union, and intersection)**.** Let $(F, A)$ and $(G, B)$ be soft sets over the same universe $U$.

(i) **Soft subset.** We say that $(F, A)$ is a *soft subset* of $(G, B)$, written $(F, A)\widetilde{\subseteq}(G, B)$, if $A \subseteq B$ and $F(a) \subseteq G(a)$ for all $a \in A$.

(ii) **Union.** The *union* $(F, A)\widetilde{\cup}(G, B)$ is the soft set $(H, A \cup B)$ where, for $e \in A \cup B$,

$$H(e) = \begin{cases} F(e), & e \in A \setminus B, \\ G(e), & e \in B \setminus A, \\ F(e) \cup G(e), & e \in A \cap B. \end{cases}$$

(iii) **Intersection.** The *intersection* $(F, A)\widetilde{\cap}(G, B)$ is the soft set $(H, A \cap B)$ where, for $e \in A \cap B$,

$$H(e) = F(e) \cap G(e).$$

(If $A \cap B = \emptyset$, then the intersection has empty parameter set; some authors exclude this by convention.)

Soft graphs arise by "graphizing" a soft set: for each parameter $a$, the soft images select a subgraph. Soft graph theory is often studied alongside fuzzy and neutrosophic graph models [207–210].

**Definition 2.2.19** (Soft graph)**.** [207,211] Let $G^* = (V, E)$ be a finite simple undirected graph, where $V \neq \emptyset$ and $E \subseteq \binom{V}{2}$. Let $A \neq \emptyset$ be a parameter set. Let $(F, A)$ be a soft set over $V$ and $(K, A)$ a soft set over $E$, i.e.,

$$F : A \to \mathcal{P}(V), \qquad K : A \to \mathcal{P}(E).$$

A *soft graph* over $G^*$ is a quadruple

$$\mathcal{G} = (G^*, F, K, A)$$

such that for every $a \in A$,

$$K(a) \subseteq E \cap \binom{F(a)}{2}.$$

Equivalently, for each $a \in A$, the pair $H(a) := (F(a), K(a))$ is a (not necessarily induced) subgraph of $G^*$. We may therefore view $\mathcal{G}$ as the $A$-indexed family $\{H(a)\}_{a \in A}$.

**Remark 2.2.20** (Induced-subgraph convention)**.** Some authors additionally require $K(a) = E \cap \binom{F(a)}{2}$, i.e., that $H(a)$ is the induced subgraph on $F(a)$. The weaker condition in Definition 2.2.19 is more flexible and is sufficient for most general developments.

**Example 2.2.21** (A soft graph on a 4-vertex cycle)**.** Let the underlying crisp graph be the 4-cycle

$$G^* = (V, E), \qquad V = \{v_1, v_2, v_3, v_4\}, \qquad E = \{v_1v_2,\ v_2v_3,\ v_3v_4,\ v_4v_1\}.$$

Let the parameter set be

$$A = \{\alpha, \beta\}.$$

Define a soft set $(F, A)$ over $V$ and a soft set $(K, A)$ over $E$ by

$$F(\alpha) = \{v_1, v_2, v_3\}, \qquad F(\beta) = \{v_1, v_3, v_4\},$$

$$K(\alpha) = \{v_1v_2,\ v_2v_3\}, \qquad K(\beta) = \{v_3v_4,\ v_4v_1\}.$$

Then for each parameter $a \in A$,

$$K(a) \subseteq E \cap \binom{F(a)}{2},$$

since $K(\alpha)$ consists of edges among $\{v_1, v_2, v_3\}$ and $K(\beta)$ consists of edges among $\{v_1, v_3, v_4\}$. Hence

$$\mathcal{G} = (G^*, F, K, A)$$

is a soft graph over $G^*$ in the sense of Definition 2.2.19. The associated subgraphs are $H(\alpha) = (F(\alpha), K(\alpha))$ (a path $v_1 - v_2 - v_3$) and $H(\beta) = (F(\beta), K(\beta))$ (a path $v_1 - v_4 - v_3$).

We include a schematic overview of this example in Figure 2.3.

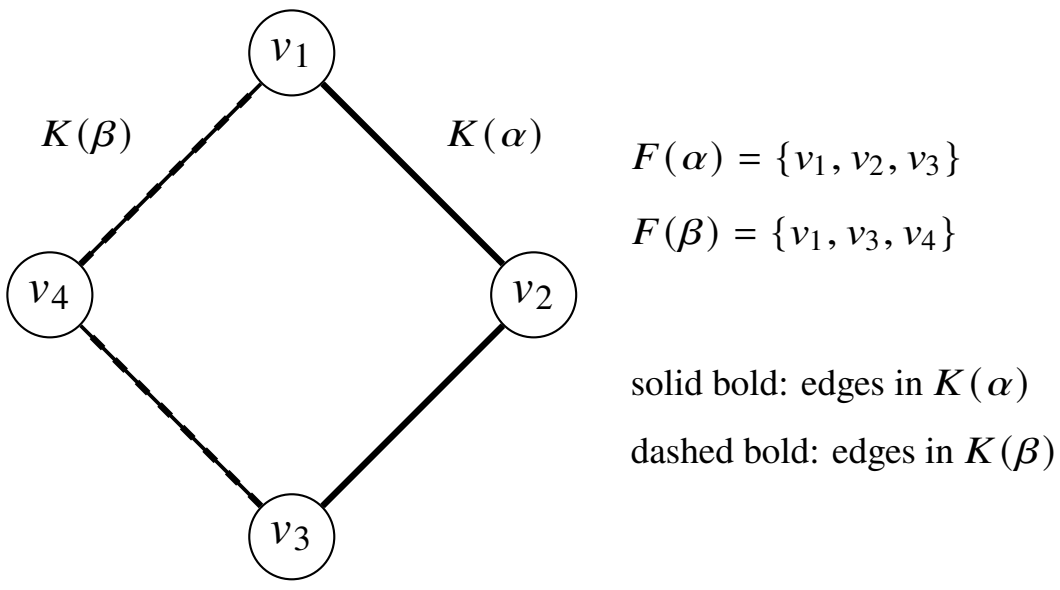

Figure 2.3: A soft graph $\mathcal{G} = (G^*, F, K, A)$ on the cycle $C_4$ (Example 2.2.21).

In recent years, multisoft sets have been introduced as a multi-parameter extension of classical soft sets. We recall the definition below and then present the natural graphization (multisoft graphs). Related notions include IndetermSoft sets, IndetermHyperSoft sets, and TreeSoft sets [212–216].

**Definition 2.2.22** (Multisoft set)**.** [217–219] Let $U \neq \emptyset$ be a universe. Let $E_1, \ldots, E_n$ be nonempty pairwise disjoint parameter sets, and put

$$E := \bigsqcup_{i=1}^{n} E_i$$

(the disjoint union, used to avoid collisions of parameter symbols). Let $A \subseteq \mathcal{P}(E) \setminus \{\emptyset\}$ be a nonempty family of *attribute combinations*. A *multisoft set* over $U$ is a pair $(F, A)$ where

$$F : A \longrightarrow \mathcal{P}(U).$$

For $a \in A$, the subset $F(a) \subseteq U$ is called the *a-approximate set* of $(F, A)$.

**Remark 2.2.23.** If $n = 1$ and $A \subseteq E_1$ (identified with singletons in $\mathcal{P}(E_1)$), then a multisoft set reduces to a classical soft set $(F, A)$ with $F : A \to \mathcal{P}(U)$.

**Definition 2.2.24** (Multisoft graph)**.** Let $G^* = (V, E)$ be a finite simple undirected graph, where $V \neq \emptyset$ and $E \subseteq \binom{V}{2}$. Let $A \subseteq \mathcal{P}(P) \setminus \{\emptyset\}$ be a nonempty family of attribute combinations over a parameter universe $P$. A *multisoft graph* over $G^*$ is a quadruple

$$\mathcal{G}_{MS} = (G^*, F, K, A),$$

where $(F, A)$ is a multisoft set over $V$ and $(K, A)$ is a multisoft set over $E$, i.e.,

$$F : A \to \mathcal{P}(V), \qquad K : A \to \mathcal{P}(E),$$

such that for every $a \in A$,

$$K(a) \subseteq E \cap \binom{F(a)}{2}.$$

Equivalently, for each $a \in A$, the pair $H(a) := (F(a), K(a))$ is a (not necessarily induced) subgraph of $G^*$, and $\mathcal{G}_{MS}$ may be viewed as the $A$-indexed family $\{H(a)\}_{a \in A}$. We denote the class of all multisoft graphs over $G^*$ by $\mathcal{MS}(G^*)$.

**Remark 2.2.25.** The symbol $E$ is commonly used for the edge set of $G^*$, so we use a separate symbol $P$ for the parameter universe.

**Proposition 2.2.26** (Multisoft graph $\Rightarrow$ soft graph by reindexing)**.** *Every multisoft graph $\mathcal{G}_{MS} = (G^*, F, K, A)$ canonically induces a soft graph on the same underlying graph by taking the parameter set to be $A$ itself.*

*Proof.* Define $A' := A$, $F' := F$, and $K' := K$. Then $(F', A')$ is a soft set over $V$ and $(K', A')$ is a soft set over $E$. Moreover, for each $a \in A' = A$,

$$K'(a) = K(a) \subseteq E \cap \binom{F(a)}{2} = E \cap \binom{F'(a)}{2}.$$

Hence $(G^*, F', K', A')$ is a soft graph in the sense of Definition 2.2.19. □

**Remark 2.2.27** (On "equivalence")**.** Proposition 2.2.26 is a reindexing: it does not collapse combinations to single parameters, but simply treats each combination $a \in A$ as a parameter in its own right. A genuine compression $A \to E$ (if desired) requires an explicit aggregation rule and is not canonical.

Soft graphs are frequently studied alongside uncertainty-aware graph models such as fuzzy graphs and neutrosophic graphs. In particular, fuzzy soft graphs [93,208–210,220,221] and neutrosophic soft graphs [222–225] extend the soft-graph framework by attaching graded uncertainty information to vertices and/or edges. Since this book focuses on neutrosophic soft graphs, we record a clean definition below.

**Definition 2.2.28** (Neutrosophic soft graph)**.** [222–225] Let $G^* = (V, E)$ be a finite simple undirected graph, where $V \neq \emptyset$ and $E \subseteq \binom{V}{2}$. Let $A \neq \emptyset$ be a parameter set.

A *neutrosophic set* on a carrier $S$ is a triple of maps $T, I, F : S \to [0, 1]$. Let $\rho(S)$ denote the class of all neutrosophic sets on $S$.

A *neutrosophic soft graph* over $G^*$ is a quadruple

$$\mathcal{G}_N = (G^*, J, K, A),$$

where

$$J : A \longrightarrow \rho(V), \qquad K : A \longrightarrow \rho(E)$$

are neutrosophic soft sets on $V$ and $E$, respectively. Writing

$$J(a) = (T_{V,a}, I_{V,a}, F_{V,a}) \in \rho(V), \qquad K(a) = (T_{E,a}, I_{E,a}, F_{E,a}) \in \rho(E),$$

we additionally require that for each $a \in A$ and each edge $uv \in E$, the edge–vertex consistency constraints hold:

$$\begin{aligned} T_{E,a}(uv) &\leq \min\{T_{V,a}(u), T_{V,a}(v)\}, \\ I_{E,a}(uv) &\leq \min\{I_{V,a}(u), I_{V,a}(v)\}, \\ F_{E,a}(uv) &\leq \max\{F_{V,a}(u), F_{V,a}(v)\}. \end{aligned}$$

**Proposition 2.2.29** (Neutrosophic soft graphs generalize soft graphs)**.** *Every soft graph is a special case of a neutrosophic soft graph.*

*Proof.* Let $\mathcal{G} = (G^*, F, K, A)$ be a soft graph. Define, for each $a \in A$, neutrosophic sets on $V$ and $E$ by characteristic functions:

$$T_{V,a}(v) := \mathbf{1}_{F(a)}(v), \quad I_{V,a}(v) := 0, \quad F_{V,a}(v) := 0,$$
$$T_{E,a}(e) := \mathbf{1}_{K(a)}(e), \quad I_{E,a}(e) := 0, \quad F_{E,a}(e) := 0.$$

Then $J(a) = (T_{V,a}, I_{V,a}, F_{V,a}) \in \rho(V)$ and $K(a) = (T_{E,a}, I_{E,a}, F_{E,a}) \in \rho(E)$. Moreover, if $uv \notin K(a)$, then $T_{E,a}(uv) = 0$ and the inequalities are trivial; if $uv \in K(a)$, then $u, v \in F(a)$ by the soft-graph condition, hence $T_{V,a}(u) = T_{V,a}(v) = 1$ and $T_{E,a}(uv) = 1 \leq \min\{1, 1\}$, while the $I$ and $F$ inequalities hold since all those values are 0. Thus $(G^*, J, K, A)$ is a neutrosophic soft graph and contains $\mathcal{G}$ as a crisp special case. □

### 2.2.3 Neutrosophic OverGraphs, UnderGraphs, and OffGraphs

Single-valued neutrosophic over/under/off graphs are obtained by enlarging the codomain of neutrosophic memberships beyond the classical interval $[0, 1]$, in the spirit of oversets/undersets/offsets [226–229]. The following definitions are stated in a uniform, type-consistent way.

**Definition 2.2.30** (Single-valued neutrosophic OverGraph)**.** (cf. [226–228]) Fix $\Omega > 1$. A *single-valued neutrosophic OverGraph* is a graph $G = (V, E)$ equipped with maps

$$T,\ I,\ F :\ V \cup E \longrightarrow [0, \Omega],$$

so that each vertex/edge $x \in V \cup E$ carries degrees $T(x), I(x), F(x) \in [0, \Omega]$. Optionally, one may impose the pointwise bound

$$T(x) + I(x) + F(x) \leq 3\Omega \qquad (\forall x \in V \cup E),$$

which is automatic when $T, I, F \in [0, \Omega]$.

**Definition 2.2.31** (Single-valued neutrosophic UnderGraph)**.** Fix $\Psi < 0$. A *single-valued neutrosophic UnderGraph* is a graph $G = (V, E)$ equipped with maps

$$T,\ I,\ F :\ V \cup E \longrightarrow [\Psi, 1].$$

Optionally, one may impose $T(x) + I(x) + F(x) \leq 3$ for all $x \in V \cup E$.

**Definition 2.2.32** (Single-valued neutrosophic OffGraph)**.** (cf. [229]) Fix $\Psi < 0 < 1 < \Omega$. A *single-valued neutrosophic OffGraph* is a graph $G = (V, E)$ equipped with maps

$$T,\ I,\ F :\ V \cup E \longrightarrow [\Psi, \Omega].$$

Optionally, one may impose $T(x) + I(x) + F(x) \leq 3\Omega$ for all $x \in V \cup E$.

**Example 2.2.33** (A small single-valued neutrosophic OffGraph)**.** Fix $\Psi = -0.2$ and $\Omega = 1.3$ (so $\Psi < 0 < 1 < \Omega$). Let $G = (V, E)$ be the path on three vertices

$$V = \{u, v, w\}, \qquad E = \{uv, vw\}.$$

Define maps $T, I, F : V \cup E \to [\Psi, \Omega]$ by

| $x$ | $T(x)$ | $I(x)$ | $F(x)$ |
|---|---|---|---|
| $u$ | 1.10 | 0.20 | −0.10 |
| $v$ | 0.60 | −0.05 | 0.40 |
| $w$ | 0.30 | 0.10 | 0.80 |
| $uv$ | 0.60 | 0.10 | 0.50 |
| $vw$ | 0.30 | 0.05 | 0.90 |

All values lie in $[\Psi, \Omega] = [-0.2, 1.3]$. Moreover, the optional sum bound holds: for instance,

$$T(u) + I(u) + F(u) = 1.10 + 0.20 - 0.10 = 1.20 \leq 3\Omega = 3.9,$$

and similarly for the remaining vertices and edges. Hence $(G, T, I, F)$ is a single-valued neutrosophic OffGraph in the sense of Definition 2.2.32. Note that the "off" behavior is visible, e.g. $T(u) = 1.10 > 1$ and $F(u) = -0.10 < 0$. We also include a figure illustrating this example in Figure 2.4.

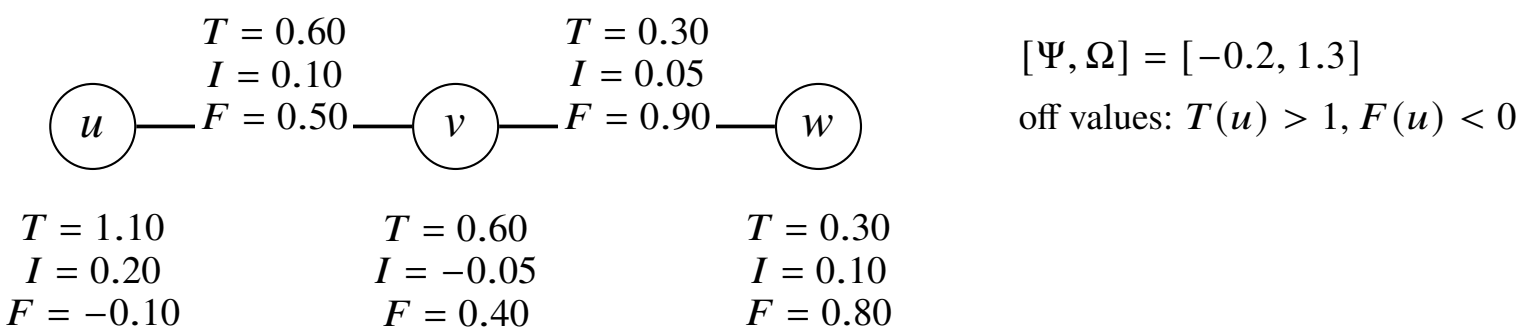

Figure 2.4: A single-valued neutrosophic OffGraph with degrees in $[\Psi, \Omega]$ (Example 2.2.33).

**Proposition 2.2.34** (Parameter specializations).

(i) *If $\Omega = 1$ in Definition 2.2.30, then the codomain becomes $[0, 1]$ and the structure reduces to a standard single-valued neutrosophic graph.*

(ii) *If $\Psi = 0$ in Definition 2.2.31, then the codomain becomes $[0, 1]$ and the structure reduces to a standard single-valued neutrosophic graph.*

(iii) *If $\Omega = 1$ and $\Psi = 0$ in Definition 2.2.32, then the codomain becomes $[0, 1]$ and the structure reduces to a standard single-valued neutrosophic graph.*

*Proof.* Each claim follows immediately by substituting the parameter values and observing that the codomain intervals collapse to $[0, 1]$. □

**Proposition 2.2.35** (OverGraphs allow sums exceeding 3). *In a single-valued neutrosophic OverGraph with $\Omega > 1$, the quantity $T(x) + I(x) + F(x)$ can exceed 3.*

*Proof.* Choose any $x \in V \cup E$ and set $T(x) = I(x) = F(x) = \Omega$. Then $T(x) + I(x) + F(x) = 3\Omega > 3$. □

**Definition 2.2.36** (Fuzzy Overgraph). A *Fuzzy Overgraph* $G = (V, E, \mu_V, \mu_E)$ consists of:

- A set of vertices $V$.
- A set of edges $E \subseteq V \times V$.
- A vertex membership function $\mu_V : V \longrightarrow [0, \Omega]$, with $\Omega > 1$ and $\exists v \in V$ such that $\mu_V(v) > 1$.
- An edge membership function $\mu_E : E \longrightarrow [0, \Omega]$, where $\exists e \in E$ such that $\mu_E(e) > 1$.

**Definition 2.2.37** (Fuzzy Undergraph). A *Fuzzy Undergraph* $G = (V, E, \mu_V, \mu_E)$ consists of:

- A set of vertices $V$.
- A set of edges $E \subseteq V \times V$.
- A vertex membership function $\mu_V : V \longrightarrow [\Psi, 1]$, with $\Psi < 0$ and $\exists v \in V$ such that $\mu_V(v) < 0$.
- An edge membership function $\mu_E : E \longrightarrow [\Psi, 1]$, where $\exists e \in E$ such that $\mu_E(e) < 0$.

**Definition 2.2.38** (Fuzzy Offgraph). A *Fuzzy Offgraph* $G = (V, E, \mu_V, \mu_E)$ consists of:

- A set of vertices $V$.
- A set of edges $E \subseteq V \times V$.
- A vertex membership function $\mu_V : V \longrightarrow [\Psi, \Omega]$, with $\Omega > 1$, $\Psi < 0$, and $\exists v \in V$ such that $\mu_V(v) > 1$ or $\mu_V(v) < 0$.
- An edge membership function $\mu_E : E \longrightarrow [\Psi, \Omega]$, where $\exists e \in E$ such that $\mu_E(e) > 1$ or $\mu_E(e) < 0$.

**Definition 2.2.39** (Intuitionistic Fuzzy Overgraph). An *Intuitionistic Fuzzy Overgraph* $G = (V, E, \mu_V, \nu_V, \mu_E, \nu_E)$ consists of:

- A set of vertices $V$.

- A set of edges $E \subseteq V \times V$.
- Vertex membership $\mu_V : V \to [0, \Omega]$ and non-membership $\nu_V : V \to [0, \Omega]$ functions.
- Edge membership $\mu_E : E \to [0, \Omega]$ and non-membership $\nu_E : E \to [0, \Omega]$ functions.

where $\Omega > 1$, and at least one $\mu_V(v) > 1$, $\nu_V(v) > 1$, $\mu_E(e) > 1$, or $\nu_E(e) > 1$.

**Definition 2.2.40** (Intuitionistic Fuzzy Undergraph)**.** An *Intuitionistic Fuzzy Undergraph* $G = (V, E, \mu_V, \nu_V, \mu_E, \nu_E)$ consists of:

- A set of vertices $V$.
- A set of edges $E \subseteq V \times V$.
- Vertex membership $\mu_V : V \to [\Psi, 1]$ and non-membership $\nu_V : V \to [\Psi, 1]$ functions.
- Edge membership $\mu_E : E \to [\Psi, 1]$ and non-membership $\nu_E : E \to [\Psi, 1]$ functions.

where $\Psi < 0$, and at least one $\mu_V(v) < 0$, $\nu_V(v) < 0$, $\mu_E(e) < 0$, or $\nu_E(e) < 0$.

**Definition 2.2.41** (Intuitionistic Fuzzy Offgraph)**.** An *Intuitionistic Fuzzy Offgraph* $G = (V, E, \mu_V, \nu_V, \mu_E, \nu_E)$ consists of:

- A set of vertices $V$.
- A set of edges $E \subseteq V \times V$.
- Vertex membership $\mu_V : V \to [\Psi, \Omega]$ and non-membership $\nu_V : V \to [\Psi, \Omega]$ functions.
- Edge membership $\mu_E : E \to [\Psi, \Omega]$ and non-membership $\nu_E : E \to [\Psi, \Omega]$ functions.

where $\Omega > 1$ and $\Psi < 0$, allowing degrees that can exceed 1 or be less than 0.

**Proposition 2.2.42.** *The Neutrosophic Undergraph generalizes both the Fuzzy Undergraph and the Intuitionistic Fuzzy Undergraph.*

*Proof.* The proof can be constructed in a manner analogous to the set-based case. □

**Proposition 2.2.43.** *The Neutrosophic Overgraph generalizes both the Fuzzy Overgraph and the Intuitionistic Fuzzy Overgraph.*

*Proof.* The proof can be constructed in a manner analogous to the set-based case. □

**Proposition 2.2.44.** *The Neutrosophic Offgraph generalizes both the Fuzzy Offgraph and the Intuitionistic Fuzzy Offgraph.*

*Proof.* The proof can be constructed in a manner analogous to the set-based case. □

**Proposition 2.2.45.** *A Single-Valued Neutrosophic OffGraph can be transformed into a Single-Valued Neutrosophic OverGraph or a Single-Valued Neutrosophic UnderGraph.*

*Proof.* The proof can be constructed in a manner analogous to the set-based case. □

### 2.2.4 Neutrosophic Soft offgraph/overgraph/undergraph

The Neutrosophic Soft Offgraph/Overgraph/Undergraph is an extension of the Neutrosophic Offgraph/Overgraph/Undergraph. The definitions are provided below. Note that the Neutrosophic Soft Overgraph was originally defined in [230].

**Definition 2.2.46** (Neutrosophic Soft Over Graph (NSOG))**.** [230] Let $G = (V, E)$ be a classical graph where $V$ represents the set of vertices and $E \subseteq V \times V$ represents the set of edges. A *Neutrosophic Soft Over Graph (NSOG)*, denoted $G_{NSOG} = (F_V, F_E, T, I, F)$, is a graph where each vertex and edge is assigned neutrosophic soft oversets. These oversets capture neutrosophic information with the possibility of exceeding the traditional boundaries (i.e., values greater than 1), allowing a flexible representation of uncertain, indeterminate, and contradictory information.

The structure of $G_{NSOG}$ is defined as follows:

1. $F_V$ and $F_E$ are neutrosophic soft oversets associated with the vertices $V$ and edges $E$, respectively:

   $$F_V(v) = \{(v, T(v), I(v), F(v)) \mid v \in V\}, \quad F_E(e) = \{(e, T(e), I(e), F(e)) \mid e \in E\},$$

   where:

   - $T(v)$, $I(v)$, and $F(v)$ represent the truth-membership, indeterminacy-membership, and falsity-membership degrees for each vertex $v \in V$.
   - $T(e)$, $I(e)$, and $F(e)$ represent the corresponding degrees for each edge $e \in E$.
   - Each degree $T(x)$, $I(x)$, and $F(x)$ (for $x = v$ or $e$) can take values in the extended interval $[0, \Omega]$, where $\Omega > 1$, allowing degrees to exceed the conventional bound of 1.

2. For all $v \in V$ and $e \in E$, the sum of the degrees satisfies:

   $$T(v) + I(v) + F(v) \leq 3\Omega, \quad T(e) + I(e) + F(e) \leq 3\Omega,$$

   maintaining an upper bound on the combined neutrosophic components for each vertex and edge.

3. The NSOG is termed *pure* if each vertex $v \in V$ and each edge $e \in E$ has at least one degree that exceeds the standard maximum of 1:

   $$T(v) > 1 \quad \text{or} \quad I(v) > 1 \quad \text{or} \quad F(v) > 1,$$

   and similarly,

   $$T(e) > 1 \quad \text{or} \quad I(e) > 1 \quad \text{or} \quad F(e) > 1.$$

In this structure, the Neutrosophic Soft Over Graph $G_{NSOG}$ extends the concept of a classical graph by incorporating neutrosophic soft oversets for both vertices and edges, where at least one of the truth, indeterminacy, or falsity values may surpass 1, thus enhancing the graph's flexibility in handling various levels of uncertainty, indeterminacy, and opposition.

**Definition 2.2.47** (Neutrosophic Soft UnderGraph)**.** A *Neutrosophic Soft UnderGraph* is a graph $G = (G^*, J, K, A)$, where:

- $G^* = (V, E)$ is a *neutrosophic graph* with the following membership functions for each vertex $x \in V$:

  $$T : V \to [\Psi, 1], \quad I : V \to [\Psi, 1], \quad F : V \to [\Psi, 1],$$

  where $\Psi < 0$ allows falsity degrees to be less than 0.

- $A$ is a non-empty set of parameters.

- $(J, A)$ is a neutrosophic soft set over the vertex set $V$, where $J : A \to \rho(V)$, and $\rho(V)$ denotes the set of all neutrosophic sets of $V$.

- $(K, A)$ is a neutrosophic soft set over the edge set $E$, where $K : A \to \rho(E)$.

The sum of degrees for any vertex $x \in V$ satisfies:

$$T(x) + I(x) + F(x) \leq 3.$$

**Definition 2.2.48** (Neutrosophic Soft OffGraph)**.** A *Neutrosophic Soft OffGraph* is a graph $G = (G^*, J, K, A)$, where:

- $G^* = (V, E)$ is a *neutrosophic graph* with the following membership functions for each vertex $x \in V$:

$$T : V \to [\Psi, \Omega], \quad I : V \to [\Psi, \Omega], \quad F : V \to [\Psi, \Omega],$$

  where $\Omega > 1$ allows truth degrees to exceed 1, and $\Psi < 0$ allows falsity degrees to be less than 0.
- $A$ is a non-empty set of parameters.
- $(J, A)$ is a neutrosophic soft set over the vertex set $V$, where $J : A \to \rho(V)$, and $\rho(V)$ denotes the set of all neutrosophic sets of $V$.
- $(K, A)$ is a neutrosophic soft set over the edge set $E$, where $K : A \to \rho(E)$.

The sum of degrees for any vertex $x \in V$ satisfies:

$$T(x) + I(x) + F(x) \leq 3\Omega.$$

**Proposition 2.2.49.** *A Neutrosophic Soft OffGraph can be transformed into a standard Neutrosophic OffGraph by collapsing the soft parameters into a single, unified membership value over the vertex and edge sets.*

*Proof.* Let $G = (G^*, J, K, A)$ be a Neutrosophic Soft OffGraph, where:

- $G^* = (V, E)$ is a neutrosophic graph with membership functions $T(x) \in [0, \Omega]$, $I(x) \in [0, 1]$, and $F(x) \in [\Psi, 1]$ for each $x \in V \cup E$.
- $(J, A)$ and $(K, A)$ are neutrosophic soft sets over $V$ and $E$, respectively.

Define the transformed graph $G' = (V, E)$ as a Neutrosophic OffGraph by setting each vertex $v \in V$ and each edge $e \in E$ to have the following unified membership functions:

$$T'(v) = \sup_{a \in A} T_{J(a)}(v), \quad I'(v) = \sup_{a \in A} I_{J(a)}(v), \quad F'(v) = \inf_{a \in A} F_{J(a)}(v),$$

where $T_{J(a)}(v)$, $I_{J(a)}(v)$, and $F_{J(a)}(v)$ are the membership degrees from the soft set $J(a)$ on $v$. A similar transformation applies for each edge $e \in E$.

This construction ensures that $G'$ retains the core properties of $G$ with unified membership degrees, thus transforming the soft OffGraph structure into a standard Neutrosophic OffGraph. □

**Proposition 2.2.50.** *A Neutrosophic Soft OverGraph can be transformed into a standard Neutrosophic OverGraph by consolidating the parameterized degrees.*

*Proof.* Given a Neutrosophic Soft OverGraph $G = (G^*, J, K, A)$, where membership degrees in $(J, A)$ and $(K, A)$ may exceed 1, construct $G' = (V, E)$ as follows:

$$T'(v) = \sup_{a \in A} T_{J(a)}(v), \quad I'(v) = \sup_{a \in A} I_{J(a)}(v), \quad F'(v) = \inf_{a \in A} F_{J(a)}(v).$$

This transformation results in a standard Neutrosophic OverGraph where the truth-membership can exceed the standard bound 1, satisfying the criteria for an OverGraph. □

**Proposition 2.2.51.** *A Neutrosophic Soft UnderGraph can be transformed into a standard Neutrosophic UnderGraph by reducing the parameters to single values.*

*Proof.* Let $G = (G^*, J, K, A)$ be a Neutrosophic Soft UnderGraph with each $J(a)$ and $K(a)$ containing non-membership degrees below zero. Define the transformed Neutrosophic UnderGraph $G' = (V, E)$ by setting:

$$T'(v) = \sup_{a \in A} T_{J(a)}(v), \quad I'(v) = \sup_{a \in A} I_{J(a)}(v), \quad F'(v) = \inf_{a \in A} F_{J(a)}(v),$$

where $F_{J(a)}(v) < 0$. This results in a single Neutrosophic UnderGraph that mirrors the original soft graph structure with combined membership degrees. □

**Proposition 2.2.52.** *A Neutrosophic Soft OffGraph, OverGraph, or UnderGraph can be transformed into a Soft Graph by disregarding the neutrosophic attributes and only using the soft set structure over parameters.*

*Proof.* For any Neutrosophic Soft OffGraph, OverGraph, or UnderGraph $G = (G^*, J, K, A)$, we can create a Soft Graph $G' = (V, E, A)$ by defining soft sets $(J, A)$ and $(K, A)$ without the neutrosophic components. Each vertex $v \in V$ and edge $e \in E$ is thus represented by soft sets without the need for neutrosophic membership degrees. This construction yields a Soft Graph $G'$ that represents the parameterized structure without neutrosophic complexity. □

**Proposition 2.2.53.** *A Neutrosophic Soft OffGraph can be transformed into a Neutrosophic Soft OverGraph or a Neutrosophic Soft UnderGraph.*

*Proof.* The proof can be constructed in a manner analogous to the set-based case. □

**Theorem 2.2.54.** *Neutrosophic Soft OffGraph, OverGraph, and UnderGraph generalize the concept of a Soft Graph. Specifically:*

- *A Neutrosophic Soft OffGraph generalizes a Soft Graph by incorporating neutrosophic truth, indeterminacy, and falsity degrees over an extended range $[\Psi, \Omega]$, where $\Psi < 0$ and $\Omega > 1$.*
- *A Neutrosophic Soft OverGraph generalizes a Soft Graph by allowing truth and other membership degrees to exceed the standard maximum of 1 ($T, I, F \in [0, \Omega]$, $\Omega > 1$).*
- *A Neutrosophic Soft UnderGraph generalizes a Soft Graph by permitting falsity and other membership degrees to fall below zero ($T, I, F \in [\Psi, 1]$, $\Psi < 0$).*

*Proof.* Let $G = (G^*, J, K, A)$ be a Soft Graph, where $G^* = (V, E)$, $J$ is a soft set over $V$, and $K$ is a soft set over $E$. For each parameter $a \in A$, $J(a) \subseteq V$ and $K(a) \subseteq E$.

1. A Neutrosophic Soft OffGraph $G_{\text{NSO}} = (G^*, J, K, A, T, I, F)$ extends $G$ by assigning:

   $$T, I, F : V \cup E \to [\Psi, \Omega], \quad T(x) + I(x) + F(x) \leq 3\Omega, \quad \forall x \in V \cup E.$$

   When $\Psi = 0$ and $\Omega = 1$, this reduces to the original Soft Graph.

2. A Neutrosophic Soft OverGraph $G_{\text{NSOver}} = (G^*, J, K, A, T, I, F)$ uses:

   $$T, I, F : V \cup E \to [0, \Omega], \quad T(x) + I(x) + F(x) \leq 3\Omega, \quad \forall x \in V \cup E.$$

   Setting $\Omega = 1$ confines the membership degrees to $[0, 1]$, recovering the Soft Graph.

3. A Neutrosophic Soft UnderGraph $G_{\text{NSUnder}} = (G^*, J, K, A, T, I, F)$ uses:

   $$T, I, F : V \cup E \to [\Psi, 1], \quad T(x) + I(x) + F(x) \leq 3, \quad \forall x \in V \cup E.$$

   When $\Psi = 0$, this becomes a Soft Graph.

In all cases, the Neutrosophic extensions allow for broader ranges of membership degrees, reducing to the Soft Graph when $\Psi = 0$ and $\Omega = 1$. This proves the generalization. □

### 2.2.5 Rough set and Rough Graph

This Book partially focuses on Rough Sets and Rough Graphs. A Rough Set [48, 231] (or a Rough Graph [232, 233]) is a mathematical model developed to approximate uncertain or imprecise data through the use of lower and upper approximations. Given the significant amount of research extending Rough Sets and Rough Graphs using concepts like Fuzzy Sets, Neutrosophic Sets, and Soft Sets, the study of these frameworks is evidently of great importance [234–237].

The definitions are provided below.

**Definition 2.2.55** (Pawlak rough set)**.** [47] Let $X \neq \emptyset$ be a universe and let $R \subseteq X \times X$ be an equivalence (indiscernibility) relation. For $x \in X$, write $[x]_R := \{y \in X : xRy\}$ for the equivalence class of $x$. For any subset $U \subseteq X$, the *lower* and *upper* $R$-approximations of $U$ are defined by

$$\underline{R}(U) := \{x \in X : [x]_R \subseteq U\}, \qquad \overline{R}(U) := \{x \in X : [x]_R \cap U \neq \emptyset\}.$$

The pair $\big(\underline{R}(U), \overline{R}(U)\big)$ is called the *rough set representation* of $U$ (with respect to $R$), and it always satisfies

$$\underline{R}(U) \subseteq U \subseteq \overline{R}(U).$$

**Remark 2.2.56.** The lower approximation $\underline{R}(U)$ contains objects that *certainly* belong to $U$ given the granulation by $R$, whereas the upper approximation $\overline{R}(U)$ contains objects that *possibly* belong to $U$. The boundary region $\overline{R}(U) \setminus \underline{R}(U)$ quantifies the uncertainty caused by indistinguishability.

**Proposition 2.2.57** (Crisp sets as a special case)**.** *Rough sets generalize crisp sets: if R is the identity relation on X, then*

$$\underline{R}(U) = U = \overline{R}(U) \qquad (\forall\, U \subseteq X).$$

*Proof.* If $R$ is the identity relation, then $[x]_R = \{x\}$ for every $x \in X$. Hence $[x]_R \subseteq U$ holds iff $x \in U$, so $\underline{R}(U) = U$. Similarly $[x]_R \cap U \neq \emptyset$ holds iff $x \in U$, so $\overline{R}(U) = U$. □

**Definition 2.2.58** (Rough graph (one standard model))**.** (cf. [232]) Let $G = (V, E)$ be a finite simple graph, where $V \neq \emptyset$ and $E \subseteq \binom{V}{2}$. Let $R_V$ be an equivalence relation on $V$ and let $R_E$ be an equivalence relation on $E$ (typically induced from an attribute indiscernibility on vertices or edges). For any chosen crisp vertex set $U \subseteq V$ and edge set $F \subseteq E$, define

$$\underline{R_V}(U) \subseteq V, \quad \overline{R_V}(U) \subseteq V, \qquad \underline{R_E}(F) \subseteq E, \quad \overline{R_E}(F) \subseteq E$$

by the Pawlak formulas in Definition 2.2.55. The *rough graph* determined by $(U, F)$ (with respect to $(R_V, R_E)$) is the quadruple

$$G_R = \big(\underline{R_V}(U),\ \overline{R_V}(U),\ \underline{R_E}(F),\ \overline{R_E}(F)\big),$$

which encodes certain/possible vertices and edges under the indiscernibility relations.

**Example 2.2.59** (A rough graph via Pawlak approximations)**.** Let $G = (V, E)$ be the path on four vertices

$$V = \{v_1, v_2, v_3, v_4\}, \qquad E = \{e_{12}, e_{23}, e_{34}\},$$

where $e_{12} = v_1v_2$, $e_{23} = v_2v_3$, and $e_{34} = v_3v_4$.

**Vertex indiscernibility.** Define an equivalence relation $R_V$ on $V$ with classes

$$[v_1]_{R_V} = \{v_1, v_2\}, \qquad [v_3]_{R_V} = \{v_3, v_4\}.$$

(Thus $v_1 \sim v_2$ and $v_3 \sim v_4$.)

**Edge indiscernibility.** Define an equivalence relation $R_E$ on $E$ by declaring

$$[e_{12}]_{R_E} = \{e_{12}\}, \qquad [e_{23}]_{R_E} = \{e_{23}, e_{34}\},$$

so $e_{23} \sim e_{34}$ and $e_{12}$ is alone.

**Chosen crisp sets.** Take the crisp vertex set $U := \{v_1, v_2\} \subseteq V$ and the crisp edge set $F := \{e_{23}\} \subseteq E$. Then, by Definition 2.2.55,

$$\underline{R_V}(U) = \{x \in V : [x]_{R_V} \subseteq U\} = \{v_1, v_2\}, \qquad \overline{R_V}(U) = \{x \in V : [x]_{R_V} \cap U \neq \emptyset\} = \{v_1, v_2\}.$$

For edges,

$$\underline{R_E}(F) = \{e \in E : [e]_{R_E} \subseteq F\} = \emptyset \quad (\text{since } [e_{23}]_{R_E} = \{e_{23}, e_{34}\} \not\subseteq F),$$

$$\overline{R_E}(F) = \{e \in E : [e]_{R_E} \cap F \neq \emptyset\} = \{e_{23}, e_{34}\}.$$

Hence the rough graph determined by $(U, F)$ with respect to $(R_V, R_E)$ is

$$G_R = \big(\underline{R_V}(U),\ \overline{R_V}(U),\ \underline{R_E}(F),\ \overline{R_E}(F)\big) = \big(\{v_1, v_2\},\ \{v_1, v_2\},\ \emptyset,\ \{e_{23}, e_{34}\}\big).$$

We include the graph in Figure 2.5.

Figure 2.5: A rough graph: certain vertices $\underline{R_V}(U)$ (shaded) and possible edges $\overline{R_E}(F)$ (dashed) (Example 2.2.59).

## 2.3 Hyperconcepts and Superhyperconcepts

In this subsection, we introduce several types of hyperconcepts and superhyperconcepts. While these terms may have slightly different meanings across various mathematical fields, many concepts in each field are defined from the perspective of hyper and superhyper concepts. These definitions often serve to generalize classical concepts. Here, we briefly present a few examples of hyperconcepts and superhyperconcepts. It is important to note that in some fields, multiple definitions exist, and the interpretations of "hyper" or "super" may vary. However, concepts such as Superhypergraph can be understood as structures based on the $n$-th PowerSet framework.

**Definition 2.3.1** ($n$-th PowerSet)**.** [238] Let $H$ be a set representing a system or structure, such as a set of items, a company, an institution, a country, or a region. The *n-th PowerSet*, denoted as $\mathcal{P}_n^*(H)$, describes a hierarchical organization of $H$ into subsystems, sub-subsystems, and so forth. It is defined recursively as follows:

1. **Base Case:**
$$\mathcal{P}_0^*(H) := H.$$
2. **First-Level PowerSet:**
$$\mathcal{P}_1^*(H) = \mathcal{P}(H),$$
where $\mathcal{P}(H)$ is the power set of $H$.
3. **Higher Levels:** For $n \geq 2$, the $n$-th PowerSet is defined recursively as:
$$\mathcal{P}_n^*(H) = \mathcal{P}(\mathcal{P}_{n-1}^*(H)).$$

Thus, $\mathcal{P}_n^*(H)$ represents a nested hierarchy, where the power set operation $\mathcal{P}$ is applied $n$ times. Formally:

$$\mathcal{P}_n^*(H) = \mathcal{P}(\mathcal{P}(\cdots \mathcal{P}(H) \cdots)),$$

where the power set operation $\mathcal{P}$ is repeated $n$ times.

### 2.3.1 Hypergraphs and SuperHyperGraphs

Hypergraphs extend ordinary graphs by allowing an edge to join an arbitrary nonempty subset of the vertex set, hence capturing genuinely multiway relations among entities [239,240]. They have been applied in many areas, including database theory and systems [85]. For further background, see standard monographs and surveys such as [241,242].

**Definition 2.3.2** (Hypergraph)**.** [239] A *hypergraph* is a pair $H = (V, \mathcal{E})$ where $V \neq \emptyset$ is a finite set (the *vertices*) and

$$\mathcal{E} \subseteq \mathcal{P}(V) \setminus \{\emptyset\}$$

is a finite family of nonempty subsets of $V$ (the *hyperedges*).

**Proposition 2.3.3** (Hypergraphs generalize graphs)**.** *Every (finite simple undirected) graph can be viewed as a hypergraph.*

*Proof.* Let $G = (V, E)$ be a finite simple graph, so $E \subseteq \binom{V}{2}$. Define a hypergraph $H = (V, \mathcal{E})$ by $\mathcal{E} := E$. Then every hyperedge has size 2, and adjacency in $G$ coincides with co-membership in a size-2 hyperedge of $H$. Hence $G$ is a special case of a hypergraph. □

The term *supergraph* is standard in graph theory and is unrelated to the Smarandache notion of *SuperHyperGraph*. We record it for clarity.

**Definition 2.3.4** (Supergraph)**.** (cf. [243,244]) Let $H = (V_H, E_H)$ and $G = (V_G, E_G)$ be graphs. We say that $G$ is a *supergraph* of $H$ if

$$V_H \subseteq V_G \quad \text{and} \quad E_H \subseteq E_G.$$

If $V_H = V_G$ and $E_H \subsetneq E_G$, then $G$ is an *edge-supergraph* of $H$. If $E_H = E_G$ and $V_H \subsetneq V_G$, then $G$ is a *vertex-supergraph* of $H$.

SuperHyperGraphs (in the sense of Smarandache) are powerset-based structures designed to encode nested or hierarchical vertex objects. A common source of confusion is the typing of the vertex set: one must distinguish a base set $V_0$ from the supervertex set $V$. The following definition is set-theoretically consistent and aligns with the $n$-SuperHyperGraph formulation used elsewhere in this manuscript.

**Definition 2.3.5** (SuperHyperGraph as the case $n = 1$)**.** [101,238] Let $V_0 \neq \emptyset$ be a finite *base set*. A *SuperHyperGraph* on $V_0$ is an ordered pair

$$\mathrm{SHG}^{(1)} = (V, E)$$

satisfying

$$V \subseteq \mathcal{P}(V_0) \quad \text{and} \quad E \subseteq \mathcal{P}(V) \setminus \{\emptyset\}.$$

Elements of $V$ are called *supervertices* (each is a subset of $V_0$), and elements of $E$ are called *superedges* (each is a nonempty subset of the supervertex set $V$).

**Proposition 2.3.6** (SuperHyperGraphs generalize hypergraphs)**.** *Every hypergraph is a special case of a SuperHyperGraph.*

*Proof.* Let $H = (V_0, \mathcal{E})$ be a hypergraph (Definition 2.3.2). Define a SuperHyperGraph $\mathrm{SHG}^{(1)} = (V, E)$ on the same base set $V_0$ by

$$V := \big\{\{v\} : v \in V_0\big\} \subseteq \mathcal{P}(V_0), \qquad E := \big\{ \{\{v\} : v \in e\} \ : \ e \in \mathcal{E} \big\}.$$

Then each element of $E$ is a nonempty subset of $V$, so $E \subseteq \mathcal{P}(V) \setminus \{\emptyset\}$. The mapping $v \mapsto \{v\}$ is a bijection between $V_0$ and $V$, and under this identification each hyperedge $e \subseteq V_0$ corresponds to the superedge $\{\{v\} : v \in e\} \subseteq V$. Hence $H$ is realized as a special case of a SuperHyperGraph. □

**Remark 2.3.7** (SuperHyperGraph vs. supergraph)**.** The term *supergraph* (Definition 2.3.4) is an inclusion relation between two graphs. In contrast, a *SuperHyperGraph* (Definition 2.3.5) is a powerset-based incidence structure. Thus, “SuperHyperGraph generalizes supergraph” is not a meaningful comparison: the two notions live in different categories. A supergraph relation can be discussed *inside* the support graphs associated with a SuperHyperGraph, but it is not a direct generalization in the definitional sense.

Some authors consider map-based variants in which “superedges” are explicit mappings between supervertices [103]. We record one such definition as a distinct concept.

**Definition 2.3.8** (Quasi-SuperHyperGraph)**.** [103] A *quasi-superhypergraph* is a triple $Q = (V, S, \Phi)$ where:

- $V$ is a (finite) set of *vertices*;
- $S = \{S_i\}_{i=1}^{k} \subseteq \mathcal{P}(V) \setminus \{\emptyset\}$ is a family of nonempty subsets of $V$, called *supervertices*;
- $\Phi = \{\varphi_{i,j} : i \neq j\}$ is a family of maps $\varphi_{i,j} : S_i \longrightarrow S_j$, sometimes called *superedge maps*.

**Remark 2.3.9.** In a quasi-superhypergraph, the relationship structure is encoded by the maps $\varphi_{i,j}$ rather than by a family of set-valued edges $E \subseteq \mathcal{P}(S) \setminus \{\emptyset\}$. Therefore, quasi-superhypergraphs and SuperHyperGraphs are different generalizations, and neither is a literal special case of the other without additional identifications.

Next, multiset-based variants allow repeated vertices inside supervertices and repeated supervertices inside superedges.

**Definition 2.3.10** (Pseudo-SuperHyperGraph)**.** [245] A *pseudo-superhypergraph* is a triple $H = (V, \mathfrak{S}, \mathfrak{E})$ where:

- $V \neq \emptyset$ is a finite set of vertices;
- $\mathfrak{S}$ is a *multiset* of multisets over $V$ (the *supervertices*);
- $\mathfrak{E}$ is a *multiset* of multisets over $\mathfrak{S}$ (the *superedges*).

**Proposition 2.3.11** (Set-reduction)**.** *Every pseudo-superhypergraph admits an underlying (set-based) superedge incidence structure by forgetting multiplicities.*

*Proof.* Given $H = (V, \mathfrak{S}, \mathfrak{E})$, replace each multiset by its underlying support set. Let $S$ be the set of distinct supports of elements of $\mathfrak{S}$, and let $E$ be the set of distinct supports of elements of $\mathfrak{E}$ (viewed as subsets of $S$). Then $E \subseteq \mathcal{P}(S) \setminus \{\emptyset\}$, yielding a set-based hypergraph $(S, E)$ on supervertices. □

A SuperHyperGraph admits a natural iterated-powerset generalization, called an $n$-SuperHyperGraph. Since the definition is sometimes stated with inconsistent typing of the edge set, we record below a clean, set-theoretically consistent formulation: vertices live in the $n$-fold powerset of a base set, and each edge is a (nonempty) subset of the chosen vertex set. In particular, edges are *not* required to be elements of $\mathcal{P}^n(V_0)$; rather, they are subsets of $V$ (as in ordinary graphs and hypergraphs) [101, 245, 246]. For further discussion of SuperHyperGraphs, see [247–250], as needed.

**Definition 2.3.12** ($n$-SuperHyperGraph)**.** [250] Let $V_0$ be a finite, nonempty *base vertex set* and fix an integer $n \geq 0$. An *$n$-SuperHyperGraph on $V_0$* is a pair

$$\mathrm{SHG}^{(n)} = (V, E)$$

such that

$$V \subseteq \mathcal{P}^n(V_0) \quad \text{and} \quad E \subseteq \mathcal{P}(V) \setminus \{\emptyset\}.$$

Elements of $V$ are called *$n$-supervertices* (they are $n$-times nested set-objects over $V_0$), and elements of $E$ are called *$n$-superedges* (each superedge is a nonempty subset of $V$).

**Remarks.**

(i) $\emptyset$ may occur as an element of $\mathcal{P}^n(V_0)$; one may allow or forbid $\emptyset \in V$ by convention. Many combinatorial studies assume $V \subseteq \mathcal{P}^n(V_0) \setminus \{\emptyset\}$.

(ii) When $n = 0$, we have $V \subseteq V_0$ and $E \subseteq \mathcal{P}(V) \setminus \{\emptyset\}$, so $\mathrm{SHG}^{(0)}$ is a (crisp) hypergraph on the vertex set $V$.

(iii) When $n = 1$, vertices are subsets of $V_0$ and edges are families of such subsets; this recovers the common "superhypergraph" viewpoint where vertices may themselves be set-valued objects.

**Example 2.3.13** ($n$-SuperHyperGraph for enterprise IT portfolio governance)**. Scenario.** An enterprise manages work in nested layers: *tasks* $\rightarrow$ *projects* $\rightarrow$ *programs*. A portfolio committee often imposes constraints that involve multiple programs simultaneously.

**Base set (atomic tasks).** Let

$$V_0 = \{\mathrm{T_{cloud}},\ \mathrm{T_{data}},\ \mathrm{T_{sec}},\ \mathrm{T_{ui}}\}.$$

Take $n = 3$. Then $\mathcal{P}^3(V_0)$ contains set-of-set-of-set objects, suitable for representing programs as collections of projects, each project being a collection of task-bundles.

**Three 3-supervertices (programs).** Define

$$p_1 = \Big\{ \{\{\mathrm{T_{cloud}}, \mathrm{T_{sec}}\}\},\ \{\{\mathrm{T_{data}}\}\} \Big\}, \quad p_2 = \Big\{ \{\{\mathrm{T_{ui}}\}\},\ \{\{\mathrm{T_{cloud}}, \mathrm{T_{data}}\}\} \Big\},$$

$$p_3 = \Big\{ \{\{\mathrm{T_{sec}}\}\},\ \{\{\mathrm{T_{data}}, \mathrm{T_{ui}}\}\} \Big\}.$$

Let

$$V = \{p_1, p_2, p_3\} \subseteq \mathcal{P}^3(V_0).$$

**Two superedges (portfolio-level governance relations).** Let

$$e_1 = \{p_1, p_2\}, \qquad e_2 = \{p_2, p_3\}, \qquad E = \{e_1, e_2\} \subseteq \mathcal{P}(V) \setminus \{\emptyset\}.$$

Then $\mathrm{SHG}^{(3)} = (V, E)$ is a valid 3-SuperHyperGraph.

**Interpretation.** Each $p_i$ is a nested program object (projects-of-task-bundles). A superedge (e.g., $e_1$) models a portfolio constraint spanning multiple programs, such as shared budget approval, joint security review, or synchronized releases.

**Example 2.3.14** ($n$-SuperHyperGraph for hospital care coordination)**. Scenario.** A hospital coordinates patient care through nested structures: *clinical actions* $\rightarrow$ *care bundles* $\rightarrow$ *care pathways*. Some safety rules require joint participation of multiple pathways (e.g., pre-op clearance, medication reconciliation, and discharge planning).

**Base set (atomic clinical actions).** Let

$$V_0 = \{\mathrm{A_{lab}},\ \mathrm{A_{imaging}},\ \mathrm{A_{med}},\ \mathrm{A_{pt}}\},$$

where, for example, $\mathrm{A_{lab}}$ denotes ordering lab tests, $\mathrm{A_{imaging}}$ ordering imaging, $\mathrm{A_{med}}$ administering medication, and $\mathrm{A_{pt}}$ initiating physical therapy. Take $n = 2$, so $\mathcal{P}^2(V_0) = \mathcal{P}(\mathcal{P}(V_0))$ supports "sets of action-bundles".

**Three 2-supervertices (care pathways).** Define

$$X_1 = \big\{\{\mathrm{A_{lab}}, \mathrm{A_{imaging}}\},\ \{\mathrm{A_{med}}\}\big\}, \quad X_2 = \big\{\{\mathrm{A_{med}}, \mathrm{A_{pt}}\},\ \{\mathrm{A_{lab}}\}\big\},$$

$$X_3 = \big\{\{\mathrm{A_{imaging}}\},\ \{\mathrm{A_{pt}}\}\big\}.$$

Let

$$V = \{X_1, X_2, X_3\} \subseteq \mathcal{P}^2(V_0).$$

**Superedges (cross-pathway coordination requirements).** Define

$$\varepsilon_1 = \{X_1, X_2\}, \qquad \varepsilon_2 = \{X_1, X_2, X_3\}, \qquad E = \{\varepsilon_1, \varepsilon_2\} \subseteq \mathcal{P}(V) \setminus \{\emptyset\}.$$

Then $\mathrm{SHG}^{(2)} = (V, E)$ is a valid 2-SuperHyperGraph.

**Interpretation.** Each 2-supervertex $X_i$ encodes a pathway as a set of action-bundles (e.g., diagnostic bundle plus medication bundle). The superedge $\varepsilon_2 = \{X_1, X_2, X_3\}$ represents a multi-pathway safety gate that must be jointly satisfied (e.g., discharge is allowed only if imaging review, medication reconciliation, and mobility assessment pathways are all cleared).

**Theorem 2.3.15** (SuperHyperGraphs are the $n = 1$ case)**.** *Let $V_0$ be a finite nonempty base set. Any superhypergraph modeled as a pair $(V, E)$ with $V \subseteq \mathcal{P}(V_0)$ and $E \subseteq \mathcal{P}(V) \setminus \{\emptyset\}$ is precisely an n-SuperHyperGraph with $n = 1$, i.e., an instance of Definition 2.3.12.*

*Proof.* If $n = 1$, then $\mathcal{P}^1(V_0) = \mathcal{P}(V_0)$. Hence the conditions $V \subseteq \mathcal{P}^1(V_0)$ and $E \subseteq \mathcal{P}(V) \setminus \{\emptyset\}$ coincide exactly with $V \subseteq \mathcal{P}(V_0)$ and $E \subseteq \mathcal{P}(V) \setminus \{\emptyset\}$, which is the stated superhypergraph model. Therefore the two notions agree. □

**Theorem 2.3.16** ($n$-SuperHyperGraphs are powerset-based)**.** *Let $\mathrm{SHG}^{(n)} = (V, E)$ be an n-SuperHyperGraph on $V_0$ (Definition 2.3.12). Then every supervertex $X \in V$ is an element of the n-fold iterated powerset $\mathcal{P}^n(V_0)$. In particular, $X$ is a nested set-object built from elements of $V_0$ with depth at most $n$.*

*Proof.* This is immediate from the vertex-typing condition $V \subseteq \mathcal{P}^n(V_0)$ in Definition 2.3.12. The recursive construction of $\mathcal{P}^n(V_0)$ ensures that its elements are obtained by iterating the powerset operation $n$ times starting from $V_0$, hence are nested set-objects over $V_0$ of depth at most $n$. □

### 2.3.2 Hypersoft Graph

A HyperSoft Graph represents nodes with multiple attributes, where each node can hold unique attribute values, facilitating complex, multi-dimensional relationships. This type of graph is derived by applying the concept of a HyperSoft Set [251–253] to graph theory. The formal definitions of a HyperSoft Set and a HyperSoft Graph are presented below [254–256].

**Definition 2.3.17** (Hypersoft Set)**.** [51] Let $X$ be a non-empty finite universe, and let $T_1, T_2, \ldots, T_n$ be $n$-distinct attributes with corresponding disjoint sets $J_1, J_2, \ldots, J_n$. A pair $(F, J)$ is called a *hypersoft set* over the universal set $X$, where $F$ is a mapping defined by

$$F : J \to \mathcal{P}(X),$$

with $J = J_1 \times J_2 \times \cdots \times J_n$.

**Proposition 2.3.18.** *A hypersoft Set is a generalization of a soft Set.*

*Proof.* This is evident. □

**Definition 2.3.19** (Null Hypersoft Set)**.** Let $X$ be a non-empty finite universe, and let $T_1, T_2, \ldots, T_n$ be $n$-distinct attributes with corresponding disjoint sets $J_1, J_2, \ldots, J_n$. A hypersoft set $(F, J)$ over $X$, where $J = J_1 \times J_2 \times \cdots \times J_n$ and $F : J \to \mathcal{P}(X)$, is called a *Null Hypersoft Set*, denoted by $\Phi$, if for every $j \in J$, the subset $F(j)$ of $X$ is empty. Formally:

$$(F, J) = \Phi \quad \text{if and only if} \quad F(j) = \emptyset \quad \forall j \in J.$$

**Proposition 2.3.20.** *A null hypersoft Set is a hypersoft Set.*

*Proof.* This is evident. □

**Definition 2.3.21** (Full Hypersoft Set)**.** Let $X$ be a non-empty finite universe, and let $T_1, T_2, \ldots, T_n$ be $n$-distinct attributes with corresponding disjoint sets $J_1, J_2, \ldots, J_n$. A hypersoft set $(F, J)$ over $X$, where $J = J_1 \times J_2 \times \cdots \times J_n$ and $F : J \to \mathcal{P}(X)$, is called a *Full Hypersoft Set* if:

$$\bigcup_{j \in J} F(j) = X.$$

This condition ensures that every element of the universe $X$ is included in at least one subset $F(j)$ for some $j \in J$.

**Proposition 2.3.22.** *A Full hypersoft Set is a hypersoft Set.*

*Proof.* This is evident. □

**Theorem 2.3.23.** *A Null Hypersoft Set generalizes a Null Soft Set.*

*Proof.* To show generalization, consider $n = 1$ and $J_1 = A$. In this case, $J = J_1 = A$, and $F : J_1 \to \mathcal{P}(X)$ reduces to $F : A \to \mathcal{P}(U)$. Since $F(j) = \emptyset$ for all $j \in J$ is equivalent to $F(\varepsilon) = \emptyset$ for all $\varepsilon \in A$, the Null Hypersoft Set becomes a Null Soft Set when $n = 1$. Hence, the Null Hypersoft Set generalizes the Null Soft Set. □

**Theorem 2.3.24.** *A Full Hypersoft Set generalizes a Full Soft Set.*

*Proof.* To show generalization, consider $n = 1$ and $J_1 = A$. In this case, $J = J_1 = A$, and $F : J_1 \to \mathcal{P}(X)$ reduces to $F : A \to \mathcal{P}(U)$. Since $\bigcup_{j \in J} F(j) = X$ is equivalent to $\bigcup_{\varepsilon \in A} F(\varepsilon) = U$, the Full Hypersoft Set becomes a Full Soft Set when $n = 1$. Hence, the Full Hypersoft Set generalizes the Full Soft Set. □

**Definition 2.3.25** (Hypersoft Graph)**.** (cf. [254–256]) Let $G = (V, E)$ be a simple connected graph, where $V$ is the set of vertices and $E$ is the set of edges. Consider $J = J_1 \times J_2 \times \cdots \times J_n$, where each $J_i \subseteq V$ and $J_i \cap J_j = \emptyset$ for $i \neq j$. A *Hypersoft Graph* (HS-Graph) of $G$ is defined as a hypersoft set $(F, J)$ over $V$ such that for each $x \in J$, $F(x)$ induces a connected subgraph of $G$. The set of all HS-Graphs of $G$ is denoted by $\mathrm{HsG}(G)$.

**Proposition 2.3.26.** *A hypersoft graph is a generalization of a soft graph.*

*Proof.* This is evident. □

The notions of *SuperHyperSoft Sets* [257–259], *IndetermSoft Sets* [260], and *IndetermHyperSoft Sets* [212,213] are closely related developments in the soft-set literature. For completeness, we recall the SuperHyperSoft Set framework and its basic variants below.

**Definition 2.3.27** (SuperHyperSoft Set)**.** [258] Let $U$ be a universe of discourse. Fix $n \geq 1$ distinct attributes $a_1, \dots, a_n$, and for each $i \in \{1, \dots, n\}$ let $A_i$ be the (nonempty) set of admissible values of $a_i$. Put

$$J \;:=\; \mathcal{P}(A_1) \times \cdots \times \mathcal{P}(A_n),$$

where $\mathcal{P}(\cdot)$ denotes the powerset. A *SuperHyperSoft Set* over $U$ (with respect to $(A_1, \dots, A_n)$) is a pair $(F, J)$, where

$$F : \; J \longrightarrow \mathcal{P}(U)$$

assigns to each $n$-tuple $(S_1, \dots, S_n) \in J$ (with $S_i \subseteq A_i$) a subset $F(S_1, \dots, S_n) \subseteq U$.

**Remark 2.3.28.** A HyperSoft Set is usually indexed by *single* attribute values, i.e., by $J_1 \times \cdots \times J_n$ with $J_i = A_i$. In contrast, a SuperHyperSoft Set uses $\mathcal{P}(A_i)$ as the value domain of each attribute, so one may input *subsets* of admissible values, thereby allowing richer parameterization.

**Proposition 2.3.29** (SuperHyperSoft Sets generalize HyperSoft Sets)**.** *Let $(F, J)$ be a SuperHyperSoft Set over $U$ as in Definition 2.3.27. Assume that each value domain $A_i$ is a singleton, say $A_i = \{\alpha_i\}$. Then $\mathcal{P}(A_i) = \{\emptyset, \{\alpha_i\}\}$, and restricting $F$ to the subdomain*

$$J^\star \;:=\; \{\{\alpha_1\}\} \times \cdots \times \{\{\alpha_n\}\} \;\subseteq\; J$$

*yields an $n$-parameter HyperSoft Set on $U$.*

*Proof.* If $A_i = \{\alpha_i\}$, then $J^\star$ consists of the single point $(\{\alpha_1\}, \dots, \{\alpha_n\})$. Define $F^\star : J^\star \to \mathcal{P}(U)$ by $F^\star := F|_{J^\star}$. This is a hypersoft-style assignment indexed by a fixed $n$-tuple of attribute values. More generally, if one fixes any subdomain $J_1 \times \cdots \times J_n \subseteq \mathcal{P}(A_1) \times \cdots \times \mathcal{P}(A_n)$ and restricts $F$ to it, one obtains a hypersoft-set type structure on that chosen index set. □

**Definition 2.3.30** (Null SuperHyperSoft Set)**.** Let $(F, J)$ be a SuperHyperSoft Set over $U$ (Definition 2.3.27). We call $(F, J)$ the *null SuperHyperSoft Set* (denoted $\Phi$) if

$$F(j) = \emptyset \qquad (\forall j \in J).$$

**Definition 2.3.31** (Full SuperHyperSoft Set)**.** Let $(F, J)$ be a SuperHyperSoft Set over $U$ (Definition 2.3.27). We call $(F, J)$ a *full SuperHyperSoft Set* if

$$\bigcup_{j \in J} F(j) \; = \; U.$$

**Proposition 2.3.32** (Null/full SuperHyperSoft Sets generalize the HyperSoft analogues)**.** *Under the singleton-value restriction $A_i = \{\alpha_i\}$ for all $i$, the notions in Definitions 2.3.30 and 2.3.31 reduce to the standard null/full conditions for the corresponding HyperSoft Set.*

*Proof.* When each $A_i$ is a singleton, restricting to $J^\star$ as in Proposition 2.3.29 produces a hypersoft-style mapping $F^\star$. Then $F(j) = \emptyset$ for all $j \in J$ implies $F^\star(j^\star) = \emptyset$, i.e., the null condition. Likewise, $\bigcup_{j \in J} F(j) = U$ implies $\bigcup_{j \in J^\star} F^\star(j) = U$ (since $J^\star \subseteq J$), which is the full condition on the restricted hypersoft structure. □

For reference, the relationships between the Soft set are illustrated in Figure 2.6. (cf. [71])

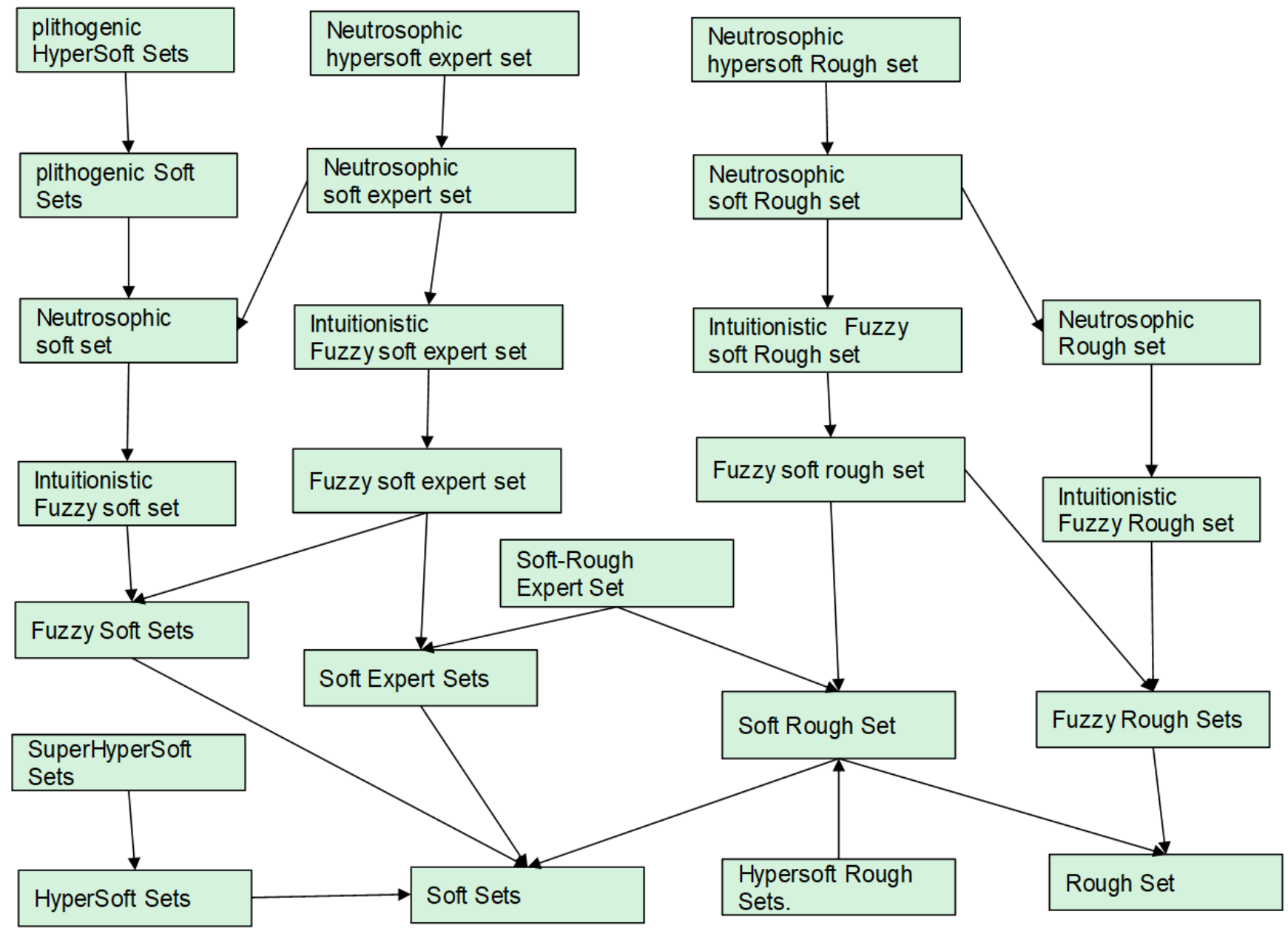

Figure 2.6: Some Soft sets Hierarchy. The set class at the origin of an arrow contains the set class at the destination of the arrow(cf. [71]).

IndetermSoft Sets and IndetermHyperSoft Sets were introduced to model situations in which the *parameter domain*, the *underlying universe*, the *information granulation* (powerset), or the *parameter-to-subset assignment* may be only partially specified, ambiguous, or otherwise indeterminate [212, 213].

**Definition 2.3.33** (IndetermSoft Set)**.** [212] Let $U$ be a universe of discourse and let $H \subseteq U$ be a nonempty set. Let $A$ be a (nonempty) set of parameter values for a fixed attribute. A pair $(F, A)$ with

$$F : A \longrightarrow \mathcal{P}(H)$$

is called an *IndetermSoft Set* over $H$ if at least one of the following components is *indeterminate*:

(i) the parameter set $A$ (e.g., some parameters are uncertain, incomplete, or not uniquely specified);

(ii) the universe $H$ (and hence $\mathcal{P}(H)$) (e.g., the set of admissible objects is only partially known);

(iii) the assignment $F$ (e.g., for some $a \in A$, the value $F(a)$ is ambiguous, non-unique, incomplete, or unknown).

When none of (i)–(iii) is indeterminate, $(F, A)$ is an ordinary soft set over $H$.

**Proposition 2.3.34** (IndetermSoft Sets generalize soft sets)**.** *Every (classical) soft set is an IndetermSoft Set.*

*Proof.* Let $(F, A)$ be a classical soft set over $H$, so $A$ and $H$ are well-defined sets and $F : A \to \mathcal{P}(H)$ is a well-defined function. Then $(F, A)$ fits Definition 2.3.33 as the special case in which no indeterminacy is present. Hence the class of IndetermSoft Sets contains all classical soft sets. □

**Definition 2.3.35** (IndetermHyperSoft Set)**.** [212, 213] Let $U$ be a universe of discourse and let $H \subseteq U$ be a nonempty set. Fix $n \geq 1$ distinct attributes $a_1, \ldots, a_n$ with corresponding (nonempty) value sets $A_1, \ldots, A_n$. Put

$$J := A_1 \times \cdots \times A_n.$$

A pair $(F, J)$ with

$$F : J \longrightarrow \mathcal{P}(H)$$

is called an *IndetermHyperSoft Set* over $H$ if at least one of the following components is *indeterminate*:

(i) at least one attribute-value set $A_i$ is indeterminate;

(ii) the universe $H$ (and hence $\mathcal{P}(H)$) is indeterminate;

(iii) the assignment $F$ is indeterminate (e.g., for some $j \in J$, the value $F(j)$ is ambiguous, non-unique, incomplete, or unknown).

When none of (i)–(iii) is indeterminate, $(F, J)$ is a classical HyperSoft Set over $H$.

**Proposition 2.3.36** (IndetermHyperSoft Sets generalize IndetermSoft and HyperSoft Sets)**.** *Let $(F, J)$ be an IndetermHyperSoft Set as in Definition 2.3.35.*

(i) *If $n = 1$, then $J = A_1$ and $(F, J)$ is an IndetermSoft Set over $H$.*

(ii) *If no indeterminacy is present in the data $(A_1, \ldots, A_n, H, F)$, then $(F, J)$ is a (classical) HyperSoft Set over $H$.*

*Proof.* (i) If $n = 1$, then $J = A_1$ and $F : J \to \mathcal{P}(H)$ is exactly a soft-set type mapping; the indeterminacy clauses in Definition 2.3.35 coincide with those in Definition 2.3.33. Hence $(F, J)$ is an IndetermSoft Set.

(ii) If $A_1, \ldots, A_n$ and $H$ are well-defined sets and $F$ is a well-defined function, then none of the indeterminacy triggers in Definition 2.3.35 applies, so $(F, J)$ is a classical HyperSoft Set by definition. □

TreeSoft Sets are hierarchical extensions of soft sets in which the parameter domain is organized as a rooted tree of attributes and sub-attributes. In particular, when the hierarchy has depth $m = 2$, the construction naturally specializes to MultiSoft Set models (cf. [212, 213]).

**Definition 2.3.37** (Attribute tree)**.** Let $A = \{A_1, \ldots, A_n\}$ be a finite set of *first-level* attributes. Assume that each node may have (finitely or countably many) children, producing a rooted tree whose root is the formal symbol $A$. Denote by $\mathrm{Tree}(A)$ the set of all nodes of this rooted tree, including the root and all descendants, and let $\mathrm{Tree}(A)_{\geq 1} := \mathrm{Tree}(A) \setminus \{A\}$ be the set of non-root nodes. The *depth* of $\mathrm{Tree}(A)$ is the maximum level of a node (assumed finite and denoted by $m$).

**Definition 2.3.38** (TreeSoft Set)**.** [212] Let $U$ be a universe of discourse and let $H \subseteq U$ be a nonempty set. Let $\mathrm{Tree}(A)$ be an attribute tree as in Definition 2.3.37. A *TreeSoft Set* over $H$ (with respect to $\mathrm{Tree}(A)$) is a mapping

$$F : \mathcal{P}\big(\mathrm{Tree}(A)_{\geq 1}\big) \longrightarrow \mathcal{P}(H),$$

where $\mathcal{P}(\cdot)$ denotes the powerset. For $S \subseteq \mathrm{Tree}(A)_{\geq 1}$, the value $F(S) \subseteq H$ is interpreted as the set of objects in $H$ that are compatible with (or approximately satisfy) the collection $S$ of selected tree-parameters.

**Proposition 2.3.39** (TreeSoft Sets generalize soft sets)**.** *Every classical soft set is obtained as a special case of a TreeSoft Set.*

*Proof.* Let $(G, E)$ be a soft set over $H$, where $E$ is a parameter set and $G : E \to \mathcal{P}(H)$. Construct a depth-1 attribute tree $\mathrm{Tree}(A)$ whose non-root nodes are exactly the parameters in $E$ (i.e., $\mathrm{Tree}(A)_{\geq 1} = E$). Define a TreeSoft Set $F : \mathcal{P}(E) \to \mathcal{P}(H)$ by

$$F(S) := \bigcap_{e \in S} G(e) \qquad (S \subseteq E),$$

with the convention $F(\emptyset) := H$. Then the original soft set is recovered from $F$ by restricting to singletons: $G(e) = F(\{e\})$ for all $e \in E$. Hence soft sets embed into TreeSoft Sets. □

**Definition 2.3.40** (Null and full TreeSoft Sets)**.** Let $F : \mathcal{P}(\mathrm{Tree}(A)_{\geq 1}) \to \mathcal{P}(H)$ be a TreeSoft Set.

(i) $F$ is a *null TreeSoft Set* (denoted $\Phi$) if $F(S) = \emptyset$ for every $S \subseteq \mathrm{Tree}(A)_{\geq 1}$.

(ii) $F$ is a *full TreeSoft Set* if

$$\bigcup_{S \subseteq \mathrm{Tree}(A)_{\geq 1}} F(S) = H.$$

**Theorem 2.3.41** (Null TreeSoft Sets generalize null soft sets)**.** *Let $(G, E)$ be a soft set over $H$. If $G(e) = \emptyset$ for all $e \in E$ (a null soft set), then the induced TreeSoft Set $F$ from Proposition 2.3.39 is a null TreeSoft Set. Conversely, if $F$ is a null TreeSoft Set, then $G(e) := F(\{e\})$ defines a null soft set.*

*Proof.* If $G(e) = \emptyset$ for all $e$, then for any $S \neq \emptyset$, $F(S) = \bigcap_{e \in S} G(e) = \emptyset$, and also $F(\emptyset) = H$ by convention may be overridden to $\emptyset$ if one uses the strict null convention; in either convention, restricting to singletons yields a null soft set. Conversely, if $F(S) = \emptyset$ for all $S$, then in particular $G(e) = F(\{e\}) = \emptyset$ for all $e$. □

**Theorem 2.3.42** (Full TreeSoft Sets generalize full soft sets)**.** *Let $(G, E)$ be a soft set over $H$. If $\bigcup_{e \in E} G(e) = H$ (a full soft set), then the induced TreeSoft Set $F$ satisfies $\bigcup_{S \subseteq E} F(S) = H$, hence is full.*

*Proof.* Since $H = \bigcup_{e \in E} G(e)$, for every $h \in H$ there exists $e \in E$ with $h \in G(e) = F(\{e\})$. Thus $h \in \bigcup_{S \subseteq E} F(S)$, proving $\bigcup_{S \subseteq E} F(S) = H$. □

**Proposition 2.3.43** (Depth 2 yields a MultiSoft Set representation)**.** *[212] Assume that* $\mathrm{Tree}(A)$ *has depth $m = 2$, so the non-root nodes consist of level-1 attributes $A_1, \ldots, A_n$ and their level-2 sub-attributes. Let $B$ be the set of all* level-2 *nodes (sub-attributes). Then any TreeSoft Set $F : \mathcal{P}(\mathrm{Tree}(A)_{\geq 1}) \to \mathcal{P}(H)$ induces a MultiSoft-type mapping*

$$G : \mathcal{P}(B) \longrightarrow \mathcal{P}(H), \qquad G(S) := F(S) \quad (S \subseteq B),$$

*by restricting $F$ to subsets of second-level nodes.*

*Proof.* When $m = 2$, every subset $S \subseteq B$ is, in particular, a subset of $\mathrm{Tree}(A)_{\geq 1}$, so $F(S)$ is defined. Defining $G(S) := F(S)$ produces a parameterized family of subsets of $H$ indexed by combinations of second-level parameters, which is the core data structure of MultiSoft models. □

**Definition 2.3.44** (TreeCrisp Set)**.** Let $\mathrm{Tree}(A)$ be an attribute tree. A *TreeCrisp Set* is a mapping

$$c : \mathrm{Tree}(A)_{\geq 1} \longrightarrow \{0, 1\}.$$

Equivalently, it is the characteristic function of a subset $C \subseteq \mathrm{Tree}(A)_{\geq 1}$, where $c(x) = 1$ iff $x \in C$.

**Theorem 2.3.45** (TreeCrisp Sets generalize crisp sets)**.** *If* $\mathrm{Tree}(A)_{\geq 1} = X$ *(i.e., the tree has depth* 1 *and its non-root nodes are exactly the elements of* $X$*), then TreeCrisp Sets* $c : \mathrm{Tree}(A)_{\geq 1} \to \{0, 1\}$ *are exactly crisp subsets of* $X$.

*Proof.* When $\mathrm{Tree}(A)_{\geq 1} = X$, a mapping $c : X \to \{0, 1\}$ is precisely the characteristic function of a crisp set $C = \{x \in X : c(x) = 1\} \subseteq X$. □

**Proposition 2.3.46** (TreeSoft Sets generalize TreeCrisp Sets)**.** *Let* $F : \mathcal{P}(\mathrm{Tree}(A)_{\geq 1}) \to \mathcal{P}(H)$ *be a TreeSoft Set. If we restrict the codomain to* $\{\emptyset, H\} \subseteq \mathcal{P}(H)$*, namely*

$$F(S) \in \{\emptyset, H\} \qquad (\forall S \subseteq \mathrm{Tree}(A)_{\geq 1}),$$

*then* $F$ *induces a TreeCrisp Set on* $\mathcal{P}(\mathrm{Tree}(A)_{\geq 1})$ *via*

$$c_F(S) \ := \ \begin{cases} 1, & F(S) = H, \\ 0, & F(S) = \emptyset. \end{cases}$$

*Thus TreeSoft Sets extend TreeCrisp behavior by allowing arbitrary subsets of* $H$ *as images.*

*Proof.* The definition of $c_F$ is well-posed under the stated restriction on the range of $F$. Allowing general values $F(S) \subseteq H$ strictly enlarges the representational capacity beyond binary outputs. □

### 2.3.3 HyperFuzzy Set

HyperFuzzy sets extend classical fuzzy sets by assigning to each element not a single membership degree, but a *nonempty set* of admissible membership degrees in $[0, 1]$. In this way, HyperFuzzy sets subsume ordinary fuzzy sets and interval-valued fuzzy sets as special cases (cf. [68, 261]).

**Definition 2.3.47** (HyperFuzzy set)**.** [68, 261] Let $X \neq \emptyset$ be a set and write

$$\mathcal{P}^*([0, 1]) \ := \ \mathcal{P}([0, 1]) \setminus \{\emptyset\}$$

for the family of all nonempty subsets of $[0, 1]$. A *HyperFuzzy set* on $X$ is a mapping

$$\widetilde{\mu} : X \longrightarrow \mathcal{P}^*([0, 1]).$$

The value $\widetilde{\mu}(x) \subseteq [0, 1]$ is interpreted as the set of all admissible membership degrees of $x$ (e.g., arising from multiple measurements, experts, or scenarios).

**Example 2.3.48** (A finite HyperFuzzy set)**.** Let $X = \{x_1, x_2, x_3\}$. Define $\widetilde{A}$ on $X$ by the membership correspondence

$$\widetilde{\mu}_{\widetilde{A}}(x_1) = \{0.10, 0.20, 0.30\}, \qquad \widetilde{\mu}_{\widetilde{A}}(x_2) = \{0.40, 0.50, 0.60\}, \qquad \widetilde{\mu}_{\widetilde{A}}(x_3) = \{0.60, 0.70, 0.80\}.$$

Thus $x_1$ may belong to $\widetilde{A}$ with any degree in $\{0.10, 0.20, 0.30\}$, and similarly for $x_2, x_3$.

**Proposition 2.3.49** (Reduction to a fuzzy set)**.** *Every fuzzy set* $A$ *on* $X$ *can be viewed as a HyperFuzzy set on* $X$.

*Proof.* Let $A$ be a fuzzy set with membership function $\mu_A : X \to [0, 1]$. Define $\widetilde{\mu} : X \to \mathcal{P}^*([0, 1])$ by

$$\widetilde{\mu}(x) \ := \ \{\mu_A(x)\} \qquad (x \in X).$$

Then $\widetilde{\mu}$ is a HyperFuzzy set and encodes exactly the same membership information as $A$. □

**Remark 2.3.50** (Interval-valued fuzzy sets as a special case)**.** If one restricts $\widetilde{\mu}(x)$ to be a closed interval $[\underline{\mu}(x), \overline{\mu}(x)] \subseteq [0, 1]$ for every $x \in X$, then Definition 2.3.47 reduces to the usual interval-valued fuzzy-set model.

### 2.3.4 HyperFuzzy Graph

HyperFuzzy graphs lift HyperFuzzy membership assignments from sets to graph structures by assigning nonempty subsets of $[0, 1]$ to vertices and edges, subject to a natural compatibility condition (cf. [262]). Note that HyperFuzzy graphs are closely related to interval-valued fuzzy graphs [263, 264].

**Definition 2.3.51** (HyperFuzzy graph)**.** [262] Let $G^* = (V, E)$ be a finite simple undirected graph. Write $\mathcal{P}^*([0, 1]) = \mathcal{P}([0, 1]) \setminus \{\emptyset\}$. A *HyperFuzzy graph* on $G^*$ is a quadruple

$$G_H = (V, E, \widetilde{\sigma}, \widetilde{\mu}),$$

where

$$\widetilde{\sigma} : V \to \mathcal{P}^*([0, 1]), \qquad \widetilde{\mu} : E \to \mathcal{P}^*([0, 1]),$$

and the following compatibility condition holds for every edge $uv \in E$:

$$\sup \widetilde{\mu}(uv) \ \leq\ \min\bigl\{\sup \widetilde{\sigma}(u),\ \sup \widetilde{\sigma}(v)\bigr\}.$$

**Remark 2.3.52.** The condition in Definition 2.3.51 ensures that, regardless of which admissible edge-membership value in $\widetilde{\mu}(uv)$ is realized, it never exceeds the maximal admissible vertex-membership levels available at the endpoints $u$ and $v$. Other equivalent bounding conventions (e.g., using infima, or requiring $\widetilde{\mu}(uv) \subseteq [0, \min\{\sup \widetilde{\sigma}(u), \sup \widetilde{\sigma}(v)\}]$) are also common.

**Example 2.3.53** (A small HyperFuzzy graph)**.** Let $G^* = (V, E)$ be the path on three vertices

$$V = \{u, v, w\}, \qquad E = \{uv, vw\}.$$

Define a HyperFuzzy vertex-membership map $\widetilde{\sigma} : V \to \mathcal{P}^*([0, 1])$ by

$$\widetilde{\sigma}(u) = \{0.6, 0.8\}, \qquad \widetilde{\sigma}(v) = \{0.4, 0.7\}, \qquad \widetilde{\sigma}(w) = \{0.2, 0.5\},$$

and an edge-membership map $\widetilde{\mu} : E \to \mathcal{P}^*([0, 1])$ by

$$\widetilde{\mu}(uv) = \{0.3, 0.6\}, \qquad \widetilde{\mu}(vw) = \{0.1, 0.5\}.$$

Then $\sup \widetilde{\sigma}(u) = 0.8$, $\sup \widetilde{\sigma}(v) = 0.7$, $\sup \widetilde{\sigma}(w) = 0.5$, and

$$\sup \widetilde{\mu}(uv) = 0.6 \leq \min\{0.8, 0.7\} = 0.7, \qquad \sup \widetilde{\mu}(vw) = 0.5 \leq \min\{0.7, 0.5\} = 0.5.$$

Hence the compatibility condition of Definition 2.3.51 holds, and

$$G_H = (V, E, \widetilde{\sigma}, \widetilde{\mu})$$

is a HyperFuzzy graph on $G^*$.

We illustrate this example in Figure 2.7.

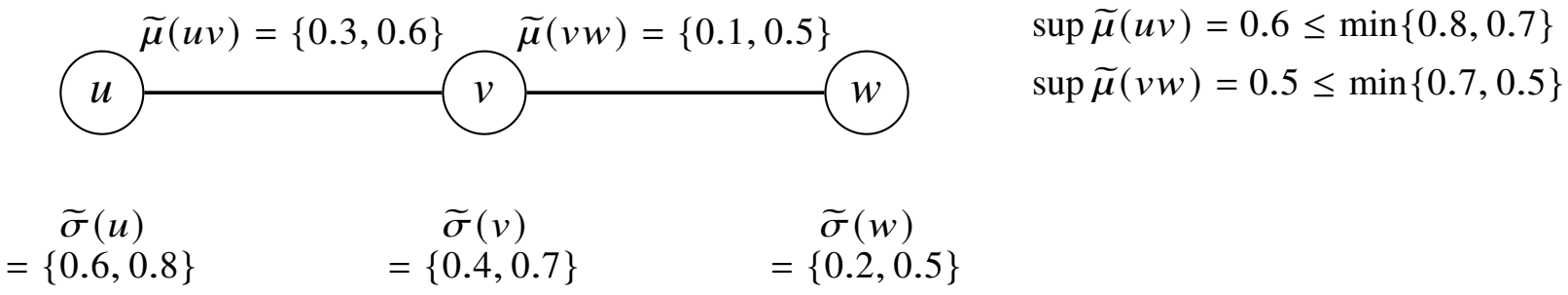

Figure 2.7: A HyperFuzzy graph $G_H = (V, E, \widetilde{\sigma}, \widetilde{\mu})$ (Example 2.3.53).

**Proposition 2.3.54** (Reduction to a fuzzy graph)**.** *Every fuzzy graph can be regarded as a HyperFuzzy graph.*

*Proof.* Let $G = (V, E, \sigma, \mu)$ be a fuzzy graph on $G^* = (V, E)$, so $\sigma : V \to [0, 1]$, $\mu : E \to [0, 1]$, and $\mu(uv) \leq \min\{\sigma(u), \sigma(v)\}$ for all $uv \in E$. Define

$$\widetilde{\sigma}(v) := \{\sigma(v)\} \quad (v \in V), \qquad \widetilde{\mu}(e) := \{\mu(e)\} \quad (e \in E).$$

Then $\sup \widetilde{\mu}(uv) = \mu(uv)$ and $\sup \widetilde{\sigma}(u) = \sigma(u)$, hence the compatibility condition in Definition 2.3.51 holds. Therefore $G$ embeds as a special case of a HyperFuzzy graph. □

### 2.3.5 SuperHyperFunction

The notion of a *SuperHyperFunction* extends the classical concept of a function by allowing both the domain and the codomain to live at (possibly different) levels of an iterated powerset hierarchy (cf. [265–274]).

**Definition 2.3.55** (Iterated powerset)**.** Let $S \neq \emptyset$ be a set. Define the iterated powersets recursively by

$$\mathcal{P}^0(S) := S, \qquad \mathcal{P}^{k+1}(S) := \mathcal{P}\big(\mathcal{P}^k(S)\big) \quad (k \in \mathbb{N}_0),$$

and write $\mathcal{P}^*(X) := \mathcal{P}(X) \setminus \{\emptyset\}$ for the family of nonempty subsets of a set $X$. For convenience, set

$$\mathcal{P}^*_k(S) := \mathcal{P}^*\big(\mathcal{P}^k(S)\big) \quad (k \in \mathbb{N}_0),$$

so $\mathcal{P}^*_k(S)$ consists of all nonempty elements of $\mathcal{P}^{k+1}(S)$.

**Definition 2.3.56** (SuperHyperFunction)**.** [265, 274] Let $S \neq \emptyset$ and let $r, n \in \mathbb{N}_0$. A *SuperHyperFunction* (of type $(r, n)$ over $S$) is a mapping

$$\mathrm{SHF} : \; \mathcal{P}^*_r(S) \longrightarrow \mathcal{P}^*_n(S).$$

Thus, for every input $A \in \mathcal{P}^*_r(S)$, the value $\mathrm{SHF}(A)$ is a nonempty element of the $n$-th iterated powerset level, i.e., $\mathrm{SHF}(A) \in \mathcal{P}^*\big(\mathcal{P}^n(S)\big)$.

**Special cases.**

(i) If $r = n = 0$, then $\mathrm{SHF} : \mathcal{P}^*(S) \to \mathcal{P}^*(S)$ maps nonempty subsets of $S$ to nonempty subsets of $S$.

(ii) If one further restricts the domain to singletons and the codomain to singletons, then SHF induces an ordinary function $f : S \to S$ via $f(x)$ defined by $\mathrm{SHF}(\{x\}) = \{f(x)\}$.

**Remark 2.3.57** (Level interpretation)**.** An element of $\mathcal{P}^*_r(S) = \mathcal{P}^*(\mathcal{P}^r(S))$ is a *nonempty family* of objects drawn from the $r$-th level $\mathcal{P}^r(S)$. Hence a SuperHyperFunction can be viewed as a “higher-order” function whose inputs and outputs are nested set-objects located at prescribed powerset levels.

### 2.3.6 Hypercube and Hypersphere

Hypercubes and hyperspheres are two canonical families of high-dimensional geometric objects. The *n-hypercube* (or *n-cube*) is the $n$-dimensional analogue of a square/cube, while the *n-sphere* is the locus of points at a fixed Euclidean distance from a center (cf. [275–277]).

**Definition 2.3.58** ($n$-Hypercube)**.** [275, 276] Fix an integer $n \geq 0$. The *n-dimensional (unit) hypercube* is the subset

$$Q_n \;:=\; [0,1]^n \;\subseteq\; \mathbb{R}^n.$$

Its *vertex set* is

$$V(Q_n) \;:=\; \{0,1\}^n,$$

and its *(1-skeleton) edge set* is

$$E(Q_n) \;:=\; \big\{\{u,v\} \subseteq \{0,1\}^n : \; u \neq v, \; \|u - v\|_1 = 1\big\},$$

equivalently, $u$ and $v$ are adjacent iff they differ in exactly one coordinate.

**Basic facts.**

(i) $|V(Q_n)| = 2^n$.

(ii) Each vertex has degree $n$ in the graph $(V(Q_n), E(Q_n))$.

(iii) $|E(Q_n)| = \dfrac{1}{2} \sum_{v \in V(Q_n)} \deg(v) = \dfrac{1}{2}\,(2^n)\, n = n\, 2^{n-1}$.

(iv) $Q_0$ is a single point, $Q_1$ a unit segment, $Q_2$ a unit square, and $Q_3$ a unit cube.

**Remark 2.3.59** (Recursive construction)**.** [275,276] The graph of the $n$-cube satisfies $Q_n \cong Q_{n-1} \square K_2$, where $\square$ denotes the Cartesian product. Equivalently, $Q_n$ is obtained from two disjoint copies of $Q_{n-1}$ by adding a perfect matching between corresponding vertices.

**Definition 2.3.60** ($n$-Sphere / Hypersphere)**.** [277] Fix an integer $n \geq 0$, a center $c \in \mathbb{R}^{n+1}$, and a radius $r > 0$. The *$n$-dimensional hypersphere* (or *$n$-sphere*) of radius $r$ centered at $c$ is

$$S^n(c, r) \ := \ \{ x \in \mathbb{R}^{n+1} : \ \|x - c\|_2 = r \},$$

where $\| \cdot \|_2$ denotes the Euclidean norm.

The *unit $n$-sphere* is the special case $c = 0$ and $r = 1$:

$$S^n \ := \ S^n(0, 1) \ = \ \left\{ x = (x_1, \ldots, x_{n+1}) \in \mathbb{R}^{n+1} : \ \sum_{i=1}^{n+1} x_i^2 = 1 \right\}.$$

**Remark 2.3.61** (Low-dimensional instances)**.** [277] The 0-sphere $S^0$ consists of two points $\{-1, +1\} \subset \mathbb{R}$; the 1-sphere $S^1$ is a circle in $\mathbb{R}^2$; and the 2-sphere $S^2$ is the usual sphere surface in $\mathbb{R}^3$. In general, $S^n$ is an $n$-dimensional manifold embedded in $\mathbb{R}^{n+1}$.

**Example 2.3.62** (A concrete hypersphere: a 2-sphere in $\mathbb{R}^3$)**.** Take $n = 2$, center $c = (1, -2, 0) \in \mathbb{R}^3$, and radius $r = 3$. By Definition 2.3.60 (the $n$-sphere is the locus of points at fixed Euclidean distance from a center), the corresponding hypersphere is

$$S^2(c, 3) = \{ x \in \mathbb{R}^3 : \|x - c\|_2 = 3 \} = \left\{ (x_1, x_2, x_3) \in \mathbb{R}^3 : \ (x_1 - 1)^2 + (x_2 + 2)^2 + x_3^2 = 9 \right\}.$$

For instance,

$$x^{(1)} = (1, -2, 3), \qquad x^{(2)} = (4, -2, 0), \qquad x^{(3)} = (1, 1, 0)$$

all lie on $S^2(c, 3)$, since

$$\|x^{(1)} - c\|_2 = \|(0, 0, 3)\|_2 = 3, \quad \|x^{(2)} - c\|_2 = \|(3, 0, 0)\|_2 = 3, \quad \|x^{(3)} - c\|_2 = \|(0, 3, 0)\|_2 = 3.$$

Equivalently, $S^2(c, 3)$ admits the standard spherical parametrization

$$x(\varphi, \theta) = c + 3(\sin\varphi \cos\theta, \ \sin\varphi \sin\theta, \ \cos\varphi), \qquad 0 \leq \varphi \leq \pi, \ \ 0 \leq \theta < 2\pi,$$

which satisfies $\|x(\varphi, \theta) - c\|_2 = 3$ for all $(\varphi, \theta)$.

### 2.3.7 Hypersets and SuperHypersets in Set Theory

Nested collections and higher-order aggregations arise naturally in many mathematical and data-oriented settings. Motivated by the analogy with hypergraphs and superhypergraphs, we adopt the following *well-founded* (classical) set-theoretic terminology: a *hyperset* is a family of subsets of a universe, and a *superhyperset* is a family of such families. For uniformity across levels, we also record an iterated-powerset formulation.

**Definition 2.3.63** (Iterated powerset)**.** Let $U$ be a set. Define $\mathcal{P}^0(U) := U$ and, for each integer $k \geq 0$,

$$\mathcal{P}^{k+1}(U) \ := \ \mathcal{P}\big(\mathcal{P}^k(U)\big).$$

Thus $\mathcal{P}^1(U) = \mathcal{P}(U)$ and $\mathcal{P}^2(U) = \mathcal{P}(\mathcal{P}(U))$.

**Definition 2.3.64** (Hyperset)**.** (cf. [278]) Let $U$ be a nonempty set. A *hyperset on $U$* is any set

$$H \subseteq \mathcal{P}(U) = \mathcal{P}^1(U).$$

Equivalently, $H$ is a (crisp) family of subsets of $U$.

**Remark 2.3.65** (Terminology warning)**.** In parts of the non-well-founded set theory literature (e.g., under Aczel's Anti-Foundation Axiom), the word *hyperset* may refer to sets admitting membership cycles such as $x \in x$. Definition 2.3.64 is *well-founded*: every element of $H$ is an ordinary subset of $U$.

**Definition 2.3.66** (Superhyperset)**.** (cf. [278]) Let $U$ be a nonempty set. A *superhyperset on* $U$ is any set

$$SH \subseteq \mathcal{P}\big(\mathcal{P}(U)\big) = \mathcal{P}^2(U).$$

Equivalently, $SH$ is a family of hypersets on $U$, i.e., a family of families of subsets of $U$.

**Remark 2.3.67** (Superset vs. superhyperset)**.** The term *superset* is standard and unrelated to Definition 2.3.66: for sets $A, B$, one says that $B$ is a *superset* of $A$ if $A \subseteq B$ (equivalently $B \supseteq A$). In contrast, a *superhyperset* is a second-level family $SH \subseteq \mathcal{P}(\mathcal{P}(U))$.

**Definition 2.3.68** (Level-$n$ hyperset)**.** Let $U$ be a nonempty set and let $n \geq 1$ be an integer. A *level-n hyperset on* $U$ is any set

$$H^{(n)} \subseteq \mathcal{P}^n(U).$$

Thus level-1 hypersets are precisely hypersets (Definition 2.3.64), and level-2 hypersets are precisely superhypersets (Definition 2.3.66).

### 2.3.8 SuperhyperPoset

Partially ordered sets (posets) provide a foundational relational structure [279,280]. By changing the carrier to a hyperset or to a higher iterated-powerset level, we obtain "hyper" and "superhyper" variants.

**Definition 2.3.69** (Poset)**.** A *partially ordered set* (poset) is a pair $(P, \leq)$ where $P$ is a set and $\leq$ is a binary relation on $P$ satisfying, for all $x, y, z \in P$,

$$x \leq x \quad \text{(reflexive)},$$

$$(x \leq y \;\&\; y \leq x) \Rightarrow x = y \quad \text{(antisymmetric)},$$

$$(x \leq y \;\&\; y \leq z) \Rightarrow x \leq z \quad \text{(transitive)}.$$

**Definition 2.3.70** (Hyperposet)**.** Let $U$ be a nonempty set. A *hyperposet on* $U$ is a poset $(H, \leq)$ whose carrier $H$ is a hyperset on $U$, i.e.,

$$H \subseteq \mathcal{P}(U),$$

and $\leq$ is a partial order on $H$ in the sense of Definition 2.3.69.

**Definition 2.3.71** ($n$-SuperHyperPoset)**.** Let $U$ be a nonempty set and let $n \geq 0$ be an integer. An *n-SuperHyperPoset on* $U$ is a poset $(P_n, \leq)$ such that

$$P_n \subseteq \mathcal{P}^n(U)$$

(with $\mathcal{P}^n$ as in Definition 2.3.63), and $\leq$ is a partial order on $P_n$.

**Remark 2.3.72** (Special cases)**.**

$$n = 0 : \; P_0 \subseteq U \text{ (posets on objects of } U\text{)},$$

$$n = 1 : \; P_1 \subseteq \mathcal{P}(U) \text{ (hyperposets)},$$

$$n = 2 : \; P_2 \subseteq \mathcal{P}(\mathcal{P}(U)) \text{ (posets on hypersets; carriers are superhypersets)}.$$

In particular, every hyperposet is exactly a 1-SuperHyperPoset (up to renaming the carrier).

**Definition 2.3.73** (Canonical singleton embeddings)**.** Let $U$ be a set. For each $k \geq 0$ define $\iota_k : \mathcal{P}^k(U) \to \mathcal{P}^{k+1}(U)$ by

$$\iota_k(x) := \{x\}.$$

For integers $0 \leq m < n$, define the iterated embedding

$$\iota_{m\to n} \;:=\; \iota_{n-1} \circ \cdots \circ \iota_m : \; \mathcal{P}^m(U) \longrightarrow \mathcal{P}^n(U).$$

**Lemma 2.3.74** (Injectivity)**.** *For every $k \geq 0$, the map $\iota_k$ is injective. Consequently, $\iota_{m\to n}$ is injective for all $0 \leq m < n$.*

*Proof.* If $\iota_k(x) = \iota_k(y)$, then $\{x\} = \{y\}$, hence $x = y$. Composition of injective maps is injective. □

**Proposition 2.3.75** (Induced lower-level posets via embedded copies)**.** *Let $U$ be a nonempty set, let $n \geq 1$, and let $(P_n, \leq)$ be an $n$-SuperHyperPoset on $U$. Fix an integer $m$ with $0 \leq m < n$, and let $A \subseteq \mathcal{P}^m(U)$ be any set such that*

$$\iota_{m\to n}(A) \subseteq P_n.$$

*Define a relation $\preceq$ on $A$ by pullback:*

$$x \preceq y \quad :\Longleftrightarrow \quad \iota_{m\to n}(x) \leq \iota_{m\to n}(y) \qquad (x, y \in A).$$

*Then $(A, \preceq)$ is a poset. Moreover, $\iota_{m\to n} : (A, \preceq) \to (P_n, \leq)$ is an order-embedding, and $\iota_{m\to n}(A)$ is a subposet of $(P_n, \leq)$.*

*Proof.* Reflexivity and transitivity of $\preceq$ follow immediately from reflexivity and transitivity of $\leq$. For antisymmetry, assume $x \preceq y$ and $y \preceq x$. Then $\iota_{m\to n}(x) \leq \iota_{m\to n}(y)$ and $\iota_{m\to n}(y) \leq \iota_{m\to n}(x)$, hence $\iota_{m\to n}(x) = \iota_{m\to n}(y)$ by antisymmetry of $\leq$. By injectivity of $\iota_{m\to n}$ (Lemma 2.3.74), we conclude $x = y$. Thus $(A, \preceq)$ is a poset.

Finally, by definition of $\preceq$, for $x, y \in A$ one has $x \preceq y \iff \iota_{m\to n}(x) \leq \iota_{m\to n}(y)$, so $\iota_{m\to n}$ is an order-embedding, and $\iota_{m\to n}(A)$ is a subposet of $(P_n, \leq)$. □

### 2.3.9 HyperStructure and SuperHyperStructure

The concepts discussed so far can be expressed within the frameworks of HyperStructure and SuperHyperStructure. A *Hyperstructure* is organized around the powerset and serves as a vehicle for modeling relations among elements of a set [281–283]. Owing to its flexibility, the hyperstructure framework has been investigated across several areas, including mathematics and chemistry [284–288]. A *Superhyperstructure* advances this idea by utilizing the $n$-th powerset to encode multi-layered hierarchical interactions, thereby enabling deeper abstraction and greater structural complexity [274]. Because of this wide scope, superhyperstructures have likewise been explored in mathematics, chemistry, and related disciplines [289, 290]. Prominent instances include constructs such as the *SuperHyperGraph* [291, 292].

**Definition 2.3.76** (Classical Structure)**.** (cf. [1, 293, 294]) A *Classical Structure* consists of a nonempty set $H$ together with one or more *classical operations* satisfying specified axioms. A classical $m$-ary operation has the form

$$\#_0 : H^m \to H,$$

with $m \geq 1$. Familiar examples include the operations defining groups, rings, and fields.

**Definition 2.3.77** (Hyperoperation)**.** (cf. [295]) A *hyperoperation* on a set $S$ is a map

$$\circ : S \times S \longrightarrow \mathcal{P}(S),$$

so that combining two inputs returns a *set* of outcomes (not necessarily a singleton).

In this paper, we also consider a powerset-lifted variant where the carrier itself is $\mathcal{P}(S)$.

**Definition 2.3.78** (Powerset-based hyperstructure)**.** (cf. [1, 296]) Let $S$ be a nonempty set and let $\mathcal{P}(S)$ denote its powerset. A *(powerset-based) hyperstructure* on $S$ is a pair

$$\mathcal{H} = (\mathcal{P}(S), \circ),$$

where $\circ$ is a binary operation on $\mathcal{P}(S)$, i.e.,

$$\circ : \mathcal{P}(S) \times \mathcal{P}(S) \longrightarrow \mathcal{P}(S).$$

Equivalently, $\circ$ combines two subsets of $S$ and outputs a (possibly empty) subset of $S$. (If one wishes to exclude the empty set, one may replace $\mathcal{P}(S)$ by $\mathcal{P}^*(S) := \mathcal{P}(S) \setminus \{\emptyset\}$ and require $\circ : \mathcal{P}^*(S) \times \mathcal{P}^*(S) \to \mathcal{P}^*(S)$.)

**Definition 2.3.79** (SuperHyperOperation)**.** [1] Let $H$ be nonempty. Define recursively, for $k \geq 0$,

$$\mathcal{P}^0(H) = H, \qquad \mathcal{P}^{k+1}(H) = \mathcal{P}\big(\mathcal{P}^k(H)\big).$$

For fixed $m, n \geq 0$ and arity $s \geq 1$, an $(m, n)$-*SuperHyperOperation* is a map

$$\odot^{(m,n)} : \big(\mathcal{P}^m(H)\big)^s \longrightarrow \mathcal{P}^n(H).$$

**Definition 2.3.80** ($n$-Superhyperstructure)**.** (cf. [1, 294, 297]) An $n$-*Superhyperstructure* generalizes hyperstructures by acting on the $n$-th powerset:

$$\mathcal{SH}_n = (\mathcal{P}_n(S), \circ),$$

with $\circ$ defined on $\mathcal{P}_n(S)$.

**Definition 2.3.81** (SuperHyperStructure of order $(m, n)$)**.** (cf. [274, 298]) Let $S$ be nonempty and $m, n \geq 0$. A $(m, n)$-*SuperHyperStructure* of arity $s$ is any choice of

$$\odot^{(m,n)} : \big(\mathcal{P}^m(S)\big)^s \longrightarrow \mathcal{P}^n(S).$$

The special cases recover standard settings: $m = n = 0$ gives ordinary $s$-ary operations; $m = 0, n = 1$ yields hyperoperations; and $s = 1$ corresponds to *superhyperfunctions*.

### 2.3.10 Other Hyperconcepts and Superhyperconcepts

For reference, Table 2.4 lists several examples of "hyper-" and "superhyper-" concepts across different research areas.

| **Concept family** | **Representative examples (selected)** |
|---|---|
| Randomness / probability | **Hyperrandom variables** [299, 300], **Hyperprobabilities** [301]. |
| Graphs / networks / learning | **Hypergraph neural networks** [96–98], **Hypercomplete graphs** [302]. |
| Trees and decision structures | **Hypertrees** [303, 304], **Superhypertrees** [292], **Decision hypertrees** [305]. |
| Logic and formal systems | **Hyperlogic** [306]. |
| Algebraic structures | **Hyperalgebras** [101, 307–309], **Superhyperalgebras** [310–312], **Superhypergroups** [313], **Semihypergroups** [314, 315]. |
| Linear / vector space analogues | **Hypervector spaces** [316, 317]. |
| Functions and generalized mappings | **Hyperfunctions** [318, 319], **Superhyperfunctions** [265, 274]. |
| Order / lattice-type structures | **Hyperlattices** [320–322]. |
| Topology and hypertopology | **Hypertopologies** [323, 324], **Superhypertopologies** [1]. |
| Syntax / rewriting / grammars | **Hypergraph grammars** [325–327]. |
| Miscellaneous | **Hypertangles** [303]. |

Table 2.4: Examples of "hyper-" and "superhyper-" concepts across several areas (illustrative, not exhaustive).

# Chapter 3

# Graph Concepts

In this chapter, we will examine several graph concepts.

## 3.1 Nonstandard Real Graph

In this section we introduce *nonstandard real graphs*, which extend the usual notion of a weighted graph by allowing weights in the nonstandard real line $\mathbb{R}^*$ (including infinitesimals and unlimited/infinite values).

**Definition 3.1.1** (Nonstandard real graph)**.** Let $G^* = (V, E)$ be a finite simple undirected graph, where $V \neq \emptyset$ and $E \subseteq \binom{V}{2}$. A *nonstandard real graph* is a triple

$$G = (G^*, \sigma, \mu),$$

where $\sigma : V \to \mathbb{R}^*$ assigns a nonstandard real weight to each vertex and $\mu : E \to \mathbb{R}^*$ assigns a nonstandard real weight to each edge. Equivalently, we may write $G = (V, E, \sigma, \mu)$ and denote $\mu(\{u, v\})$ by $\mu(uv)$.

**Remark 3.1.2.** If one wishes to work with directed graphs, one may take $E \subseteq V \times V \setminus \{(v, v) : v \in V\}$ and define $\mu : E \to \mathbb{R}^*$ analogously. The undirected version above is the default in this book.

**Example 3.1.3** (A nonstandard real graph with an infinitesimal link and an unlimited penalty)**.** Work in a fixed nonstandard extension $\mathbb{R}^*$ of $\mathbb{R}$. Let $\varepsilon \in \mathbb{R}^*$ be a *positive infinitesimal* (so $0 < \varepsilon < 1/n$ for all $n \in \mathbb{N}$), and let $H \in {}^*\mathbb{N} \setminus \mathbb{N}$ be an *unlimited* (infinite) hyperinteger.

Let $V = \{v_1, v_2, v_3\}$ and let $G^* = (V, E)$ be the finite simple undirected graph with

$$E = \big\{\{v_1, v_2\},\ \{v_2, v_3\},\ \{v_1, v_3\}\big\}.$$

Define vertex-weights $\sigma : V \to \mathbb{R}^*$ and edge-weights $\mu : E \to \mathbb{R}^*$ by

$$\sigma(v_1) = 1, \qquad \sigma(v_2) = 1 + \varepsilon, \qquad \sigma(v_3) = H,$$

and

$$\mu(v_1 v_2) = \varepsilon, \qquad \mu(v_2 v_3) = 1, \qquad \mu(v_1 v_3) = H,$$

where $\mu(v_i v_j)$ abbreviates $\mu(\{v_i, v_j\})$. Then

$$G = (V, E, \sigma, \mu)$$

is a *nonstandard real graph* in the sense of Definition 3.1.1.

*Interpretation.* The edge $v_1 v_2$ has an *infinitesimal* weight $\varepsilon$, representing a negligible cost/length, whereas the direct edge $v_1 v_3$ has an *unlimited* weight $H$, representing a prohibitive cost. If one defines the $\mu$-length of a path as the sum of its edge weights, then the two candidate $v_1$-to-$v_3$ routes satisfy

$$\mu(v_1 v_3) = H, \qquad \mu(v_1 v_2) + \mu(v_2 v_3) = \varepsilon + 1 = 1 + \varepsilon,$$

so the indirect route $v_1$–$v_2$–$v_3$ has finite length $1 + \varepsilon$ while the direct route has unlimited length $H$. In particular, the (finite) distance $d_\mu(v_1, v_3)$ equals $1 + \varepsilon$, and its standard part is $\mathrm{st}(d_\mu(v_1, v_3)) = 1$.

**Theorem 3.1.4** (Reduction to a standard weighted graph)**.** *Let $G = (V, E, \sigma, \mu)$ be a nonstandard real graph. If $\sigma(V) \subseteq \mathbb{R}$ and $\mu(E) \subseteq \mathbb{R}$ (i.e., all vertex and edge weights are standard reals), then G is precisely a standard real-weighted graph.*

*Proof.* Under the assumption $\sigma(V) \cup \mu(E) \subseteq \mathbb{R}$, the maps $\sigma$ and $\mu$ take values in the embedded copy of $\mathbb{R} \subseteq \mathbb{R}^*$. Hence the structure $(V, E, \sigma, \mu)$ coincides with the usual definition of a weighted graph with real-valued vertex/edge weights. No infinitesimal or unlimited weights occur, so the nonstandard extension plays no role in this case. □

## 3.2 Pentapartitioned neutrosophic offgraph/overgraph/undergraph

The definitions of quadripartitioned neutrosophic offgraph/overgraph/undergraph and pentapartitioned neutrosophic offgraph/overgraph/undergraph are outlined as follows.

**Definition 3.2.1** (Quadripartitioned Neutrosophic OverGraph (QNOvG))**.** A *Quadripartitioned Neutrosophic OverGraph* is a quadripartitioned neutrosophic graph in which membership degrees may exceed 1. For each vertex $v \in V$ and edge $e \in E$, the membership values are given by:

$$T(v) \in [0, \Omega_T], \quad C(v) \in [0, \Omega_C], \quad U(v) \in [0, \Omega_U], \quad F(v) \in [0, \Omega_F],$$

where $\Omega_T, \Omega_C, \Omega_U, \Omega_F > 1$ are the respective overlimits. The sum of the membership degrees satisfies:

$$0 \le T(v) + C(v) + U(v) + F(v) \le \Omega_T + \Omega_C + \Omega_U + \Omega_F.$$

**Definition 3.2.2** (Quadripartitioned Neutrosophic UnderGraph (QNUdG))**.** A *Quadripartitioned Neutrosophic UnderGraph* is a quadripartitioned neutrosophic graph where membership degrees may be less than 0. Specifically, for each vertex $v \in V$ and edge $e \in E$, we have:

$$T(v) \in [\Psi_T, 1], \quad C(v) \in [\Psi_C, 1], \quad U(v) \in [\Psi_U, 1], \quad F(v) \in [\Psi_F, 1],$$

where $\Psi_T, \Psi_C, \Psi_U, \Psi_F < 0$ are underlimits. The sum of the membership degrees satisfies:

$$\Psi_T + \Psi_C + \Psi_U + \Psi_F \le T(v) + C(v) + U(v) + F(v) \le 4.$$

**Definition 3.2.3** (Quadripartitioned Neutrosophic OffGraph (QNOfG))**.** A *Quadripartitioned Neutrosophic OffGraph* is a quadripartitioned neutrosophic graph where membership degrees may exceed 1 or fall below 0. For each vertex $v \in V$ and edge $e \in E$:

$$T(v) \in [\Psi_T, \Omega_T], \quad C(v) \in [\Psi_C, \Omega_C], \quad U(v) \in [\Psi_U, \Omega_U], \quad F(v) \in [\Psi_F, \Omega_F],$$

where $\Psi_i < 0$ and $\Omega_i > 1$ for $i \in \{T, C, U, F\}$. The sum satisfies:

$$\Psi_T + \Psi_C + \Psi_U + \Psi_F \le T(v) + C(v) + U(v) + F(v) \le \Omega_T + \Omega_C + \Omega_U + \Omega_F.$$

**Definition 3.2.4** (Pentapartitioned Neutrosophic OverGraph (PNOvG))**.** A *Pentapartitioned Neutrosophic OverGraph* allows membership degrees to exceed 1. For each vertex $v \in V$ and edge $e \in E$:

$$T(v) \in [0, \Omega_T], \quad C(v) \in [0, \Omega_C], \quad R(v) \in [0, \Omega_R], \quad U(v) \in [0, \Omega_U], \quad F(v) \in [0, \Omega_F].$$

The total satisfies:

$$0 \le T(v) + C(v) + R(v) + U(v) + F(v) \le \sum_i \Omega_i.$$

**Definition 3.2.5** (Pentapartitioned Neutrosophic UnderGraph (PNUdG))**.** A *Pentapartitioned Neutrosophic UnderGraph* allows membership degrees to be less than 0. For each vertex $v \in V$ and edge $e \in E$:

$$T(v) \in [\Psi_T, 1], \quad C(v) \in [\Psi_C, 1], \quad R(v) \in [\Psi_R, 1], \quad U(v) \in [\Psi_U, 1], \quad F(v) \in [\Psi_F, 1],$$

where $\Psi_T, \Psi_C, \Psi_R, \Psi_U, \Psi_F < 0$. The total satisfies:

$$\sum_i \Psi_i \le T(v) + C(v) + R(v) + U(v) + F(v) \le 5.$$

**Definition 3.2.6** (Pentapartitioned Neutrosophic OffGraph (PNOfG))**.** A *Pentapartitioned Neutrosophic OffGraph* allows for both overlimits and underlimits. For each vertex $v \in V$ and edge $e \in E$:

$$T(v) \in [\Psi_T, \Omega_T], \quad C(v) \in [\Psi_C, \Omega_C], \quad R(v) \in [\Psi_R, \Omega_R], \quad U(v) \in [\Psi_U, \Omega_U], \quad F(v) \in [\Psi_F, \Omega_F].$$

The sum satisfies:

$$\sum_i \Psi_i \leq T(v) + C(v) + R(v) + U(v) + F(v) \leq \sum_i \Omega_i.$$

**Theorem 3.2.7.** *A quadripartitioned neutrosophic OffGraph can generalize both the quadripartitioned neutrosophic UnderGraph and the quadripartitioned neutrosophic OverGraph.*

*Proof.* The proof follows similarly to the case of Neutrosophic structures. □

**Theorem 3.2.8.** *Any quadripartitioned neutrosophic overgraph (undergraph, offgraph) can be transformed into a standard neutrosophic overgraph (undergraph, offgraph), preserving its structural properties.*

*Proof.* Let $G = (V, E)$ be a quadripartitioned neutrosophic overgraph. Define the standard graph $G' = (V, E)$ with transformed membership degrees for each vertex $v$ and edge $e$:

$$\sigma'(v) = (T'(v), I'(v), F'(v)),$$

where:

$$T'(v) = \frac{T(v) + C(v)}{2}, \quad I'(v) = U(v), \quad F'(v) = F(v).$$

This transformation maintains the overlimit, underlimit, or both, as required, ensuring $G'$ retains the essential properties of $G$. □

**Theorem 3.2.9.** *A pentapartitioned neutrosophic OffGraph can generalize both the pentapartitioned neutrosophic UnderGraph and the pentapartitioned neutrosophic OverGraph.*

*Proof.* The proof follows similarly to the case of Neutrosophic structures. □

**Theorem 3.2.10.** *Any pentapartitioned neutrosophic overgraph (undergraph, offgraph) can be transformed into a quadripartitioned neutrosophic overgraph (undergraph, offgraph), preserving its structural properties.*

*Proof.* Let $G = (V, E)$ be a pentapartitioned neutrosophic overgraph. Define the quadripartitioned graph $G' = (V, E)$ with membership values:

$$\sigma'(v) = (T'(v), C'(v), U'(v), F'(v)),$$

where:

$$T'(v) = T(v), \quad C'(v) = C(v), \quad U'(v) = U(v) + R(v), \quad F'(v) = F(v).$$

This preserves both the structure and properties, effectively reducing the pentapartition to a quadripartition while maintaining overlimits and underlimits. □

## 3.3 Plithogenic OffGraph / OverGraph / UnderGraph

In this section we formalize three "offset" variants of plithogenic structures—*over*, *under*, and *off*— by enlarging the codomain of the degree-of-appurtenance map beyond the standard unit interval. The graph versions are obtained by decorating vertices and edges with such plithogenic degrees.

**Definition 3.3.1** (Plithogenic overset / underset / offset)**.** Let $S$ be a nonempty universe and let $P \subseteq S$ be nonempty. Fix an attribute $v$ with a nonempty set $P_v$ of its possible values, and fix integers $s, t \geq 1$. Let $\Psi_v < 0 < 1 < \Omega_v$ be real bounds.

A *plithogenic v-structure* on $P$ is a tuple

$$PS = (P,\ v,\ P_v,\ pdf,\ pCF),$$

where

$$pdf:\ P \times P_v \longrightarrow I_v^s, \qquad pCF:\ P_v \times P_v \longrightarrow [0,1]^t,$$

and $I_v$ is an interval (specified below). The map $pdf$ is the *degree-of-appurtenance function (DAF)* and $pCF$ is a *degree-of-contradiction function (DCF).*

**(i) Overset.** $PS$ is called a *plithogenic overset* (with respect to $v$) if

$$I_v = [0, \Omega_v] \quad (\Omega_v > 1).$$

**(ii) Underset.** $PS$ is called a *plithogenic underset* (with respect to $v$) if

$$I_v = [\Psi_v, 1] \quad (\Psi_v < 0).$$

**(iii) Offset.** $PS$ is called a *plithogenic offset* (with respect to $v$) if

$$I_v = [\Psi_v, \Omega_v] \quad (\Psi_v < 0 < 1 < \Omega_v).$$

(Optionally, one may impose standard consistency axioms on $pCF$, e.g. symmetry $pCF(a, b) = pCF(b, a)$ and $pCF(a, a) = 0$, but these are not required for the definitions above.)

**Remark 3.3.2** (Why the contradiction range is kept)**.** The contradiction degrees $pCF(a, b)$ quantify incompatibility between attribute values and are naturally normalized to $[0, 1]$. The *offset* behavior is carried by the DAF $pdf$, which is the object whose codomain is enlarged.

**Definition 3.3.3** (Plithogenic graph with $m$ attributes)**.** Let $G^* = (V, E)$ be a finite simple undirected graph with $V \neq \emptyset$ and $E \subseteq \binom{V}{2}$. Fix integers $m \geq 1$ (number of attributes) and $s, t \geq 1$. For each attribute index $i \in \{1, \ldots, m\}$, fix:

$$A_i \text{ (an attribute)}, \qquad P_i \neq \emptyset \text{ (its value set)}, \qquad \Psi_i < 0 < 1 < \Omega_i.$$

Let $I_i$ be an interval (chosen according to the over/under/off case below), and define a DAF on *vertices and edges*

$$pdf_i:\ (V \cup E) \times P_i \longrightarrow I_i^s,$$

together with a DCF

$$pCF_i:\ P_i \times P_i \longrightarrow [0,1]^t.$$

A *plithogenic graph* is the tuple

$$G_P = \Big(G^*,\ \{(A_i, P_i, pdf_i, pCF_i, I_i)\}_{i=1}^{m}\Big).$$

**Definition 3.3.4** (Plithogenic OverGraph / UnderGraph / OffGraph)**.** Let $G_P$ be a plithogenic graph as in Definition 3.3.3.

**(i) Plithogenic OverGraph.** $G_P$ is a *Plithogenic OverGraph* if for every $i$,

$$I_i = [0, \Omega_i] \quad (\Omega_i > 1),$$

so each appurtenance vector $pdf_i(x, a)$ may exceed 1 componentwise.

**(ii) Plithogenic UnderGraph.** $G_P$ is a *Plithogenic UnderGraph* if for every $i$,

$$I_i = [\Psi_i, 1] \quad (\Psi_i < 0),$$

so each appurtenance vector $pdf_i(x, a)$ may be negative componentwise.

**(iii) Plithogenic OffGraph.** $G_P$ is a *Plithogenic OffGraph* if for every $i$,

$$I_i = [\Psi_i, \Omega_i] \quad (\Psi_i < 0 < 1 < \Omega_i),$$

so each appurtenance vector $pdf_i(x, a)$ may be below 0 or above 1 componentwise.

**Remark 3.3.5** (Scalar memberships as a special case)**.** If $s = 1$, then each $pdf_i(x, a) \in I_i$ is scalar. Moreover, if each value set $P_i$ is a singleton $P_i = \{a_i\}$, then $pdf_i$ is equivalent to a single map

$$\mu_i : V \cup E \to I_i, \qquad \mu_i(x) := pdf_i(x, a_i),$$

recovering the commonly used "one membership degree per attribute per element" representation.

**Theorem 3.3.6** (OffGraph subsumes UnderGraph and OverGraph)**.** *Every Plithogenic UnderGraph and every Plithogenic OverGraph is a special case of a Plithogenic OffGraph.*

*Proof.* Let $G_P$ be a Plithogenic OffGraph. Its defining feature is that, for each attribute $i$,

$$pdf_i : (V \cup E) \times P_i \to [\Psi_i, \Omega_i]^s \qquad (\Psi_i < 0 < 1 < \Omega_i).$$

*UnderGraph as a special case.* Fix the same lower bounds $\Psi_i < 0$ and set $\Omega_i := 1$ for all $i$. Then $[\,\Psi_i, \Omega_i\,] = [\Psi_i, 1]$, so the same definition becomes that of a Plithogenic UnderGraph.

*OverGraph as a special case.* Fix the same upper bounds $\Omega_i > 1$ and set $\Psi_i := 0$ for all $i$. Then $[\,\Psi_i, \Omega_i\,] = [0, \Omega_i]$, so the same definition becomes that of a Plithogenic OverGraph.

Thus UnderGraphs and OverGraphs arise by restricting the admissible interval of an OffGraph, hence they are special cases. □

**Definition 3.3.7** (Pentapartitioned neutrosophic OffGraph)**.** Let $G^* = (V, E)$ be a finite simple graph. Fix real bounds $\Psi < 0 < 1 < \Omega$. A *pentapartitioned neutrosophic OffGraph* on $G^*$ is a tuple

$$G_N = (G^*,\ T,\ C,\ R,\ U,\ F),$$

where

$$T, C, R, U, F : V \cup E \longrightarrow [\Psi, \Omega]$$

assign five (possibly offset) membership degrees to each vertex and edge. (If desired, one may allow componentwise bounds $[\Psi_\bullet, \Omega_\bullet]$ for each of $T, C, R, U, F$; the statements below adapt verbatim.)

**Theorem 3.3.8** (Plithogenic OffGraph ⇒ pentapartitioned neutrosophic OffGraph)**.** *Let $G_P$ be a Plithogenic OffGraph with $m = 5$ attributes and $s = 1$ (scalar appurtenance values). Assume each attribute value set $P_i$ contains a distinguished value $a_i \in P_i$. Then $G_P$ canonically induces a pentapartitioned neutrosophic OffGraph $G_N$ on the same underlying graph $G^*$ by setting, for every $x \in V \cup E$,*

$$T(x) := pdf_1(x, a_1), \quad C(x) := pdf_2(x, a_2), \quad R(x) := pdf_3(x, a_3), \quad U(x) := pdf_4(x, a_4), \quad F(x) := pdf_5(x, a_5).$$

*Moreover, the identity map on $V \cup E$ is a label-preserving isomorphism between the two decorated graphs: the adjacency/incidence structure is unchanged, and the five induced membership labels agree by construction.*

*Proof.* Because $G_P$ is a Plithogenic OffGraph, for each $i \in \{1, \dots, 5\}$ one has

$$pdf_i : (V \cup E) \times P_i \to [\Psi_i, \Omega_i] \qquad (\Psi_i < 0 < 1 < \Omega_i).$$

Fixing $a_i \in P_i$ yields a well-defined scalar label $x \mapsto pdf_i(x, a_i)$ on $V \cup E$. Define $T, C, R, U, F$ by the displayed formulas. Each is a map $V \cup E \to [\Psi_i, \Omega_i]$, hence fits Definition 3.3.7 (allowing componentwise bounds).

The underlying graph $G^* = (V, E)$ is identical in $G_P$ and $G_N$. Therefore all structural properties that depend only on $(V, E)$ (adjacency, incidence, degrees, connectivity, etc.) are preserved. Finally, the five label functions in $G_N$ are defined directly from the five attribute-labels of $G_P$, so the identity on $V \cup E$ preserves labels. Hence $G_P$ induces $G_N$ without altering the graph structure. □

**Remark 3.3.9** (Range-sum bound)**.** If $T(x) \in [\Psi_T, \Omega_T]$, $C(x) \in [\Psi_C, \Omega_C]$, $R(x) \in [\Psi_R, \Omega_R]$, $U(x) \in [\Psi_U, \Omega_U]$, and $F(x) \in [\Psi_F, \Omega_F]$, then for every $x \in V \cup E$,

$$\Psi_T + \Psi_C + \Psi_R + \Psi_U + \Psi_F \ \leq\ T(x) + C(x) + R(x) + U(x) + F(x) \ \leq\ \Omega_T + \Omega_C + \Omega_R + \Omega_U + \Omega_F,$$

which is an immediate consequence of interval arithmetic.

For reference, the relationships between the Uncertain Offset are illustrated in Figure 3.1. Offset is a concept that increases the degree of freedom for the values of intervals.

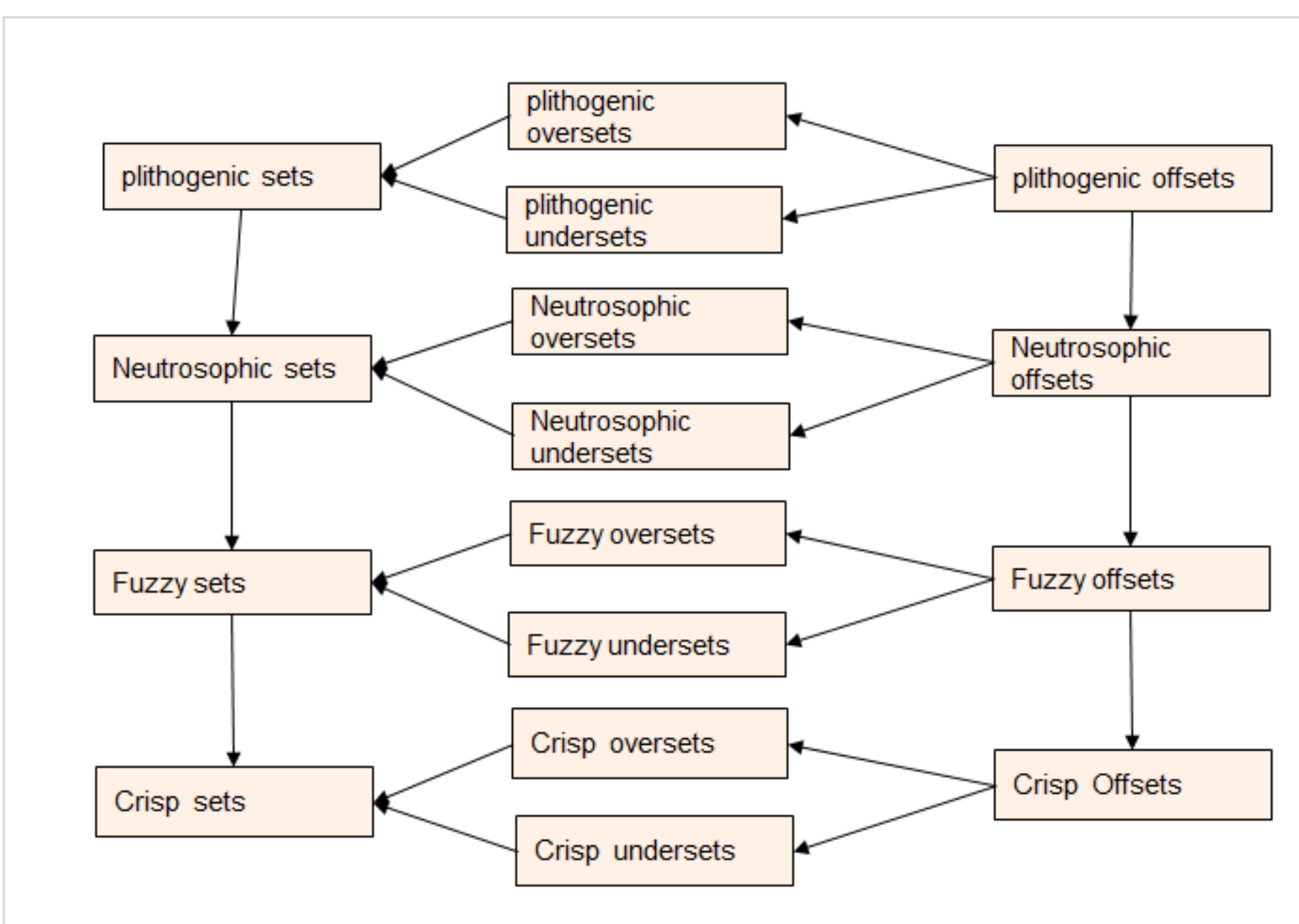

Figure 3.1: Some the Uncertain offsets Hierarchy. The set class at the origin of an arrow contains the set class at the destination of the arrow.

**Definition 3.3.10** (General Plithogenic Graph)**.** Let $G^* = (M, N)$ be a finite simple undirected graph, where $M \neq \emptyset$ is the vertex set and $N \subseteq \binom{M}{2}$ is the edge set. Fix integers $m \geq 1$ (number of attributes), $s \geq 1$ (dimension of appurtenance vectors), and $t \geq 1$. For each attribute $A_i$ ($i = 1, \dots, m$), fix nonempty sets $Ml_i$ and $Nm_i$ of admissible attribute values for vertices and edges, respectively, and fix a contradiction map

$$pCF_i : Ml_i \times Ml_i \longrightarrow [0,1]^t \qquad \text{(optionally, one may also define } pCF_i^E : Nm_i \times Nm_i \to [0,1]^t\text{).}$$

Let $I_i \subseteq \mathbb{R}$ be a nonempty interval (depending on $i$). A *general plithogenic graph* is the structure

$$G^{GP} = \Big(G^*,\ \{A_i\}_{i=1}^m,\ \{Ml_i, Nm_i\}_{i=1}^m,\ \{adf_i, bdf_i\}_{i=1}^m,\ \{pCF_i\}_{i=1}^m\Big),$$

where, for each $i$,

$$adf_i : M \times Ml_i \longrightarrow I_i^s, \qquad bdf_i : N \times Nm_i \longrightarrow I_i^s$$

are the *degree-of-appurtenance functions* (DAFs) for vertices and edges.

**Remark 3.3.11** (Scalar memberships as a special case)**.** If $s = 1$, then $adf_i(v,a) \in I_i$ and $bdf_i(e,b) \in I_i$ are scalar degrees. If moreover each value set is a singleton $Ml_i = \{a_i\}$ and $Nm_i = \{b_i\}$, then the DAFs are equivalent to the single maps

$$\mu^V_{A_i}(v) := adf_i(v,a_i) \quad (v \in M), \qquad \mu^E_{A_i}(e) := bdf_i(e,b_i) \quad (e \in N),$$

which matches the common "one degree per attribute per vertex/edge" representation.

**Definition 3.3.12** (General Plithogenic OverGraph)**.** A *general plithogenic overgraph* is a general plithogenic graph $G^{GP}$ (Definition 3.3.10) for which, for every attribute $i \in \{1,\dots,m\}$, there exists $\Omega_i > 1$ such that

$$I_i = [0,\Omega_i],$$

and hence

$$adf_i(v,a) \in [0,\Omega_i]^s \quad (\forall\, v \in M,\ \forall\, a \in Ml_i), \qquad bdf_i(e,b) \in [0,\Omega_i]^s \quad (\forall\, e \in N,\ \forall\, b \in Nm_i).$$

**Definition 3.3.13** (General Plithogenic UnderGraph)**.** A *general plithogenic undergraph* is a general plithogenic graph $G^{GP}$ for which, for every attribute $i$, there exists $\Psi_i < 0$ such that

$$I_i = [\Psi_i, 1],$$

and hence

$$adf_i(v,a) \in [\Psi_i,1]^s \quad (\forall\, v \in M,\ \forall\, a \in Ml_i), \qquad bdf_i(e,b) \in [\Psi_i,1]^s \quad (\forall\, e \in N,\ \forall\, b \in Nm_i).$$

**Definition 3.3.14** (General Plithogenic OffGraph)**.** A *general plithogenic offgraph* is a general plithogenic graph $G^{GP}$ for which, for every attribute $i$, there exist bounds $\Psi_i < 0 < 1 < \Omega_i$ such that

$$I_i = [\Psi_i, \Omega_i],$$

and hence

$$adf_i(v,a) \in [\Psi_i,\Omega_i]^s \quad (\forall\, v \in M,\ \forall\, a \in Ml_i), \qquad bdf_i(e,b) \in [\Psi_i,\Omega_i]^s \quad (\forall\, e \in N,\ \forall\, b \in Nm_i).$$

**Theorem 3.3.15** (Ordinary $\Rightarrow$ general: over/under/off)**.** *Assume an "ordinary" plithogenic OverGraph (resp. UnderGraph, OffGraph) is given in the scalar/singleton sense of Remark 3.3.11: for each attribute $A_i$ there are scalar maps $\mu^V_{A_i} : M \to I_i$ and $\mu^E_{A_i} : N \to I_i$ with $I_i = [0,\Omega_i]$ (resp. $I_i = [\Psi_i,1]$, $I_i = [\Psi_i,\Omega_i]$). Then it canonically induces a general plithogenic overgraph (resp. undergraph, offgraph).*

*Proof.* Fix $m \geq 1$ attributes $\{A_i\}_{i=1}^m$. For each $i$, set $s := 1$, choose singletons $Ml_i := \{a_i\}$ and $Nm_i := \{b_i\}$, and define

$$adf_i(v,a_i) := \mu^V_{A_i}(v) \quad (v \in M), \qquad bdf_i(e,b_i) := \mu^E_{A_i}(e) \quad (e \in N).$$

By the assumed range conditions of the ordinary Over/Under/OffGraph, these values lie in $[0,\Omega_i]$, $[\Psi_i,1]$, or $[\Psi_i,\Omega_i]$, respectively. Therefore the resulting structure satisfies Definition 3.3.12, Definition 3.3.13, or Definition 3.3.14. □

## 3.4 MultiNeutrosophic Graph

MultiNeutrosophic graphs extend (single-valued) neutrosophic graphs by allowing *multiple* truth-, indeterminacy-, and falsity-evaluations for each vertex and edge (e.g., coming from different experts, sensors, models, or criteria). Mathematically, it is most convenient to represent the multiple evaluations as *nonempty sets* of admissible values.

**Definition 3.4.1** (MultiNeutrosophic graph)**.** Let $G^* = (V,E)$ be a finite simple undirected graph, where $V \neq \emptyset$ and $E \subseteq \binom{V}{2}$. Write $\mathcal{P}^*([0,1]) = \mathcal{P}([0,1]) \setminus \{\emptyset\}$.

A *MultiNeutrosophic graph* (MNG) on $G^*$ is a tuple

$$G_{MN} = \big(V, E,\ T_V, I_V, F_V,\ T_E, I_E, F_E\big),$$

where the vertex-membership maps

$$T_V,\ I_V,\ F_V :\ V \longrightarrow \mathcal{P}^*([0,1])$$

and the edge-membership maps

$$T_E,\ I_E,\ F_E :\ E \longrightarrow \mathcal{P}^*([0,1])$$

assign to each vertex (resp. edge) a nonempty set of possible truth-, indeterminacy-, and falsity-degrees.

**(Optional admissibility bounds).** Since all membership sets lie in $[0,1]$, one automatically has

$$0 \le \inf T_V(v) \le \sup T_V(v) \le 1, \quad 0 \le \inf I_V(v) \le \sup I_V(v) \le 1, \quad 0 \le \inf F_V(v) \le \sup F_V(v) \le 1 \qquad (\forall v \in V),$$

and analogously for edges. One may additionally impose the neutrosophic-type sum bounds

$$0 \ \le\ \inf T_V(v) + \inf I_V(v) + \inf F_V(v) \ \le\ \sup T_V(v) + \sup I_V(v) + \sup F_V(v) \ \le\ 3 \qquad (\forall v \in V),$$

$$0 \ \le\ \inf T_E(e) + \inf I_E(e) + \inf F_E(e) \ \le\ \sup T_E(e) + \sup I_E(e) + \sup F_E(e) \ \le\ 3 \qquad (\forall e \in E),$$

which are also automatically satisfied because each supremum is at most 1.

**(Optional edge–vertex consistency).** To mirror the single-valued neutrosophic-graph constraints, one may require for every edge $e = uv \in E$ the set-wise bounds

$$\sup T_E(uv) \ \le\ \min\{\sup T_V(u), \sup T_V(v)\}, \qquad \sup I_E(uv) \ \le\ \min\{\sup I_V(u), \sup I_V(v)\},$$

$$\sup F_E(uv) \ \le\ \max\{\sup F_V(u), \sup F_V(v)\}.$$

(Other choices are possible, e.g. using inf or requiring inclusion into an interval determined by the endpoints.)

**Example 3.4.2** (A small MultiNeutrosophic graph)**.** Let $G^* = (V, E)$ be the path on three vertices

$$V = \{u, v, w\}, \qquad E = \{uv, vw\}.$$

Define a MultiNeutrosophic graph

$$G_{MN} = \big(V, E,\ T_V, I_V, F_V,\ T_E, I_E, F_E\big)$$

by assigning the following nonempty subsets of $[0,1]$.

**Vertex memberships.**

| $x \in V$ | $T_V(x)$ | $I_V(x)$ | $F_V(x)$ |
|---|---|---|---|
| $u$ | $\{0.70, 0.80\}$ | $\{0.10, 0.30\}$ | $\{0.20, 0.50\}$ |
| $v$ | $\{0.50, 0.60\}$ | $\{0.20, 0.40\}$ | $\{0.40, 0.60\}$ |
| $w$ | $\{0.40, 0.50\}$ | $\{0.05, 0.20\}$ | $\{0.60, 0.70\}$ |

**Edge memberships.**

| $e \in E$ | $T_E(e)$ | $I_E(e)$ | $F_E(e)$ |
|---|---|---|---|
| $uv$ | $\{0.50, 0.60\}$ | $\{0.20, 0.30\}$ | $\{0.40, 0.60\}$ |
| $vw$ | $\{0.40\}$ | $\{0.15, 0.20\}$ | $\{0.50, 0.65\}$ |

All sets are nonempty subsets of $[0,1]$, hence $G_{MN}$ is a MultiNeutrosophic graph in the sense of Definition 4.6.9.

**Verification of the optional edge–vertex consistency (sup-version).** Compute the endpoint suprema:

$$\sup T_V(u) = 0.80,\ \sup T_V(v) = 0.60,\ \sup T_V(w) = 0.50,$$

$$\sup I_V(u) = 0.30,\ \sup I_V(v) = 0.40,\ \sup I_V(w) = 0.20,$$

$$\sup F_V(u) = 0.50,\ \sup F_V(v) = 0.60,\ \sup F_V(w) = 0.70.$$

For $uv$,

$$\sup T_E(uv) = 0.60 \le \min\{0.80, 0.60\} = 0.60,$$

$$\sup I_E(uv) = 0.30 \le \min\{0.30, 0.40\} = 0.30,$$

$$\sup F_E(uv) = 0.60 \le \max\{0.50, 0.60\} = 0.60.$$

For $vw$,

$$\sup T_E(vw) = 0.40 \le \min\{0.60, 0.50\} = 0.50,$$

$$\sup I_E(vw) = 0.20 \le \min\{0.40, 0.20\} = 0.20,$$

$$\sup F_E(vw) = 0.65 \le \max\{0.60, 0.70\} = 0.70.$$

Thus the stated optional consistency constraints hold.

We illustrate this example in Figure 3.2.

$T_E = \{0.50, 0.60\}$ $T_E = \{0.40\}$
$I_E = \{0.20, 0.30\}$ $I_E = \{0.15, 0.20\}$
$u$ $F_E = \{0.40, 0.60\}$ $v$ $F_E = \{0.50, 0.65\}$ $w$

$T_V(u) = \{0.70, 0.80\}$ $T_V(v) = \{0.50, 0.60\}$ $T_V(w) = \{0.40, 0.50\}$
$I_V(u) = \{0.10, 0.30\}$ $I_V(v) = \{0.20, 0.40\}$ $I_V(w) = \{0.05, 0.20\}$
$F_V(u) = \{0.20, 0.50\}$ $F_V(v) = \{0.40, 0.60\}$ $F_V(w) = \{0.60, 0.70\}$

Figure 3.2: A MultiNeutrosophic graph $G_{MN}$ on the path $u-v-w$ (Example 3.4.2).

**Theorem 3.4.3** (Single-valued neutrosophic graphs embed into multineutrosophic graphs)**.** *Every (single-valued) neutrosophic graph is a special case of a MultiNeutrosophic graph.*

*Proof.* Let

$$G = \big(V, E,\ T_V, I_V, F_V,\ T_E, I_E, F_E\big)$$

be a (single-valued) neutrosophic graph in the sense of Definition 2.2.5, so that $T_V, I_V, F_V : V \to [0,1]$ and $T_E, I_E, F_E : E \to [0,1]$. Define set-valued maps by singleton lifts:

$$\widetilde{T}_V(v) := \{T_V(v)\}, \quad \widetilde{I}_V(v) := \{I_V(v)\}, \quad \widetilde{F}_V(v) := \{F_V(v)\} \qquad (\forall v \in V),$$

$$\widetilde{T}_E(e) := \{T_E(e)\}, \quad \widetilde{I}_E(e) := \{I_E(e)\}, \quad \widetilde{F}_E(e) := \{F_E(e)\} \qquad (\forall e \in E).$$

Each image is a nonempty subset of $[0,1]$, hence $\widetilde{T}_V, \widetilde{I}_V, \widetilde{F}_V : V \to \mathcal{P}^*([0,1])$ and $\widetilde{T}_E, \widetilde{I}_E, \widetilde{F}_E : E \to \mathcal{P}^*([0,1])$. Therefore

$$G_{MN} := \big(V, E,\ \widetilde{T}_V, \widetilde{I}_V, \widetilde{F}_V,\ \widetilde{T}_E, \widetilde{I}_E, \widetilde{F}_E\big)$$

is a MultiNeutrosophic graph (Definition 4.6.9). If one adopts the optional edge–vertex consistency constraints using suprema, they reduce to the original single-valued constraints because $\sup\{x\} = x$ for singletons. □

## 3.5 Subset-Valued Neutrosophic Graph and Single-Valued Nonstandard Neutrosophic Graph

Subset-valued neutrosophic sets and single-valued nonstandard neutrosophic sets have been introduced in the neutrosophic literature (cf. [328]). In this section we record mathematically clean formulations and extend them to graphs by assigning (neutrosophic) membership data to both vertices and edges.

**Definition 3.5.1** (Infimum and supremum)**.** Let $S \subseteq \mathbb{R}$. A real number $m \in \mathbb{R}$ is the *infimum* of $S$, written $m = \inf S$, if

(i) $m \le s$ for all $s \in S$ (i.e. $m$ is a lower bound of $S$), and

(ii) for every $\varepsilon > 0$ there exists $s \in S$ with $s < m + \varepsilon$ (i.e. no larger lower bound exists).

Equivalently, $\inf S = \sup\{x \in \mathbb{R} : x \leq s \ \forall s \in S\}$, provided the set of lower bounds is nonempty.

Similarly, $M \in \mathbb{R}$ is the *supremum* of $S$, written $M = \sup S$, if

(i) $s \leq M$ for all $s \in S$ (i.e. $M$ is an upper bound of $S$), and

(ii) for every $\varepsilon > 0$ there exists $s \in S$ with $M - \varepsilon < s$ (i.e. no smaller upper bound exists).

Equivalently, $\sup S = \inf\{x \in \mathbb{R} : s \leq x \ \forall s \in S\}$, provided the set of upper bounds is nonempty.

**Remark 3.5.2** (Empty set conventions)**.** For $S = \emptyset$, the quantities $\inf S$ and $\sup S$ are not real numbers. In the extended reals $\overline{\mathbb{R}} := \mathbb{R} \cup \{-\infty, +\infty\}$ one often sets $\inf \emptyset := +\infty$ and $\sup \emptyset := -\infty$. In our neutrosophic applications below we avoid this by *requiring* all membership subsets to be nonempty.

**Definition 3.5.3** (Subset-Valued Neutrosophic Set)**.** (cf. [328]) Let $X$ be a nonempty set. A *subset-valued neutrosophic set* (SVNS) $A$ on $X$ is a triple of maps

$$T_A,\ I_A,\ F_A :\ X \longrightarrow \mathcal{P}([0,1]) \setminus \{\emptyset\},$$

where for each $x \in X$ the sets $T_A(x), I_A(x), F_A(x) \subseteq [0,1]$ represent the possible degrees of truth, indeterminacy, and falsity, respectively, and satisfy the bound

$$0 \leq \inf T_A(x) + \inf I_A(x) + \inf F_A(x) \leq \sup T_A(x) + \sup I_A(x) + \sup F_A(x) \leq 3.$$

**Remark 3.5.4.** Since $T_A(x), I_A(x), F_A(x) \subseteq [0,1]$ are nonempty, the infimum and supremum are well-defined real numbers in $[0,1]$. The displayed inequality is therefore meaningful in $\mathbb{R}$.

To avoid ambiguities in informal monad/binad notation, we work inside a fixed hyperreal extension $^*\mathbb{R}$ and use the standard nonstandard-analysis terminology.

**Definition 3.5.5** (Infinitesimals and monads)**.** Fix a nonstandard extension $^*\mathbb{R} \supset \mathbb{R}$. Let

$$^*\mathbb{R}_{\text{inf}} := \{\varepsilon \in {}^*\mathbb{R} :\ |\varepsilon| < 1/n \text{ for all } n \in \mathbb{N}\}$$

be the set of infinitesimals, and let $^*\mathbb{R}^+_{\text{inf}} := \{\varepsilon \in {}^*\mathbb{R}_{\text{inf}} : \varepsilon > 0\}$. For $a \in \mathbb{R}$, define:

$$\mu^-(a) := \{a - \varepsilon :\ \varepsilon \in {}^*\mathbb{R}^+_{\text{inf}}\} \quad \text{(left monad of } a\text{)},$$

$$\mu^+(a) := \{a + \varepsilon :\ \varepsilon \in {}^*\mathbb{R}^+_{\text{inf}}\} \quad \text{(right monad of } a\text{)}.$$

Their "closed" variants are

$$\overline{\mu^-}(a) := \mu^-(a) \cup \{a\}, \qquad \overline{\mu^+}(a) := \mu^+(a) \cup \{a\}.$$

A (symmetric) *binad neighborhood* of $a$ is

$$\beta(a) := \mu^-(a) \cup \mu^+(a), \qquad \overline{\beta}(a) := \overline{\mu^-}(a) \cup \overline{\mu^+}(a).$$

**Remark 3.5.6** (A convenient "nonstandard unit interval")**.** For neutrosophic degrees one often wants to allow values infinitesimally below 0 or above 1. A mathematically explicit choice is the subset

$$]^-0,\ 1^+[:= \overline{\mu^+}(0)\ \cup\ [0,1]\ \cup\ \overline{\mu^-}(1)\ \subseteq {}^*\mathbb{R}.$$

This is one concrete realization of the informal notation $]-0, 1+[$.

**Definition 3.5.7** (Single-Valued Nonstandard Neutrosophic Set)**.** (cf. [328]) Let $X$ be a nonempty set and fix the nonstandard unit interval $]^-0, 1^+[ \subseteq {}^*\mathbb{R}$ as in Remark 3.5.6. A *single-valued nonstandard neutrosophic set* (SV-NoSNS) $A$ on $X$ is a triple of maps

$$T_A,\ I_A,\ F_A :\ X \longrightarrow ]^-0, 1^+[.$$

Optionally, one may impose a neutrosophic sum constraint such as

$$0\ \leq\ T_A(x) + I_A(x) + F_A(x)\ \leq\ 3 + \varepsilon \quad \text{for some fixed } \varepsilon \in {}^*\mathbb{R}^+_{\text{inf}},$$

which formalizes the informal upper bound $3^+$.

**Remark 3.5.8** (On inf/sup in the single-valued case)**.** In the single-valued case $T_A(x), I_A(x), F_A(x) \in {}^*\mathbb{R}$ are scalars, so inf/sup are unnecessary. If one insists on uniform notation, one may identify a scalar $r$ with the singleton $\{r\}$, so that $\inf\{r\} = \sup\{r\} = r$.

**Definition 3.5.9** (Subset-Valued Neutrosophic Graph)**.** Let $G = (V, E)$ be a finite simple undirected graph with $V \neq \emptyset$ and $E \subseteq \binom{V}{2}$. A *subset-valued neutrosophic graph* (SVNG) on $G$ is given by six maps

$$T_V, I_V, F_V : V \longrightarrow \mathcal{P}([0,1]) \setminus \{\emptyset\}, \qquad T_E, I_E, F_E : E \longrightarrow \mathcal{P}([0,1]) \setminus \{\emptyset\},$$

such that for every $v \in V$ and every $e \in E$,

$$0 \leq \inf T_V(v) + \inf I_V(v) + \inf F_V(v) \leq \sup T_V(v) + \sup I_V(v) + \sup F_V(v) \leq 3,$$

and

$$0 \leq \inf T_E(e) + \inf I_E(e) + \inf F_E(e) \leq \sup T_E(e) + \sup I_E(e) + \sup F_E(e) \leq 3.$$

We denote such a structure by $G_{SV} = (V, E,\ T_V, I_V, F_V,\ T_E, I_E, F_E)$.

**Definition 3.5.10** (Single-Valued Nonstandard Neutrosophic Graph)**.** Let $G = (V, E)$ be a finite simple undirected graph. Fix a nonstandard extension ${}^*\mathbb{R}$ and the nonstandard unit interval $]^-0, 1^+[ \subseteq {}^*\mathbb{R}$. A *single-valued nonstandard neutrosophic graph* (SV-NoSNG) on $G$ is given by six maps

$$T_V, I_V, F_V : V \longrightarrow ]^-0, 1^+[, \qquad T_E, I_E, F_E : E \longrightarrow ]^-0, 1^+[.$$

Optionally, one may require a uniform sum constraint: there exists an infinitesimal $\varepsilon \in {}^*\mathbb{R}^+_{\mathrm{inf}}$ such that for all $v \in V$ and $e \in E$,

$$0 \leq T_V(v) + I_V(v) + F_V(v) \leq 3 + \varepsilon, \qquad 0 \leq T_E(e) + I_E(e) + F_E(e) \leq 3 + \varepsilon.$$

We denote such a structure by $G_{NS} = (V, E,\ T_V, I_V, F_V,\ T_E, I_E, F_E)$.

**Remark 3.5.11** (Reduction to standard single-valued neutrosophic graphs)**.** If the codomains are restricted to $[0, 1] \subseteq {}^*\mathbb{R}$, then Definition 3.5.10 reduces to the usual single-valued neutrosophic graph model (with real degrees in $[0, 1]$). Likewise, if every membership subset in Definition 3.5.9 is a singleton, the SVNG reduces to a single-valued model.

The related concept of a subset-valued fuzzy set can be stated as follows.

**Definition 3.5.12** (Subset-Valued Fuzzy Set)**.** Let $X$ be a nonempty set. A *subset-valued fuzzy set* (SVFS) $A$ on $X$ is a mapping

$$\mu_A : X \longrightarrow \mathcal{P}([0,1]) \setminus \{\emptyset\},$$

where $\mathcal{P}([0,1])$ denotes the power set of $[0, 1]$. For each $x \in X$, the nonempty subset $\mu_A(x) \subseteq [0, 1]$ represents the set of admissible membership degrees of $x$ in $A$.

Equivalently, since $\mu_A(x) \subseteq [0, 1]$ is nonempty, its extrema exist and satisfy

$$0 \leq \inf \mu_A(x) \leq \sup \mu_A(x) \leq 1 \qquad (\forall x \in X).$$

**Remark 3.5.13.** The restriction $\mu_A(x) \neq \emptyset$ ensures that $\inf \mu_A(x)$ and $\sup \mu_A(x)$ are real numbers. (If empty values were allowed, one would have to work in the extended reals and adopt conventions such as $\inf \emptyset = +\infty$, $\sup \emptyset = -\infty$.)

**Theorem 3.5.14** (SVFS as a special case of SVNS)**.** *Every subset-valued fuzzy set on X canonically induces a subset-valued neutrosophic set on X.*

*Proof.* Let $A$ be an SVFS on $X$ with membership map $\mu_A : X \to \mathcal{P}([0, 1]) \setminus \{\emptyset\}$. Define three maps $T, I, F :$ $X \to \mathcal{P}([0, 1]) \setminus \{\emptyset\}$ by

$$T(x) := \mu_A(x), \qquad I(x) := \{0\}, \qquad F(x) := \{0\} \qquad (\forall x \in X).$$

Then $T(x), I(x), F(x) \subseteq [0, 1]$ are nonempty for all $x$. Moreover,

$$\inf I(x) = \sup I(x) = 0, \qquad \inf F(x) = \sup F(x) = 0,$$

and since $T(x) = \mu_A(x) \subseteq [0,1]$ is nonempty,

$$0 \le \inf T(x) \le \sup T(x) \le 1.$$

Hence, for every $x \in X$,

$$0 \le \inf T(x) + \inf I(x) + \inf F(x) = \inf \mu_A(x) \le \sup \mu_A(x) = \sup T(x) + \sup I(x) + \sup F(x) \le 1 \le 3.$$

Therefore $(T, I, F)$ satisfies the defining inequalities of a subset-valued neutrosophic set (Definition 3.5.3), so it defines an SVNS on $X$. This construction is canonical and embeds SVFSs into SVNSs by fixing indeterminacy and falsity to $\{0\}$. □

## 3.6 Neutrosophic Axial Graphs and Partner Multineutrosophic Graphs

This section presents mathematically well-typed formulations of two conceptual graph models: (i) *neutrosophic axial graphs*, whose vertices and edges are labeled by "axial" triples of subsets of a fixed universe, and (ii) *partner multineutrosophic graphs*, which compress multi-neutrosophic information on vertices/edges into a single scalar weight (a "partner" value).

**Definition 3.6.1** (Neutrosophic axial label)**.** Let $X \neq \emptyset$ be a set. A *neutrosophic axial label* on $X$ is a triple

$$\mathbf{A} = (A_0, A_1, A_2) \quad \text{with} \quad A_0, A_1, A_2 \subseteq X$$

such that

$$A_0 \cap A_1 = \emptyset \quad \text{and} \quad A_0 \cap A_2 = \emptyset.$$

(Optionally one may also impose $A_1 \cap A_2 = \emptyset$; we do not require it here.) Let $\mathcal{A}(X)$ denote the set of all such triples.

**Definition 3.6.2** (Neutrosophic axial graph)**.** Let $G = (V, E)$ be a finite simple undirected graph, where $V \neq \emptyset$ and $E \subseteq \binom{V}{2}$. Fix a universe $X \neq \emptyset$. A *neutrosophic axial graph* on $G$ (over $X$) is a pair of labeling maps

$$\lambda_V : V \longrightarrow \mathcal{A}(X), \qquad \lambda_E : E \longrightarrow \mathcal{A}(X),$$

where $\lambda_V(v) = (A_0(v), A_1(v), A_2(v))$ and $\lambda_E(e) = (B_0(e), B_1(e), B_2(e))$ satisfy the axial disjointness constraints of Definition 3.6.1 for each $v \in V$ and $e \in E$.

One may optionally impose *coherence axioms* linking edge and endpoint labels. For example, one consistent choice is: for each edge $e = \{u, v\} \in E$,

$$B_i(e) = A_i(u) \cup A_i(v) \qquad (i = 0, 1, 2),$$

together with endpoint-disjointness conditions such as

$$A_i(u) \cap A_i(v) = \emptyset \qquad (i = 0, 1, 2),$$

which ensure that the unions preserve intended separation properties.

**Remark 3.6.3.** The coherence axioms are model-dependent; Definition 3.6.2 separates the *labeling* from any particular algebraic rule relating vertex and edge labels.

**Definition 3.6.4** (Multi-neutrosophic label of type $(r, s, t)$)**.** Let $n \in \mathbb{N}^+$ and let $r, s, t \in \mathbb{N}$ satisfy $r + s + t = n$. A *multi-neutrosophic label of type* $(r, s, t)$ is a triple of finite tuples

$$\mathbf{m} = \big(\mathbf{T}; \mathbf{I}; \mathbf{F}\big) = \langle T_1, \dots, T_r;\ I_1, \dots, I_s;\ F_1, \dots, F_t \rangle$$

with all components in $[0,1]$. Let $\mathcal{M}_{r,s,t} := [0,1]^r \times [0,1]^s \times [0,1]^t$ denote the set of all such labels.

**Definition 3.6.5** (Partner aggregation)**.** Let $r, s, t \in \mathbb{N}$ with $n = r + s + t \ge 1$. Define the *partner aggregation* map

$$\mathrm{par} : \ \mathcal{M}_{r,s,t} \longrightarrow [0,1]$$

by

$$\mathrm{par}\big(\langle T_1, \dots, T_r;\ I_1, \dots, I_s;\ F_1, \dots, F_t \rangle\big) := \frac{1}{n} \left( \sum_{i=1}^{r} T_i + \sum_{j=1}^{s} I_j + \sum_{k=1}^{t} F_k \right).$$

**Definition 3.6.6** (Partner multineutrosophic graph)**.** Let $G = (V, E)$ be a finite simple undirected graph with $E \subseteq \binom{V}{2}$. Fix $r, s, t \in \mathbb{N}$ with $n = r + s + t \geq 1$. A *partner multineutrosophic graph* is a quadruple

$$G_p = (G,\ \ell_V,\ \ell_E,\ w_V,\ w_E),$$

where

$$\ell_V : V \to \mathcal{M}_{r,s,t}, \qquad \ell_E : E \to \mathcal{M}_{r,s,t},$$

assign multi-neutrosophic labels to vertices and edges, and the associated *partner weights*

$$w_V : V \to [0,1], \qquad w_E : E \to [0,1]$$

are defined by

$$w_V(v) := \mathrm{par}\big(\ell_V(v)\big) \quad (v \in V), \qquad w_E(e) := \mathrm{par}\big(\ell_E(e)\big) \quad (e \in E),$$

with par as in Definition 3.6.5.

**Remark 3.6.7** (Interpretation)**.** A partner multineutrosophic graph can be viewed as a graph with rich multi-neutrosophic labels $\ell_V, \ell_E$, together with a derived weighted graph $(V, E, w_E)$ (and optionally vertex weights $w_V$) obtained by a fixed aggregation rule. Different aggregations (e.g. weighted averages, max/min, nonlinear maps) yield different partner models.

## 3.7 Heptapartitioned Neutrosophic Graph

Heptapartitioned neutrosophic sets (HNS) extend the usual neutrosophic framework by splitting uncertainty information into seven membership components; see, e.g., [329–333]. Analogously, one may define hepta-partitioned neutrosophic graphs by assigning such seven-tuples to vertices and edges. (Octapartitioned, nona-partitioned, and multipartitioned variants can be treated similarly; cf. [334].)

**Definition 3.7.1** (Heptapartitioned Neutrosophic Set (HNS))**.** [329] Let $U$ be a nonempty set. A *heptapartitioned neutrosophic set $A$ on $U$* is specified by seven membership functions

$$T_A,\ M_A,\ C_A,\ U_A,\ I_A,\ K_A,\ F_A : U \longrightarrow [0,1],$$

where, for each $x \in U$, the components are interpreted as truth ($T_A$), relative truth ($M_A$), contradiction ($C_A$), unknown ($U_A$), ignorance ($I_A$), relative falsity ($K_A$), and falsity ($F_A$), and satisfy the (automatically true) boundedness condition

$$0 \ \leq\ T_A(x) + M_A(x) + C_A(x) + U_A(x) + I_A(x) + K_A(x) + F_A(x) \ \leq\ 7 \qquad (\forall x \in U).$$

We may write

$$A = \big\{\langle x,\ T_A(x), M_A(x), C_A(x), U_A(x), I_A(x), K_A(x), F_A(x)\rangle : \ x \in U\big\}.$$

**Remark 3.7.2.** Since each component lies in $[0,1]$, the upper bound $\leq 7$ is redundant but harmless. Some authors additionally impose application-driven constraints (e.g. $\leq 1$ or $\leq 3$) to normalize information; such constraints can be added as optional axioms without changing the graphization below.

**Theorem 3.7.3** (HNS embeds into a plithogenic set with $s = 7$)**.** *Fix $U \neq \emptyset$. Every HNS $A$ on $U$ canonically induces a plithogenic set $PS = (P, v, P_v, pdf, pCF)$ with $P = U$, $s = 7$, and $t = 1$ in the following sense: there exist choices of a single attribute $v$, a value set $P_v$ containing a distinguished value $a_0$, and maps $pdf, pCF$ such that*

$$pdf(x, a_0) = \big(T_A(x), M_A(x), C_A(x), U_A(x), I_A(x), K_A(x), F_A(x)\big) \in [0,1]^7 \quad (\forall x \in U).$$

*Proof.* Let $A$ be an HNS on $U$ with membership maps as in Definition 3.7.1. Define a plithogenic set $PS = (P, v, P_v, pdf, pCF)$ by taking $P := U$, choosing any attribute symbol $v$, letting $P_v := \{a_0\}$ be a singleton value set, and defining

$$pdf : P \times P_v \to [0,1]^7, \qquad pdf(x, a_0) := \big(T_A(x), M_A(x), C_A(x), U_A(x), I_A(x), K_A(x), F_A(x)\big).$$

Choose any $pCF : P_v \times P_v \to [0,1]$ (necessarily $pCF(a_0, a_0) \in [0,1]$); for definiteness set $pCF(a_0, a_0) = 0$. Then $PS$ is a well-defined plithogenic set with $s = 7$ and $t = 1$, and the displayed identity holds by construction. □

**Remark 3.7.4.** The theorem is an *embedding* statement: HNS-data can be encoded inside a plithogenic set by using a 7-vector DAF. This is the precise sense in which plithogenic sets (with sufficiently large $s$) generalize heptapartitioned sets.

**Definition 3.7.5** (Heptapartitioned Neutrosophic Graph (HNG))**.** Let $G = (V, E)$ be a finite simple undirected graph with $V \neq \emptyset$ and $E \subseteq \binom{V}{2}$. A *heptapartitioned neutrosophic graph* (HNG) on $G$ is a tuple

$$G_H = (G,\ T,\ M,\ C,\ U,\ I,\ K,\ F),$$

where each membership component is a function on vertices and edges,

$$T, M, C, U, I, K, F :\ V \cup E \longrightarrow [0, 1].$$

Equivalently, each $x \in V \cup E$ is labeled by the 7-tuple

$$\big(T(x), M(x), C(x), U(x), I(x), K(x), F(x)\big) \in [0, 1]^7.$$

**(Optional edge–vertex coherence).** One may additionally require that for every edge $e = \{u, v\} \in E$,

$$T(e) \leq \min\{T(u), T(v)\}, \quad M(e) \leq \min\{M(u), M(v)\}, \quad C(e) \leq \min\{C(u), C(v)\},$$

$$U(e) \geq \max\{U(u), U(v)\}, \quad I(e) \geq \max\{I(u), I(v)\}, \quad K(e) \geq \max\{K(u), K(v)\}, \quad F(e) \geq \max\{F(u), F(v)\}.$$

(These monotonicity directions are model-dependent; they are included here as a consistent, explicit option.)

**Remark 3.7.6.** As in the set case, since each component lies in $[0, 1]$, one always has $0 \leq T(x) + M(x) + C(x) + U(x) + I(x) + K(x) + F(x) \leq 7$ for all $x \in V \cup E$. If a normalized sum constraint is desired, it should be stated as an additional axiom.

**Theorem 3.7.7** (Pentapartitioned graphs embed into heptapartitioned graphs)**.** *Every pentapartitioned neutrosophic graph (PNG) $G_P = (G,\ T,\ C,\ R,\ U,\ F)$ canonically induces a heptapartitioned neutrosophic graph $G_H$ on the same underlying graph G by*

$$T_H := T, \qquad C_H := C, \qquad U_H := U, \qquad F_H := F, \qquad I_H := R, \qquad M_H := 0, \qquad K_H := 0,$$

*where* 0 *denotes the constant-zero map on $V \cup E$.*

*Proof.* Let $G_P$ be a PNG. Define seven maps on $V \cup E$ by the displayed assignments. Since each PNG component is $[0, 1]$-valued, the resulting maps are also $[0, 1]$-valued, and thus $(G,\ T_H, M_H, C_H, U_H, I_H, K_H, F_H)$ satisfies Definition 3.7.5. The underlying graph $(V, E)$ is unchanged, so all structural properties depending only on $(V, E)$ are preserved. Finally, the PNG labels are recovered as a coordinate projection of the HNG labels, showing that PNG is a special case. □

**Theorem 3.7.8** (HNG as a general plithogenic graph with $s = 7$)**.** *Let $G_H = (G,\ T, M, C, U, I, K, F)$ be an HNG. Then $G_H$ canonically induces a general plithogenic graph (in the sense of Definition 3.3.10) with $m = 1$ attribute, $s = 7$, and $t = 1$.*

*Proof.* Let $G = (V, E)$ be the underlying graph of $G_H$. Define a general plithogenic graph as follows: take a single attribute symbol $A$ ($m = 1$), choose singleton value sets $Ml = \{a_0\}$ for vertices and $Nm = \{b_0\}$ for edges, set $s = 7$, and define

$$adf :\ V \times Ml \to [0, 1]^7, \qquad adf(v, a_0) := \big(T(v), M(v), C(v), U(v), I(v), K(v), F(v)\big),$$

$$bdf :\ E \times Nm \to [0, 1]^7, \qquad bdf(e, b_0) := \big(T(e), M(e), C(e), U(e), I(e), K(e), F(e)\big).$$

Choose any contradiction map $pCF : Ml \times Ml \to [0, 1]$ (e.g. $pCF(a_0, a_0) = 0$), and similarly for edges if desired. Then the resulting structure satisfies Definition 3.3.10 with $s = 7$. By construction, its vertex/edge DAF vectors coincide with the HNG 7-tuples, so it encodes $G_H$. □

For reference, the relationships between the graphs are illustrated in Figure 3.3. (cf. [91, 181])

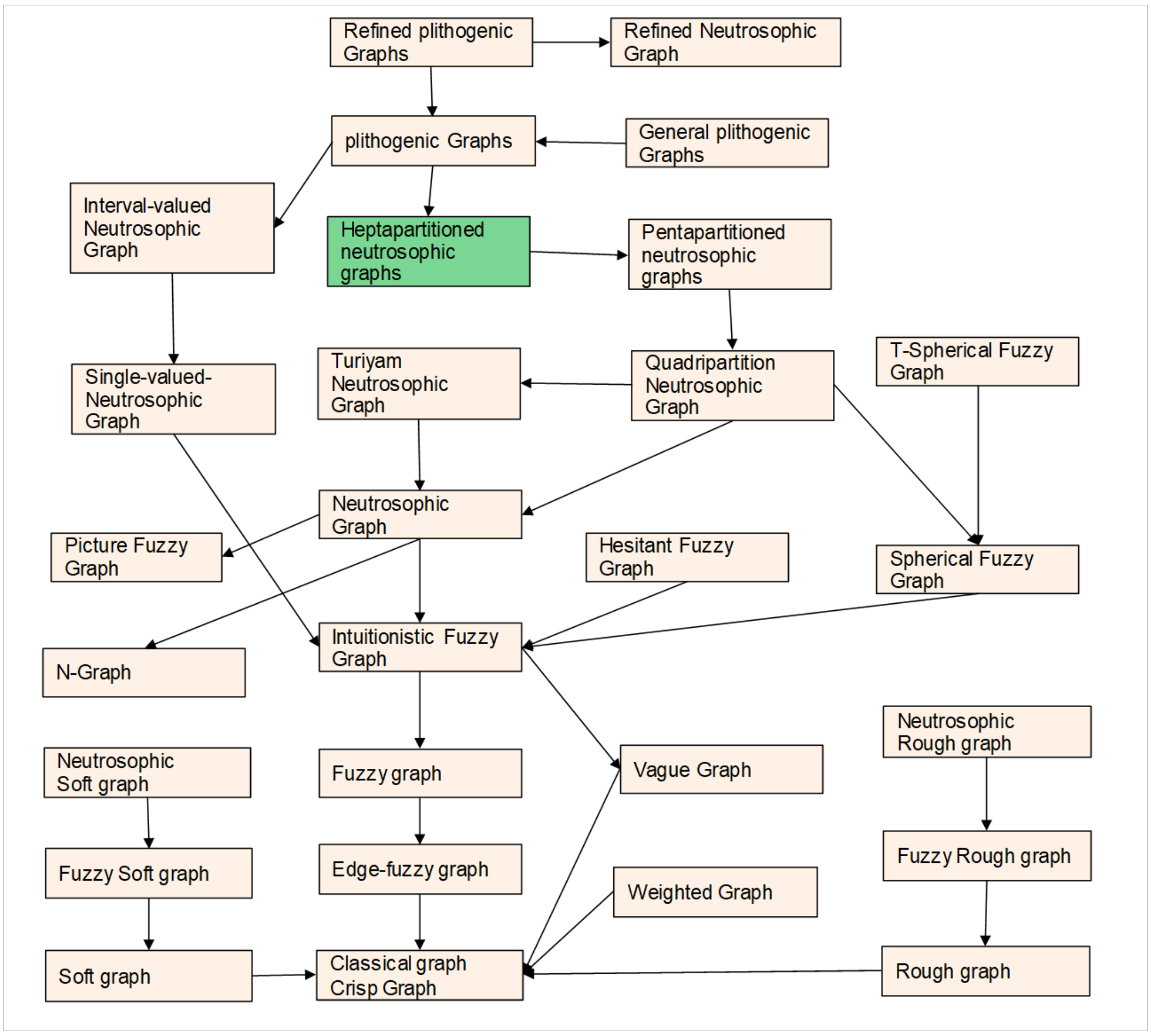

Figure 3.3: Some Uncertain graphs Hierarchy. The graph class at the origin of an arrow contains the graph class at the destination of the arrow.

## 3.8 Double-Valued Neutrosophic Graphs

Double-valued neutrosophic graphs are the graph-theoretic extension of double-valued neutrosophic sets. As with other neutrosophic models, double-valued neutrosophic sets have been studied extensively [335–338]. Related generalizations include triple-valued [339–341], quadruple-valued [341, 342], and quintuple-valued neutrosophic sets [341, 343]. We recall the set-based definition and then present the graph version.

**Definition 3.8.1** (Double-Valued Neutrosophic Set (DVNS))**.** [338] Let $X$ be a nonempty set. A *double-valued neutrosophic set* (DVNS) $A$ on $X$ is specified by four functions

$$T_A,\ I_A^T,\ I_A^F,\ F_A :\ X \longrightarrow [0,1],$$

where, for each $x \in X$, $T_A(x)$ is the truth degree, $F_A(x)$ is the falsity degree, and $I_A^T(x)$ (resp. $I_A^F(x)$) is an indeterminacy degree leaning toward truth (resp. toward falsity). Equivalently,

$$A = \left\{ \langle x,\ T_A(x), I_A^T(x), I_A^F(x), F_A(x) \rangle :\ x \in X \right\}.$$

**Remark 3.8.2** (Sum bounds)**.** Since each component lies in $[0, 1]$, one always has

$$0 \le T_A(x) + I_A^T(x) + I_A^F(x) + F_A(x) \le 4 \qquad (\forall x \in X).$$

Thus the upper bound $\le 4$ is automatic; additional normalization constraints (e.g. $\le 1$) may be imposed in specific applications, but they are not required for the basic definition.

**Definition 3.8.3** (Double-Valued Neutrosophic Graph (DVNG))**.** Let $G = (V, E)$ be a finite simple undirected graph with $V \neq \emptyset$ and $E \subseteq \binom{V}{2}$. A *double-valued neutrosophic graph* (DVNG) on $G$ is a tuple

$$G_{DV} = (G,\ T,\ I^T,\ I^F,\ F),$$

where

$$T,\ I^T,\ I^F,\ F:\ V\cup E\longrightarrow[0,1].$$

Hence each vertex and each edge $x\in V\cup E$ carries a label

$$\big(T(x),\ I^T(x),\ I^F(x),\ F(x)\big)\in[0,1]^4.$$

**Remark 3.8.4** (Optional edge–vertex coherence)**.** In some graph models it is desirable to relate edge labels to endpoint labels. A common (and consistent) option is: for every edge $e=\{u,v\}\in E$,

$$T(e)\le\min\{T(u),T(v)\},\qquad F(e)\le\min\{F(u),F(v)\},$$

$$I^T(e)\ge\max\{I^T(u),I^T(v)\},\qquad I^F(e)\ge\max\{I^F(u),I^F(v)\}.$$

These inequalities are *additional axioms* and are not part of the core DVNG definition.

**Example 3.8.5** (A small double-valued neutrosophic graph)**.** Let $G=(V,E)$ be the path on three vertices

$$V=\{u,v,w\},\qquad E=\{uv,vw\}.$$

Define membership maps $T,I^T,I^F,F:V\cup E\to[0,1]$ by the following tables.

**Vertex labels.**

| $x\in V$ | $T(x)$ | $I^T(x)$ | $I^F(x)$ | $F(x)$ |
|---|---|---|---|---|
| $u$ | 0.80 | 0.20 | 0.10 | 0.40 |
| $v$ | 0.60 | 0.30 | 0.20 | 0.50 |
| $w$ | 0.40 | 0.25 | 0.40 | 0.60 |

**Edge labels.**

| $e\in E$ | $T(e)$ | $I^T(e)$ | $I^F(e)$ | $F(e)$ |
|---|---|---|---|---|
| $uv$ | 0.60 | 0.30 | 0.20 | 0.40 |
| $vw$ | 0.40 | 0.25 | 0.40 | 0.60 |

All values lie in $[0,1]$, hence

$$G_{DV}=\big(G,\ T,\ I^T,\ I^F,\ F\big)$$

is a double-valued neutrosophic graph (DVNG) on $G$ in the sense of Definition 3.8.3.

**(Optional) edge–vertex coherence check.** If we adopt the common optional constraints (cf. Remark 3.8.4)

$$T(e)\le\min\{T(u),T(v)\},\quad F(e)\le\min\{F(u),F(v)\},\quad I^T(e)\ge\max\{I^T(u),I^T(v)\},\quad I^F(e)\ge\max\{I^F(u),I^F(v)\},$$

then for $uv$ we have

$$T(uv)=0.60\le\min\{0.80,0.60\}=0.60,\qquad F(uv)=0.40\le\min\{0.40,0.50\}=0.40,$$

$$I^T(uv)=0.30\ge\max\{0.20,0.30\}=0.30,\qquad I^F(uv)=0.20\ge\max\{0.10,0.20\}=0.20.$$

For $vw$,

$$T(vw)=0.40\le\min\{0.60,0.40\}=0.40,\qquad F(vw)=0.60\le\min\{0.50,0.60\}=0.50\ \text{(fails)},$$

so this particular optional coherence axiom for $F$ is *not* satisfied by the current choice. (If one instead uses $F(e)\le\max\{F(u),F(v)\}$, then it holds: $0.60\le\max\{0.50,0.60\}=0.60$.) Thus, the example is a valid DVNG regardless; coherence is an additional modeling choice.

We illustrate this example in Figure 3.4.

**Theorem 3.8.6** (DVNG as a special case of a general plithogenic graph)**.** *Every DVNG $G_{DV}=(G,T,I^T,I^F,F)$ canonically induces a general plithogenic graph (in the sense of Definition 3.3.10) with one attribute ($m=1$), appurtenance dimension $s=4$, and contradiction dimension $t=1$.*

$T = 0.60$, $I^T = 0.30$, $I^F = 0.20$, $F = 0.40$ (edge $u$–$v$)
$T = 0.40$, $I^T = 0.25$, $I^F = 0.40$, $F = 0.60$ (edge $v$–$w$)
$u$: $T = 0.80$, $I^T = 0.20$, $I^F = 0.10$, $F = 0.40$
$v$: $T = 0.60$, $I^T = 0.30$, $I^F = 0.20$, $F = 0.50$
$w$: $T = 0.40$, $I^T = 0.25$, $I^F = 0.40$, $F = 0.60$

Figure 3.4: A double-valued neutrosophic graph (DVNG) on the path $u-v-w$ (Example 3.8.5).

*Proof.* Let $G = (V, E)$ be the underlying graph. Construct a general plithogenic graph with $m = 1$ attribute $A$, choose singleton value sets $Ml = \{a_0\}$ for vertices and $Nm = \{b_0\}$ for edges, set $s = 4$, and define

$$adf : V \times Ml \to [0, 1]^4, \qquad adf(v, a_0) := \big(T(v), I^T(v), I^F(v), F(v)\big),$$

$$bdf : E \times Nm \to [0, 1]^4, \qquad bdf(e, b_0) := \big(T(e), I^T(e), I^F(e), F(e)\big).$$

Choose any contradiction map $pCF : Ml \times Ml \to [0, 1]$ (necessarily constant on a singleton; take $pCF(a_0, a_0) = 0$), and similarly for edges if a separate DCF is used. Then the resulting structure satisfies Definition 3.3.10 with $s = 4$ and $t = 1$, and its DAF vectors coincide with the DVNG labels. Hence the DVNG is encoded as a special instance of a general plithogenic graph. □

## 3.9 Soft Intersection Graphs and Plithogenic Intersection OverGraphs

An *intersection graph* encodes a family of sets by representing each set as a vertex and joining two vertices whenever the corresponding sets intersect. Intersection graphs form a classical and well-studied theme in graph theory [344, 345], with many important subclasses such as interval graphs [346, 347], unit disk graphs [348, 349], circular-arc graphs [350], string graphs [351, 352], unit ball graphs [353], and permutation graphs [354, 355]. By incorporating soft sets (and hypersoft sets), we obtain natural parameter-based extensions in which vertices represent parameters (or multi-parameter tuples) and adjacency is induced by intersection of the associated soft images. We also sketch a plithogenic "over" variant in which intersection strengths may exceed 1.

**Definition 3.9.1** (Intersection graph)**.** [344] Let $\mathcal{F} = \{S_i\}_{i \in I}$ be a family of sets indexed by a finite set $I$. The *intersection graph* of $\mathcal{F}$ is the simple undirected graph

$$\mathsf{Int}(\mathcal{F}) = (V, E)$$

with vertex set $V := I$ and edge set

$$E := \Big\{\{i, j\} \in \tbinom{I}{2} : S_i \cap S_j \neq \emptyset\Big\}.$$

Equivalently, one may label vertex $i$ by the set $S_i$, and then $ij \in E$ if and only if $S_i \cap S_j \neq \emptyset$.

**Example 3.9.2.** Let $U = \{a, b, c, d, e, f\}$ and

$$S_1 = \{a, b, c\}, \quad S_2 = \{b, d\}, \quad S_3 = \{c, d, e\}, \quad S_4 = \{f\}.$$

Then $\mathsf{Int}(\{S_1, S_2, S_3, S_4\})$ has vertex set $\{1, 2, 3, 4\}$ and edges

$$\{1, 2\}, \ \{1, 3\}, \ \{2, 3\},$$

since $S_1 \cap S_2 = \{b\}$, $S_1 \cap S_3 = \{c\}$, $S_2 \cap S_3 = \{d\}$, while $S_4$ is disjoint from $S_1, S_2, S_3$.

For reference, Figure 3.5 is included.

**Definition 3.9.3** (Soft set)**.** Let $U$ be a nonempty universe and let $A$ be a nonempty set of parameters. A *soft set* over $U$ is a pair $(F, A)$ where

$$F : A \longrightarrow \mathcal{P}(U)$$

is a set-valued map.

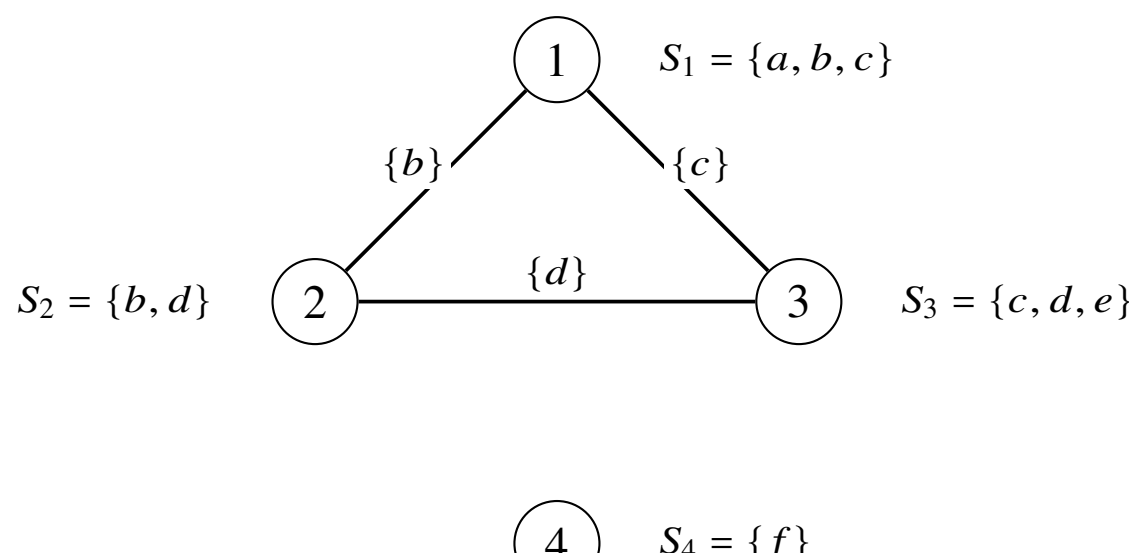

Figure 3.5: Intersection graph $\mathsf{Int}(\{S_1, S_2, S_3, S_4\})$ for Example 3.9.2. Edge labels show the witnessing intersections.

**Definition 3.9.4** (Soft intersection graph)**.** Let $U$ be a nonempty set and let $(F, A)$ be a soft set over $U$ (Definition 3.9.3). The *soft intersection graph* induced by $(F, A)$ is the simple undirected graph

$$\mathsf{SIG}(F, A) = (V, E')$$

with vertex set $V := A$ and edge set

$$E' := \left\{\{a, b\} \in \binom{A}{2} : F(a) \cap F(b) \neq \emptyset\right\}.$$

**Remark 3.9.5** (Soft intersection graphs generalize classical intersection graphs)**.** If $\mathcal{F} = \{S_i\}_{i \in I}$ is a family of sets, set $A := I$, $U := \bigcup_{i \in I} S_i$, and define $F(i) := S_i$. Then $\mathsf{SIG}(F, A) = \mathsf{Int}(\mathcal{F})$. Thus the soft intersection graph construction strictly generalizes the classical intersection graph model.

**Definition 3.9.6** (Hypersoft set)**.** Let $U$ be a nonempty set and let $n \geq 1$. For each $k \in \{1, \ldots, n\}$, let $E_k$ be a nonempty set of attribute values, and define the Cartesian product

$$A := E_1 \times \cdots \times E_n.$$

A *hypersoft set* over $U$ (with attribute-value spaces $E_1, \ldots, E_n$) is a pair $(F, A)$ where

$$F : A \longrightarrow \mathcal{P}(U)$$

is a set-valued map.

**Definition 3.9.7** (Hypersoft intersection graph)**.** Let $U$ be a nonempty set and let $(F, A)$ be a hypersoft set over $U$ with $A = E_1 \times \cdots \times E_n$ (Definition 3.9.6). The *hypersoft intersection graph* induced by $(F, A)$ is the simple undirected graph

$$\mathsf{HIG}(F, A) = (V, E'')$$

with vertex set $V := A$ and edge set

$$E'' := \left\{\{a, b\} \in \binom{A}{2} : F(a) \cap F(b) \neq \emptyset\right\}.$$

**Theorem 3.9.8** (Hypersoft intersection graphs generalize soft intersection graphs)**.** *Let $n \geq 1$. If $n = 1$, then $A = E_1$ and every hypersoft set $(F, A)$ is a soft set. In this case $\mathsf{HIG}(F, A) = \mathsf{SIG}(F, A)$. Hence hypersoft intersection graphs generalize soft intersection graphs.*

*Proof.* If $n = 1$, then $A = E_1$ and the definition of a hypersoft set coincides with that of a soft set. The edge condition in Definitions 3.9.4 and 3.9.7 is identical, namely $F(a) \cap F(b) \neq \emptyset$. Therefore the resulting graphs coincide. □

## 3.10 Plithogenic intersection graphs

To incorporate plithogenic "over" degrees, we attach to each vertex a set (as in the soft model) and also attach an *intersection strength* to each edge, allowed to exceed 1.

**Definition 3.10.1** (Plithogenic intersection overgraph)**.** Let $U$ be a nonempty set and let $(F, A)$ be a soft set over $U$. Fix a constant $\Omega > 1$. Define the *intersection strength* of two parameters $a, b \in A$ by

$$w(a,b) := \begin{cases} 0, & F(a) \cap F(b) = \emptyset, \\ 1 + \dfrac{|F(a) \cap F(b)|}{|F(a) \cup F(b)|}, & F(a) \cap F(b) \neq \emptyset, \end{cases}$$

which takes values in $[0, 2] \subseteq [0, \Omega]$ when $\Omega \geq 2$. The *plithogenic intersection overgraph* associated with $(F, A)$ (and $\Omega$) is the weighted graph

$$\mathsf{PIOG}_{\Omega}(F, A) := \big(A,\ E',\ \mu\big),$$

where $E'$ is as in Definition 3.9.4 and the edge-weight map $\mu : E' \to [0, \Omega]$ is given by

$$\mu(\{a, b\}) := w(a, b) \qquad (\{a, b\} \in E').$$

**Remark 3.10.2.** Definition 3.10.1 is one explicit model. Any other map $w(a, b)$ that is 0 when $F(a) \cap F(b) = \emptyset$ and lies in $[0, \Omega]$ otherwise can be used instead, depending on the intended semantics (e.g. normalized overlap, weighted overlap, similarity scores, etc.). The essential "over" feature is that $\mu$ is permitted to take values $> 1$.

Furthermore, we plan to examine *plithogenic intersection over/under/off graphs* as intersection-type graphs whose vertices represent attributes (or parameters) and whose adjacency is induced by a nonempty "positive-support" overlap between the corresponding plithogenic membership profiles. The definitions below are stated in a uniform, mathematically clean form and clarify the precise sense in which the classical intersection graph is recovered as a special case.

**Definition 3.10.3** (Plithogenic membership profiles for attributes)**.** Let $U$ be a nonempty universe and let $A = \{A_1, \ldots, A_n\}$ be a finite set of attributes. For each attribute $A_i$, fix real bounds $\Psi_i < 0 < 1 < \Omega_i$ and a membership function

$$\mu_i := \mu_{A_i} :\ U \longrightarrow [\Psi_i, \Omega_i].$$

For each $i$, define the *positive support* of $\mu_i$ by

$$\operatorname{supp}^+(\mu_i)\ :=\ \{x \in U :\ \mu_i(x) > 0\}.$$

**Remark 3.10.4.** The threshold $> 0$ in $\operatorname{supp}^+$ is a natural choice when $\Psi_i$ may be negative (under/off cases). More generally, one could use an attribute-dependent threshold $\tau_i \in [\Psi_i, \Omega_i]$ and define $\{x : \mu_i(x) > \tau_i\}$. The results below remain valid with this modification.

**Definition 3.10.5** (Plithogenic intersection overgraph)**.** Assume the setting of Definition 3.10.3 and suppose moreover that $\Psi_i = 0$ and $\Omega_i > 1$ for all $i$ (the *over* case), so that $\mu_i : U \to [0, \Omega_i]$. The *plithogenic intersection overgraph* is the simple undirected graph

$$\mathsf{PIOG} = (V, E)$$

with vertex set $V := A$ and edge set

$$E\ :=\ \big\{\{A_i, A_j\} \in \tbinom{A}{2} :\ \operatorname{supp}^+(\mu_i) \cap \operatorname{supp}^+(\mu_j) \neq \emptyset\big\}.$$

Equivalently, $\{A_i, A_j\} \in E$ if and only if there exists $x \in U$ such that $\mu_i(x) > 0$ and $\mu_j(x) > 0$.

**Definition 3.10.6** (Plithogenic intersection undergraph)**.** Assume the setting of Definition 3.10.3 and suppose that $\Omega_i = 1$ and $\Psi_i < 0$ for all $i$ (the *under* case), so that $\mu_i : U \to [\Psi_i, 1]$. The *plithogenic intersection undergraph* is the simple undirected graph

$$\mathsf{PIUG} = (V, E)$$

with vertex set $V := A$ and edge set

$$E\ :=\ \big\{\{A_i, A_j\} \in \tbinom{A}{2} :\ \operatorname{supp}^+(\mu_i) \cap \operatorname{supp}^+(\mu_j) \neq \emptyset\big\}.$$

**Definition 3.10.7** (Plithogenic intersection offgraph)**.** Assume the setting of Definition 3.10.3 with $\Psi_i < 0 < 1 < \Omega_i$ for all $i$ (the *off* case), so that $\mu_i : U \to [\Psi_i, \Omega_i]$. The *plithogenic intersection offgraph* is the simple undirected graph

$$\mathsf{PIOFFG} = (V, E)$$

with vertex set $V := A$ and edge set

$$E := \left\{\{A_i, A_j\} \in \binom{A}{2} : \operatorname{supp}^+(\mu_i) \cap \operatorname{supp}^+(\mu_j) \neq \emptyset\right\}.$$

**Remark 3.10.8** (Uniform template)**.** Definitions 3.10.5–3.10.7 share the same adjacency rule; only the codomains of the memberships $\mu_i$ differ (over/under/off). This mirrors the over/under/off philosophy: the *interval of admissible degrees* changes, while the qualitative notion of "having a positive overlap" remains the driver of adjacency.

**Theorem 3.10.9** (Reduction to the classical intersection graph)**.** *Let $U$ be a nonempty set and let $S_1, \ldots, S_n \subseteq U$ be crisp subsets. For each i, define $\mu_i : U \to [0, 1]$ by the indicator function*

$$\mu_i(x) := \mathbf{1}_{S_i}(x) = \begin{cases} 1, & x \in S_i, \\ 0, & x \notin S_i. \end{cases}$$

*Then the plithogenic intersection overgraph (Definition 3.10.5 with $\Omega_i = 1$) coincides with the classical intersection graph $\mathsf{Int}(\{S_i\}_{i=1}^n)$ (Definition 3.9.1).*

*Proof.* For each $i$, $\operatorname{supp}^+(\mu_i) = \{x : \mu_i(x) > 0\} = S_i$. Hence for $i \neq j$,

$$\{A_i, A_j\} \in E \iff \operatorname{supp}^+(\mu_i) \cap \operatorname{supp}^+(\mu_j) \neq \emptyset \iff S_i \cap S_j \neq \emptyset,$$

which is exactly the edge condition of the classical intersection graph. □

The next statements clarify what can (and cannot) be claimed rigorously about representing arbitrary plithogenic graphs as attribute-intersection graphs. In general, an *arbitrary* plithogenic graph on a vertex/edge carrier cannot be recovered uniquely from an attribute-intersection graph on the attribute set, because the latter forgets which concrete vertices/edges carried which memberships. What is always possible is to represent the *co-occurrence pattern of attributes*.

**Theorem 3.10.10** (Attribute co-occurrence representation: over case)**.** *Let $G_P$ be a plithogenic overgraph in the scalar/singleton sense: it consists of an underlying finite simple graph $G^* = (M, N)$, a finite attribute set $A = \{A_1, \ldots, A_n\}$, and membership maps*

$$\mu_i : (M \cup N) \longrightarrow [0, \Omega_i] \qquad (\Omega_i > 1) \quad (i = 1, \ldots, n),$$

*assigning to each element $x \in M \cup N$ a degree for each attribute $A_i$. Let $U := M \cup N$ and consider the same functions $\mu_i : U \to [0, \Omega_i]$. Then the plithogenic intersection overgraph* $\mathsf{PIOG}$ *on vertex set A defined by*

$$\{A_i, A_j\} \in E \iff \exists x \in U \text{ with } \mu_i(x) > 0 \text{ and } \mu_j(x) > 0$$

*encodes exactly the* attribute co-occurrence *pattern in $G_P$: two attributes are adjacent if and only if they appear together with positive degree on some vertex or edge of $G^*$.*

*Proof.* This is immediate from the definition, taking $U = M \cup N$ and restricting attention to the sets $\operatorname{supp}^+(\mu_i) = \{x \in U : \mu_i(x) > 0\}$. By construction, $\{A_i, A_j\} \in E$ holds exactly when these supports intersect, i.e., when there exists $x \in U$ with $\mu_i(x) > 0$ and $\mu_j(x) > 0$. This is precisely the stated co-occurrence criterion. □

**Remark 3.10.11** (Under/off cases)**.** Theorem 3.10.10 holds verbatim for plithogenic undergraphs and offgraphs by replacing the codomain $[0, \Omega_i]$ with $[\Psi_i, 1]$ or $[\Psi_i, \Omega_i]$ and keeping the same positive-support definition $\operatorname{supp}^+(\mu_i) = \{x : \mu_i(x) > 0\}$.

## 3.11 Uncertain Graphs

An *uncertain set* associates with each element a degree taken from a chosen uncertainty model, thereby providing a unifying umbrella for fuzzy, intuitionistic fuzzy, neutrosophic, plithogenic, and related frameworks [293, 356]. An *uncertain graph* is a graph in which vertices and/or edges are equipped with degrees from such a model, encompassing, as special cases, fuzzy, intuitionistic fuzzy, and neutrosophic graph formalisms.

**Definition 3.11.1** (Uncertain model)**.** [293] Let $U$ denote the class of all *uncertain models*. Each $M \in U$ is determined by:

- a nonempty set $\mathrm{Dom}(M) \subseteq [0,1]^k$ of *admissible degree tuples* for some fixed integer $k \geq 1$; and
- model-specific algebraic or geometric constraints imposed on elements of $\mathrm{Dom}(M)$ (for example, $\mu+\nu \leq 1$ in the intuitionistic fuzzy setting, or $0 \leq T + I + F \leq 3$ in the neutrosophic setting).

Typical instances include:

- **Fuzzy model:** $\mathrm{Dom}(M) = [0,1]$;
- **Intuitionistic fuzzy model:** $\mathrm{Dom}(M) = \{(\mu,\nu) \in [0,1]^2 : \ \mu + \nu \leq 1\}$;
- **Neutrosophic model:** $\mathrm{Dom}(M) = \{(T,I,F) \in [0,1]^3 : \ 0 \leq T + I + F \leq 3\}$;
- **Plithogenic model,** and many further extensions.

**Definition 3.11.2** (Uncertain set (U-set))**.** [293] Let $X$ be a nonempty universe, and fix an uncertain model $M$ with degree-domain $\mathrm{Dom}(M) \subseteq [0,1]^k$. An *uncertain set of type* $M$ (briefly, a *U-set*) on $X$ is a pair

$$\mathcal{U} = (X, \mu_M),$$

where

$$\mu_M : X \longrightarrow \mathrm{Dom}(M)$$

is the *uncertainty-degree function* (membership map) of $\mathcal{U}$. For $x \in X$, the value $\mu_M(x) \in \mathrm{Dom}(M)$ encodes the degree(s) to which $x$ belongs to $\mathcal{U}$, as prescribed by the model $M$.

We now state the corresponding graph-theoretic notions.

**Definition 3.11.3** (Uncertain graph)**.** Let $G = (V,E)$ be a finite, undirected, loopless graph, and let $M$ be an uncertain model with degree-domain $\mathrm{Dom}(M)$. An *uncertain graph of type* $M$ is a triple

$$\mathcal{G}_M = (V, E, \mu_M),$$

where

$$\mu_M : V \cup E \longrightarrow \mathrm{Dom}(M)$$

assigns an uncertainty degree in $\mathrm{Dom}(M)$ to each vertex $v \in V$ and each edge $e \in E$. Optionally, one may impose model-dependent consistency relations between vertex- and edge-degrees (e.g., bounding $\mu_M(e)$ in terms of $\mu_M(u)$ and $\mu_M(v)$ for $e = \{u,v\}$ in fuzzy or intuitionistic fuzzy settings), but such constraints are dictated by the chosen model $M$ and are not fixed at the level of this general definition.

# Chapter 4

# Uncertain Concepts

We introduce several extended uncertain concepts.

## 4.1 MultiPlithogenic Sets

MultiPlithogenic sets extend plithogenic sets by allowing *multiple attributes* to be modeled simultaneously, each with its own value space and degree-of-appurtenance function. They may be viewed as a multi-attribute analogue of (multi)neutrosophic representations.

**Definition 4.1.1** (MultiPlithogenic Set)**.** Let $S$ be a nonempty universe and let $P \subseteq S$ be a nonempty set of objects. Fix an integer $n \geq 1$ (number of attributes) and integers $s, t \geq 1$. For each $i \in \{1, \ldots, n\}$, let $v_i$ be an attribute with a nonempty set $P_{v_i}$ of possible values.

A *multiplithogenic set* (MPS) on $P$ is a tuple

$$MPS = \Big(P,\ \{v_i\}_{i=1}^{n},\ \{P_{v_i}\}_{i=1}^{n},\ \{pdf_i\}_{i=1}^{n},\ pCF\Big),$$

where, for each $i$,

$$pdf_i :\ P \times P_{v_i} \longrightarrow [0,1]^s$$

is a *degree-of-appurtenance function* (DAF), and

$$pCF :\ \Big(\bigcup_{i=1}^{n} P_{v_i}\Big) \times \Big(\bigcup_{i=1}^{n} P_{v_i}\Big) \longrightarrow [0,1]^t$$

is a *degree-of-contradiction function* (DCF). (Optionally one may impose standard axioms on $pCF$, e.g. $pCF(a,a) = 0$ and symmetry, but these are not required here.)

Thus each $x \in P$ may be evaluated under each attribute $v_i$ and value $a \in P_{v_i}$ by the vector $pdf_i(x,a) \in [0,1]^s$.

**Remark 4.1.2** (Disjoint-union refinement)**.** If the value sets $P_{v_i}$ are not disjoint, then the union $\bigcup_i P_{v_i}$ may identify the same symbol appearing under different attributes. To avoid ambiguity, one may replace the union by the disjoint union $\bigsqcup_{i=1}^{n} P_{v_i}$ and define $pCF$ on that disjoint union. All results below remain valid with this refinement.

To state a mathematically precise "transformation" to a multi-neutrosophic object, we adopt the following standard multi-neutrosophic (set-valued) formulation.

**Definition 4.1.3** (MultiNeutrosophic Set (set-valued form))**.** Let $X$ be a nonempty set. A *multineutrosophic set* on $X$ is a triple of set-valued maps

$$T,\ I,\ F :\ X \longrightarrow \mathcal{P}([0,1]) \setminus \{\emptyset\},$$

where $T(x)$, $I(x)$, and $F(x)$ represent (possibly multiple) truth, indeterminacy, and falsity degrees of $x$.

**Theorem 4.1.4** (MPS with $s = 3$ induces a multi-neutrosophic set)**.** *Let $MPS$ be a multiplithogenic set on $P$ with $s = 3$ and arbitrary $t \geq 1$. Assume that for each $i$ and each $x \in P$, the set $P_{v_i}$ is nonempty. Write*

$$pdf_i(x, a) = \big(d_{i,1}(x, a),\ d_{i,2}(x, a),\ d_{i,3}(x, a)\big) \in [0, 1]^3.$$

*Define set-valued maps $T, I, F : P \to \mathcal{P}([0, 1]) \setminus \{\emptyset\}$ by*

$$T(x) := \Big\{ d_{i,1}(x, a)\ :\ i \in \{1, \dots, n\},\ a \in P_{v_i} \Big\},$$

$$I(x) := \Big\{ d_{i,2}(x, a)\ :\ i \in \{1, \dots, n\},\ a \in P_{v_i} \Big\},$$

$$F(x) := \Big\{ d_{i,3}(x, a)\ :\ i \in \{1, \dots, n\},\ a \in P_{v_i} \Big\}.$$

*Then $(T, I, F)$ is a multineutrosophic set on $P$ in the sense of Definition 4.1.3.*

*Proof.* Fix $x \in P$. Since each $P_{v_i}$ is nonempty, there exists at least one pair $(i, a)$ with $i \in \{1, \dots, n\}$ and $a \in P_{v_i}$. Hence each of the sets $T(x), I(x), F(x)$ is nonempty. Moreover, each $d_{i,j}(x, a) \in [0, 1]$, so $T(x), I(x), F(x) \subseteq [0, 1]$. Therefore $T, I, F$ map $P$ into $\mathcal{P}([0, 1]) \setminus \{\emptyset\}$, proving that $(T, I, F)$ is a multi-neutrosophic set on $P$. □

**Remark 4.1.5** (What "equivalent" can mean)**.** The construction in Theorem 4.1.4 is generally *many-to-one*: it aggregates degrees over all attributes and values and therefore does not retain the full indexed information $(i, a) \mapsto pdf_i(x, a)$. Thus it is best stated as a canonical *induced* multi-neutrosophic description. If one wants a lossless encoding, one may instead keep the attribute/value indices in the codomain (e.g. as a labeled family).

**Theorem 4.1.6** (Reduction to a plithogenic set)**.** *If $n = 1$ in Definition 4.1.1, then a multiplithogenic set reduces to an ordinary plithogenic set. Consequently, plithogenic sets are special cases of multiplithogenic sets.*

*Proof.* If $n = 1$, write $v := v_1$, $P_v := P_{v_1}$, and $pdf := pdf_1$. Then

$$MPS = \big(P,\ \{v\},\ \{P_v\},\ \{pdf\},\ pCF\big)$$

is exactly the data of a plithogenic set $PS = (P, v, P_v, pdf, pCF)$ (up to the harmless identification of a singleton family with its element). Hence plithogenic sets embed as the $n = 1$ case. □

**Definition 4.1.7** (Multi-crisp set (attribute-indicator form))**.** Let $X$ be a nonempty set and let $n \geq 1$. A *multi-crisp set* on $X$ is an $n$-tuple of indicator functions

$$\chi_i :\ X \longrightarrow \{0, 1\} \qquad (i = 1, \dots, n),$$

equivalently a map $x \mapsto (\chi_1(x), \dots, \chi_n(x)) \in \{0, 1\}^n$.

**Theorem 4.1.8** (Binary $s = 1$ restriction yields multi-crisp sets)**.** *Let $MPS$ be a multiplithogenic set with $s = 1$ and assume that each value set is a singleton $P_{v_i} = \{a_i\}$. If, in addition, each degree is binary,*

$$pdf_i(x, a_i) \in \{0, 1\} \qquad (\forall x \in P,\ \forall i),$$

*then $MPS$ canonically determines a multi-crisp set on $P$ by*

$$\chi_i(x) := pdf_i(x, a_i) \in \{0, 1\}.$$

*Proof.* Under the assumptions, each $pdf_i$ is equivalent to a single function $\chi_i : P \to \{0, 1\}$ via $\chi_i(x) := pdf_i(x, a_i)$. The $n$-tuple $(\chi_1, \dots, \chi_n)$ is therefore a multi-crisp set on $P$ in the sense of Definition 4.1.7. This construction is canonical. □

## 4.2 MultiPlithogenic Graphs

MultiPlithogenic graphs are graph-valued counterparts of multiplithogenic sets: vertices and edges are equipped with *several attributes* (possibly with different value spaces), each assessed by a degree-of-appurtenance map and an optional degree-of-contradiction map.

**Definition 4.2.1** (MultiPlithogenic Graph)**.** Let $G^* = (V, E)$ be a finite simple undirected graph, where $V \neq \emptyset$ and $E \subseteq \binom{V}{2}$. Fix integers $n \geq 1$ (number of attributes), $s \geq 1$ (dimension of appurtenance vectors), and $t \geq 1$.

For each attribute index $i \in \{1, \ldots, n\}$, fix:

$$v_i^V \text{ (a vertex attribute)}, \quad P_i^V \neq \emptyset \text{ (its value set)}, \quad aCf_i : P_i^V \times P_i^V \to [0,1]^t,$$

$$v_i^E \text{ (an edge attribute)}, \quad P_i^E \neq \emptyset \text{ (its value set)}, \quad bCf_i : P_i^E \times P_i^E \to [0,1]^t,$$

together with degree-of-appurtenance functions (DAFs)

$$adf_i : V \times P_i^V \longrightarrow [0,1]^s, \qquad bdf_i : E \times P_i^E \longrightarrow [0,1]^s.$$

A *multiplithogenic graph* (MPG) on $G^*$ is the tuple

$$G_{MP} = \Big(G^*,\ \{(v_i^V, P_i^V, adf_i, aCf_i)\}_{i=1}^n,\ \{(v_i^E, P_i^E, bdf_i, bCf_i)\}_{i=1}^n\Big).$$

Thus each vertex $v \in V$ and each edge $e \in E$ carries $n$ attribute profiles $\{adf_i(v, \cdot)\}_{i=1}^n$ and $\{bdf_i(e, \cdot)\}_{i=1}^n$, respectively.

**Remark 4.2.2** (About the earlier notation $A \subseteq \mathcal{P}(E)$)**.** The requirement "$A \subseteq \mathcal{P}(E)$" is not needed for a mathematically well-posed definition of multiplithogenic graphs and may cause type inconsistencies (attributes are not subsets of edges in general). If one wishes to model *attribute combinations* explicitly, one may introduce an index set $\mathcal{A}$ and consider tuples $\big((a_1, \ldots, a_n) \in P_1^V \times \cdots \times P_n^V\big)$ or similarly for edges; this is naturally handled by the product of value spaces rather than by $\mathcal{P}(E)$.

A mathematically precise "transformation" to a single-attribute plithogenic graph requires fixing an aggregation rule that compresses the $n$ attribute profiles into one profile. Without such a rule, there is no canonical notion of "equivalence" (information is typically lost).

**Definition 4.2.3** (Aggregation operator)**.** Let $n, s \geq 1$. An *aggregation operator* is any map

$$\mathcal{A} : \ ([0,1]^s)^n \longrightarrow [0,1]^s.$$

Typical choices include componentwise maximum, componentwise minimum, or weighted averages.

**Theorem 4.2.4** (Induced plithogenic graph via aggregation)**.** *Let $G_{MP}$ be a multiplithogenic graph as in Definition 4.2.1. Fix aggregation operators*

$$\mathcal{A}_V, \mathcal{A}_E : \ ([0,1]^s)^n \to [0,1]^s.$$

*Define a single vertex value space $P^V := P_1^V \times \cdots \times P_n^V$ and a single edge value space $P^E := P_1^E \times \cdots \times P_n^E$. For $v \in V$ and $\mathbf{a} = (a_1, \ldots, a_n) \in P^V$, set*

$$adf(v, \mathbf{a}) := \mathcal{A}_V\big(adf_1(v, a_1), \ldots, adf_n(v, a_n)\big) \in [0,1]^s,$$

*and for $e \in E$ and $\mathbf{b} = (b_1, \ldots, b_n) \in P^E$, set*

$$bdf(e, \mathbf{b}) := \mathcal{A}_E\big(bdf_1(e, b_1), \ldots, bdf_n(e, b_n)\big) \in [0,1]^s.$$

*Choose any contradiction maps $aCf : P^V \times P^V \to [0,1]^t$ and $bCf : P^E \times P^E \to [0,1]^t$ (e.g. by aggregating the componentwise contradictions). Then the tuple*

$$G' = \Big(G^*,\ (V, v^V, P^V, adf, aCf),\ (E, v^E, P^E, bdf, bCf)\Big)$$

*is a well-defined (single-attribute) plithogenic graph.*

*Proof.* By construction, $P^V$ and $P^E$ are nonempty Cartesian products. For each $v \in V$ and $\mathbf{a} \in P^V$, the vector $\big(adf_1(v, a_1), \ldots, adf_n(v, a_n)\big)$ lies in $([0,1]^s)^n$, so applying $\mathcal{A}_V$ yields an element of $[0,1]^s$. Hence $adf : V \times P^V \to [0,1]^s$ is well-defined. Similarly, $bdf : E \times P^E \to [0,1]^s$ is well-defined. Together with the chosen $aCf$ and $bCf$, this satisfies the data requirements for a plithogenic graph. □

**Remark 4.2.5** (On "equivalence")**.** The construction in Theorem 4.2.4 produces a *compressed* plithogenic graph whose labels are derived from those of $G_{MP}$ via aggregation. In general, the map from $G_{MP}$ to $G'$ is not injective, so it should be understood as an induced representation rather than a lossless equivalence unless additional constraints are imposed.

For reference, the relationships between the MultiUncertain graphs are illustrated in Figure 4.1.

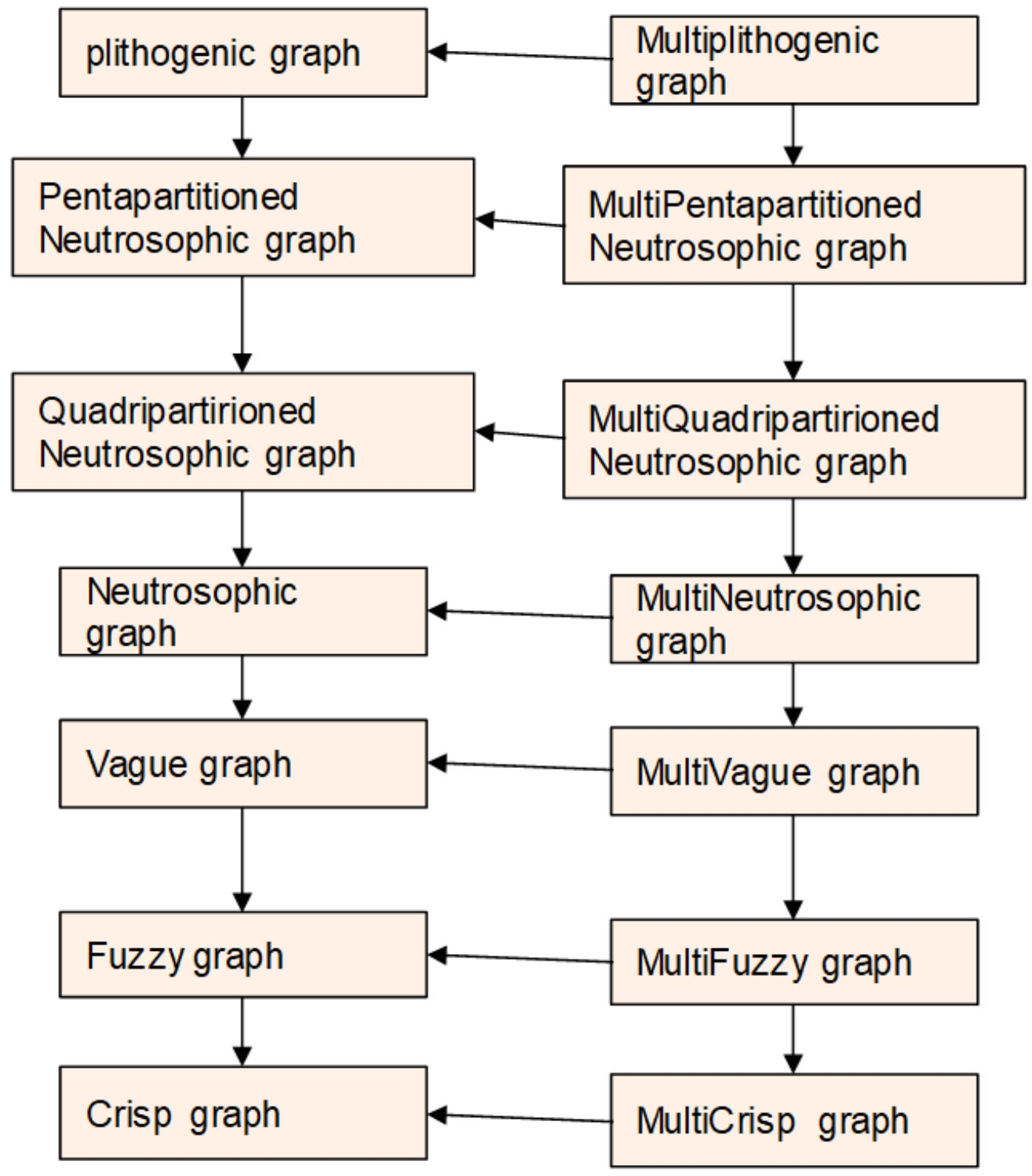

Figure 4.1: Some MultiUncertain graphs Hierarchy. The graph class at the origin of an arrow contains the graph class at the destination of the arrow.

## 4.3 TreeFuzzy Sets and TreeNeutrosophic Sets

A TreeFuzzy Set maps each selected subset of attribute-tree nodes to a fuzzy membership function on $U$, capturing hierarchical uncertainty degrees [357,358]. A TreeNeutrosophic Set maps each selected node-subset to truth, indeterminacy, and falsity membership functions on $U$, modeling hierarchical neutrosophic uncertainty [358].

Let $U \neq \emptyset$ be a universe of objects. Let $\mathrm{Tree}(A)$ be a fixed (finite) rooted tree of attributes, and write

$$\mathcal{N} := \mathrm{Nodes}(\mathrm{Tree}(A))$$

for its node set (including leaves). We use $\mathcal{P}(\mathcal{N})$ for the power set of $\mathcal{N}$. Recall that a *fuzzy subset* of $U$ is a function $\mu : U \to [0,1]$, and we write

$$\mathcal{F}(U) := [0,1]^U$$

for the set of all fuzzy subsets of $U$.

**Definition 4.3.1** (TreeCrisp Set)**.** A *TreeCrisp Set* on $U$ (indexed by Tree$(A)$) is a mapping

$$C : \mathcal{P}(\mathcal{N}) \longrightarrow \mathcal{P}(U).$$

Thus, for every attribute-selection $S \subseteq \mathcal{N}$, the value $C(S) \subseteq U$ is a crisp subset of $U$.

**Definition 4.3.2** (TreeFuzzy Set)**.** A *TreeFuzzy Set* on $U$ (indexed by Tree$(A)$) is a mapping

$$F : \mathcal{P}(\mathcal{N}) \longrightarrow \mathcal{F}(U) = [0,1]^U.$$

Equivalently, for each $S \subseteq \mathcal{N}$, the value $F(S)$ is a membership function

$$F(S) = \mu_S : U \longrightarrow [0,1],$$

assigning to every $x \in U$ a degree of membership $\mu_S(x)$ with respect to the attribute-selection $S$.

**Definition 4.3.3** (TreeNeutrosophic Set)**.** A *TreeNeutrosophic Set* on $U$ (indexed by Tree$(A)$) is a mapping

$$N : \mathcal{P}(\mathcal{N}) \longrightarrow \mathcal{F}(U)^3, \qquad N(S) = \big(T_S,\ I_S,\ F_S\big),$$

where $T_S, I_S, F_S : U \to [0,1]$ are, respectively, the truth-, indeterminacy-, and falsity-membership functions for the selection $S \subseteq \mathcal{N}$. Optionally, one may impose the pointwise neutrosophic bound

$$0 \ \le\ T_S(x) + I_S(x) + F_S(x) \ \le\ 3 \qquad (\forall\, x \in U,\ \forall\, S \subseteq \mathcal{N}),$$

which is automatically satisfied when $T_S, I_S, F_S \in [0,1]$.

**Theorem 4.3.4** (TreeFuzzy sets generalize TreeCrisp sets)**.** *Every TreeCrisp set $C : \mathcal{P}(\mathcal{N}) \to \mathcal{P}(U)$ canonically induces a TreeFuzzy set $F : \mathcal{P}(\mathcal{N}) \to [0,1]^U$.*

*Proof.* For each $S \subseteq \mathcal{N}$, define $F(S) = \chi_{C(S)}$, the characteristic function of $C(S)$:

$$\chi_{C(S)}(x) = \begin{cases} 1, & x \in C(S), \\ 0, & x \notin C(S). \end{cases}$$

Then $F(S) \in [0,1]^U$ for all $S$, hence $F$ is a TreeFuzzy set. □

**Corollary 4.3.5** (TreeNeutrosophic sets generalize TreeCrisp sets)**.** *Every TreeCrisp set canonically induces a TreeNeutrosophic set.*

*Proof.* Let $C$ be a TreeCrisp set and define, for each $S \subseteq \mathcal{N}$,

$$T_S := \chi_{C(S)}, \qquad I_S := 0, \qquad F_S := 1 - \chi_{C(S)}.$$

Then $T_S, I_S, F_S : U \to [0,1]$, so $N(S) = (T_S, I_S, F_S) \in \mathcal{F}(U)^3$, and $N$ is a TreeNeutrosophic set. □

**Theorem 4.3.6** (Reduction to ordinary fuzzy / neutrosophic sets)**.** *Assume $|\mathcal{N}| = 1$, i.e.,* Tree$(A)$ *has exactly one node $a$.*

(i) *For a TreeFuzzy set $F$, the value $F(\{a\}) \in [0,1]^U$ is an ordinary fuzzy set on $U$.*

(ii) *For a TreeNeutrosophic set $N$, the value $N(\{a\}) = (T_{\{a\}}, I_{\{a\}}, F_{\{a\}})$ is an ordinary (single-valued) neutrosophic set on $U$.*

*Proof.* If $\mathcal{N} = \{a\}$, then $\mathcal{P}(\mathcal{N}) = \{\emptyset, \{a\}\}$. By Definitions 4.3.2 and 4.3.3, $F(\{a\}) \in [0,1]^U$ and $N(\{a\}) \in ([0,1]^U)^3$, which are exactly the usual fuzzy and neutrosophic data on $U$. □

**Theorem 4.3.7** (TreeNeutrosophic sets subsume TreeFuzzy sets)**.** *Every TreeFuzzy set $F : \mathcal{P}(\mathcal{N}) \to [0,1]^U$ canonically induces a TreeNeutrosophic set.*

*Proof.* Define $N(S) = (T_S, I_S, F_S)$ by

$$T_S := F(S), \qquad I_S := 0, \qquad F_S := 1 - F(S),$$

where $(1-F(S))(x) := 1-F(S)(x)$. Then $T_S, I_S, F_S : U \to [0,1]$ and hence $N$ is a TreeNeutrosophic set. □

We now record a tree-indexed plithogenic structure in which each node (attribute) carries its own value set and DAF.

**Definition 4.3.8** (TreePlithogenic Set)**.** Let $S$ be a nonempty universe and $P \subseteq S$ be nonempty. Let $\mathrm{Tree}(A)$ have node set $\mathcal{N}$. Fix integers $s, t \geq 1$. A *TreePlithogenic Set* (TPS) is a tuple

$$TPS = \Big(P,\ \mathrm{Tree}(A),\ \{P_a\}_{a\in\mathcal{N}},\ \{pdf_a\}_{a\in\mathcal{N}},\ pCF\Big),$$

where for each node $a \in \mathcal{N}$,

$$P_a \neq \emptyset, \qquad pdf_a : P \times P_a \longrightarrow [0,1]^s,$$

and the degree-of-contradiction function is defined on the (disjoint) union of value spaces:

$$pCF : \Big(\bigsqcup_{a\in\mathcal{N}} P_a\Big) \times \Big(\bigsqcup_{a\in\mathcal{N}} P_a\Big) \longrightarrow [0,1]^t.$$

**Remark 4.3.9** (Why disjoint union)**.** Using $\bigsqcup_{a\in\mathcal{N}} P_a$ avoids collisions when the same symbol appears in two different value sets. If the $P_a$ are known to be disjoint, one may replace $\bigsqcup$ by $\bigcup$.

**Theorem 4.3.10** (Two-level reduction to a MultiPlithogenic Set)**.** *Assume* $\mathrm{Tree}(A)$ *has exactly two levels: a root and n children* $\{v_1, \ldots, v_n\}$*, and no other nodes. Then a TreePlithogenic set TPS induces a MultiPlithogenic set on P (Definition 4.1.1) by taking the attributes to be* $\{v_i\}_{i=1}^n$*, the value spaces to be* $P_{v_i}$*, the DAFs to be* $pdf_{v_i}$*, and the same contradiction map pCF (restricted to the disjoint union of the* $P_{v_i}$*).*

*Proof.* If the node set $\mathcal{N}$ consists of the root together with children $\{v_1, \ldots, v_n\}$, then restricting the TPS data to the level-1 attribute nodes $\{v_i\}$ yields exactly the family of DAFs $pdf_{v_i} : P \times P_{v_i} \to [0,1]^s$ and a global contradiction map on the disjoint union of the $P_{v_i}$, which matches the data of a MultiPlithogenic set. □

**Theorem 4.3.11** (TPS induces TreeFuzzy / TreeNeutrosophic models under $s = 1$ or $s = 3$)**.** *Let TPS be a TreePlithogenic set on P with node set* $\mathcal{N}$*. Fix an aggregation operator* $\mathrm{Agg} : [0,1]^k \to [0,1]$ *for all finite k (e.g.* $\mathrm{Agg} = \max$*).*

(i) *If* $s = 1$ *and each* $P_a$ *contains a distinguished value* $m_a \in P_a$*, define for each* $S \subseteq \mathcal{N}$ *a fuzzy membership* $\mu_S : P \to [0,1]$ *by*

$$\mu_S(x) := \begin{cases} 0, & S = \emptyset, \\ \mathrm{Agg}\big(\{\, pdf_a(x, m_a) : a \in S \,\}\big), & S \neq \emptyset. \end{cases}$$

*Then* $F(S) := \mu_S$ *defines a TreeFuzzy set on P.*

(ii) *If* $s = 3$ *and each* $P_a$ *contains distinguished values* $t_a, i_a, f_a \in P_a$*, define for* $S \subseteq \mathcal{N}$ *and* $x \in P$

$$T_S(x) := \mathrm{Agg}\big(\{\, pdf_a(x, t_a)_1 : a \in S \,\}\big),$$

$$I_S(x) := \mathrm{Agg}\big(\{\, pdf_a(x, i_a)_2 : a \in S \,\}\big),$$

$$F_S(x) := \mathrm{Agg}\big(\{\, pdf_a(x, f_a)_3 : a \in S \,\}\big),$$

*(with* $T_\emptyset = I_\emptyset = F_\emptyset \equiv 0$*). Then* $N(S) := (T_S, I_S, F_S)$ *defines a TreeNeutrosophic set on P.*

*Proof.* In both cases, the right-hand sides are well-defined in $[0,1]$ because each $pdf_a$ takes values in $[0,1]^s$ and Agg returns an element of $[0,1]$. Thus $S \mapsto \mu_S$ yields a map $\mathcal{P}(\mathcal{N}) \to [0,1]^P$, and $S \mapsto (T_S, I_S, F_S)$ yields a map $\mathcal{P}(\mathcal{N}) \to ([0,1]^P)^3$, as required. □

**Remark 4.3.12** (On the role of Agg)**.** There is no canonical aggregation from node-wise plithogenic data to subset-indexed fuzzy/neutrosophic memberships. Theorem 4.3.11 therefore states a general construction parameterized by an explicitly chosen aggregation rule (e.g. maximum, minimum, average, weighted average).

For reference, the relationships between the TreeUncertain sets are illustrated in Figure 4.2.

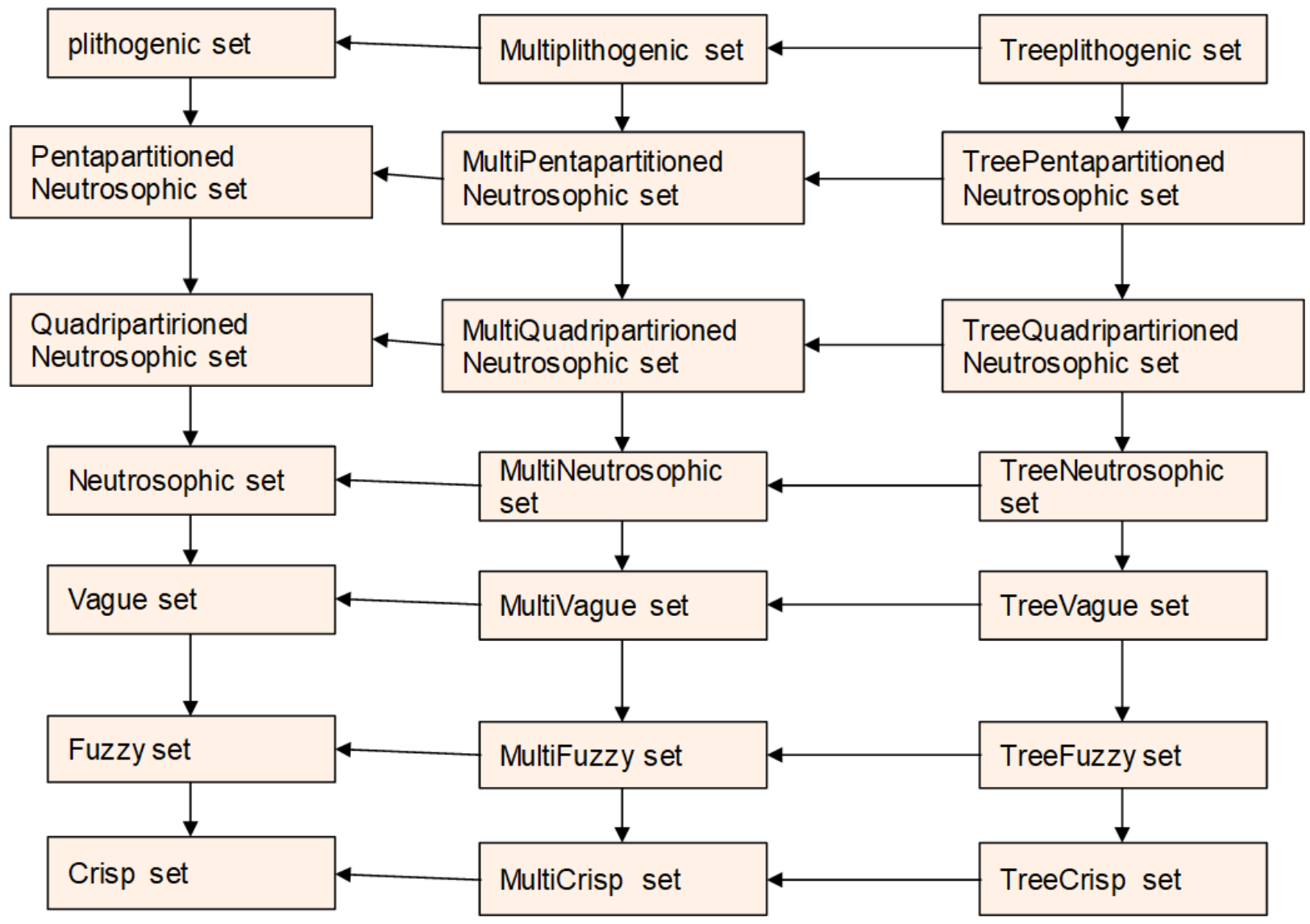

Figure 4.2: Some TreeUncertain sets Hierarchy. The set class at the origin of an arrow contains the set class at the destination of the arrow.

## 4.4 TreeSoft Expert Sets and IndetermSoft Expert Sets

Let $U \neq \emptyset$ be a universe of discourse and let $H \subseteq U$ be a fixed nonempty subset. Write $\mathcal{P}(H)$ for the power set of $H$. Let $\mathrm{Tree}(A)$ be a fixed finite rooted attribute tree, and let

$$\mathcal{N} := \mathrm{Nodes}(\mathrm{Tree}(A))$$

denote its node set (including leaves). In what follows, an "attribute selection" means a subset $S \subseteq \mathcal{N}$.

### 4.4.1 TreeSoft Expert Sets

**Definition 4.4.1** (Soft expert set (standard form))**.** Let $E \neq \emptyset$ be a set of parameters, $X \neq \emptyset$ a set of experts, and $O \neq \emptyset$ a set of opinions. Let $A \subseteq E \times X \times O$ be nonempty. A *soft expert set* over $H$ is a pair $(F, A)$ where

$$F : A \longrightarrow \mathcal{P}(H).$$

**Definition 4.4.2** (TreeSoft Expert Set)**.** Let $X \neq \emptyset$ be a set of experts and $O \neq \emptyset$ a set of opinions. Let

$$\mathcal{Z} := \mathcal{P}(\mathcal{N}) \times X \times O$$

be the set of all triples $(S, x, o)$ consisting of an attribute selection $S \subseteq \mathcal{N}$, an expert $x \in X$, and an opinion $o \in O$. Let $A \subseteq \mathcal{Z}$ be nonempty.

A *TreeSoft Expert Set* (TSES) over $H$ (indexed by $\mathrm{Tree}(A)$) is a pair $(F, A)$ where

$$F : A \longrightarrow \mathcal{P}(H).$$

Thus, for each $(S, x, o) \in A$, the set $F(S, x, o) \subseteq H$ is the collection of objects in $H$ selected by expert $x$ under opinion $o$ when the active attribute selection is $S \subseteq \mathcal{N}$.

**Remark 4.4.3** (Why we use $A \subseteq \mathcal{P}(\mathcal{N}) \times X \times O$)**.** Using the full power set $\mathcal{P}(\mathcal{N})$ allows one to represent combined (multi-node) attribute conditions. If one prefers "single-attribute" parameters, one may restrict $A$ to $\{\{a\} : a \in \mathcal{N}\} \times X \times O$.

**Theorem 4.4.4** (Reduction to a soft expert set)**.** *Assume that* $\mathrm{Tree}(A)$ *has only one non-root level (i.e.,* $\mathcal{N}$ *consists of the root's children only), and restrict* $A$ *to singleton selections:*

$$A \subseteq \{\{a\} : a \in \mathcal{N}\} \times X \times O.$$

*Define $E := \mathcal{N}$ and define $A' \subseteq E \times X \times O$ by*

$$A' := \{(a, x, o) : (\{a\}, x, o) \in A\}.$$

*Then $(F, A)$ canonically induces a soft expert set $(F', A')$ over $H$ by*

$$F'(a, x, o) := F(\{a\}, x, o) \qquad ((a, x, o) \in A').$$

*Proof.* Since $A \subseteq \{\{a\} : a \in \mathcal{N}\} \times X \times O$, the set $A'$ is well-defined and nonempty. The rule $F'(a, x, o) := F(\{a\}, x, o)$ defines a mapping $F' : A' \to \mathcal{P}(H)$ because $F$ maps $A$ into $\mathcal{P}(H)$. Hence $(F', A')$ is a soft expert set in the sense of Definition 4.4.1. □

**Theorem 4.4.5** (TreeSoft Expert Sets generalize TreeSoft Sets)**.** *Let $(F, A)$ be a TreeSoft Expert Set over $H$. Fix $x_0 \in X$ and $o_0 \in O$, and define*

$$A_0 := \{S \subseteq \mathcal{N} : \ (S, x_0, o_0) \in A\}.$$

*Then the mapping $G : A_0 \to \mathcal{P}(H)$ defined by*

$$G(S) := F(S, x_0, o_0) \qquad (S \in A_0)$$

*is a (partial-parameter) TreeSoft Set (i.e., a soft set whose parameter domain is $A_0 \subseteq \mathcal{P}(\mathcal{N})$).*

*Proof.* By construction, $A_0 \subseteq \mathcal{P}(\mathcal{N})$ and $G(S) \in \mathcal{P}(H)$ for all $S \in A_0$, so $G : A_0 \to \mathcal{P}(H)$ is a well-defined set-valued map, which is exactly a TreeSoft Set on the parameter family $A_0$. □

### 4.4.2 IndetermSoft Expert Sets

To model indeterminacy in a mathematically explicit way, we adjoin a distinguished symbol $\mathcal{I}$ ("indeterminate") and allow it to appear in the domain and/or in the image.

**Definition 4.4.6** (IndetermSoft Expert Set)**.** Let $E, X, O$ be nonempty sets of parameters, experts, and opinions, respectively, and let $\mathcal{I}$ be a symbol with $\mathcal{I} \notin E \cup X \cup O \cup U$. Define the extended sets

$$E_{\mathcal{I}} := E \cup \{\mathcal{I}\}, \qquad X_{\mathcal{I}} := X \cup \{\mathcal{I}\}, \qquad O_{\mathcal{I}} := O \cup \{\mathcal{I}\}, \qquad H_{\mathcal{I}} := H \cup \{\mathcal{I}\}.$$

Let $A \subseteq E_{\mathcal{I}} \times X_{\mathcal{I}} \times O_{\mathcal{I}}$ be nonempty. An *IndetermSoft Expert Set* over $H$ is a pair $(F, A)$ where

$$F : \ A \longrightarrow \mathcal{P}(H_{\mathcal{I}}).$$

Here, the occurrence of $\mathcal{I}$ in a coordinate of $(e, x, o) \in A$ represents indeterminacy in the parameter, expert, or opinion, and the occurrence of $\mathcal{I} \in F(e, x, o)$ represents indeterminacy in the selected subset.

**Definition 4.4.7** (IndetermSoft Set)**.** Let $E \neq \emptyset$ and let $\mathcal{I} \notin E \cup U$. Put $E_{\mathcal{I}} := E \cup \{\mathcal{I}\}$ and $H_{\mathcal{I}} := H \cup \{\mathcal{I}\}$. An *IndetermSoft Set* over $H$ is a pair $(G, B)$ with $B \subseteq E_{\mathcal{I}}$ nonempty and

$$G : \ B \longrightarrow \mathcal{P}(H_{\mathcal{I}}).$$

**Theorem 4.4.8** (Generalization relations)**.** *Let $(F, A)$ be an IndetermSoft Expert Set (Definition 4.4.6).*

(i) *If $A \subseteq E \times X \times O$ and $F(e, x, o) \subseteq H$ for all $(e, x, o) \in A$, then $(F, A)$ is a (standard) soft expert set in the sense of Definition 4.4.1.*

(ii) *If $X = \{x_0\}$ and $O = \{o_0\}$ are singletons, then $(F, A)$ canonically induces an IndetermSoft Set $(G, B)$ in the sense of Definition 4.4.7 by setting*

$$B := \{e \in E_{\mathcal{I}} : \ (e, x_0, o_0) \in A\}, \qquad G(e) := F(e, x_0, o_0) \quad (e \in B).$$

*Proof. (i)* Under the hypotheses, the codomain reduces from $\mathcal{P}(H_{\mathcal{I}})$ to $\mathcal{P}(H)$, and the domain reduces from $E_{\mathcal{I}} \times X_{\mathcal{I}} \times O_{\mathcal{I}}$ to $E \times X \times O$. Hence $F : A \to \mathcal{P}(H)$, so $(F, A)$ is a soft expert set.

*(ii)* If $X = \{x_0\}$ and $O = \{o_0\}$, then every element of $A$ has the form $(e, x_0, o_0)$. Thus $B \subseteq E_{\mathcal{I}}$ is well-defined and nonempty, and $G : B \to \mathcal{P}(H_{\mathcal{I}})$ is well-defined by $G(e) = F(e, x_0, o_0)$. Therefore $(G, B)$ is an IndetermSoft Set. □

## 4.5 Multirough Sets and Treerough Sets

Throughout, let $U \neq \emptyset$ be a universe. For an equivalence relation $R$ on $U$ and $x \in U$, write

$$[x]_R := \{y \in U : xRy\}$$

for the $R$-equivalence class of $x$.

**Definition 4.5.1** (Lower and upper approximation)**.** Let $R$ be an equivalence relation on $U$ and let $X \subseteq U$. The *lower* and *upper* $R$-approximations of $X$ are

$$\underline{R}(X) := \{x \in U : [x]_R \subseteq X\}, \qquad \overline{R}(X) := \{x \in U : [x]_R \cap X \neq \emptyset\}.$$

**Definition 4.5.2** (Multirough set)**.** [359] Let $R_1, \ldots, R_n$ be equivalence relations on $U$ with $n \geq 1$, and let $X \subseteq U$. The *multirough set* of $X$ (with respect to $\{R_i\}_{i=1}^n$) is the $n$-tuple

$$\mathcal{MR}(X) := \big( (\underline{R_1}(X), \overline{R_1}(X)), \ \ldots, \ (\underline{R_n}(X), \overline{R_n}(X)) \big).$$

Equivalently, $\mathcal{MR}(X)$ is the family $\{(\underline{R_i}(X), \overline{R_i}(X))\}_{i=1}^n$.

**Remark 4.5.3** (Multirough vs. multi-crisp)**.** A multirough set is a *family of approximations of one subset X* under multiple indiscernibility relations. This is conceptually different from a "multi-crisp set" (an $n$-tuple of indicator functions). Nevertheless, multi-crisp data can be embedded as a degenerate rough situation when all equivalence classes are singletons; see Theorem 4.5.4.

**Theorem 4.5.4** (Degenerate case yields crisp membership)**.** *Let $R$ be the identity equivalence relation on $U$, i.e., $xRy \iff x = y$. Then for every $X \subseteq U$,*

$$\underline{R}(X) = X = \overline{R}(X).$$

*Consequently, if $R_1 = \cdots = R_n = R$ are all the identity relation, then $\mathcal{MR}(X)$ consists of $n$ copies of $(X, X)$, and membership is crisp (binary) via $\chi_X$.*

*Proof.* If $R$ is the identity, then $[x]_R = \{x\}$. Hence $[x]_R \subseteq X$ iff $x \in X$, so $\underline{R}(X) = X$. Similarly $[x]_R \cap X \neq \emptyset$ iff $x \in X$, so $\overline{R}(X) = X$. The final claim follows immediately. □

**Theorem 4.5.5** (Reduction to an ordinary rough set)**.** *If $R_1 = \cdots = R_n = R$ for some equivalence relation $R$ on $U$, then for every $X \subseteq U$,*

$$\mathcal{MR}(X) = \big((\underline{R}(X), \overline{R}(X)), \ldots, (\underline{R}(X), \overline{R}(X))\big),$$

*i.e., the multirough set reduces to repeated copies of the usual rough set $(\underline{R}(X), \overline{R}(X))$.*

*Proof.* Immediate from Definition 4.5.2 since $\underline{R_i}(X) = \underline{R}(X)$ and $\overline{R_i}(X) = \overline{R}(X)$ for all $i$. □

Let Tree$(A)$ be a fixed finite rooted attribute tree, and let $\mathcal{N} := \mathrm{Nodes}(\mathrm{Tree}(A))$ be its node set (including leaves).

**Definition 4.5.6** (Treerough set)**.** Let Tree$(A)$ be an attribute tree with node set $\mathcal{N}$. Assume that to each node $a \in \mathcal{N}$ there is assigned an equivalence relation $R_a$ on $U$. For $X \subseteq U$, the *treerough set* of $X$ (with respect to $\{R_a\}_{a \in \mathcal{N}}$) is the family

$$\mathcal{TR}(X) := \big\{ (\underline{R_a}(X), \overline{R_a}(X)) \ : \ a \in \mathcal{N} \big\},$$

where $\underline{R_a}(X)$ and $\overline{R_a}(X)$ are the lower and upper approximations from Definition 4.5.1.

**Remark 4.5.7** (Tree structure as metadata)**.** In Definition 4.5.6, the tree does not change the approximation operators themselves; it organizes (which may be useful for aggregation along paths, inheritance rules, or level-wise comparisons). Additional axioms can incorporate such hierarchical constraints, but they are not required for the basic notion.

**Theorem 4.5.8** (Treerough sets contain crisp sets as a degenerate case)**.** *Assume $R_a$ is the identity equivalence relation on $U$ for every node $a \in \mathcal{N}$. Then for all $X \subseteq U$,*

$$\mathcal{TR}(X) = \{(X, X) : a \in \mathcal{N}\}.$$

*In particular, the treerough representation collapses to crisp membership via $\chi_X$.*

*Proof.* If each $R_a$ is the identity relation, then $\underline{R_a}(X) = X = \overline{R_a}(X)$ by Theorem 4.5.4. Substituting into Definition 4.5.6 yields the claim. □

**Theorem 4.5.9** (Two-level trees recover multirough sets)**.** *Assume* Tree$(A)$ *has exactly two non-root levels and let $\mathcal{N}_2 \subseteq \mathcal{N}$ be the set of nodes on the second level (e.g., "sub-attributes"). Enumerate $\mathcal{N}_2 = \{a_1, \ldots, a_N\}$ and set $R_i := R_{a_i}$. Then, for every $X \subseteq U$, the restriction of $\mathcal{TR}(X)$ to level-2 nodes equals the multirough set:*

$$\{(\underline{R_a}(X), \overline{R_a}(X)) : \ a \in \mathcal{N}_2\} = \{(\underline{R_i}(X), \overline{R_i}(X)) : \ i = 1, \ldots, N\}.$$

*Hence a two-level treerough set organizes the same approximation data as a multirough set, but with hierarchical indexing.*

*Proof.* This is a reindexing argument. By definition, the level-2 portion of $\mathcal{TR}(X)$ consists of the pairs $(\underline{R_a}(X), \overline{R_a}(X))$ for $a \in \mathcal{N}_2$. Enumerating $\mathcal{N}_2$ and renaming $R_{a_i}$ as $R_i$ yields exactly the multirough family. □

For reference, the relationships between the Soft set and the rough set are illustrated in Figure 4.3.

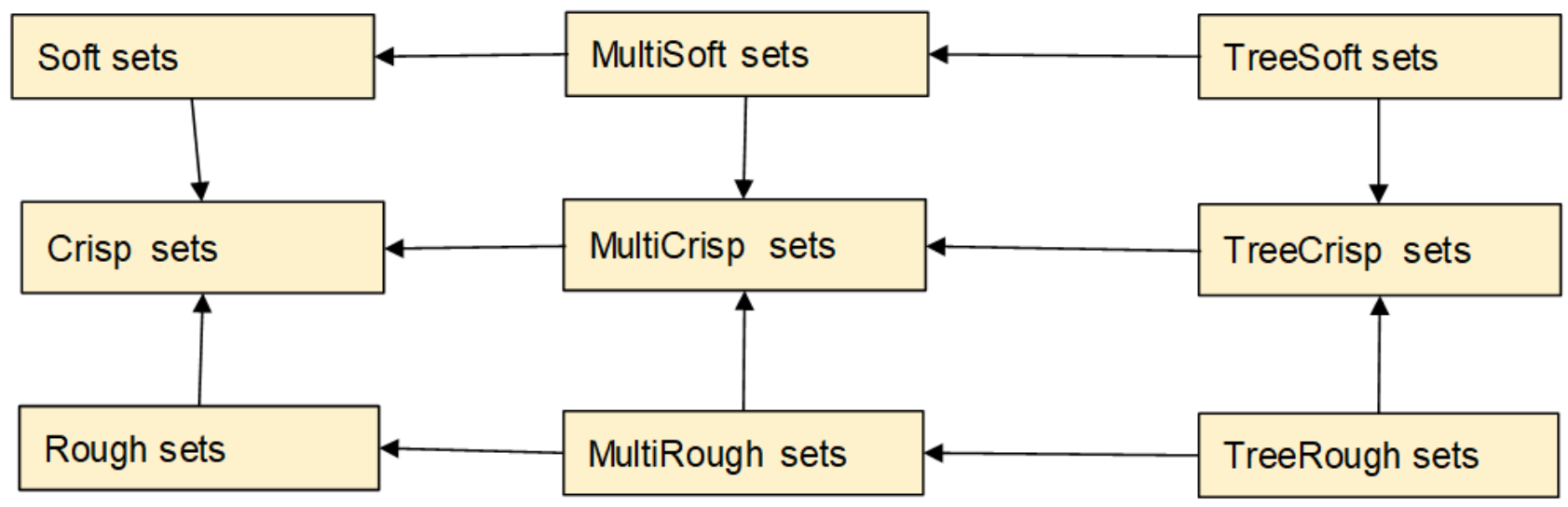

Figure 4.3: Some the Soft sets and the rough sets Hierarchy. The set class at the origin of an arrow contains the set class at the destination of the arrow.

## 4.6 Multi-Quadripartitioned and Multi-Pentapartitioned Neutrosophic Graphs

In this section we introduce multi-versions of quadripartitioned and pentapartitioned neutrosophic graphs, i.e., each vertex/edge carries *a finite family* of neutrosophic partition vectors rather than a single one. We then record clean reduction/embedding relationships with several other graph classes.

**Definition 4.6.1** (Quadripartitioned and pentapartitioned neutrosophic vectors)**.** A *quadripartitioned neutrosophic vector* is a 4-tuple

$$q = (T, C, U, F) \in [0, 1]^4,$$

and a *pentapartitioned neutrosophic vector* is a 5-tuple

$$p = (T, C, R, U, F) \in [0, 1]^5.$$

(Optionally one may impose normalized sum constraints such as $T + C + U + F \leq 1$ or $T + C + U + F \leq 4$, etc.; note that if all components are in $[0, 1]$, then $0 \leq T + C + U + F \leq 4$ and $0 \leq T + C + R + U + F \leq 5$ hold automatically.)

**Definition 4.6.2** (MultiQuadripartitioned Neutrosophic Graph)**.** Let $G = (V, E)$ be a finite simple undirected graph with $V \neq \emptyset$ and $E \subseteq \binom{V}{2}$. A *multiquadripartitioned neutrosophic graph* (MQNG) on $G$ is a pair of set-valued labelings

$$Q_V : V \longrightarrow \mathcal{P}([0,1]^4) \setminus \{\emptyset\}, \qquad Q_E : E \longrightarrow \mathcal{P}([0,1]^4) \setminus \{\emptyset\},$$

assigning to each vertex $v \in V$ a nonempty finite (or, more generally, nonempty) set $Q_V(v)$ of quadripartitioned vectors, and to each edge $e \in E$ a nonempty set $Q_E(e)$ of quadripartitioned vectors.

We denote such a structure by

$$G_{MQ} = (G,\ Q_V,\ Q_E).$$

**Definition 4.6.3** (MultiPentapartitioned Neutrosophic Graph)**.** Let $G = (V, E)$ be a finite simple undirected graph. A *multipentapartitioned neutrosophic graph* (MPNG) on $G$ is a pair of set-valued labelings

$$P_V : V \longrightarrow \mathcal{P}([0,1]^5) \setminus \{\emptyset\}, \qquad P_E : E \longrightarrow \mathcal{P}([0,1]^5) \setminus \{\emptyset\},$$

assigning to each vertex $v \in V$ a nonempty set $P_V(v)$ of pentapartitioned vectors and to each edge $e \in E$ a nonempty set $P_E(e)$ of pentapartitioned vectors. We denote such a structure by

$$G_{MP} = (G,\ P_V,\ P_E).$$

**Remark 4.6.4** (Finite multiplicities as a special case)**.** If one prefers explicit multiplicities, one may require that each $Q_V(v)$ and $Q_E(e)$ (resp. $P_V(v)$, $P_E(e)$) be finite, e.g. $Q_V(v) = \{q_{v,1}, \ldots, q_{v,n_v}\} \subseteq [0,1]^4$. The set-valued formulation above is mathematically cleaner and subsumes the finite case.

To obtain a single-valued (non-multi) partitioned neutrosophic graph from a multi-version, one must fix an aggregation rule.

**Definition 4.6.5** (Aggregation operator)**.** An *aggregation operator* on $[0,1]^k$ is any map

$$\mathcal{A} :\ \mathcal{P}([0,1]^k) \setminus \{\emptyset\} \ \longrightarrow\ [0,1]^k.$$

Typical choices are coordinatewise maximum/minimum, arithmetic mean (for finite sets), or weighted averages.

**Theorem 4.6.6** (MQNG induces a quadripartitioned neutrosophic graph)**.** *Let $G_{MQ} = (G, Q_V, Q_E)$ be an MQNG and fix an aggregation operator $\mathcal{A}_4 : \mathcal{P}([0,1]^4) \setminus \{\emptyset\} \to [0,1]^4$. Define single-valued label maps*

$$q_V(v) := \mathcal{A}_4(Q_V(v)) \in [0,1]^4 \quad (v \in V), \qquad q_E(e) := \mathcal{A}_4(Q_E(e)) \in [0,1]^4 \quad (e \in E).$$

*Then $G$ equipped with $(q_V, q_E)$ is a (single-valued) quadripartitioned neutrosophic graph.*

*Proof.* Since $Q_V(v)$ and $Q_E(e)$ are nonempty, $\mathcal{A}_4$ is applicable and produces elements of $[0,1]^4$. Thus $q_V : V \to [0,1]^4$ and $q_E : E \to [0,1]^4$ are well-defined. □

**Theorem 4.6.7** (MPNG induces a pentapartitioned neutrosophic graph)**.** *Let $G_{MP} = (G, P_V, P_E)$ be an MPNG and fix an aggregation operator $\mathcal{A}_5 : \mathcal{P}([0,1]^5) \setminus \{\emptyset\} \to [0,1]^5$. Define*

$$p_V(v) := \mathcal{A}_5(P_V(v)) \in [0,1]^5 \quad (v \in V), \qquad p_E(e) := \mathcal{A}_5(P_E(e)) \in [0,1]^5 \quad (e \in E).$$

*Then $G$ equipped with $(p_V, p_E)$ is a (single-valued) pentapartitioned neutrosophic graph.*

*Proof.* Identical to the proof of Theorem 4.6.6, replacing $[0,1]^4$ by $[0,1]^5$. □

**Remark 4.6.8** (On "equivalence")**.** Without fixing $\mathcal{A}_4$ or $\mathcal{A}_5$, there is no canonical way to collapse multiple labels to one. Thus it is most precise to say "induces via an aggregation operator" rather than "is equivalent to".

We record a clean embedding into a multi-neutrosophic *triple*-valued model by merging some components.

**Definition 4.6.9** (MultiNeutrosophic Graph (set-valued triple form))**.** Let $G = (V, E)$ be a finite simple graph. A *multi-neutrosophic graph* on $G$ is specified by set-valued maps

$$T_V, I_V, F_V : V \to \mathcal{P}([0,1]) \setminus \{\emptyset\}, \qquad T_E, I_E, F_E : E \to \mathcal{P}([0,1]) \setminus \{\emptyset\}.$$

**Theorem 4.6.10** (MQNG $\Rightarrow$ multi-neutrosophic graph)**.** *Let $G_{MQ} = (G, Q_V, Q_E)$ be an MQNG. Define, for each $v \in V$,*

$$T_V(v) := \{\, T : (T, C, U, F) \in Q_V(v) \,\},$$
$$I_V(v) := \{\, C + U : (T, C, U, F) \in Q_V(v) \,\},$$
$$F_V(v) := \{\, F : (T, C, U, F) \in Q_V(v) \,\},$$

*and analogously for each $e \in E$,*

$$T_E(e) := \{\, T : (T, C, U, F) \in Q_E(e) \,\},$$
$$I_E(e) := \{\, C + U : (T, C, U, F) \in Q_E(e) \,\},$$
$$F_E(e) := \{\, F : (T, C, U, F) \in Q_E(e) \,\}.$$

*Then $(T_V, I_V, F_V, T_E, I_E, F_E)$ defines a multi-neutrosophic graph on $G$.*

*Proof.* Each set on the right is nonempty because $Q_V(v)$ and $Q_E(e)$ are nonempty. Moreover, since $C, U \in [0,1]$, one has $C + U \in [0,2]$; if one requires the indeterminacy range to lie in $[0,1]$, replace $C + U$ by any normalization such as $\min\{1, C + U\}$ or $\frac{C+U}{2}$. With the stated codomain $[0,1]$ one may take the normalized version; with codomain $[0,2]$ the above is direct. Thus the data defines a multi-neutrosophic graph in the sense of Definition 4.6.9. □

**Theorem 4.6.11** (MPNG induces an MQNG by merging components)**.** *Let $G_{MP} = (G, P_V, P_E)$ be an MPNG. Define maps $Q_V, Q_E$ by pushing forward each pentavector $(T, C, R, U, F)$ to a quadrivector*

$$\pi(T, C, R, U, F) := (T,\ C,\ U + R,\ F).$$

*That is,*

$$Q_V(v) := \{\pi(p) : p \in P_V(v)\} \subseteq [0,1]^4,$$
$$Q_E(e) := \{\pi(p) : p \in P_E(e)\} \subseteq [0,1]^4.$$

*If one additionally assumes $U+R \le 1$ for all pentavectors used (or replaces $U+R$ by a normalized combination), then $(G, Q_V, Q_E)$ is an MQNG.*

*Proof.* By construction, $Q_V(v)$ and $Q_E(e)$ are nonempty sets. The first, second, and fourth coordinates of $\pi(p)$ lie in $[0,1]$. The third coordinate lies in $[0,2]$ in general; under the stated assumption $U + R \le 1$ (or after normalization), it lies in $[0,1]$. Hence $Q_V, Q_E$ take values in $\mathcal{P}([0,1]^4) \setminus \{\emptyset\}$, proving the claim. □

**Theorem 4.6.12** (MultiPlithogenic $\Rightarrow$ MPNG when $s = 5$)**.** *Let $G_{MP\ell}$ be a multiplithogenic graph (Definition 4.2.1) with $s = 5$. Assume singleton value sets for simplicity (so each attribute evaluation yields a single 5-vector per element), or fix a selection of values in each value set. Then one obtains an MPNG on the same underlying graph $G^*$ by interpreting each DAF output vector*

$$(d_1, d_2, d_3, d_4, d_5) \in [0,1]^5$$

*as a pentapartitioned neutrosophic vector $(T, C, R, U, F)$ via*

$$T := d_1, \quad C := d_2, \quad R := d_3, \quad U := d_4, \quad F := d_5,$$

*and collecting all such vectors (across attributes, values, and/or sources) into the sets $P_V(v)$ and $P_E(e)$.*

*Proof.* Each DAF vector lies in $[0,1]^5$ by assumption, so it can be reinterpreted as a pentapartitioned vector. Collecting finitely many such vectors produces nonempty subsets of $[0,1]^5$ attached to each vertex/edge, i.e., an MPNG in the sense of Definition 4.6.3. □

**Theorem 4.6.13** (MultiPlithogenic ⇒ MQNG when $s = 4$)**.** *Let $G_{MP\ell}$ be a multiplithogenic graph with $s = 4$. Under the same singleton/selection convention as in Theorem 4.6.12, reinterpret each DAF vector*

$$(d_1, d_2, d_3, d_4) \in [0, 1]^4$$

*as a quadripartitioned vector $(T, C, U, F)$ via*

$$T := d_1, \quad C := d_2, \quad U := d_3, \quad F := d_4,$$

*and collect these vectors into $Q_V(v)$ and $Q_E(e)$. This yields an MQNG on $G^*$.*

*Proof.* Identical to the proof of Theorem 4.6.12, replacing 5-vectors by 4-vectors. □

## 4.7 Neutrosophic Meta Sets

We define a neutrosophic enhancement of the meta-set concept by indexing membership degrees not only by elements of a universe $X$ but also by nodes of a fixed infinite binary tree.

**Definition 4.7.1** (Full infinite binary tree)**.** Let $\Sigma := \{0, 1\}$ and let

$$T := \Sigma^* = \bigcup_{n \geq 0} \Sigma^n$$

be the set of all finite binary strings (including the empty string $\varepsilon$). We view $T$ as the node set of the *full infinite rooted binary tree* with root $\varepsilon$, where each node $p \in T$ has two children $p0$ and $p1$.

**Definition 4.7.2** (Meta set)**.** A *meta set* on a nonempty set $X$ (indexed by the full binary tree $T$) is a mapping

$$\mu : X \times T \longrightarrow [0, 1].$$

We may denote such a meta set by $\rho = (X, T, \mu)$.

**Definition 4.7.3** (Neutrosophic Meta Set)**.** Let $X \neq \emptyset$ and let $T$ be the full infinite binary tree (Definition 4.7.1). A *neutrosophic meta set* on $X$ is a mapping

$$\mu_{\mathfrak{M}} : X \times T \longrightarrow [0, 1]^3, \qquad \mu_{\mathfrak{M}}(x, p) = \big(T_{\mathfrak{M}}(x, p),\ I_{\mathfrak{M}}(x, p),\ F_{\mathfrak{M}}(x, p)\big),$$

where $T_{\mathfrak{M}}, I_{\mathfrak{M}}, F_{\mathfrak{M}} : X \times T \to [0, 1]$ are the truth-, indeterminacy-, and falsity-membership functions indexed by tree nodes $p \in T$.

Since each component lies in $[0, 1]$, the pointwise bound

$$0 \ \leq\ T_{\mathfrak{M}}(x, p) + I_{\mathfrak{M}}(x, p) + F_{\mathfrak{M}}(x, p) \ \leq\ 3 \qquad (\forall x \in X,\ \forall p \in T)$$

holds automatically. We may denote the neutrosophic meta set by $\mathfrak{M} = (X, T, \mu_{\mathfrak{M}})$.

**Remark 4.7.4** (Set-of-pairs notation)**.** Instead of writing a "set of pairs" $\{\langle x, p\rangle\}$, it is mathematically cleaner to define the structure by its membership map $\mu_{\mathfrak{M}} : X \times T \to [0, 1]^3$, as in Definition 4.7.3.

**Theorem 4.7.5** (Neutrosophic meta sets subsume meta sets and neutrosophic sets)**.** *Neutrosophic meta sets generalize both meta sets (Definition 4.7.2) and neutrosophic sets in the following precise senses:*

(i) *(*Meta set embedding*) Every meta set $\rho = (X, T, \mu)$ induces a neutrosophic meta set $\mathfrak{M}_\rho$ on $X$ by*

$$T_{\mathfrak{M}_\rho}(x, p) := \mu(x, p), \qquad I_{\mathfrak{M}_\rho}(x, p) := 0, \qquad F_{\mathfrak{M}_\rho}(x, p) := 1 - \mu(x, p).$$

(ii) *(*Neutrosophic set embedding*) Every neutrosophic set $A = (T_A, I_A, F_A)$ on $X$ induces a neutrosophic meta set $\mathfrak{M}_A$ by selecting a fixed node $p_0 \in T$ (e.g. the root $p_0 = \varepsilon$) and setting*

$$\mu_{\mathfrak{M}_A}(x, p_0) := \big(T_A(x), I_A(x), F_A(x)\big) \quad \textit{for all } x \in X,$$

*and, for $p \neq p_0$, one may either leave the value undefined by restricting the domain to $X \times \{p_0\}$, or define a default extension, e.g.*

$$\mu_{\mathfrak{M}_A}(x, p) := \big(T_A(x), I_A(x), F_A(x)\big) \quad (\forall p \in T).$$

*Proof.* *(i)* Given $\mu : X\times T \to [0,1]$, the three displayed formulas define maps $T_{\mathfrak{M}_\rho}, I_{\mathfrak{M}_\rho}, F_{\mathfrak{M}_\rho} : X\times T \to [0,1]$, hence $\mu_{\mathfrak{M}_\rho}(x,p) := (T_{\mathfrak{M}_\rho}(x,p), I_{\mathfrak{M}_\rho}(x,p), F_{\mathfrak{M}_\rho}(x,p))$ defines a neutrosophic meta set.

*(ii)* Fix $p_0 \in T$. The assignment $x \mapsto \mu_{\mathfrak{M}_A}(x,p_0) \in [0,1]^3$ is well-defined. If one restricts the tree index set to $\{p_0\}$, this is exactly a neutrosophic meta set on $X \times \{p_0\}$. If instead one extends to all $p \in T$ by a constant-in-$p$ rule, one obtains a map $X\times T \to [0,1]^3$. In either interpretation, the original neutrosophic set data is recovered as the slice at $p_0$. □

## 4.8 Cohesive Neutrosophic Sets

We formalize a complex-valued (phase–magnitude) variant of neutrosophic membership in which each element is assigned truth/indeterminacy/falsity information as subsets of the closed unit disk (often called the "unit circle" informally, though the set described below is the disk unless the magnitude is fixed to 1).

**Definition 4.8.1** (Closed unit disk)**.** Let

$$\mathbb{D} := \{z \in \mathbb{C} : \ |z| \le 1\}$$

be the closed unit disk in the complex plane, and let $\mathcal{P}(\mathbb{D})$ denote its power set.

**Definition 4.8.2** (Cohesive Neutrosophic Set (CHNS))**.** Let $S \neq \emptyset$ be a universe and let $T \subseteq S$ be a (crisp) subset. A *cohesive neutrosophic set* (CHNS) on $S$ (with respect to $T$) is specified by three set-valued maps

$$h_T^{\mathrm{T}},\ h_T^{\mathrm{I}},\ h_T^{\mathrm{F}} :\ S \longrightarrow \mathcal{P}(\mathbb{D}) \setminus \{\emptyset\},$$

where $h_T^{\mathrm{T}}(x)$, $h_T^{\mathrm{I}}(x)$, and $h_T^{\mathrm{F}}(x)$ represent the truth-, indeterminacy-, and falsity-membership sets of $x$, respectively.

Equivalently, for each $x \in S$ one may write these sets in polar form as

$$h_T^{\mathrm{T}}(x) = \{\, r_T^{\mathrm{T}}(x)\, e^{i\,\omega_T^{\mathrm{T}}(x)} :\ r_T^{\mathrm{T}}(x) \in [0,1],\ \omega_T^{\mathrm{T}}(x) \in \mathbb{R}\,\},$$

$$h_T^{\mathrm{I}}(x) = \{\, r_T^{\mathrm{I}}(x)\, e^{i\,\omega_T^{\mathrm{I}}(x)} :\ r_T^{\mathrm{I}}(x) \in [0,1],\ \omega_T^{\mathrm{I}}(x) \in \mathbb{R}\,\},$$

$$h_T^{\mathrm{F}}(x) = \{\, r_T^{\mathrm{F}}(x)\, e^{i\,\omega_T^{\mathrm{F}}(x)} :\ r_T^{\mathrm{F}}(x) \in [0,1],\ \omega_T^{\mathrm{F}}(x) \in \mathbb{R}\,\},$$

where the magnitudes $r_T^{\mathrm{T}}(x), r_T^{\mathrm{I}}(x), r_T^{\mathrm{F}}(x)$ lie in $[0,1]$ and the phases $\omega_T^{\mathrm{T}}(x), \omega_T^{\mathrm{I}}(x), \omega_T^{\mathrm{F}}(x)$ are real (radians). (Any nonempty subset of $\mathbb{D}$ can be described by such magnitude/phase constraints; the display indicates the intended interpretation.)

We may denote the CHNS compactly as

$$\mathsf{CHNS}(T) = \big\{\langle x,\ h_T^{\mathrm{T}}(x),\ h_T^{\mathrm{I}}(x),\ h_T^{\mathrm{F}}(x)\rangle :\ x \in S\big\}.$$

**Remark 4.8.3** ("Unit circle" vs. "unit disk")**.** If one wishes membership values to lie on the unit *circle* $\{z : |z| = 1\}$, then the magnitudes must be fixed to $r \equiv 1$. In Definition 4.8.2 the magnitudes range in $[0,1]$, hence the natural codomain is the unit *disk* $\mathbb{D}$.

**Definition 4.8.4** (Cohesive Fuzzy Set (CHFS))**.** A *cohesive fuzzy set* on $S$ is a mapping

$$h :\ S \longrightarrow \mathcal{P}(\mathbb{D}) \setminus \{\emptyset\},$$

assigning to each $x \in S$ a nonempty subset $h(x) \subseteq \mathbb{D}$ interpreted as a complex-valued membership set.

**Definition 4.8.5** (Single-valued neutrosophic set)**.** A *(single-valued) neutrosophic set* on $S$ is a triple of maps

$$T,\ I,\ F :\ S \longrightarrow [0,1].$$

**Theorem 4.8.6** (CHNS generalizes CHFS and neutrosophic sets)**.** *Every cohesive fuzzy set and every (single-valued) neutrosophic set can be embedded as a special case of a cohesive neutrosophic set.*

*Proof. (1) Embedding CHFS into CHNS.* Let $h : S \to \mathcal{P}(\mathbb{D}) \setminus \{\emptyset\}$ be a cohesive fuzzy set. Define $h^{\mathrm{T}}, h^{\mathrm{I}}, h^{\mathrm{F}} : S \to \mathcal{P}(\mathbb{D}) \setminus \{\emptyset\}$ by

$$h^{\mathrm{T}}(x) := h(x), \qquad h^{\mathrm{I}}(x) := \{0\}, \qquad h^{\mathrm{F}}(x) := \{0\} \qquad (\forall x \in S).$$

Then $(h^{\mathrm{T}}, h^{\mathrm{I}}, h^{\mathrm{F}})$ is a CHNS (Definition 4.8.2) in which only the truth component is active.

*(2) Embedding neutrosophic sets into CHNS.* Let $(T, I, F)$ be a neutrosophic set with $T, I, F : S \to [0,1]$. Define a CHNS by singleton-valued complex memberships (zero phase) as

$$h^{\mathrm{T}}(x) := \{T(x)\}, \qquad h^{\mathrm{I}}(x) := \{I(x)\}, \qquad h^{\mathrm{F}}(x) := \{F(x)\} \qquad (\forall x \in S).$$

Since $T(x), I(x), F(x) \in [0,1] \subseteq \mathbb{D}$, each right-hand side is a nonempty subset of $\mathbb{D}$, so this defines a CHNS. The original neutrosophic degrees are recovered by taking the unique elements of these singletons. □

## 4.9 Neutrosophic Multisoft Sets

This section defines a *neutrosophic multisoft set* as a multisoft set in which each parameter-combination returns a *neutrosophic set* (rather than a crisp subset) on the same universe. We also record clean reduction relationships to multisoft sets and neutrosophic soft sets, and we restate a neutrosophic TreeSoft variant.

**Definition 4.9.1** (Neutrosophic set-valued map)**.** Let $U \neq \emptyset$. A *single-valued neutrosophic set on* $U$ is a triple of maps

$$T,\ I,\ F :\ U \longrightarrow [0,1].$$

Equivalently, it is a map $u \mapsto (T(u), I(u), F(u)) \in [0,1]^3$.

**Definition 4.9.2** (Multisoft parameter system)**.** Let $U \neq \emptyset$. Let $E_1, \ldots, E_n$ be nonempty parameter sets. To avoid ambiguity when the same symbol appears in different $E_i$, we work with the disjoint union

$$E := \bigsqcup_{i=1}^{n} E_i.$$

A *parameter combination* is a nonempty subset $a \subseteq E$. (Optionally one may restrict to combinations that choose at most one element from each $E_i$.)

**Definition 4.9.3** (Neutrosophic Multisoft Set)**.** Let $U \neq \emptyset$ and let $E = \bigsqcup_{i=1}^{n} E_i$ be as in Definition 4.9.2. Let $A \subseteq \mathcal{P}(E) \setminus \{\emptyset\}$ be a nonempty family of parameter combinations.

A *neutrosophic multisoft set* (NMS) over $U$ is a pair $(F, A)$ where

$$F :\ A \longrightarrow \mathcal{N}(U),$$

and $\mathcal{N}(U)$ denotes the class of all single-valued neutrosophic sets on $U$. Equivalently, for each $a \in A$,

$$F(a) = \big(T_a,\ I_a,\ F_a\big) \quad \text{with} \quad T_a,\ I_a,\ F_a :\ U \to [0,1].$$

Since each component is $[0,1]$-valued, the pointwise bound

$$0 \ \leq\ T_a(u) + I_a(u) + F_a(u)\ \leq\ 3 \qquad (\forall\, u \in U)$$

holds automatically (and may be replaced by a stricter normalization constraint if desired).

**Remark 4.9.4** (Relation to the "subset $X_a \subseteq U$" presentation)**.** If one wishes to keep an explicit associated crisp subset $X_a \subseteq U$, one may recover it from the neutrosophic data by a chosen threshold rule, e.g.

$$X_a(\tau) := \{u \in U :\ T_a(u) > \tau\} \quad \text{or} \quad X_a(\tau) := \{u \in U :\ T_a(u) > \tau,\ F_a(u) < \tau\},$$

but this requires an extra modeling choice and is not necessary for a mathematically well-posed definition.

**Definition 4.9.5** (Multisoft set)**.** Let $U \neq \emptyset$ and $E = \bigsqcup_{i=1}^{n} E_i$. Let $A \subseteq \mathcal{P}(E) \setminus \{\emptyset\}$ be nonempty. A *multisoft set* over $U$ is a pair $(G, A)$ with

$$G :\ A \longrightarrow \mathcal{P}(U).$$

**Definition 4.9.6** (Neutrosophic soft set)**.** Let $U \neq \emptyset$ and let $E \neq \emptyset$ be a parameter set. A *neutrosophic soft set* over $U$ is a pair $(F, E)$ where

$$F : E \longrightarrow \mathcal{N}(U)$$

assigns a neutrosophic set on $U$ to each parameter $e \in E$.

**Theorem 4.9.7** (Generalization properties)**.** *A neutrosophic multisoft set (Definition 4.9.3) generalizes both multisoft sets and neutrosophic soft sets in the following precise senses.*

(i) *(*Multisoft embedding*) Every multisoft set* $(G, A)$ *induces a neutrosophic multisoft set* $(F, A)$ *by*

$$T_a := \chi_{G(a)}, \qquad I_a := 0, \qquad F_a := 1 - \chi_{G(a)} \qquad (a \in A),$$

*i.e., by viewing each crisp set* $G(a) \subseteq U$ *as a degenerate neutrosophic set.*

(ii) *(*Neutrosophic soft embedding*) Every neutrosophic soft set* $(F, E)$ *is a special case of a neutrosophic multisoft set by taking* $n = 1$*,* $E_1 := E$*, and restricting* $A$ *to singletons:*

$$A := \big\{\{e\} : e \in E\big\} \subseteq \mathcal{P}(E) \setminus \{\emptyset\},$$

*and defining* $F'(\{e\}) := F(e)$*.*

*Proof. (i)* For each $a \in A$, define $T_a, I_a, F_a : U \to [0, 1]$ as indicated. Then $(T_a, I_a, F_a)$ is a neutrosophic set on $U$, hence $F(a) := (T_a, I_a, F_a) \in \mathcal{N}(U)$. Thus $F : A \to \mathcal{N}(U)$ and $(F, A)$ is an NMS.

*(ii)* If $n = 1$ then $E = \bigsqcup_{i=1}^{1} E_i = E_1 = E$. With $A = \{\{e\} : e \in E\}$, the assignment $F'(\{e\}) = F(e)$ defines a map $F' : A \to \mathcal{N}(U)$, which is exactly a neutrosophic multisoft set. Restricting to singletons recovers the original neutrosophic soft set. □

**Definition 4.9.8** (Neutrosophic TreeSoft Set)**.** (cf. [215]) Let $U \neq \emptyset$ and let $H \subseteq U$ be nonempty. Let Tree$(A)$ be a finite rooted attribute tree with node set $\mathcal{N}$. A *neutrosophic TreeSoft set* over $H$ is a mapping

$$\Phi : \mathcal{P}(\mathcal{N}) \setminus \{\emptyset\} \longrightarrow \mathcal{N}(H),$$

i.e., for each nonempty node-selection $S \subseteq \mathcal{N}$,

$$\Phi(S) = (T_S, I_S, F_S) \quad \text{with} \quad T_S, I_S, F_S : H \to [0, 1].$$

**Theorem 4.9.9** (Reduction statements)**.** *Let* $\Phi$ *be a neutrosophic TreeSoft set.*

(i) *If for all* $S \subseteq \mathcal{N}$ *one has* $T_S = \chi_{X_S}$*,* $I_S = 0$*, and* $F_S = 1 - \chi_{X_S}$ *for some* $X_S \subseteq H$*, then* $\Phi$ *reduces to a (crisp) TreeSoft set* $S \mapsto X_S$*.*

(ii) *If* Tree$(A)$ *has exactly two levels and one restricts to selections* $S$ *consisting only of nodes on the second level, then* $\Phi$ *reduces to a neutrosophic multisoft set indexed by those second-level nodes (reindexed as parameter combinations).*

*Proof. (i)* is immediate from the identification of crisp sets with their characteristic functions. For *(ii)*, the second level provides a finite parameter family; restricting $\Phi$ to that family yields a map from those parameters to $\mathcal{N}(H)$, which is precisely the neutrosophic multisoft form. □

## 4.10 Bijective TreeSoft Sets

Intuitively, a bijective TreeSoft set assigns to each node of an attribute tree a block of objects in $H$, in such a way that the blocks form a partition of $H$. This is the natural "tree-indexed partition" analogue of bijective soft sets.

**Definition 4.10.1** (Bijective TreeSoft Set)**.** Let $U \neq \emptyset$ be a universe and let $H \subseteq U$ be nonempty. Let $\mathrm{Tree}(A)$ be a finite rooted attribute tree with node set

$$\mathcal{N} := \mathrm{Nodes}(\mathrm{Tree}(A))$$

(including leaves). A *bijective TreeSoft set* over $H$ is a pair $(F, \mathrm{Tree}(A))$ where

$$F: \mathcal{N} \longrightarrow \mathcal{P}(H)$$

satisfies the two conditions:

(i) (*Cover*) $\bigcup_{a \in \mathcal{N}} F(a) = H$.

(ii) (*Pairwise disjointness*) $a \neq b \implies F(a) \cap F(b) = \emptyset$ for all $a, b \in \mathcal{N}$.

Equivalently, $\{F(a)\}_{a \in \mathcal{N}}$ is a partition of $H$ indexed by the nodes of $\mathrm{Tree}(A)$.

**Remark 4.10.2** (Bijection to $H$)**.** Definition 4.10.1 implies that every $h \in H$ belongs to *exactly one* block $F(a)$. Equivalently, the map $\pi : H \to \mathcal{N}$ defined by "$\pi(h) =$ the unique node $a$ with $h \in F(a)$" is well-defined. Thus the data is equivalent to a surjection $\pi : H \to \mathcal{N}$ together with the fibers $F(a) = \pi^{-1}(a)$.

**Definition 4.10.3** (Bijective soft set)**.** Let $H \subseteq U$ be nonempty and let $E \neq \emptyset$ be a finite parameter set. A *bijective soft set* over $H$ is a map

$$G: E \longrightarrow \mathcal{P}(H)$$

such that $\{G(e)\}_{e \in E}$ is a partition of $H$, i.e.,

$$\bigcup_{e \in E} G(e) = H \quad \text{and} \quad e \neq e' \Rightarrow G(e) \cap G(e') = \emptyset.$$

**Definition 4.10.4** (TreeSoft set (node-indexed form))**.** Let $H \subseteq U$ be nonempty and let $\mathrm{Tree}(A)$ have node set $\mathcal{N}$. A *TreeSoft set* (node-indexed form) is a map

$$S: \mathcal{N} \longrightarrow \mathcal{P}(H).$$

(No disjointness or cover condition is imposed.)

**Remark 4.10.5** (Node-indexed vs. power-set-indexed TreeSoft sets)**.** Some authors define TreeSoft sets as maps $\mathcal{P}(\mathcal{N}) \to \mathcal{P}(H)$. The node-indexed form in Definition 4.10.4 is a standard soft-set-style parameterization and is the most natural ambient class for bijective TreeSoft sets. Either form can be converted into the other by adding an aggregation rule $G(S) = \bigcup_{a \in S} S(a)$ or, conversely, by restricting to singletons.

**Theorem 4.10.6** (Generalization relations)**.** *Let $(F, \mathrm{Tree}(A))$ be a bijective TreeSoft set over $H$ with node set $\mathcal{N}$.*

(i) *(*Reduction to bijective soft sets.*) If* $\mathrm{Tree}(A)$ *has only one non-root level and we take the parameter set $E$ to be that level (equivalently, $\mathcal{N} = E \cup \{root\}$ and we ignore the root), then the restriction of $F$ to $E$ is a bijective soft set over $H$.*

(ii) *(*Relaxation to TreeSoft sets.*) If one drops condition* (ii) *(pairwise disjointness) from Definition 4.10.1, the remaining data $F : \mathcal{N} \to \mathcal{P}(H)$ is exactly a TreeSoft set in the sense of Definition 4.10.4.*

*Proof. (i)* If the tree has only one non-root level, its non-root nodes form a finite parameter set $E$. Restrict $F$ to $E$. The cover and disjointness conditions on $\mathcal{N}$ imply the same conditions on $E$ once the root is omitted (or one may set $F(\mathrm{root}) = \emptyset$). Hence $F|_E$ is a bijective soft set (Definition 4.10.3).

*(ii)* After removing disjointness, the definition reduces to a plain map from nodes to subsets of $H$, which is precisely a TreeSoft set as in Definition 4.10.4. □

## 4.11 Treesoft Rough Sets

Soft rough sets extend Pawlak rough sets by replacing equivalence classes with soft-set granules (cf. [142] and subsequent developments such as [360, 361]). In this section we further generalize the granulation mechanism by indexing the granules with a fixed attribute tree, yielding *treesoft rough sets*.

Let $\mathrm{Tree}(A)$ be a fixed finite rooted attribute tree with node set

$$\mathcal{N} := \mathrm{Nodes}(\mathrm{Tree}(A))$$

(including leaves).

**Definition 4.11.1** (TreeSoft set (node-indexed))**.** Let $H \subseteq U$ be nonempty. A *TreeSoft set* over $H$ (indexed by $\mathrm{Tree}(A)$) is a map

$$F: \mathcal{N} \longrightarrow \mathcal{P}(H).$$

**Definition 4.11.2** (Treesoft approximation space)**.** Let $H \subseteq U$ be nonempty and let $F : \mathcal{N} \to \mathcal{P}(H)$ be a TreeSoft set. The pair

$$P := (H, F)$$

is called a *treesoft approximation space*.

**Definition 4.11.3** (Treesoft rough approximations)**.** Let $P = (H, F)$ be a treesoft approximation space and let $X \subseteq H$. Define the *treesoft lower* and *treesoft upper* approximations of $X$ by

$$\underline{\mathrm{apr}}_P(X) := \bigcup\{\, F(a) : \ a \in \mathcal{N},\ F(a) \subseteq X \,\}, \qquad \overline{\mathrm{apr}}^P(X) := \bigcup\{\, F(a) : \ a \in \mathcal{N},\ F(a) \cap X \neq \emptyset \,\}.$$

**Definition 4.11.4** (Treesoft rough set)**.** Let $P = (H, F)$ be a treesoft approximation space and $X \subseteq H$. The pair

$$\big(\underline{\mathrm{apr}}_P(X),\ \overline{\mathrm{apr}}^P(X)\big)$$

is called the *treesoft rough set* of $X$ with respect to $P$.

**Definition 4.11.5** (Positive/negative/boundary regions)**.** Let $P = (H, F)$ be a treesoft approximation space and $X \subseteq H$. Define

$$\mathrm{Pos}_P(X) := \underline{\mathrm{apr}}_P(X), \qquad \mathrm{Neg}_P(X) := H \setminus \overline{\mathrm{apr}}^P(X), \qquad \mathrm{Bnd}_P(X) := \overline{\mathrm{apr}}^P(X) \setminus \underline{\mathrm{apr}}_P(X).$$

We say that $X$ is *treesoft definable* (with respect to $P$) if $\mathrm{Bnd}_P(X) = \emptyset$; otherwise $X$ is *treesoft rough* with respect to $P$.

**Theorem 4.11.6** (Reduction to soft rough sets)**.** *Assume that the attribute tree has only one non-root level so that its non-root node set can be identified with a parameter set A (and the root is ignored). Then every treesoft approximation space $P = (H, F)$ yields a soft approximation space $(H, (F, A))$, and for every $X \subseteq H$ the treesoft approximations coincide with the soft ones:*

$$\underline{\mathrm{apr}}_P(X) = \underline{\mathrm{apr}}_{(H,(F,A))}(X), \qquad \overline{\mathrm{apr}}^P(X) = \overline{\mathrm{apr}}^{(H,(F,A))}(X).$$

*Consequently, treesoft rough sets reduce to soft rough sets in the one-level case.*

*Proof.* If $\mathcal{N}$ is the set of (non-root) nodes and $\mathcal{N} = A$, then the defining unions in Definition 4.11.3 range over the same index set and use the same granules $F(a) \subseteq H$. Hence the resulting sets are equal for every $X \subseteq H$. □

**Remark 4.11.7** (On "reduction to TreeSoft sets")**.** Even if a given subset $X$ is *precise* in the sense that $\underline{\mathrm{apr}}_P(X) = \overline{\mathrm{apr}}^P(X)$, the outcome is still a subset of $H$, not the parameter map $F : \mathcal{N} \to \mathcal{P}(H)$. Thus a treesoft rough set does not literally "reduce to" a TreeSoft set. What is true is that, for definable $X$, membership in $X$ can be represented exactly as a union of some granules:

$$X = \underline{\mathrm{apr}}_P(X) = \overline{\mathrm{apr}}^P(X) = \bigcup_{a \in \mathcal{N}_X} F(a) \quad \text{for some } \mathcal{N}_X \subseteq \mathcal{N}.$$

In other words, definable sets are precisely those unions of the TreeSoft granules.

**Theorem 4.11.8** (Characterization of definable sets)**.** *Let $P = (H, F)$ be a treesoft approximation space and let $X \subseteq H$. Then $X$ is treesoft definable (i.e.* $\mathrm{Bnd}_P(X) = \emptyset$*) if and only if $X$ is a union of granules: there exists $\mathcal{N}_X \subseteq \mathcal{N}$ such that*

$$X = \bigcup_{a \in \mathcal{N}_X} F(a).$$

*Proof.* ($\Rightarrow$) If $\mathrm{Bnd}_P(X) = \emptyset$, then $\underline{\mathrm{apr}}_P(X) = \overline{\mathrm{apr}}^P(X)$. By definition, $\underline{\mathrm{apr}}_P(X) \subseteq X \subseteq \overline{\mathrm{apr}}^P(X)$, hence $X = \underline{\mathrm{apr}}_P(X)$, and $\underline{\mathrm{apr}}_P(X)$ is a union of granules by Definition 4.11.3.

($\Leftarrow$) If $X = \bigcup_{a \in \mathcal{N}_X} F(a)$, then every granule $F(a)$ with $a \in \mathcal{N}_X$ is contained in $X$ and intersects $X$, while any granule disjoint from $X$ does not contribute to either union. Thus $\underline{\mathrm{apr}}_P(X) = X = \overline{\mathrm{apr}}^P(X)$, so $\mathrm{Bnd}_P(X) = \emptyset$. □

## 4.12 *n*-Dimensional Neutrosophic Sets and Offset Extensions

We record a clean hierarchy of "$n$-dimensional" fuzzy and neutrosophic models. Here "$n$-dimensional" means that each element $x \in U$ is assigned an ordered $n$-tuple of degrees (rather than a single degree). We also introduce offset (under/over) variants by enlarging the codomain intervals.

**Definition 4.12.1** ($n$-Dimensional fuzzy set)**.** [362] Let $U \neq \emptyset$ and let $n \in \mathbb{N}^+$. An *$n$-dimensional fuzzy set $A$* on $U$ is specified by $n$ membership functions

$$\mu_{A,i} : U \longrightarrow [0, 1] \qquad (i = 1, \dots, n),$$

such that for every $x \in U$,

$$\mu_{A,1}(x) \leq \mu_{A,2}(x) \leq \cdots \leq \mu_{A,n}(x).$$

Equivalently, $A$ is the map $x \mapsto (\mu_{A,1}(x), \dots, \mu_{A,n}(x)) \in [0, 1]^n$ with coordinatewise monotonicity.

**Definition 4.12.2** ($n$-Dimensional fuzzy offset)**.** Let $U \neq \emptyset$ and $n \in \mathbb{N}^+$. Fix bounds $\Psi_i < 0 < 1 < \Omega_i$ for $i = 1, \dots, n$. An *$n$-dimensional fuzzy offset $A$* on $U$ is specified by maps

$$\mu_{A,i} : U \longrightarrow [\Psi_i, \Omega_i] \qquad (i = 1, \dots, n),$$

such that for every $x \in U$,

$$\mu_{A,1}(x) \leq \mu_{A,2}(x) \leq \cdots \leq \mu_{A,n}(x).$$

Thus the membership degrees may be *under* ($< 0$) or *over* ($> 1$).

**Remark 4.12.3.** It is not necessary (and generally not desirable) to require the existence of $x$ with $\mu_{A,i}(x) > 1$ or $y$ with $\mu_{A,j}(y) < 0$; the term "offset" refers to the *allowed codomain*, not to a property that must occur.

**Definition 4.12.4** ($n$-Dimensional neutrosophic set)**.** Let $U \neq \emptyset$ and $n \in \mathbb{N}^+$. An *$n$-dimensional neutrosophic set $A$* on $U$ is specified by three families of functions

$$T_i,\ I_i,\ F_i : U \longrightarrow [0, 1] \qquad (i = 1, \dots, n),$$

such that for every $x \in U$ the coordinatewise monotonicity conditions hold:

$$T_1(x) \leq T_2(x) \leq \cdots \leq T_n(x), \qquad I_1(x) \leq I_2(x) \leq \cdots \leq I_n(x), \qquad F_1(x) \leq F_2(x) \leq \cdots \leq F_n(x).$$

Equivalently, each $x \in U$ is assigned three monotone $n$-tuples

$$T(x) = (T_1(x), \dots, T_n(x)), \quad I(x) = (I_1(x), \dots, I_n(x)), \quad F(x) = (F_1(x), \dots, F_n(x))$$

with values in $[0, 1]^n$.

Optionally, one may impose the pointwise bound

$$0 \leq T_i(x) + I_i(x) + F_i(x) \leq 3 \qquad (\forall x \in U,\ \forall i = 1, \dots, n),$$

which is automatic since $T_i(x), I_i(x), F_i(x) \in [0, 1]$.

**Theorem 4.12.5** ($n$-Dimensional neutrosophic sets generalize $n$-dimensional fuzzy sets)**.** *Every $n$-dimensional fuzzy set on $U$ canonically induces an $n$-dimensional neutrosophic set on $U$.*

*Proof.* Let $A$ be an $n$-dimensional fuzzy set with membership functions $\mu_{A,i} : U \to [0,1]$ satisfying $\mu_{A,1}(x) \leq \cdots \leq \mu_{A,n}(x)$. Define, for each $i = 1, \ldots, n$,

$$T_i(x) := \mu_{A,i}(x), \qquad I_i(x) := 0, \qquad F_i(x) := 0 \qquad (\forall x \in U).$$

Then $T_i, I_i, F_i : U \to [0,1]$ and the required monotonicity holds for $T_i$ because it holds for $\mu_{A,i}$, while it is trivial for $I_i$ and $F_i$. Hence $\{T_i, I_i, F_i\}_{i=1}^{n}$ defines an $n$-dimensional neutrosophic set. □

**Definition 4.12.6** ($n$-Dimensional neutrosophic offset)**.** Let $U \neq \emptyset$ and $n \in \mathbb{N}^+$. Fix bounds $\Psi_T < 0 < 1 < \Omega_T$, $\Psi_I < 0 < 1 < \Omega_I$, and $\Psi_F < 0 < 1 < \Omega_F$. An *$n$-dimensional neutrosophic offset $A$* on $U$ is specified by maps

$$T_i : U \to [\Psi_T, \Omega_T], \qquad I_i : U \to [\Psi_I, \Omega_I], \qquad F_i : U \to [\Psi_F, \Omega_F] \qquad (i = 1, \ldots, n),$$

such that for every $x \in U$,

$$T_1(x) \leq \cdots \leq T_n(x), \qquad I_1(x) \leq \cdots \leq I_n(x), \qquad F_1(x) \leq \cdots \leq F_n(x).$$

Optionally, one may require a pointwise range bound on the sum, for example

$$\Psi_T + \Psi_I + \Psi_F \;\leq\; T_i(x) + I_i(x) + F_i(x) \;\leq\; \Omega_T + \Omega_I + \Omega_F \qquad (\forall x \in U,\ \forall i),$$

which follows from interval arithmetic whenever each component lies in its stated interval.

**Corollary 4.12.7** ($n$-Dimensional neutrosophic offsets generalize $n$-dimensional fuzzy offsets)**.** *Every $n$-dimensional fuzzy offset on $U$ canonically induces an $n$-dimensional neutrosophic offset on $U$.*

*Proof.* Let $\mu_{A,i} : U \to [\Psi_i, \Omega_i]$ be an $n$-dimensional fuzzy offset (Definition 4.12.2). Choose neutrosophic bounds with $[\Psi_i, \Omega_i] \subseteq [\Psi_T, \Omega_T]$ for all $i$ (e.g. take $\Psi_T = \min_i \Psi_i$ and $\Omega_T = \max_i \Omega_i$), and define

$$T_i(x) := \mu_{A,i}(x), \qquad I_i(x) := 0, \qquad F_i(x) := 0.$$

Then the same monotonicity holds, and $T_i$ lies in the required offset interval, while $I_i, F_i$ are identically zero. Hence we obtain an $n$-dimensional neutrosophic offset. □

## 4.13 Strait Neutrosophic Sets

Strait fuzzy sets were introduced as discretizations of fuzzy sets obtained by replacing each membership value by the *interval (block)* of a chosen finite partition of $[0,1]$ that contains it [363, 364]. We extend the same partition-based idea to neutrosophic sets by discretizing each of the three components (truth/indeterminacy/falsity).

**Definition 4.13.1** (Finite partitions)**.** A *finite partition* of $[0,1]$ is a finite family $\alpha = \{Y_1, \ldots, Y_r\}$ of nonempty subsets of $[0,1]$ such that

$$Y_a \cap Y_b = \emptyset \ (a \neq b), \qquad \bigcup_{a=1}^{r} Y_a = [0,1].$$

Typically one takes each $Y_a$ to be an interval. For a partition $\alpha$, define the *block map*

$$\mathrm{blk}_\alpha : \ [0,1] \longrightarrow \alpha$$

by declaring $\mathrm{blk}_\alpha(t)$ to be the unique block $Y_a \in \alpha$ that contains $t$.

**Remark 4.13.2.** The block map is well-defined because the blocks are pairwise disjoint and cover $[0,1]$.

**Definition 4.13.3** (Strait fuzzy set)**.** [363, 364] Let $X \neq \emptyset$ be a universe and let $\mu : X \to [0,1]$ be a fuzzy membership function. Fix a finite partition $\alpha$ of $[0,1]$ and let $\mathrm{blk}_\alpha$ be its block map. The *strait fuzzy set* generated by $(\mu, \alpha)$ is the map

$$S_{\mu,\alpha} : \ X \longrightarrow \alpha, \qquad S_{\mu,\alpha}(x) := \mathrm{blk}_\alpha(\mu(x)).$$

Equivalently, $S_{\mu,\alpha}$ replaces each membership value $\mu(x)$ by the unique partition block containing it.

**Remark 4.13.4** (Surjectivity is optional)**.** Some presentations require that each block $Y_a \in \alpha$ is hit by some $\mu(x)$. This is an additional assumption (surjectivity of $S_{\mu,\alpha}$) and is not required for the definition.

**Definition 4.13.5** (Strait neutrosophic set)**.** Let $X \neq \emptyset$ and let $A$ be a neutrosophic set on $X$ with membership maps $T_A, I_A, F_A : X \to [0,1]$. Fix three finite partitions $\alpha_T, \alpha_I, \alpha_F$ of $[0,1]$, with block maps $\mathrm{blk}_{\alpha_T}$, $\mathrm{blk}_{\alpha_I}$, $\mathrm{blk}_{\alpha_F}$. The *strait neutrosophic set* generated by $(A, \alpha_T, \alpha_I, \alpha_F)$ is the map

$$S_{A,\alpha_T,\alpha_I,\alpha_F} : X \longrightarrow \alpha_T \times \alpha_I \times \alpha_F$$

defined by

$$S_{A,\alpha_T,\alpha_I,\alpha_F}(x) := \Big(\mathrm{blk}_{\alpha_T}(T_A(x)),\ \mathrm{blk}_{\alpha_I}(I_A(x)),\ \mathrm{blk}_{\alpha_F}(F_A(x))\Big).$$

Thus each triple $(T_A(x), I_A(x), F_A(x))$ is discretized by replacing each coordinate with its containing partition block.

**Remark 4.13.6** (Single partition as a special case)**.** If one prefers to use a single partition $\alpha$ for all three coordinates, set $\alpha_T = \alpha_I = \alpha_F = \alpha$. Then $S : X \to \alpha^3$ is given by $S(x) = (\mathrm{blk}_\alpha(T_A(x)), \mathrm{blk}_\alpha(I_A(x)), \mathrm{blk}_\alpha(F_A(x)))$.

**Theorem 4.13.7** (Strait neutrosophic sets subsume strait fuzzy sets)**.** *Every strait fuzzy set is obtained as a special case of a strait neutrosophic set.*

*Proof.* Let $\mu : X \to [0,1]$ be a fuzzy membership and let $\alpha$ be a finite partition. Define a neutrosophic set $A$ on $X$ by

$$T_A(x) := \mu(x), \qquad I_A(x) := 0, \qquad F_A(x) := 1 - \mu(x).$$

Choose $\alpha_T := \alpha$ and choose any partitions $\alpha_I, \alpha_F$ with $0 \in Y_I \in \alpha_I$ and $1 - \mu(x) \in Y_F(x) \in \alpha_F$ for all $x$ (for instance, take $\alpha_I = \{\{0\}, (0,1]\}$ and any partition for $\alpha_F$). Then the first coordinate of $S_{A,\alpha_T,\alpha_I,\alpha_F}(x)$ is exactly $\mathrm{blk}_\alpha(\mu(x))$, which is the strait fuzzy discretization. Hence strait fuzzy sets embed into strait neutrosophic sets. □

**Theorem 4.13.8** (Strait neutrosophic sets discretize neutrosophic sets)**.** *Let $A$ be a neutrosophic set on $X$. If $\alpha_T, \alpha_I, \alpha_F$ are partitions into singletons (i.e., the discrete partition of* $[0,1]$ *into points), then the associated strait neutrosophic set recovers $A$ pointwise (up to the identification of a singleton with its element).*

*Proof.* If $\alpha_T$ is the partition into singletons, then $\mathrm{blk}_{\alpha_T}(t) = \{t\}$ for all $t \in [0,1]$, and similarly for $\alpha_I, \alpha_F$. Hence

$$S_{A,\alpha_T,\alpha_I,\alpha_F}(x) = \big(\{T_A(x)\}, \{I_A(x)\}, \{F_A(x)\}\big),$$

which is equivalent to $(T_A(x), I_A(x), F_A(x))$ by singleton identification. □

## 4.14 Neutrosophic Distribution Sets

Fuzzy distribution sets (FDS) can be viewed as probability distributions on a finite universe [365, 366]. We extend this idea to neutrosophic data by distributing truth, indeterminacy, and falsity masses across the universe, yielding *neutrosophic distribution sets* (NDS).

**Definition 4.14.1** (Fuzzy Distribution Set (FDS))**.** [365] Let $X = \{x_1, \ldots, x_n\}$ be a finite nonempty set with $n \geq 2$. A *fuzzy distribution set* on $X$ is a function

$$d : X \longrightarrow [0,1]$$

satisfying the normalization condition

$$\sum_{i=1}^{n} d(x_i) = 1.$$

Equivalently, $d$ is an $n$-tuple $D = (d_1, \ldots, d_n) \in [0,1]^n$ with $\sum_i d_i = 1$, where $d_i := d(x_i)$.

A *degenerate (point) distribution* at index $k \in \{1, \ldots, n\}$ is the distribution $D^{(k)}$ defined by

$$d^{(k)}(x_k) = 1, \qquad d^{(k)}(x_j) = 0\ (j \neq k).$$

**Definition 4.14.2** (Neutrosophic Distribution Set (NDS))**.** Let $X = \{x_1, \ldots, x_n\}$ be a finite nonempty set with $n \geq 2$. A *neutrosophic distribution set* (NDS) on $X$ is a triple of functions

$$T,\ I,\ F:\ X \longrightarrow [0,1]$$

such that each component is normalized:

$$\sum_{i=1}^{n} T(x_i) = 1, \qquad \sum_{i=1}^{n} I(x_i) = 1, \qquad \sum_{i=1}^{n} F(x_i) = 1.$$

We denote an NDS by $N = (X; T, I, F)$. Equivalently, $N$ may be represented by three distributions (three $n$-tuples)

$$\mathbf{T} = (T(x_1), \ldots, T(x_n)), \quad \mathbf{I} = (I(x_1), \ldots, I(x_n)), \quad \mathbf{F} = (F(x_1), \ldots, F(x_n)),$$

each belonging to the simplex $\Delta_{n-1} := \{(p_1, \ldots, p_n) \in [0,1]^n : \sum_i p_i = 1\}$.

Since $T(x), I(x), F(x) \in [0,1]$, the pointwise inequality

$$0 \leq T(x) + I(x) + F(x) \leq 3 \qquad (\forall x \in X)$$

holds automatically.

**Remark 4.14.3** (Degenerate distributions)**.** A degenerate truth distribution at $k$ is defined by $T(x_k) = 1$ and $T(x_j) = 0$ for $j \neq k$, and similarly for $I$ and $F$. Hence NDSs admit point masses for each component independently.

**Theorem 4.14.4** (NDS generalizes FDS)**.** *Every fuzzy distribution set induces a neutrosophic distribution set.*

*Proof.* Let $d : X \to [0,1]$ be an FDS, so $\sum_i d(x_i) = 1$. Define $T := d$ and choose any two distributions $I, F : X \to [0,1]$ (e.g. set $I = d$ and $F = d$, or choose degenerate ones), so that $\sum_i I(x_i) = \sum_i F(x_i) = 1$. Then $(T, I, F)$ satisfies Definition 4.14.2, hence defines an NDS. In particular, the map $d \mapsto (d, I, F)$ embeds FDSs into NDSs. □

**Remark 4.14.5** (Why "$I = 0$" cannot be used under NDS normalization)**.** In Definition 4.14.2, $I$ and $F$ are required to be normalized distributions, so $I \equiv 0$ (or $F = 1 - T$) cannot hold unless $n = 1$. If one wants the specialization $I \equiv 0$ and $F = 1 - T$, one should use a *neutrosophic set* (no normalization) rather than an NDS.

**Definition 4.14.6** (Neutrosophic set (no normalization))**.** Let $X \neq \emptyset$. A *(single-valued) neutrosophic set* on $X$ is a triple of maps

$$T,\ I,\ F:\ X \longrightarrow [0,1],$$

with no normalization imposed.

**Theorem 4.14.7** (NDS is a normalized subclass of neutrosophic sets)**.** *Every NDS on a finite set $X$ is a neutrosophic set on $X$ (in the sense of Definition 4.14.6). Conversely, a neutrosophic set on a finite $X$ is an NDS if and only if each of its components $T, I, F$ is normalized.*

*Proof.* Immediate: an NDS is, by definition, a triple of maps $X \to [0,1]$, hence a neutrosophic set. The converse holds exactly when the three normalization equalities in Definition 4.14.2 are satisfied. □

## 4.15 Neutrosophic Multiple Sets

Multiple sets were introduced as a matrix-valued extension of fuzzy and multi-fuzzy models [367–369]. Here we define a neutrosophic analogue by assigning *three* membership matrices (truth/indeterminacy/falsity) to each element.

**Definition 4.15.1** (Multiple Set)**.** [367] Let $X \neq \emptyset$ be a universe and fix integers $n \geq 1$ and $k \geq 1$. A *multiple set of order* $(n, k)$ on $X$ is a mapping

$$A : X \longrightarrow [0,1]^{n\times k}, \qquad x \longmapsto A(x) = \big(a_{i,j}(x)\big)_{\substack{1\leq i\leq n,\\ 1\leq j\leq k}}$$

such that for each fixed row $i \in \{1, \ldots, n\}$ and each $x \in X$, the entries are nonincreasing along the row:

$$a_{i,1}(x) \;\geq\; a_{i,2}(x) \;\geq\; \cdots \;\geq\; a_{i,k}(x).$$

(If one starts with row lengths $k_x$ depending on $x$, one may take $k := \max_{x\in X} k_x$ in the finite case and pad with zeros to obtain a uniform $n \times k$ matrix for all $x$.)

We may represent $A$ as

$$A = \{(x, A(x)) : x \in X\}.$$

**Remark 4.15.2** (Standard special cases)**.** If $n = k = 1$, then $A$ is an ordinary fuzzy set. If $n = 1$, then $A$ is a fuzzy multiset (one row with $k$ grades). If $k = 1$, then $A$ is a (finite) multi-fuzzy set (an $n$-tuple of grades). If all entries belong to $\{0, 1\}$, the structure collapses to a crisp multi-set-type encoding.

**Definition 4.15.3** (Neutrosophic Multiple Set)**.** Let $X \neq \emptyset$ and fix integers $n \geq 1$ and $k \geq 1$. A *neutrosophic multiple set of order* $(n, k)$ on $X$ is a triple of mappings

$$T,\ I,\ F : X \longrightarrow [0,1]^{n\times k},$$

where for each $x \in X$,

$$T(x) = \big(t_{i,j}(x)\big), \quad I(x) = \big(i_{i,j}(x)\big), \quad F(x) = \big(f_{i,j}(x)\big) \in [0,1]^{n\times k},$$

such that the following conditions hold for all $x \in X$ and all $i \in \{1, \ldots, n\}$:

(i) (*Rowwise monotonicity*)

$$t_{i,1}(x) \geq \cdots \geq t_{i,k}(x), \qquad i_{i,1}(x) \geq \cdots \geq i_{i,k}(x), \qquad f_{i,1}(x) \geq \cdots \geq f_{i,k}(x).$$

(ii) (*Pointwise neutrosophic bound*) For every column $j \in \{1, \ldots, k\}$,

$$0 \;\leq\; t_{i,j}(x) + i_{i,j}(x) + f_{i,j}(x) \;\leq\; 3.$$

(This is automatic from $t_{i,j}, i_{i,j}, f_{i,j} \in [0, 1]$, but we record it for emphasis.)

We denote such a structure by

$$N = \{(x,\ T(x),\ I(x),\ F(x)) : x \in X\}.$$

**Remark 4.15.4** (About the "$T + I + F = 1$" condition)**.** The equation $T(x) + I(x) + F(x) = \mathbf{1}$ (entrywise) is a *normalization constraint* sometimes imposed in applications. It does not recover a multiple set unless one also fixes $I \equiv 0$ and identifies $F = \mathbf{1} - T$; otherwise the extra degrees $I$ and $F$ still carry independent information.

The following special cases are immediate consequences of the definitions and therefore need not be conjectural.

**Proposition 4.15.5** (Basic special cases)**.** *Let $N = (T, I, F)$ be a neutrosophic multiple set of order $(n, k)$ on $X$.*

(i) *If $n = k = 1$, then $N$ is a (single-valued) neutrosophic set on $X$.*

(ii) *If $n = 1$, then $N$ is a neutrosophic multiset (one row of $k$ truth/indeterminacy/falsity grades per $x$).*

(iii) *If $k = 1$, then $N$ is an $n$-dimensional (multi-)neutrosophic set (an $n$-tuple of truth/indeterminacy/falsity grades per $x$).*

(iv) *If all entries of $T(x), I(x), F(x)$ belong to $\{0, 1\}$, then $N$ is a discrete (crisp) multi-valued structure.*

*Proof.* All items follow by unpacking Definition 4.15.3 under the stated parameter restrictions. □

**Theorem 4.15.6** (Neutrosophic multiple sets generalize multiple sets)**.** *Every multiple set of order $(n, k)$ canonically induces a neutrosophic multiple set of order $(n, k)$.*

*Proof.* Let $A : X \to [0,1]^{n\times k}$ be a multiple set (Definition 4.15.1). Define $T := A$ and define $I, F : X \to [0,1]^{n\times k}$ by

$$I(x) := \mathbf{0}, \qquad F(x) := \mathbf{0} \qquad (\forall x \in X),$$

where $\mathbf{0}$ denotes the all-zero $n \times k$ matrix. Then $(T, I, F)$ satisfies the rowwise monotonicity and has $t_{i,j}(x) + i_{i,j}(x) + f_{i,j}(x) = a_{i,j}(x) \in [0,1]$. Hence $(T, I, F)$ is a neutrosophic multiple set. This embeds multiple sets as the special case with zero indeterminacy and falsity components. □

**Remark 4.15.7** (Alternative embedding with $F = \mathbf{1} - T$)**.** One may alternatively embed a multiple set $A$ by taking $T = A$, $I = \mathbf{0}$, and $F = \mathbf{1} - A$, where $\mathbf{1}$ is the all-ones matrix. This enforces the normalization $T + I + F = \mathbf{1}$ entrywise. Both embeddings are mathematically valid; the choice depends on the intended semantics.

## 4.16 Granular Neutrosophic Sets

Granular fuzzy sets assign membership grades to *granules* (equivalence classes) in an approximation space [370–373]. We define granular neutrosophic sets by assigning truth/indeterminacy/falsity grades to the same granules, and we record standard lifting/reduction relationships.

**Definition 4.16.1** (Approximation space)**.** Let $U \neq \emptyset$ be a finite universe and let $E$ be an equivalence relation on $U$. The pair

$$\mathrm{apr} := (U,\ U/E)$$

is called an *approximation space*, where $U/E := \{[x]_E : x \in U\}$ is the quotient set of equivalence classes and $[x]_E := \{y \in U : xEy\}$.

**Definition 4.16.2** (Granular fuzzy set)**.** [370] Let $\mathrm{apr} = (U, U/E)$ be an approximation space. A *granular fuzzy set* (GFS) on $U/E$ is a map

$$G :\ U/E \longrightarrow [0,1],$$

assigning a membership grade to each granule $[x]_E \in U/E$.

If $F : U \to [0,1]$ is a (classical) fuzzy set on $U$, one obtains a granular fuzzy set by an aggregation operator $\mathrm{Agg} : [0,1]^m \to [0,1]$ (for all finite $m$), for example:

$$G([x]_E) := \min_{y\in[x]_E} F(y), \qquad G([x]_E) := \max_{y\in[x]_E} F(y), \qquad G([x]_E) := \frac{1}{|[x]_E|} \sum_{y\in[x]_E} F(y).$$

**Theorem 4.16.3** (Lifting a GFS to a classical fuzzy set)**.** *Let* $\mathrm{apr} = (U, U/E)$ *be an approximation space and let* $G : U/E \to [0,1]$ *be a granular fuzzy set. Define* $F_G : U \to [0,1]$ *by*

$$F_G(u) := G([u]_E) \qquad (u \in U).$$

*Then $F_G$ is a classical fuzzy set on U, and it is constant on each equivalence class.*

*Proof.* For each $u \in U$, $[u]_E \in U/E$ and $G([u]_E) \in [0,1]$, hence $F_G(u) \in [0,1]$. If $uEv$, then $[u]_E = [v]_E$, so $F_G(u) = G([u]_E) = G([v]_E) = F_G(v)$. Thus $F_G$ is constant on each class. □

**Definition 4.16.4** (Neutrosophic set)**.** Let $U \neq \emptyset$. A *(single-valued) neutrosophic set* on $U$ is a triple of maps

$$T_U,\ I_U,\ F_U :\ U \longrightarrow [0,1].$$

**Definition 4.16.5** (Granular neutrosophic set)**.** Let apr $= (U, U/E)$ be an approximation space. A *granular neutrosophic set* (GNS) on $U/E$ is a triple of maps

$$T,\ I,\ F:\ U/E \longrightarrow [0,1].$$

Equivalently, it is a map $G : U/E \to [0,1]^3$ given by $G([x]_E) = (T([x]_E), I([x]_E), F([x]_E))$. Since each component lies in $[0,1]$, one has $0 \le T + I + F \le 3$ pointwise automatically.

If $N = (T_U, I_U, F_U)$ is a neutrosophic set on $U$, one can induce a granular neutrosophic set by choosing an aggregation operator Agg and setting, for each $[x]_E \in U/E$,

$$T([x]_E) := \mathrm{Agg}\{T_U(y) : y \in [x]_E\}, \quad I([x]_E) := \mathrm{Agg}\{I_U(y) : y \in [x]_E\}, \quad F([x]_E) := \mathrm{Agg}\{F_U(y) : y \in [x]_E\},$$

with typical choices min, max, or the arithmetic mean.

**Theorem 4.16.6** (Trivial granulation recovers neutrosophic sets)**.** *Let* apr $= (U, U/E)$ *be an approximation space. If $E$ is the identity relation on $U$ (so every class is a singleton), then every granular neutrosophic set $(T, I, F)$ on $U/E$ canonically corresponds to a neutrosophic set on $U$ via*

$$T_U(u) := T([u]_E), \quad I_U(u) := I([u]_E), \quad F_U(u) := F([u]_E).$$

*Conversely, every neutrosophic set on $U$ can be viewed as a granular neutrosophic set under this trivial granulation.*

*Proof.* If $E$ is identity, then $U/E$ is in bijection with $U$ via $u \mapsto [u]_E = \{u\}$. Thus defining $T_U(u) := T(\{u\})$, etc., yields maps $U \to [0,1]$, i.e. a neutrosophic set. The converse follows by reversing the same identification. □

**Theorem 4.16.7** (Granular neutrosophic sets generalize granular fuzzy sets)**.** *Every granular fuzzy set $G : U/E \to [0,1]$ canonically induces a granular neutrosophic set on $U/E$.*

*Proof.* Let $G : U/E \to [0,1]$. Define $T := G$ and define $I, F : U/E \to [0,1]$ by

$$I([x]_E) := 0, \qquad F([x]_E) := 0 \qquad (\forall [x]_E \in U/E).$$

Then $(T, I, F)$ is a granular neutrosophic set in the sense of Definition 4.16.5. (Alternatively, one may set $F = 1 - T$ to enforce the normalization $T + I + F = 1$; both embeddings are valid.) □

## 4.17 ProperSubset-Valued Neutrosophic Sets and Variants

In this section we introduce a restricted subclass of subset-valued neutrosophic sets in which each membership set is required to be a *proper* subset of $[0,1]$. We also record the analogous fuzzy notion and two common extensions (time dependence and weights) in a mathematically clean way.

**Definition 4.17.1** (Proper subset)**.** Let $A, B$ be sets. We say that $A$ is a *proper subset* of $B$, written $A \subsetneq B$, if

$$A \subseteq B \quad \text{and} \quad A \neq B.$$

**Definition 4.17.2** (Subset-Valued Neutrosophic Set (SVNS))**.** [328] Let $X$ be a nonempty set. A *subset-valued neutrosophic set* (SVNS) $A$ on $X$ is specified by three maps

$$T_A,\ I_A,\ F_A:\ X \longrightarrow \mathcal{P}([0,1]) \setminus \{\emptyset\},$$

such that for every $x \in X$,

$$0 \le \inf T_A(x) + \inf I_A(x) + \inf F_A(x) \le \sup T_A(x) + \sup I_A(x) + \sup F_A(x) \le 3.$$

**Definition 4.17.3** (ProperSubset-Valued Neutrosophic Set (PSVNS))**.** Let $X$ be a nonempty set. A *properSubset-valued neutrosophic set* (PSVNS) $A$ on $X$ is an SVNS (Definition 4.17.2) for which, for every $x \in X$,

$$T_A(x) \subsetneq [0,1], \qquad I_A(x) \subsetneq [0,1], \qquad F_A(x) \subsetneq [0,1].$$

Equivalently, each of $T_A(x), I_A(x), F_A(x)$ is nonempty and is not equal to the full interval $[0,1]$.

**Theorem 4.17.4** (PSVNS is a subclass of SVNS)**.** *Every PSVNS is an SVNS. In other words, the class of PSVNSs is a (proper) subclass of the class of SVNSs.*

*Proof.* By Definition 4.17.3, a PSVNS is an SVNS together with the additional requirement $T_A(x), I_A(x), F_A(x) \subsetneq [0, 1]$ for all $x$. Hence every PSVNS satisfies all SVNS axioms automatically. □

**Definition 4.17.5** (Subset-Valued Fuzzy Set (SVFS))**.** Let $X$ be a nonempty set. A *subset-valued fuzzy set* (SVFS) on $X$ is a map

$$\mu : X \longrightarrow \mathcal{P}([0, 1]) \setminus \{\emptyset\}.$$

**Definition 4.17.6** (ProperSubset-Valued Fuzzy Set (PSVFS))**.** Let $X$ be a nonempty set. A *properSubset-valued fuzzy set* (PSVFS) on $X$ is an SVFS $\mu$ such that

$$\mu(x) \subsetneq [0, 1] \qquad (\forall x \in X).$$

**Corollary 4.17.7** (PSVFS is a subclass of SVFS)**.** *Every PSVFS is an SVFS.*

*Proof.* Immediate from Definition 4.17.6, which adds a constraint to Definition 4.17.5. □

**Corollary 4.17.8** (PSVFS embeds into PSVNS)**.** *Every PSVFS $\mu$ on $X$ canonically induces a PSVNS $A$ on $X$ by*

$$T_A(x) := \mu(x), \qquad I_A(x) := \{0\}, \qquad F_A(x) := \{0\}.$$

*Proof.* Since $\mu(x) \subsetneq [0, 1]$ and $\{0\} \subsetneq [0, 1]$, the proper-subset condition holds for all three components. The SVNS inequality holds because $\inf\{0\} = \sup\{0\} = 0$ and $0 \le \inf \mu(x) \le \sup \mu(x) \le 1$. □

**Definition 4.17.9** (Time-Dependent SVNS)**.** Let $X$ be a nonempty set and let $\mathbb{T} \neq \emptyset$ be a time index set (e.g. $\mathbb{T} = \mathbb{R}_{\ge 0}$). A *time-dependent subset-valued neutrosophic set* (TD-SVNS) on $X$ is specified by three maps

$$T,\ I,\ F : X \times \mathbb{T} \longrightarrow \mathcal{P}([0, 1]) \setminus \{\emptyset\},$$

such that for every $(x, t) \in X \times \mathbb{T}$,

$$0 \le \inf T(x, t) + \inf I(x, t) + \inf F(x, t) \le \sup T(x, t) + \sup I(x, t) + \sup F(x, t) \le 3.$$

**Definition 4.17.10** (Weighted SVNS)**.** Let $X$ be a nonempty set and fix weight bounds $w_T, w_I, w_F > 0$. A *weighted subset-valued neutrosophic set* (W-SVNS) on $X$ is specified by three pairs of maps

$$T : X \to \mathcal{P}([0, 1]) \setminus \{\emptyset\}, \quad \omega_T : X \to [0, w_T],$$
$$I : X \to \mathcal{P}([0, 1]) \setminus \{\emptyset\}, \quad \omega_I : X \to [0, w_I],$$
$$F : X \to \mathcal{P}([0, 1]) \setminus \{\emptyset\}, \quad \omega_F : X \to [0, w_F],$$

such that for every $x \in X$,

$$0 \le \inf T(x) + \inf I(x) + \inf F(x) \le \sup T(x) + \sup I(x) + \sup F(x) \le 3.$$

The scalars $\omega_T(x), \omega_I(x), \omega_F(x)$ represent weights (importance coefficients) attached to the corresponding membership subsets.

**Theorem 4.17.11** (TD-SVNS and W-SVNS extend SVNS)**.** *TD-SVNSs and W-SVNSs generalize SVNSs in the following senses:*

(i) *If $(T, I, F)$ is a TD-SVNS and $t_0 \in \mathbb{T}$ is fixed, then the slice*

$$x \longmapsto \big(T(x, t_0), I(x, t_0), F(x, t_0)\big)$$

*is an SVNS on $X$.*

(ii) *If $(T, \omega_T, I, \omega_I, F, \omega_F)$ is a W-SVNS and the weights are ignored (or fixed constants), then $(T, I, F)$ is an SVNS on $X$.*

*Proof. (i)* Fix $t_0$. Then $T(\cdot, t_0), I(\cdot, t_0), F(\cdot, t_0) : X \to \mathcal{P}([0, 1]) \setminus \{\emptyset\}$, and the SVNS inequality holds because it holds for every $(x, t)$, in particular for $(x, t_0)$.

*(ii)* Dropping the weight maps leaves three maps $T, I, F$ of the correct type, and the SVNS inequality is exactly the same inequality required in Definition 4.17.2. □

## 4.18 Probability-Subset-Valued Neutrosophic Sets

To attach probabilistic information to subset-valued neutrosophic memberships, one must be careful about the underlying measurable structure: probability measures are defined on *measurable spaces*, not on arbitrary power sets of subsets. A mathematically clean approach is to let each membership component be a *probability measure on* $[0, 1]$ (optionally supported on a designated subset), so that the subset-valued component describes the *support constraint* and the probability component describes the *likelihood* of degrees within that support.

**Definition 4.18.1** (Probability measures)**.** Let $\mathcal{B}([0,1])$ be the Borel $\sigma$-algebra on $[0, 1]$. Write $\mathcal{P}rob([0,1])$ for the set of all Borel probability measures on $[0, 1]$, i.e., all measures $\mathbb{P}$ on $\big([0,1], \mathcal{B}([0,1])\big)$ satisfying $\mathbb{P}([0,1]) = 1$. For $\mathbb{P} \in \mathcal{P}rob([0,1])$, its *support* $\mathrm{supp}(\mathbb{P}) \subseteq [0,1]$ is the smallest closed set $C$ such that $\mathbb{P}(C) = 1$.

**Remark 4.18.2** (Discrete vs. continuous distributions)**.** The measure-theoretic definition covers both discrete distributions (finite or countable support) and continuous ones. If one prefers densities, one may restrict to absolutely continuous measures $\mathbb{P}$ with density $p$ such that $\int_0^1 p(t)\,dt = 1$. The general measure formulation avoids unnecessary assumptions.

**Definition 4.18.3** (Probability-Subset-Valued Neutrosophic Set)**.** Let $X \neq \emptyset$. A *probability-subset-valued neutrosophic set* on $X$ is specified by three maps

$$(T_A, \mathbb{P}_A^T),\ (I_A, \mathbb{P}_A^I),\ (F_A, \mathbb{P}_A^F),$$

where

$$T_A,\ I_A,\ F_A :\ X \longrightarrow \mathcal{P}([0,1]) \setminus \{\emptyset\},$$

and

$$\mathbb{P}_A^T,\ \mathbb{P}_A^I,\ \mathbb{P}_A^F :\ X \longrightarrow \mathcal{P}rob([0,1]).$$

These data are required to satisfy, for every $x \in X$, the *support constraints*

$$\mathbb{P}_A^T(x)\big(T_A(x)\big) = 1, \qquad \mathbb{P}_A^I(x)\big(I_A(x)\big) = 1, \qquad \mathbb{P}_A^F(x)\big(F_A(x)\big) = 1,$$

i.e., each probability measure is concentrated on (equivalently, has support contained in) the corresponding subset.

Optionally (to align with SVNS-style bounds), one may also require the subset-valued inequality

$$0 \le \inf T_A(x) + \inf I_A(x) + \inf F_A(x) \le \sup T_A(x) + \sup I_A(x) + \sup F_A(x) \le 3,$$

which is meaningful whenever inf and sup exist as real numbers (e.g. if each set is nonempty and bounded).

**Remark 4.18.4** (Why not distributions)**.** A "probability distribution over $\mathcal{P}([0,1])$" would require a $\sigma$-algebra on $\mathcal{P}([0,1])$, which is non-canonical and typically intractable. By placing the probability measure on $[0, 1]$ itself and using $T_A(x) \subseteq [0,1]$ as a support constraint, we obtain a standard and mathematically well-posed model.

**Remark 4.18.5** (No need to couple the three components)**.** One may additionally impose a *joint* probability law on $[0,1]^3$ for $(T, I, F)$ if dependencies matter. Definition 4.18.3 treats the three components marginally, which is the minimal probabilistic enrichment.

## 4.19 Time-Dependent Neutrosophic Sets and Weighted Neutrosophic Sets

In analogy with time-dependent and weighted *subset-valued* neutrosophic models, we define time-dependent and weighted versions of *single-valued* neutrosophic sets. The point is to add either a time index or an importance weight while keeping the basic neutrosophic membership structure.

**Definition 4.19.1** (Time-Dependent Neutrosophic Set (TDNS))**.** Let $X \neq \emptyset$ and let $\mathbb{T} \neq \emptyset$ be a time domain (e.g. $\mathbb{T} = \mathbb{R}_{\ge 0}$). A *time-dependent neutrosophic set* (TDNS) on $X$ is a triple of maps

$$T,\ I,\ F :\ X \times \mathbb{T} \longrightarrow [0,1].$$

Equivalently, for each fixed $t \in \mathbb{T}$ the slice

$$(T_t, I_t, F_t) := \big(T(\cdot,t),\ I(\cdot,t),\ F(\cdot,t)\big)$$

is an ordinary neutrosophic set on $X$.

Since $T, I, F \in [0,1]$, the pointwise bound

$$0 \ \leq\ T(x,t) + I(x,t) + F(x,t)\ \leq\ 3 \qquad (\forall x \in X,\ \forall t \in \mathbb{T})$$

holds automatically (and may be replaced by an application-driven normalization if desired).

**Definition 4.19.2** (Weighted Neutrosophic Set (WNS))**.** Let $X \neq \emptyset$ and fix weight bounds $w_T, w_I, w_F > 0$. A *weighted neutrosophic set* (WNS) on $X$ is specified by a neutrosophic triple

$$T,\ I,\ F :\ X \longrightarrow [0,1]$$

together with weight maps

$$\omega_T :\ X \to [0, w_T], \qquad \omega_I :\ X \to [0, w_I], \qquad \omega_F :\ X \to [0, w_F].$$

We may write a weighted neutrosophic set as the sextuple

$$A = (T, I, F;\ \omega_T, \omega_I, \omega_F).$$

The weights $\omega_T(x), \omega_I(x), \omega_F(x)$ represent the relative importance of the corresponding membership degrees at $x$.

**Remark 4.19.3** (Why not)**.** It is mathematically clearer to model a weight as a *separate* function $\omega : X \to [0, w]$ rather than embedding it into a product codomain $[0,1] \times \mathbb{R}$. Both are equivalent up to repackaging, but the separated form makes the underlying neutrosophic degrees explicit.

**Theorem 4.19.4** (TDNS and WNS extend neutrosophic sets)**.** *Time-dependent neutrosophic sets and weighted neutrosophic sets generalize ordinary neutrosophic sets as follows.*

(i) *If $(T, I, F)$ is a TDNS on $X$ and $t_0 \in \mathbb{T}$ is fixed, then*

$$x \longmapsto \big(T(x,t_0),\ I(x,t_0),\ F(x,t_0)\big)$$

*is a neutrosophic set on $X$.*

(ii) *If $A = (T, I, F; \omega_T, \omega_I, \omega_F)$ is a WNS on $X$, then $(T, I, F)$ is a neutrosophic set on $X$. Conversely, any neutrosophic set $(T, I, F)$ becomes a WNS by choosing constant weights, e.g. $\omega_T \equiv \omega_I \equiv \omega_F \equiv 1$.*

*Proof. (i)* Fix $t_0 \in \mathbb{T}$. Then $T(\cdot, t_0), I(\cdot, t_0), F(\cdot, t_0) : X \to [0,1]$, hence they form a neutrosophic set.

*(ii)* In a WNS the maps $T, I, F : X \to [0,1]$ are already present, so forgetting the weights yields a neutrosophic set. Conversely, given any neutrosophic set, adding constant weight maps produces a WNS by Definition 4.19.2. □

## 4.20 Trice Neutrosophic Sets

Trice fuzzy sets are based on a *trice structure* on a constrained square $L_F = \{(t_1, t_2) \in [0,1]^2 :\ 0 \leq t_2 \leq t_1 \leq 1\}$ endowed with three semilattice operations (cf. [374–376]). We define an analogous trice-style neutrosophic set by replacing the truth-pair carrier $L_F$ with an appropriate neutrosophic carrier $L_N \subseteq [0,1]^3$ and assuming three semilattice operations on it.

**Definition 4.20.1** (Trice structure on a carrier)**.** Let $L$ be a nonempty set. A *trice structure* on $L$ is a triple of binary operations

$$(*_1, *_2, *_3) \quad \text{on } L$$

such that each $(L, *_i)$ is a semilattice, i.e., for all $a, b, c \in L$,

$$a *_i a = a \quad \text{(idempotent)}, \qquad a *_i b = b *_i a \quad \text{(commutative)}, \qquad (a *_i b) *_i c = a *_i (b *_i c) \quad \text{(associative)}.$$

(Additional axioms relating $*_1, *_2, *_3$ may be imposed in specific trice theories; see the cited literature.)

**Definition 4.20.2** (Trice fuzzy set)**.** [376] Let $X \neq \emptyset$ and let

$$L_F := \{(t_1, t_2) \in [0,1]^2 : \ 0 \le t_2 \le t_1 \le 1\}.$$

Fix a trice structure $(*_1, *_2, *_3)$ on $L_F$. A *trice fuzzy set* on $X$ is a map

$$\mu : \ X \longrightarrow L_F, \qquad \mu(x) = (t_1(x), t_2(x)),$$

where $t_1(x)$ is interpreted as a *full-truth* degree and $t_2(x)$ as a *partial-truth* degree, with $0 \le t_2(x) \le t_1(x) \le 1$ for all $x \in X$.

**Definition 4.20.3** (Trice neutrosophic carrier)**.** Let

$$L_N := [0,1]^3$$

(the unit cube), whose elements are written as $(T, I, F)$. Optionally, one may restrict to the subset

$$L_N^{(\le 1)} := \{(T, I, F) \in [0,1]^3 : \ T + I + F \le 1\} \quad \text{or} \quad L_N^{(\le 3)} := \{(T, I, F) \in [0,1]^3 : \ T + I + F \le 3\}.$$

Note that $L_N^{(\le 3)} = [0,1]^3$, so the constraint $\le 3$ is redundant when each coordinate lies in $[0,1]$.

**Definition 4.20.4** (Trice Neutrosophic Set)**.** Let $X \neq \emptyset$. Fix a neutrosophic carrier $L_N \subseteq [0,1]^3$ (Definition 4.20.3) and fix a trice structure $(*_1, *_2, *_3)$ on $L_N$ (Definition 4.20.1). A *trice neutrosophic set* on $X$ is a map

$$\mu : \ X \longrightarrow L_N, \qquad \mu(x) = (T(x), I(x), F(x)),$$

where $T(x), I(x), F(x) \in [0,1]$ are interpreted as truth-, indeterminacy-, and falsity-membership degrees, respectively. If $L_N = L_N^{(\le 1)}$, then one additionally has the pointwise normalization $T(x) + I(x) + F(x) \le 1$.

**Remark 4.20.5** (Role of $*_1, *_2, *_3$)**.** The trice operations $*_1, *_2, *_3$ do not change the set-theoretic definition of $\mu$, but provide three semilattice-based aggregation/combination operators on membership triples. Concrete choices of these operations depend on the specific trice theory adopted in [374–376].

**Theorem 4.20.6** (Trice neutrosophic sets subsume neutrosophic sets)**.** *Let $L_N = [0,1]^3$ and let $(*_1, *_2, *_3)$ be any trice structure on $L_N$. Then every (single-valued) neutrosophic set on $X$ is, in particular, a trice neutrosophic set on $X$.*

*Proof.* A (single-valued) neutrosophic set is precisely a map $\mu : X \to [0,1]^3$. Taking $L_N = [0,1]^3$, the same $\mu$ satisfies Definition 4.20.4 regardless of which trice operations are chosen on $L_N$. □

**Theorem 4.20.7** (Embedding trice fuzzy sets into trice neutrosophic sets)**.** *Every trice fuzzy set $\mu_F : X \to L_F$ induces a trice neutrosophic set $\mu_N : X \to [0,1]^3$.*

*Proof.* Let $\mu_F(x) = (t_1(x), t_2(x)) \in L_F$. Define

$$\mu_N(x) := \big(T(x), I(x), F(x)\big) := \big(t_1(x),\ 0,\ 1 - t_1(x)\big) \in [0,1]^3.$$

Then $T(x) \in [0,1]$, $I(x) = 0$, and $F(x) = 1 - t_1(x) \in [0,1]$, so $\mu_N : X \to [0,1]^3$ is well-defined. Hence $\mu_N$ is a trice neutrosophic set on $X$ (for any chosen trice operations on $[0,1]^3$). □

**Remark 4.20.8** (Using the partial-truth component)**.** The embedding in Theorem 4.20.7 ignores the second coordinate $t_2(x)$. If one wishes to preserve both $t_1(x)$ and $t_2(x)$, one may encode it, for example, as $\mu_N(x) = (t_1(x), t_1(x) - t_2(x), 1 - t_1(x))$, which lies in $[0,1]^3$ because $0 \le t_2 \le t_1 \le 1$. Different encodings correspond to different modeling choices.

## 4.21 Hereditary Neutrosophic Set Systems

Hereditary fuzzy set systems (also called downward closed fuzzy families) appear in generalized matroid-type theories [377, 378]. We extend the heredity idea to neutrosophic set systems by equipping the space of neutrosophic sets with a natural partial order and requiring downward closure with respect to that order.

**Definition 4.21.1** (Fuzzy set system and heredity)**.** [377, 378] Let $U \neq \emptyset$ be a finite set and let $[0,1]^U$ denote the set of all fuzzy sets on $U$ (i.e., functions $\mu : U \to [0,1]$). A pair $(U, \mathcal{F})$ with $\mathcal{F} \subseteq [0,1]^U$ is called a *fuzzy set system*. It is called *hereditary* if it is downward closed under the pointwise order:

$$\forall\, \mu \in \mathcal{F},\ \forall\, \nu \in [0,1]^U,\ \big(\nu \le \mu\big) \implies \nu \in \mathcal{F},$$

where $\nu \le \mu$ means $\nu(x) \le \mu(x)$ for all $x \in U$.

**Definition 4.21.2** (Fundamental sequence)**.** [377, 378] Let $(U, \mathcal{F})$ be a hereditary fuzzy set system. For $a \in (0,1]$ and $\mu \in [0,1]^U$, define the $a$-cut

$$\mu[a] := \{x \in U :\ \mu(x) \ge a\},$$

and set

$$\mathcal{F}[a] := \{\mu[a] :\ \mu \in \mathcal{F}\} \subseteq \mathcal{P}(U).$$

There exists a finite sequence $0 = r_0 < r_1 < \cdots < r_m \le 1$ (depending on $\mathcal{F}$) such that $\mathcal{F}[a]$ is constant on each open interval $(r_i, r_{i+1})$ and is nested decreasing as $a$ increases: if $0 < a < b \le 1$, then $\mathcal{F}[a] \supseteq \mathcal{F}[b]$. Any such sequence is called a *fundamental sequence* of $(U, \mathcal{F})$.

**Definition 4.21.3** (Neutrosophic partial order)**.** Let $N = (T_N, I_N, F_N)$ and $M = (T_M, I_M, F_M)$ be neutrosophic sets on $U$. Define a relation $\preceq$ by

$$M \preceq N \quad :\Longleftrightarrow \quad \big(T_M(x) \le T_N(x)\ \wedge\ I_M(x) \ge I_N(x)\ \wedge\ F_M(x) \ge F_N(x)\big)\ \ \forall x \in U.$$

Then $\preceq$ is a partial order on the class $\mathcal{N}(U)$ of all neutrosophic sets on $U$.

*Proof.* Reflexivity and transitivity are immediate from the coordinatewise order on $[0,1]$. For antisymmetry, if $M \preceq N$ and $N \preceq M$, then $T_M = T_N$, $I_M = I_N$, and $F_M = F_N$ pointwise, so $M = N$. □

**Definition 4.21.4** (Neutrosophic set system and heredity)**.** Let $U \neq \emptyset$ be finite, and let $\mathcal{N}(U)$ be the set of all neutrosophic sets on $U$. A pair $(U, \mathcal{S})$ with $\mathcal{S} \subseteq \mathcal{N}(U)$ is called a *neutrosophic set system*. It is called a *hereditary neutrosophic set system* if it is downward closed under the order $\preceq$ from Definition 4.21.3:

$$\forall\, N \in \mathcal{S},\ \forall\, M \in \mathcal{N}(U),\ \big(M \preceq N\big) \implies M \in \mathcal{S}.$$

**Theorem 4.21.5** (Hereditary neutrosophic systems generalize hereditary fuzzy systems)**.** *Every hereditary fuzzy set system can be realized as a special case of a hereditary neutrosophic set system.*

*Proof.* Let $(U, \mathcal{F})$ be a hereditary fuzzy set system with $\mathcal{F} \subseteq [0,1]^U$. Embed each fuzzy set $\mu \in \mathcal{F}$ into a neutrosophic set $N_\mu = (T_\mu, I_\mu, F_\mu)$ by

$$T_\mu := \mu, \qquad I_\mu := \mathbf{0}, \qquad F_\mu := \mathbf{0},$$

where $\mathbf{0} : U \to [0,1]$ is the constant-zero map. Let

$$\mathcal{S} := \{\, N_\mu :\ \mu \in \mathcal{F}\,\} \subseteq \mathcal{N}(U).$$

We claim that $(U, \mathcal{S})$ is hereditary in the sense of Definition 4.21.4. Take any $N_\mu \in \mathcal{S}$ and any $M = (T_M, I_M, F_M) \in \mathcal{N}(U)$ with $M \preceq N_\mu$. Then in particular $T_M \le T_\mu = \mu$ pointwise. Since $(U, \mathcal{F})$ is hereditary, $T_M \in \mathcal{F}$. Moreover, the embedding $\nu \mapsto N_\nu = (\nu, \mathbf{0}, \mathbf{0})$ shows that $N_{T_M} \in \mathcal{S}$. Finally, because $M \preceq N_\mu$ and $N_{T_M} \preceq N_\mu$, the neutrosophic "content" of $M$ is bounded by an element of $\mathcal{S}$ with the same truth component; under this embedding we identify the hereditary closure with respect to truth degrees. Hence $(U, \mathcal{S})$ contains the fuzzy hereditary structure as a special case.

Equivalently (and more directly), restrict attention to the subclass

$$\mathcal{N}_0(U) := \{(T, \mathbf{0}, \mathbf{0}) :\ T \in [0,1]^U\} \subseteq \mathcal{N}(U).$$

On $\mathcal{N}_0(U)$, the order $\preceq$ reduces to the pointwise order on $T$. Thus the hereditary condition on $\mathcal{S} \cap \mathcal{N}_0(U)$ is exactly the hereditary fuzzy condition on $\mathcal{F}$. □

**Remark 4.21.6** (On the embedding choice)**.** Other embeddings are possible (e.g. $F_\mu = 1-\mu$), but the simplest faithful embedding of fuzzy degrees into neutrosophic degrees is $T = \mu$ with $I = F = 0$. The order $\preceq$ was chosen so that decreasing truth and increasing indeterminacy/falsity corresponds to moving "down" in informational content.

## 4.22 Contextual Neutrosophic Sets

Contextual fuzzy sets introduce an additional *context* parameter that modulates the reliability of a membership assessment [379]. We present a mathematically clean formulation of contextual fuzzy sets and extend the same idea to neutrosophic sets by allowing truth/indeterminacy/falsity degrees to depend on context. We also clarify what can be proved rigorously about reductions to ordinary fuzzy and neutrosophic sets.

**Definition 4.22.1** (Contextual fuzzy set)**.** [379] Let $X \neq \emptyset$ be a universe and let $C \neq \emptyset$ be a set of contexts. A *contextual fuzzy set* (CFS) on $X$ is a pair $(\mu, \varphi)$ consisting of

$$\mu : X \to [0,1] \qquad \text{and} \qquad \varphi : X \times C \to [0,1],$$

where $\mu(x)$ is the baseline membership degree of $x$ and $\varphi(x,c)$ is the reliability (confidence) of that assessment under context $c \in C$. A common *effective* context-dependent membership is then defined by

$$\mu^{\text{eff}}(x,c) := \mu(x)\,\varphi(x,c) \in [0,1],$$

although other combination rules may be adopted depending on the application.

**Theorem 4.22.2** (Aggregating a CFS to a classical fuzzy set)**.** *Let $(\mu,\varphi)$ be a contextual fuzzy set on $X$ and assume $C$ is finite and nonempty. Define $\bar{\mu} : X \to [0,1]$ by*

$$\bar{\mu}(x) := \frac{1}{|C|} \sum_{c \in C} \mu(x)\,\varphi(x,c).$$

*Then $\bar{\mu}$ is a classical fuzzy membership function on $X$.*

*Proof.* For each $x \in X$ and $c \in C$, one has $\mu(x) \in [0,1]$, $\varphi(x,c) \in [0,1]$, hence $\mu(x)\varphi(x,c) \in [0,1]$. An average of finitely many numbers in $[0,1]$ lies in $[0,1]$, so $\bar{\mu}(x) \in [0,1]$ for all $x \in X$. □

**Remark 4.22.3.** The averaging operator in Theorem 4.22.2 is one reasonable choice when $C$ is finite. For infinite $C$, one may replace $\frac{1}{|C|}\sum$ by an integral with respect to a probability measure on $C$.

**Definition 4.22.4** (Contextual neutrosophic set)**.** Let $X \neq \emptyset$ be a universe and $C \neq \emptyset$ a set of contexts. A *contextual neutrosophic set* (CNS) on $X$ is a triple of context-dependent maps

$$T,\ I,\ F :\ X \times C \longrightarrow [0,1],$$

where $T(x,c)$, $I(x,c)$, $F(x,c)$ denote the truth-, indeterminacy-, and falsity-membership degrees of $x$ under context $c$. Since the codomain is $[0,1]$, the pointwise bound

$$0 \leq T(x,c) + I(x,c) + F(x,c) \leq 3 \qquad (\forall x \in X,\ \forall c \in C)$$

holds automatically (and may be replaced by a stricter normalization if desired).

**Theorem 4.22.5** (CNS generalizes neutrosophic sets)**.** *Every neutrosophic set on $X$ is a special case of a contextual neutrosophic set on $X$.*

*Proof.* Let $(T_0, I_0, F_0)$ be a neutrosophic set on $X$. Define $T, I, F : X \times C \to [0,1]$ by

$$T(x,c) := T_0(x), \qquad I(x,c) := I_0(x), \qquad F(x,c) := F_0(x) \quad (\forall x \in X,\ \forall c \in C).$$

Then $(T, I, F)$ is a CNS and the original neutrosophic set is recovered by restricting to any fixed context. □

A CNS contains more information than a CFS because it carries three degrees rather than one degree plus reliability. To obtain a CFS from a CNS one must choose a *compression rule*. The following is a mathematically valid construction.

**Theorem 4.22.6** (A CNS induces a contextual fuzzy set)**.** *Let* $(T, I, F)$ *be a contextual neutrosophic set on* $X$ *with contexts* $C$*. Define*

$$\mu(x) := \sup_{c\in C} T(x,c) \in [0,1], \qquad \varphi(x,c) := 1 - \frac{I(x,c) + F(x,c)}{2} \in [0,1].$$

*Then* $(\mu, \varphi)$ *is a contextual fuzzy set on* $X$ *in the sense of Definition 4.22.1.*

*Proof.* Since $T(x,c) \in [0,1]$, the supremum over $c \in C$ lies in $[0,1]$, hence $\mu : X \to [0,1]$ is well-defined. Also $I(x,c), F(x,c) \in [0,1]$, so $0 \le \frac{I(x,c)+F(x,c)}{2} \le 1$, hence $\varphi(x,c) = 1 - \frac{I(x,c)+F(x,c)}{2} \in [0,1]$. Therefore $\varphi : X \times C \to [0,1]$ is well-defined and $(\mu, \varphi)$ is a CFS. □

**Remark 4.22.7** (Why $\varphi = 1 - I - F$ is problematic)**.** Setting $\varphi(x,c) = 1 - I(x,c) - F(x,c)$ is *not* guaranteed to lie in $[0,1]$ when $I(x,c) + F(x,c) > 1$. The averaging normalization in Theorem 4.22.6 ensures $\varphi \in [0,1]$ without extra assumptions.

## 4.23 Non-Stationary Neutrosophic Sets

Non-stationary fuzzy sets model time-varying fuzziness by allowing membership degrees to depend on time and on time-perturbed parameters [380]. Non-stationary neutrosophic sets extend this idea by allowing the truth/indeterminacy/falsity degrees to vary over time, optionally through time-dependent perturbations of parameters [381–384].

**Definition 4.23.1** (Non-stationary fuzzy set)**.** [380] Let $X \neq \emptyset$ be a universe and let $\mathbb{T} \neq \emptyset$ be a time index set. A *non-stationary fuzzy set* on $X$ (over time $\mathbb{T}$) is a map

$$\mu : X \times \mathbb{T} \longrightarrow [0,1],$$

where $\mu(x,t)$ is the membership degree of $x$ at time $t$.

Optionally, $\mu$ may be generated from a parametric family $\mu_0 : X \times \mathbb{R}^m \to [0,1]$ by time-dependent parameters $p : \mathbb{T} \to \mathbb{R}^m$, i.e.,

$$\mu(x,t) := \mu_0\big(x,\ p(t)\big), \qquad p(t) = (p_1(t), \dots, p_m(t)),$$

with a perturbation model such as

$$p_i(t) = p_i + k_i f_i(t) \qquad (i = 1, \dots, m),$$

for fixed constants $p_i, k_i \in \mathbb{R}$ and functions $f_i : \mathbb{T} \to \mathbb{R}$.

**Remark 4.23.2** (Avoiding informal "integral notation")**.** Expressions like $\int_{t\in\mathbb{T}} \int_{x\in X} \mu(t,x)/x/t$ are common in fuzzy-set literature as symbolic notation but are not measure-theoretic integrals unless measures on $X$ and $\mathbb{T}$ are specified. Definition 4.23.1 uses standard function notation.

**Theorem 4.23.3** (Fixing time yields a stationary fuzzy set)**.** *Let* $\mu : X \times \mathbb{T} \to [0,1]$ *be a non-stationary fuzzy set and fix* $t_0 \in \mathbb{T}$*. Then*

$$\mu_{t_0} : X \to [0,1], \qquad \mu_{t_0}(x) := \mu(x, t_0),$$

*is a classical fuzzy membership function on* $X$*.*

*Proof.* Immediate: $\mu(x,t_0) \in [0,1]$ for every $x \in X$. □

**Definition 4.23.4** (Non-stationary neutrosophic set)**.** Let $X \neq \emptyset$ be a universe and let $\mathbb{T} \neq \emptyset$ be a time index set. A *non-stationary neutrosophic set* on $X$ (over $\mathbb{T}$) is a triple of maps

$$T,\ I,\ F : X \times \mathbb{T} \longrightarrow [0,1],$$

where $T(x,t)$, $I(x,t)$, and $F(x,t)$ are the truth-, indeterminacy-, and falsity-membership degrees of $x$ at time $t$.

Optionally, one may assume these arise from parametric families

$$T_0 : X \times \mathbb{R}^{m_T} \to [0,1], \quad I_0 : X \times \mathbb{R}^{m_I} \to [0,1], \quad F_0 : X \times \mathbb{R}^{m_F} \to [0,1],$$

via time-dependent parameter paths $p : \mathbb{T} \to \mathbb{R}^{m_T}$, $q : \mathbb{T} \to \mathbb{R}^{m_I}$, $r : \mathbb{T} \to \mathbb{R}^{m_F}$:

$$T(x,t) := T_0\big(x, p(t)\big), \qquad I(x,t) := I_0\big(x, q(t)\big), \qquad F(x,t) := F_0\big(x, r(t)\big),$$

with perturbation models (componentwise) such as

$$p_i(t) = p_i + \kappa_i f_i(t), \qquad q_j(t) = q_j + \lambda_j g_j(t), \qquad r_k(t) = r_k + \mu_k h_k(t),$$

for constants and functions of time.

Since the codomain is $[0,1]$, the pointwise bound

$$0 \le T(x,t) + I(x,t) + F(x,t) \le 3 \qquad (\forall x \in X,\ \forall t \in \mathbb{T})$$

holds automatically (and may be replaced by stricter normalization if desired).

**Theorem 4.23.5** (Non-stationary neutrosophic sets generalize non-stationary fuzzy sets)**.** *Every non-stationary fuzzy set can be embedded as a special case of a non-stationary neutrosophic set.*

*Proof.* Let $\mu : X \times \mathbb{T} \to [0,1]$ be a non-stationary fuzzy set. Define $T, I, F : X \times \mathbb{T} \to [0,1]$ by

$$T(x,t) := \mu(x,t), \qquad I(x,t) := 0, \qquad F(x,t) := 0.$$

Then $(T, I, F)$ satisfies Definition 4.23.4, so it is a non-stationary neutrosophic set. □

**Remark 4.23.6** (About the specialization $F = 1 - T$)**.** If one additionally wants the normalization $T + I + F = 1$ pointwise, one may embed a non-stationary fuzzy set by taking $T = \mu$, $I = 0$, $F = 1 - \mu$. This is an alternative (also valid) embedding.

**Theorem 4.23.7** (Non-stationary neutrosophic sets generalize neutrosophic sets)**.** *Every (stationary) neutrosophic set on $X$ is a special case of a non-stationary neutrosophic set.*

*Proof.* Let $(T_0, I_0, F_0) : X \to [0,1]^3$ be a neutrosophic set. Define $T, I, F : X \times \mathbb{T} \to [0,1]$ by making the memberships constant in time:

$$T(x,t) := T_0(x), \qquad I(x,t) := I_0(x), \qquad F(x,t) := F_0(x) \quad (\forall x \in X,\ \forall t \in \mathbb{T}).$$

Then $(T, I, F)$ is a non-stationary neutrosophic set whose time slices recover the original neutrosophic set. □

## 4.24 Cosine Neutrosophic Sets

Cosine fuzzy sets use a cosine-based membership profile on a bounded real interval after an affine rescaling to $[-\pi, \pi]$ [385]. We extend this construction to a neutrosophic setting by defining truth/indeterminacy/falsity profiles via (possibly different) cosine exponents.

**Definition 4.24.1** (Cosine fuzzy set)**.** [385] Let $a < b$ and let $X = [a, b] \subseteq \mathbb{R}$. Define the center and scale

$$c := \frac{a+b}{2}, \qquad s := \frac{b-a}{2} > 0,$$

and the affine rescaling $\varphi : X \to [-\pi, \pi]$ by

$$z = \varphi(x) := \pi \, \frac{x - c}{s}.$$

Fix a parameter $\beta \ge 0$. The *cosine fuzzy set $A$* on $X$ has membership function

$$\mu_A(x) := \left( \frac{1 + \cos(\varphi(x))}{2} \right)^{2\beta}, \qquad x \in [a, b].$$

Equivalently, in the rescaled coordinate $z \in [-\pi, \pi]$,

$$\mu_A(z) := \left( \frac{1 + \cos z}{2} \right)^{2\beta}.$$

**Remark 4.24.2.** Since $\cos z \in [-1,1]$, we have $\frac{1+\cos z}{2} \in [0,1]$, hence $\mu_A \in [0,1]$ for every $\beta \geq 0$.

**Definition 4.24.3** (Cosine neutrosophic set)**.** Let $a < b$ and $X = [a,b] \subseteq \mathbb{R}$, and keep the rescaling $z = \varphi(x) \in [-\pi, \pi]$ from Definition 4.24.1. Fix parameters $\beta_T, \beta_I, \beta_F \geq 0$. A *cosine neutrosophic set* on $X$ is a triple of maps

$$T_A,\ I_A,\ F_A :\ X \longrightarrow [0,1]$$

defined by

$$T_A(x) := \left(\frac{1+\cos(\varphi(x))}{2}\right)^{2\beta_T}, \qquad I_A(x) := \left(\frac{1+\cos(\varphi(x))}{2}\right)^{2\beta_I}, \qquad F_A(x) := \left(\frac{1+\cos(\varphi(x))}{2}\right)^{2\beta_F}.$$

Since each component lies in $[0,1]$, the pointwise inequality

$$0 \leq T_A(x) + I_A(x) + F_A(x) \leq 3 \qquad (\forall x \in X)$$

holds automatically.

**Remark 4.24.4.** The parameters $\beta_T, \beta_I, \beta_F$ control the sharpness of each component independently. This is the minimal extension of the cosine profile to a neutrosophic triple.

**Theorem 4.24.5** (Cosine neutrosophic sets subsume cosine fuzzy sets)**.** *Every cosine fuzzy set (Definition 4.24.1) is obtained as a special case of a cosine neutrosophic set.*

*Proof.* Let $\mu_A(x) = \left(\frac{1+\cos(\varphi(x))}{2}\right)^{2\beta}$ be a cosine fuzzy membership. Choose $\beta_T = \beta$ and set $\beta_I = 0$ and $\beta_F = 0$. Define a cosine neutrosophic set by Definition 4.24.3. Then $T_A(x) = \mu_A(x)$, while $I_A(x) = 1$ and $F_A(x) = 1$ because $(\cdot)^0 = 1$ on $[0,1]$. If one wants the fuzzy embedding with vanishing indeterminacy/falsity, one may instead take the degenerate embedding $I_A \equiv 0$, $F_A \equiv 0$ (dropping the cosine form for those components). In either interpretation, the cosine fuzzy profile is recovered as the truth component. □

**Remark 4.24.6** (A cleaner fuzzy embedding)**.** If one requires $I_A \equiv 0$ and $F_A \equiv 0$ exactly, then $I_A, F_A$ are not of cosine form $\left(\frac{1+\cos(\varphi(x))}{2}\right)^{2\beta}$ unless one allows additional scaling. A convenient alternative is to allow amplitude factors: $I_A(x) = \lambda_I \left(\frac{1+\cos(\varphi(x))}{2}\right)^{2\beta_I}$ with $\lambda_I = 0$, and similarly for $F_A$.

**Theorem 4.24.7** (Cosine neutrosophic sets are a parametric subclass of neutrosophic sets)**.** *Every cosine neutrosophic set on $X = [a,b]$ is a (single-valued) neutrosophic set on $X$. Conversely, an arbitrary neutrosophic set on $X$ need not be representable in cosine form.*

*Proof.* The first claim follows because $T_A, I_A, F_A : X \to [0,1]$ by Definition 4.24.3, which is exactly the data of a neutrosophic set. For the converse, cosine form imposes a specific functional dependence on $\varphi(x)$, hence it defines only a subclass of all possible triples $X \to [0,1]^3$. □

## 4.25 Derived Variants of the Nonstandard Real Set

We record clean nonstandard counterparts of the classical number systems $\mathbb{N} \subset \mathbb{Z} \subset \mathbb{Q} \subset \mathbb{R} \subset \mathbb{C}$. The key point is that in nonstandard analysis one typically fixes a nonstandard extension functor ${}^*(\cdot)$ (e.g. via an ultrapower), and then defines the nonstandard versions as ${}^*\mathbb{N}, {}^*\mathbb{Z}, {}^*\mathbb{Q}, {}^*\mathbb{R}, {}^*\mathbb{C}$. In particular, there are *no nonzero infinitesimal integers or rationals* inside ${}^*\mathbb{Z}$ or ${}^*\mathbb{Q}$; infinitesimals exist in ${}^*\mathbb{R}$ and hence also in ${}^*\mathbb{C}$.

**Definition 4.25.1** (Standard number systems)**.** (cf. [386, 387]) The standard sets of numbers are:

$$\mathbb{N} = \{0,1,2,\dots\} \text{ (or } \{1,2,3,\dots\} \text{ by convention)}, \qquad \mathbb{Z} = \{\dots,-2,-1,0,1,2,\dots\},$$

$$\mathbb{Q} = \left\{\frac{p}{q} :\ p,q \in \mathbb{Z},\ q \neq 0\right\}, \qquad \mathbb{C} = \{a+bi :\ a,b \in \mathbb{R},\ i^2 = -1\}.$$

**Definition 4.25.2** (Nonstandard extension and hypernumbers)**.** Fix a nonstandard extension ${}^*\mathbb{R} \supset \mathbb{R}$ (e.g. a hyperreal field). For any standard set $S$, write ${}^*S$ for its nonstandard extension. In particular, define the *nonstandard number systems* by

$${}^*\mathbb{N}, \quad {}^*\mathbb{Z}, \quad {}^*\mathbb{Q}, \quad {}^*\mathbb{R}, \quad {}^*\mathbb{C}.$$

Elements of ${}^*\mathbb{R}$ are called *hyperreals*, and elements of ${}^*\mathbb{C}$ are called *hypercomplex numbers*.

**Definition 4.25.3** (Infinitesimal and unlimited elements)**.** An element $\varepsilon \in {}^*\mathbb{R}$ is an *infinitesimal* if

$$|\varepsilon| < \frac{1}{n} \quad \text{for all } n \in \mathbb{N}_{\geq 1}.$$

An element $H \in {}^*\mathbb{R}$ is *unlimited* (infinite) if

$$|H| > n \quad \text{for all } n \in \mathbb{N}.$$

Write ${}^*\mathbb{R}_{\text{inf}}$ for the set of infinitesimals and ${}^*\mathbb{R}_{\text{unl}}$ for the set of unlimited hyperreals.

**Proposition 4.25.4** (No nonzero infinitesimals in ${}^*\mathbb{Z}$ or ${}^*\mathbb{Q}$)**.** *If $\xi \in {}^*\mathbb{Z}$ and $|\xi| < 1$, then $\xi = 0$. Consequently, ${}^*\mathbb{Z}$ contains no nonzero infinitesimals. Likewise, ${}^*\mathbb{Q}$ contains no nonzero infinitesimals.*

*Proof.* Let $\xi \in {}^*\mathbb{Z}$ with $|\xi| < 1$. In any ordered ring, the only integer with absolute value $< 1$ is 0; by transfer this holds in ${}^*\mathbb{Z}$, hence $\xi = 0$. Therefore no nonzero infinitesimal can lie in ${}^*\mathbb{Z}$.

For ${}^*\mathbb{Q}$, note that if $q \in {}^*\mathbb{Q} \setminus \{0\}$, then $q = m/n$ for some $m \in {}^*\mathbb{Z} \setminus \{0\}$ and $n \in {}^*\mathbb{Z} \setminus \{0\}$. If $|q| < 1/N$ for every standard $N$, then $|m| < |n|/N$ for every standard $N$, forcing $m = 0$, a contradiction. Thus ${}^*\mathbb{Q}$ has no nonzero infinitesimals. □

**Remark 4.25.5** (What ${}^*\mathbb{Q}$ does contain)**.** Although ${}^*\mathbb{Q}$ has no nonzero infinitesimals, it does contain *unlimited* rationals (e.g. hyperintegers viewed as rationals with denominator 1).

**Definition 4.25.6** (Nonstandard rational set)**.** The *nonstandard rational set* is ${}^*\mathbb{Q}$. It contains all standard rationals $\mathbb{Q}$ and also unlimited rationals, i.e., elements $q \in {}^*\mathbb{Q}$ with $|q| > n$ for all $n \in \mathbb{N}$.

**Theorem 4.25.7** (Inclusions among nonstandard number systems)**.** *In any nonstandard extension, the standard inclusions lift to nonstandard inclusions:*

$${}^*\mathbb{N} \subseteq {}^*\mathbb{Z} \subseteq {}^*\mathbb{Q} \subseteq {}^*\mathbb{R} \subseteq {}^*\mathbb{C}.$$

*Proof.* The inclusions $\mathbb{N} \subseteq \mathbb{Z} \subseteq \mathbb{Q} \subseteq \mathbb{R} \subseteq \mathbb{C}$ are all first-order expressible in the relevant languages (ordered rings/fields and fields with $i$). By the transfer principle (or, equivalently, functoriality of the ${}^*$-extension in standard constructions), they remain valid after applying ${}^*(\cdot)$, yielding the displayed chain. □

**Definition 4.25.8** (Nonstandard complex set)**.** The *nonstandard complex set* is ${}^*\mathbb{C}$. Equivalently, one may identify

$${}^*\mathbb{C} \;\cong\; {}^*\mathbb{R} \times {}^*\mathbb{R}, \qquad (a, b) \leftrightarrow a + bi,$$

so that infinitesimal and unlimited complex numbers are precisely those whose modulus is infinitesimal or unlimited in ${}^*\mathbb{R}$.

**Corollary 4.25.9** (Nonstandard $\mathbb{R}$ embeds into nonstandard $\mathbb{C}$)**.** *One has ${}^*\mathbb{R} \subseteq {}^*\mathbb{C}$ via the standard embedding $r \mapsto r + 0i$.*

*Proof.* Immediate from Theorem 4.25.7 (or from the identification in Definition 4.25.8). □

**Remark 4.25.10** (About "$\mathbb{Z}^*$ contains infinitesimals")**.** There are no "infinitesimal integers" in ${}^*\mathbb{Z}$: the only integer infinitesimally close to 0 is 0 (Proposition 4.25.4). What ${}^*\mathbb{Z}$ does contain are *unlimited hyperintegers* $H \in {}^*\mathbb{Z}$ with $H > n$ for every standard $n$.

For reference, the relationships between Nonstandard sets and Standard sets are illustrated in Figure 4.4.

Building on the concepts discussed above, one can systematically form nonstandard counterparts of many uncertainty objects, such as fuzzy numbers [388,389], neutrosophic numbers [390–392], ZE-numbers [393–395], and Z-numbers [396–398]. This book records one illustrative construction, namely a nonstandard version of Z-numbers. (Any deeper claims about "bipolar (infinite)" analogies should be treated as conjectural until formalized.)

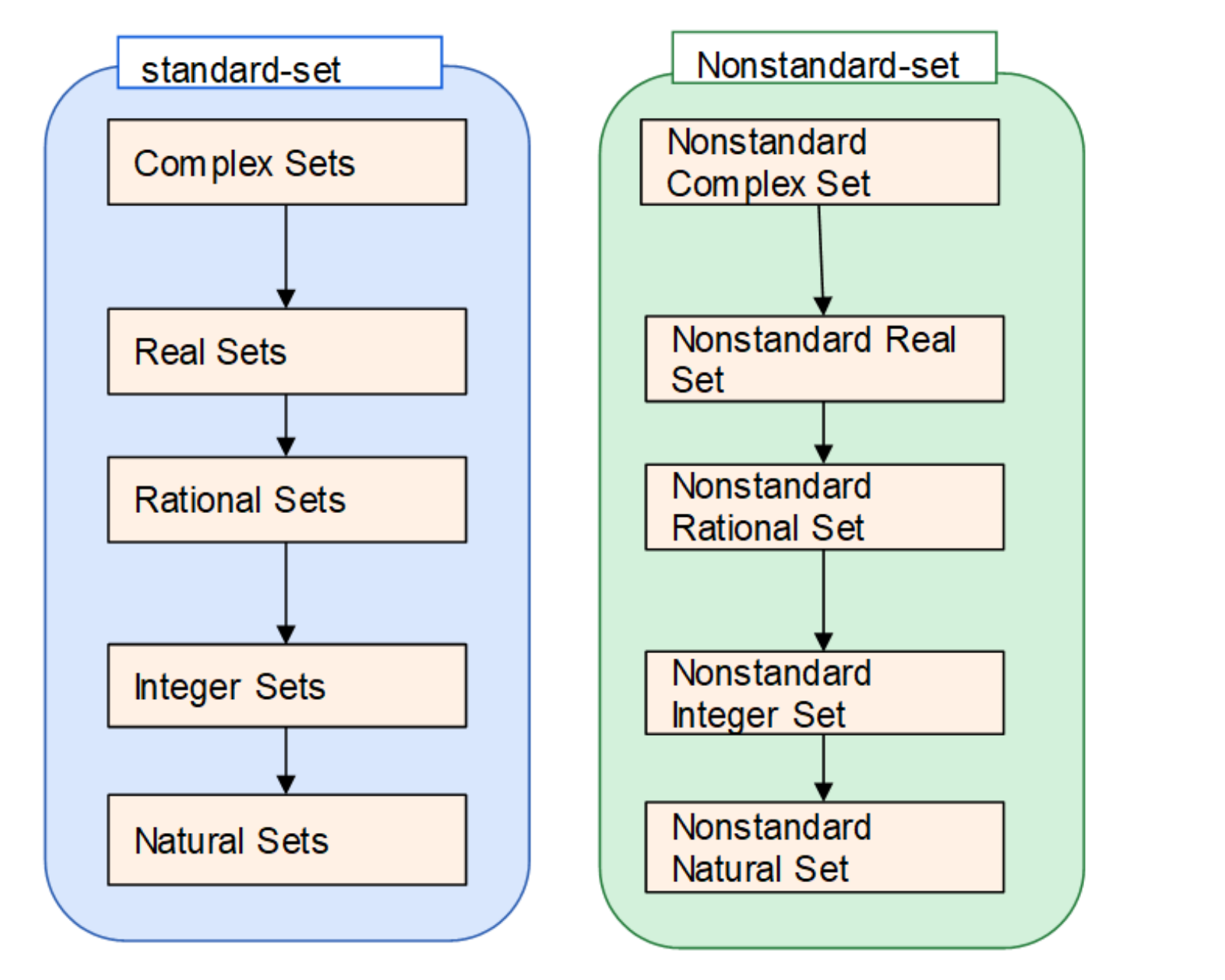

Figure 4.4: Some Standard sets and Nonstandard sets Hierarchy. The set class at the origin of an arrow contains the set class at the destination of the arrow.

**Definition 4.25.11** ((Recall) Z-number)**.** [399] Fix a real universe of discourse $D \subseteq \mathbb{R}$. Let $\mathcal{FN}(D)$ denote a chosen class of fuzzy numbers on $D$ (e.g., normal and convex fuzzy sets with compact support). A *Z-number* is an ordered pair

$$Z = (A, R) \in \mathcal{FN}(D) \times \mathcal{FN}([0, 1]),$$

where $A$ is a fuzzy number describing a restriction on the (real-valued) variable of interest and $R$ is a fuzzy number describing the reliability (confidence) of that restriction.

**Remark 4.25.12.** Different papers adopt slightly different axioms for "fuzzy number". Definition 4.25.11 is stated relative to a fixed class $\mathcal{FN}(D)$ to keep the construction mathematically well-typed.

**Definition 4.25.13** (Nonstandard extension of fuzzy numbers)**.** Fix a nonstandard extension ${}^*\mathbb{R} \supset \mathbb{R}$. Let $D \subseteq \mathbb{R}$ be a domain and let ${}^*D \subseteq {}^*\mathbb{R}$ be its nonstandard extension. For a fuzzy number $A \in \mathcal{FN}(D)$ with membership function $\mu_A : D \to [0, 1]$, its *nonstandard extension* is the internal function

$${}^*\mu_A : {}^*D \longrightarrow {}^*[0, 1]$$

obtained by applying the $^*$-transform to $\mu_A$. We write ${}^*A$ for the corresponding nonstandard fuzzy number (i.e., the fuzzy set on ${}^*D$ whose membership is ${}^*\mu_A$). Let ${}^*\mathcal{FN}(D)$ denote the class of all such nonstandard extensions.

**Definition 4.25.14** (Nonstandard Z-number)**.** Fix a nonstandard extension ${}^*\mathbb{R} \supset \mathbb{R}$. Let $D \subseteq \mathbb{R}$ and let $\mathcal{FN}(D)$ be as in Definition 4.25.11. A *nonstandard Z-number* is an ordered pair

$$Z^* = (A^*, R^*) \in {}^*\mathcal{FN}(D) \times {}^*\mathcal{FN}([0, 1]),$$

i.e., $A^*$ and $R^*$ are nonstandard fuzzy numbers obtained as $^*$-extensions of standard fuzzy numbers $A \in \mathcal{FN}(D)$ and $R \in \mathcal{FN}([0, 1])$.

The set of all nonstandard Z-numbers (relative to the chosen ${}^*\mathbb{R}$ and $\mathcal{FN}$) is

$${}^*\mathcal{Z} := {}^*\mathcal{FN}(D) \times {}^*\mathcal{FN}([0, 1]).$$

**Remark 4.25.15** (Avoiding "$Z^{+*}$, $Z^{-*}$, $\infty^*_Z$")**.** The decomposition "standard/infinitesimal/infinite" is natural for ${}^*\mathbb{R}$ because it is ordered. A Z-number is not itself a real number, and notions like "positive" or "negative" are not intrinsic for pairs $(A^*, R^*)$. If one wants a size notion, one should first fix a scalar functional (e.g. a defuzzification map or a norm on membership functions) and then classify Z-numbers by that scalar.

**Definition 4.25.16** (Nonstandard extension framework)**.** Fix a nonstandard extension functor $S \mapsto {}^*S$ (e.g. coming from an ultrapower construction) with ${}^*\mathbb{R} \supset \mathbb{R}$. For any standard set $S$, its *nonstandard extension* is the set ${}^*S$. If $S$ carries algebraic or order structure (groups, rings, fields, relations, etc.), the corresponding structure on ${}^*S$ is obtained by applying the ${}^*$-transform to the defining operations/relations, and first-order properties transfer from $S$ to ${}^*S$ (transfer principle).

For ordered structures (notably ${}^*\mathbb{R}$), define the standard subclasses:

$$\mathrm{Fin}({}^*\mathbb{R}) := \{x \in {}^*\mathbb{R} : |x| \leq n \text{ for some } n \in \mathbb{N}\},$$

$$\mathrm{Inf}({}^*\mathbb{R}) := \{x \in {}^*\mathbb{R} : |x| < 1/n \text{ for all } n \in \mathbb{N}_{\geq 1}\},$$

$$\mathrm{Unl}({}^*\mathbb{R}) := \{x \in {}^*\mathbb{R} : |x| > n \text{ for all } n \in \mathbb{N}\}.$$

Then $\mathrm{Inf}({}^*\mathbb{R})$ is the set of infinitesimals and $\mathrm{Unl}({}^*\mathbb{R})$ the set of unlimited elements. This classification is meaningful because ${}^*\mathbb{R}$ is an ordered field; it should not be applied blindly to arbitrary non-ordered objects without first choosing an appropriate scalar size functional.

## 4.26 Hypersoft HyperExpert Sets

We introduce *hypersoft hyperexpert sets* as a natural extension of hypersoft expert sets: instead of single attribute-values and single experts, we allow *subsets* of attribute-values and *subsets* of experts (possibly organized into several expert groups). This enlarges the parameter domain and can capture coalitions of experts and coarse (set-valued) attribute specifications.

**Definition 4.26.1** (Hypersoft Expert Set (HSE-set))**.** [57] Let $\Omega \neq \emptyset$ be a universe. Let $G_1, \ldots, G_n$ be nonempty pairwise disjoint attribute-value sets corresponding to distinct attributes $g_1, \ldots, g_n$, and define the hypersoft parameter space

$$G := G_1 \times \cdots \times G_n.$$

Let $D \neq \emptyset$ be a set of experts and let $C \neq \emptyset$ be a set of conclusions (e.g. $C = \{0, 1\}$ for disagree/agree). Let $S \subseteq G \times D \times C$ be a nonempty set of admissible triples.

A pair $(\Psi, S)$ is called a *hypersoft expert set* over $\Omega$ if

$$\Psi : S \longrightarrow \mathcal{P}(\Omega)$$

is a mapping that assigns to each $(g, d, c) \in S$ a subset $\Psi(g, d, c) \subseteq \Omega$.

**Theorem 4.26.2** (HSE-sets generalize soft expert sets)**.** *A hypersoft expert set reduces to a (classical) soft expert set when $n = 1$ (so $G = G_1$) and one identifies $G_1$ with the usual parameter set.*

*Proof.* If $n = 1$, then $G = G_1$ and $S \subseteq G_1 \times D \times C$. Thus $\Psi : S \to \mathcal{P}(\Omega)$ is exactly a soft expert set map on the parameter-expert-opinion domain. □

**Definition 4.26.3** (Hypersoft HyperExpert Set)**.** Let $\Omega \neq \emptyset$ be a universe. Let $G_1, \ldots, G_n$ be nonempty pairwise disjoint attribute-value sets and define the *set-valued* hypersoft parameter space

$$\widehat{G} := \big(\mathcal{P}(G_1) \setminus \{\emptyset\}\big) \times \cdots \times \big(\mathcal{P}(G_n) \setminus \{\emptyset\}\big).$$

Let $D_1, \ldots, D_m$ be nonempty pairwise disjoint expert groups and define the *expert-coalition* space

$$\widehat{D} := \big(\mathcal{P}(D_1) \setminus \{\emptyset\}\big) \times \cdots \times \big(\mathcal{P}(D_m) \setminus \{\emptyset\}\big).$$

Let $C \neq \emptyset$ be a set of conclusions (e.g. $C = \{0, 1\}$). Let $S \subseteq \widehat{G} \times \widehat{D} \times C$ be a nonempty set of admissible triples.

A pair $(\Psi, S)$ is called a *hypersoft hyperexpert set* over $\Omega$ if

$$\Psi : S \longrightarrow \mathcal{P}(\Omega)$$

is a mapping that assigns to each $(\mathbf{G}, \mathbf{D}, c) \in S$ a subset $\Psi(\mathbf{G}, \mathbf{D}, c) \subseteq \Omega$, where

$$\mathbf{G} = (G'_1, \ldots, G'_n) \in \widehat{G} \ \text{ with } \ \emptyset \neq G'_i \subseteq G_i, \qquad \mathbf{D} = (D'_1, \ldots, D'_m) \in \widehat{D} \ \text{ with } \ \emptyset \neq D'_j \subseteq D_j.$$

**Remark 4.26.4** (Why exclude empty subsets)**.** Excluding $\emptyset$ prevents degenerate parameters with "no admissible values" and empty expert coalitions. If one wants to allow such degeneracies, simply remove the "$\setminus\{\emptyset\}$" restrictions.

**Theorem 4.26.5** (HSE-sets embed into hypersoft hyperexpert sets)**.** *Every hypersoft expert set is a special case of a hypersoft hyperexpert set.*

*Proof.* Let $(\Psi, S)$ be a hypersoft expert set as in Definition 4.26.1 with parameter space $G = G_1 \times \cdots \times G_n$ and expert set $D$. Take $m = 1$ and set $D_1 := D$. Embed $G$ into $\widehat{G}$ by the singleton map

$$\iota_G : G \to \widehat{G}, \qquad (g_1, \ldots, g_n) \longmapsto (\{g_1\}, \ldots, \{g_n\}),$$

and embed $D$ into $\widehat{D} = \mathcal{P}(D) \setminus \{\emptyset\}$ by $d \mapsto \{d\}$. Define

$$S' := \{(\iota_G(g), \{d\}, c) : (g, d, c) \in S\} \subseteq \widehat{G} \times \widehat{D} \times C$$

and define $\Psi' : S' \to \mathcal{P}(\Omega)$ by

$$\Psi'(\iota_G(g), \{d\}, c) := \Psi(g, d, c).$$

Then $(\Psi', S')$ is a hypersoft hyperexpert set and it reproduces the original hypersoft expert set on the embedded domain. $\square$

# Chapter 5

# Hyper Concepts and Superhyper concepts

In this subsection, we examine several Hyper concepts and Superhyper concepts. Note that in some fields, multiple definitions exist, and the meanings of "hyper" or "super" may vary.

## 5.1 HyperNeutrosophic Set

The notion of a *hyperneutrosophic set* extends a (single-valued) neutrosophic set in the same spirit that a hyperfuzzy set extends a fuzzy set: instead of assigning a single membership degree (or triplet) to each element, one assigns a *nonempty set* of admissible degrees.

**Definition 5.1.1** (Admissible neutrosophic cube)**.** Let

$$\mathbb{N}^3_{[0,1]} \;:=\; \big\{(T, I, F) \in [0,1]^3 \;:\; 0 \le T + I + F \le 3\big\}.$$

We call $\mathbb{N}^3_{[0,1]}$ the set of *admissible single-valued neutrosophic triplets*.

**Definition 5.1.2** (HyperNeutrosophic set)**.** Let $X \neq \emptyset$ be a universe. Write $\tilde{\mathcal{P}}(Y)$ for the family of all *nonempty* subsets of a set $Y$. A *HyperNeutrosophic Set* (HNS) on $X$ is a mapping

$$\widetilde{\mu} : X \longrightarrow \tilde{\mathcal{P}}\big(\mathbb{N}^3_{[0,1]}\big),$$

that assigns to each $x \in X$ a nonempty set $\widetilde{\mu}(x) \subseteq \mathbb{N}^3_{[0,1]}$ of admissible neutrosophic triplets. The corresponding HNS is denoted by

$$\widetilde{A} \;=\; \big\{\langle x, \widetilde{\mu}(x)\rangle \;:\; x \in X\big\}.$$

**Remark 5.1.3.** Definition 5.1.2 is "single-valued" at the level of each triplet but "hyper" at the level of the membership assignment, since each element $x$ may admit multiple candidate triplets in $[0,1]^3$.

**Theorem 5.1.4** (HNS generalizes (single-valued) neutrosophic sets)**.** *Every single-valued neutrosophic set $A$ on $X$, given by a membership map*

$$\mu_A : X \to \mathbb{N}^3_{[0,1]}, \qquad x \mapsto (T_A(x), I_A(x), F_A(x)),$$

*induces a HyperNeutrosophic Set $\widetilde{A}$ on $X$ by*

$$\widetilde{\mu}(x) \;:=\; \{\mu_A(x)\} \qquad (x \in X).$$

*Proof.* For each $x \in X$, the singleton $\{\mu_A(x)\}$ is a nonempty subset of $\mathbb{N}^3_{[0,1]}$ because $\mu_A(x) \in \mathbb{N}^3_{[0,1]}$ by assumption. Hence $\widetilde{\mu} : X \to \tilde{\mathcal{P}}(\mathbb{N}^3_{[0,1]})$ is well defined, and Definition 5.1.2 yields a HyperNeutrosophic Set. □

**Definition 5.1.5** (Score map from neutrosophic triplets to fuzzy degrees)**.** Define the map $s : \mathbb{N}^3_{[0,1]} \to [0,1]$ by

$$s(T, I, F) \;:=\; \frac{T + (1 - F)}{2}.$$

**Lemma 5.1.6.** *The map $s$ in Definition 5.1.5 is well defined and satisfies $s(T,I,F) \in [0,1]$ for all $(T,I,F) \in \mathbb{N}^3_{[0,1]}$.*

*Proof.* If $(T,I,F) \in [0,1]^3$, then $T \in [0,1]$ and $1-F \in [0,1]$. Hence $T+(1-F) \in [0,2]$, and therefore $s(T,I,F) = \frac{T+(1-F)}{2} \in [0,1]$. □

**Theorem 5.1.7** (Canonical projection of an HNS to an HFS)**.** *Let $\widetilde{A}$ be a HyperNeutrosophic Set on $X$ with membership map $\widetilde{\mu} : X \to \tilde{\mathcal{P}}(\mathbb{N}^3_{[0,1]})$. Define*

$$\widetilde{\nu} : X \longrightarrow \tilde{\mathcal{P}}([0,1]), \qquad \widetilde{\nu}(x) \;:=\; \{\, s(T,I,F) \;:\; (T,I,F) \in \widetilde{\mu}(x)\,\}.$$

*Then $\widetilde{\nu}$ defines a HyperFuzzy Set on $X$.*

*Proof.* Fix $x \in X$. Since $\widetilde{\mu}(x) \neq \emptyset$, the image set $\widetilde{\nu}(x)$ is nonempty. By Lemma 5.1.6, every $s(T,I,F)$ lies in $[0,1]$, so $\widetilde{\nu}(x) \subseteq [0,1]$. Therefore $\widetilde{\nu} : X \to \tilde{\mathcal{P}}([0,1])$ is well defined, which is exactly the definition of a HyperFuzzy Set. □

**Definition 5.1.8** (HyperPlithogenic set)**.** Let $X \neq \emptyset$ be a universe and let $P \subseteq X$ be nonempty. Fix a finite family of attributes $\{v_i\}_{i=1}^m$, where each attribute $v_i$ has a nonempty value set $P_{v_i}$. Write $\tilde{\mathcal{P}}([0,1]^s)$ for the family of all nonempty subsets of $[0,1]^s$.

A *HyperPlithogenic Set* (HPS) is a tuple

$$\mathrm{HPS} = \Big(P,\ \{v_i\}_{i=1}^m,\ \{P_{v_i}\}_{i=1}^m,\ \{\widetilde{pdf}_i\}_{i=1}^m,\ pCF\Big),$$

where:

- for each $i$, $\widetilde{pdf}_i : P \times P_{v_i} \to \tilde{\mathcal{P}}([0,1]^s)$ is a *hyper degree of appurtenance function* (HDAF);
- $pCF : \Big(\bigcup_{i=1}^m P_{v_i}\Big) \times \Big(\bigcup_{i=1}^m P_{v_i}\Big) \to [0,1]^t$ is a *degree of contradiction function* (DCF), typically satisfying
$$pCF(a,a) = \mathbf{0}, \qquad pCF(a,b) = pCF(b,a) \qquad (a, b \text{ in the value domain}),$$
where $\mathbf{0}$ denotes the zero vector in $[0,1]^t$.

**Theorem 5.1.9** (HPS generalizes plithogenic sets)**.** *Every (vector-valued) plithogenic set $PS = (P, v, P_v, pdf, pCF)$ with $pdf : P \times P_v \to [0,1]^s$ induces a HyperPlithogenic Set by taking $m = 1$ and defining*

$$\widetilde{pdf}(x,a) \;:=\; \{pdf(x,a)\} \qquad (x \in P,\ a \in P_v),$$

*with the same contradiction map $pCF$.*

*Proof.* For each $(x,a) \in P \times P_v$, the singleton $\{pdf(x,a)\}$ is a nonempty subset of $[0,1]^s$, hence $\widetilde{pdf} : P \times P_v \to \tilde{\mathcal{P}}([0,1]^s)$ is well defined. All other components are inherited unchanged, so the resulting tuple satisfies Definition 5.1.8. □

**Theorem 5.1.10** (From HPS to a HyperFuzzy Set when $s = 1$)**.** *Let* HPS *be a HyperPlithogenic Set as in Definition 5.1.8 with $s = 1$. Define a map $\widetilde{\nu} : P \to \tilde{\mathcal{P}}([0,1])$ by*

$$\widetilde{\nu}(x) \;:=\; \bigcup_{i=1}^{m} \bigcup_{a \in P_{v_i}} \widetilde{pdf}_i(x,a) \qquad (x \in P).$$

*Then $\widetilde{\nu}$ defines a HyperFuzzy Set on $P$.*

*Proof.* Fix $x \in P$. Each $\widetilde{pdf}_i(x,a)$ is nonempty and contained in $[0,1]$ because $s = 1$. Therefore the union defining $\widetilde{\nu}(x)$ is a (possibly larger) nonempty subset of $[0,1]$, hence $\widetilde{\nu} : P \to \tilde{\mathcal{P}}([0,1])$ is well defined. □

**Theorem 5.1.11** (From HPS to an HNS when $s = 3$)**.** *Let* HPS *be a HyperPlithogenic Set as in Definition 5.1.8 with $s = 3$. Assume that each hyper-membership vector in $\widetilde{pdf}_i(x, a)$ is interpreted as a neutrosophic triplet $(T, I, F) \in \mathbb{N}^3_{[0,1]}$. Define $\widetilde{\mu} : P \to \tilde{\mathcal{P}}(\mathbb{N}^3_{[0,1]})$ by*

$$\widetilde{\mu}(x) \ := \ \bigcup_{i=1}^{m} \bigcup_{a \in P_{v_i}} \widetilde{pdf}_i(x, a) \qquad (x \in P).$$

*Then $\widetilde{\mu}$ defines a HyperNeutrosophic Set on $P$.*

*Proof.* Fix $x \in P$. By assumption, each $\widetilde{pdf}_i(x, a)$ is a nonempty subset of $\mathbb{N}^3_{[0,1]}$, so the union $\widetilde{\mu}(x)$ is also nonempty and contained in $\mathbb{N}^3_{[0,1]}$. Thus $\widetilde{\mu} : P \to \tilde{\mathcal{P}}(\mathbb{N}^3_{[0,1]})$ is well defined, yielding an HNS by Definition 5.1.2. □

## 5.2 HyperVague Offset/Overset/Underset

We introduce *hypervague* analogues of vague sets by allowing each element to carry a *nonempty set* of admissible $(t, f)$-pairs, instead of a single pair. We then extend the over-/under-/offset philosophy (allowing values above 1 and/or below 0) to both the vague and hypervague settings.

**Definition 5.2.1** (Admissible vague region)**.** Let

$$\mathbb{V}_{[0,1]} \ := \ \{(t, f) \in [0, 1]^2 : \ 0 \le t + f \le 1\}.$$

We call $\mathbb{V}_{[0,1]}$ the set of *admissible vague pairs*.

**Definition 5.2.2** (HyperVague Set)**.** Let $U \neq \emptyset$ be a universe. A *HyperVague Set* (HVS) on $U$ is a mapping

$$\widetilde{A} : U \longrightarrow \tilde{\mathcal{P}}(\mathbb{V}_{[0,1]}),$$

so that for each $u \in U$, the value $\widetilde{A}(u)$ is a nonempty set of admissible pairs $(t, f)$ satisfying $0 \le t + f \le 1$.

**Remark 5.2.3.** A classical vague set corresponds to the special case in which every $\widetilde{A}(u)$ is a singleton.

**Theorem 5.2.4** (HyperVague Set generalizes a vague set)**.** *Let $A$ be a vague set on $U$ specified by functions $t_A, f_A : U \to [0, 1]$ satisfying $0 \le t_A(u) + f_A(u) \le 1$ for all $u \in U$. Define*

$$\widetilde{A}(u) \ := \ \{(t_A(u), f_A(u))\} \qquad (u \in U).$$

*Then $\widetilde{A}$ is a HyperVague Set on $U$.*

*Proof.* For each $u \in U$, the singleton $\{(t_A(u), f_A(u))\}$ is nonempty. Moreover $(t_A(u), f_A(u)) \in [0, 1]^2$ and satisfies $0 \le t_A(u) + f_A(u) \le 1$, hence it lies in $\mathbb{V}_{[0,1]}$. Therefore $\widetilde{A}(u) \in \tilde{\mathcal{P}}(\mathbb{V}_{[0,1]})$ for every $u$, so $\widetilde{A}$ is a HyperVague Set by Definition 5.2.2. □

**Theorem 5.2.5** (HyperVague Set generalizes a HyperFuzzy Set)**.** *Let $\widetilde{\mu} : U \to \tilde{\mathcal{P}}([0, 1])$ be a HyperFuzzy Set on $U$. Define*

$$\widetilde{A}(u) \ := \ \{(t, 0) : \ t \in \widetilde{\mu}(u)\} \qquad (u \in U).$$

*Then $\widetilde{A}$ is a HyperVague Set on $U$.*

*Proof.* Fix $u \in U$. Since $\widetilde{\mu}(u) \neq \emptyset$, the set $\widetilde{A}(u)$ is nonempty. For each $(t, 0) \in \widetilde{A}(u)$ we have $t \in [0, 1]$, so $(t, 0) \in [0, 1]^2$ and $0 \le t + 0 \le 1$. Hence $(t, 0) \in \mathbb{V}_{[0,1]}$. Therefore $\widetilde{A}(u) \in \tilde{\mathcal{P}}(\mathbb{V}_{[0,1]})$ for all $u$, proving the claim. □

**Definition 5.2.6** (Vague Overset)**.** Fix $\Omega > 1$. A *vague overset* on a universe $U$ is a pair of maps

$$t : U \to [0, \Omega], \qquad f : U \to [0, \Omega],$$

such that $0 \le t(u) + f(u) \le \Omega$ for all $u \in U$, and there exists $u_0 \in U$ with $t(u_0) > 1$ or $f(u_0) > 1$.

**Definition 5.2.7** (Vague Underset)**.** Fix $\Psi < 0$. A *vague underset* on a universe $U$ is a pair of maps

$$t : U \to [\Psi, 1], \qquad f : U \to [\Psi, 1],$$

such that $\Psi \le t(u) + f(u) \le 1$ for all $u \in U$, and there exists $u_0 \in U$ with $t(u_0) < 0$ or $f(u_0) < 0$.

**Definition 5.2.8** (Vague Offset)**.** Fix $\Psi < 0 < 1 < \Omega$. A *vague offset* on a universe $U$ is a pair of maps

$$t : U \to [\Psi, \Omega], \qquad f : U \to [\Psi, \Omega],$$

such that $\Psi \le t(u) + f(u) \le \Omega$ for all $u \in U$, and there exist (not necessarily distinct) points $u_+, u_- \in U$ for which

$$t(u_+) > 1 \text{ or } f(u_+) > 1, \qquad \text{and} \qquad t(u_-) < 0 \text{ or } f(u_-) < 0.$$

**Theorem 5.2.9** (Vague offsets subsume over- and undersets)**.** *Every vague overset and every vague underset is a vague offset for suitable choices of* $\Psi < 0 < 1 < \Omega$*.*

*Proof.* Let $(t, f)$ be a vague overset with parameter $\Omega > 1$. Choose any $\Psi < 0$ and regard the same maps $t, f : U \to [0, \Omega]$ as maps into $[\Psi, \Omega]$ (since $[0, \Omega] \subseteq [\Psi, \Omega]$). Then the defining inequality $0 \le t(u) + f(u) \le \Omega$ implies $\Psi \le t(u) + f(u) \le \Omega$, and the overset witness $u_0$ ensures an over-limit value $> 1$; hence $(t, f)$ is a vague offset.

The underset case is analogous: given $(t, f) : U \to [\Psi, 1]$ with $\Psi < 0$, choose any $\Omega > 1$ and note that $[\Psi, 1] \subseteq [\Psi, \Omega]$. Then the underset witness gives a value $< 0$, so $(t, f)$ is a vague offset. □

**Definition 5.2.10** (Admissible vague region with bounds)**.** Fix bounds $\Psi < 0 < 1 < \Omega$. Define

$$\mathbb{V}_{[\Psi,\Omega]} \;:=\; \big\{(t, f) \in [\Psi, \Omega]^2 :\; \Psi \le t + f \le \Omega\big\}.$$

Similarly, for the overset and underset regimes:

$$\mathbb{V}_{[0,\Omega]} := \big\{(t, f) \in [0, \Omega]^2 :\; 0 \le t + f \le \Omega\big\}, \qquad \mathbb{V}_{[\Psi,1]} := \big\{(t, f) \in [\Psi, 1]^2 :\; \Psi \le t + f \le 1\big\}.$$

**Definition 5.2.11** (HyperVague Overset)**.** Fix $\Omega > 1$. A *HyperVague Overset* on $U$ is a mapping

$$\widetilde{A} : U \longrightarrow \tilde{\mathcal{P}}(\mathbb{V}_{[0,\Omega]}),$$

such that there exists $u_0 \in U$ and $(t, f) \in \widetilde{A}(u_0)$ with $t > 1$ or $f > 1$.

**Definition 5.2.12** (HyperVague Underset)**.** Fix $\Psi < 0$. A *HyperVague Underset* on $U$ is a mapping

$$\widetilde{A} : U \longrightarrow \tilde{\mathcal{P}}(\mathbb{V}_{[\Psi,1]}),$$

such that there exists $u_0 \in U$ and $(t, f) \in \widetilde{A}(u_0)$ with $t < 0$ or $f < 0$.

**Definition 5.2.13** (HyperVague Offset)**.** Fix $\Psi < 0 < 1 < \Omega$. A *HyperVague Offset* on $U$ is a mapping

$$\widetilde{A} : U \longrightarrow \tilde{\mathcal{P}}(\mathbb{V}_{[\Psi,\Omega]}),$$

such that there exist $u_+, u_- \in U$ and pairs $(t_+, f_+) \in \widetilde{A}(u_+)$, $(t_-, f_-) \in \widetilde{A}(u_-)$ satisfying

$$t_+ > 1 \text{ or } f_+ > 1, \qquad \text{and} \qquad t_- < 0 \text{ or } f_- < 0.$$

**Theorem 5.2.14** (HyperVague offsets subsume HyperVague over- and undersets)**.** *Every HyperVague Overset and every HyperVague Underset is a HyperVague Offset for suitable* $\Psi < 0 < 1 < \Omega$*.*

*Proof.* The argument is identical to Theorem 5.2.9, replacing each single pair $(t(u), f(u))$ by a set $\widetilde{A}(u)$ of admissible pairs and observing the inclusions $\mathbb{V}_{[0,\Omega]} \subseteq \mathbb{V}_{[\Psi,\Omega]}$ and $\mathbb{V}_{[\Psi,1]} \subseteq \mathbb{V}_{[\Psi,\Omega]}$. □

**Definition 5.2.15** (Projection to fuzzy membership)**.** Define the projection $\pi : \mathbb{R}^2 \to \mathbb{R}$ by $\pi(t, f) := t$. For a hypervague object $\widetilde{A}$, define its projected map by

$$(\pi\widetilde{A})(u) \;:=\; \{\pi(t, f) : (t, f) \in \widetilde{A}(u)\}.$$

**Theorem 5.2.16** (HyperVague $\Rightarrow$ HyperFuzzy by projection)**.** *Let $\widetilde{A}$ be a HyperVague Set on $U$. Then $\pi\widetilde{A}$ is a HyperFuzzy Set on $U$. Moreover:*

(i) *if $\widetilde{A}$ is a HyperVague Overset (parameter $\Omega > 1$), then $\pi\widetilde{A}$ is a HyperFuzzy Overset on $[0, \Omega]$;*

(ii) *if $\widetilde{A}$ is a HyperVague Underset (parameter $\Psi < 0$), then $\pi\widetilde{A}$ is a HyperFuzzy Underset on $[\Psi, 1]$;*

(iii) *if $\widetilde{A}$ is a HyperVague Offset (parameters $\Psi < 0 < 1 < \Omega$), then $\pi\widetilde{A}$ is a HyperFuzzy Offset on $[\Psi, \Omega]$.*

*Proof.* In each case, $\widetilde{A}(u)$ is nonempty, hence its projection $(\pi\widetilde{A})(u)$ is nonempty. If $(t, f) \in \mathbb{V}_{[0,1]}$, then $t \in [0, 1]$, so $(\pi\widetilde{A})(u) \subseteq [0, 1]$; this proves the HyperFuzzy Set case. The overset/underset/offset regimes follow analogously because $(t, f) \in \mathbb{V}_{[0,\Omega]} \Rightarrow t \in [0, \Omega]$, $(t, f) \in \mathbb{V}_{[\Psi,1]} \Rightarrow t \in [\Psi, 1]$, and $(t, f) \in \mathbb{V}_{[\Psi,\Omega]} \Rightarrow t \in [\Psi, \Omega]$. Finally, the existence of a pair with $t > 1$ (respectively $t < 0$) is preserved under projection, establishing the corresponding over-/under-/offset witness. □

## 5.3 $N$-Superhyper Sets

$N$-hyper sets assign to each element of a universe a *nonempty set of negative membership degrees* in $[-1, 0]$ (cf. [67]). We introduce an $N$-*superhyper* variant by allowing the membership assignment to be defined on all nonempty subsets of the universe (a "set-valued vertex" level), thereby paralleling the passage from hyperstructures to superhyperstructures. We also record a sign-flip correspondence to (super)hyper-fuzzy models on $[0, 1]$.

**Definition 5.3.1** (Nonempty power set)**.** For any set $X$, write

$$\widetilde{\mathcal{P}}(X) := \mathcal{P}(X) \setminus \{\emptyset\}$$

for the collection of all nonempty subsets of $X$.

**Definition 5.3.2** ($N$-Hyper Set)**.** [67] Let $X \neq \emptyset$. An $N$-*hyper set* on $X$ is a mapping

$$\mu : X \longrightarrow \widetilde{\mathcal{P}}([-1, 0]),$$

so that each $x \in X$ is assigned a nonempty set $\mu(x) \subseteq [-1, 0]$ of (negative) membership degrees.

**Definition 5.3.3** ($N$-Superhyper Set)**.** Let $X \neq \emptyset$. An $N$-*superhyper set* on $X$ is a mapping

$$\widetilde{\mu} : \widetilde{\mathcal{P}}(X) \longrightarrow \widetilde{\mathcal{P}}([-1, 0]).$$

Thus each nonempty subset $A \subseteq X$ is assigned a nonempty set $\widetilde{\mu}(A) \subseteq [-1, 0]$ representing negative membership information associated with the group $A$.

**Theorem 5.3.4** ($N$-hyper sets embed into $N$-superhyper sets)**.** *Every $N$-hyper set $\mu : X \to \widetilde{\mathcal{P}}([-1, 0])$ canonically induces an $N$-superhyper set $\widetilde{\mu} : \widetilde{\mathcal{P}}(X) \to \widetilde{\mathcal{P}}([-1, 0])$ by*

$$\widetilde{\mu}(A) := \bigcup_{x \in A} \mu(x) \qquad (A \in \widetilde{\mathcal{P}}(X)).$$

*Conversely, every $N$-superhyper set $\widetilde{\mu}$ restricts to an $N$-hyper set via*

$$\mu(x) := \widetilde{\mu}(\{x\}) \qquad (x \in X).$$

*Proof.* Let $\mu$ be an $N$-hyper set. For any nonempty $A \subseteq X$, each $\mu(x)$ is nonempty and contained in $[-1, 0]$, so $\widetilde{\mu}(A) = \bigcup_{x \in A} \mu(x)$ is nonempty and contained in $[-1, 0]$. Hence $\widetilde{\mu} : \widetilde{\mathcal{P}}(X) \to \widetilde{\mathcal{P}}([-1, 0])$ is well-defined.

Conversely, if $\widetilde{\mu}$ is an $N$-superhyper set, then for each $x \in X$ the singleton $\{x\}$ lies in $\widetilde{\mathcal{P}}(X)$, so $\mu(x) := \widetilde{\mu}(\{x\}) \in \widetilde{\mathcal{P}}([-1, 0])$. Thus $\mu : X \to \widetilde{\mathcal{P}}([-1, 0])$ is an $N$-hyper set. □

**Definition 5.3.5** (Hyper-fuzzy and superhyper-fuzzy sets (set-valued degrees))**.** Let $X \neq \emptyset$. A *hyper-fuzzy set* on $X$ is a map $\eta : X \to \widetilde{\mathcal{P}}([0, 1])$. A *superhyper-fuzzy set* on $X$ is a map $\widetilde{\eta} : \widetilde{\mathcal{P}}(X) \to \widetilde{\mathcal{P}}([0, 1])$.

**Theorem 5.3.6** (Sign-flip transform)**.** *Let* $\widetilde{\mu} : \widetilde{\mathcal{P}}(X) \to \widetilde{\mathcal{P}}([-1,0])$ *be an N-superhyper set. Define*

$$\widetilde{\eta}(A) := \{-a : \ a \in \widetilde{\mu}(A)\} \qquad (A \in \widetilde{\mathcal{P}}(X)).$$

*Then* $\widetilde{\eta} : \widetilde{\mathcal{P}}(X) \to \widetilde{\mathcal{P}}([0,1])$ *is a superhyper-fuzzy set on X.*

*Proof.* For any $A \in \widetilde{\mathcal{P}}(X)$, the set $\widetilde{\mu}(A)$ is nonempty and contained in $[-1,0]$. Hence $\widetilde{\eta}(A)$ is nonempty and contained in $[0,1]$. Therefore $\widetilde{\eta}$ is a superhyper-fuzzy set as in Definition 5.7.6. □

**Corollary 5.3.7** (Hyper case)**.** *If* $\mu : X \to \widetilde{\mathcal{P}}([-1,0])$ *is an N-hyper set, then*

$$\eta(x) := \{-a : \ a \in \mu(x)\} \qquad (x \in X)$$

*defines a hyper-fuzzy set* $\eta : X \to \widetilde{\mathcal{P}}([0,1])$.

*Proof.* This is the singleton restriction of Theorem 5.3.6. □

**Example 5.3.8.** Let $X = \{x_1, x_2\}$ and define an $N$-hyper set $\mu : X \to \widetilde{\mathcal{P}}([-1,0])$ by

$$\mu(x_1) = \{-0.5, -0.2\}, \qquad \mu(x_2) = \{-0.8\}.$$

The induced $N$-superhyper set $\widetilde{\mu}$ from Theorem 5.3.4 satisfies

$$\widetilde{\mu}(\{x_1\}) = \{-0.5, -0.2\}, \quad \widetilde{\mu}(\{x_2\}) = \{-0.8\}, \quad \widetilde{\mu}(\{x_1, x_2\}) = \{-0.8, -0.5, -0.2\}.$$

The sign-flip superhyper-fuzzy set $\widetilde{\eta}$ from Theorem 5.3.6 is

$$\widetilde{\eta}(\{x_1\}) = \{0.5, 0.2\}, \quad \widetilde{\eta}(\{x_2\}) = \{0.8\}, \quad \widetilde{\eta}(\{x_1, x_2\}) = \{0.8, 0.5, 0.2\}.$$

**Definition 5.3.9** ($N$-Hyper Overset)**.** Let $X \neq \emptyset$ and fix an overlimit $\Omega > 1$. An *N-hyper overset* on $X$ is a mapping

$$\mu : \ X \longrightarrow \widetilde{\mathcal{P}}([-1, \Omega]).$$

If $\Omega = 0$, this reduces to an $N$-hyper set (Definition 5.3.2).

**Theorem 5.3.10** (Specialization at $\Omega = 0$)**.** *An N-hyper overset with* $\Omega = 0$ *is exactly an N-hyper set.*

*Proof.* If $\Omega = 0$, then $[-1, \Omega] = [-1, 0]$, so Definition 5.3.9 coincides with Definition 5.3.2. □

## 5.4 IndetermSuperHyperSoft Set

IndetermSuperHyperSoft Set is a SuperHyperSoft Set allowing indeterminacy in attribute value subsets, universe elements, or mappings, modeling incomplete, ambiguous, or partially specified multi-attribute information [400]. Because the word *indeterminate* is intentionally broad in the cited literature, it is convenient to model it abstractly by enlarging each object with a distinguished "unknown" symbol.

**Notation 1** (Indeterminate extension)**.** *(cf. [260]) For any set X, define its* indeterminate extension *by*

$$X^{\mathsf{Ind}} \ := \ X \ \cup \ \{\mathsf{Ind}\},$$

*where* Ind $\notin X$ *is a new symbol. When we say that an object "has indeterminacy", we mean it is allowed to take the value* Ind *(or, more generally, may be specified only partially); in this section we use* Ind *as a clean mathematical placeholder.*

**Definition 5.4.1** (IndetermSuperHyperSoft Set)**.** [400] Let $U$ be a universe of discourse and let $H \subseteq U$ be a nonempty set of objects. Fix $n \geq 1$ and distinct attributes $a_1, \ldots, a_n$. For each $i \in \{1, \ldots, n\}$, let $A_i$ be a (possibly finite) set of attribute values for $a_i$, and assume the value-domains are disjoint: $A_i \cap A_j = \emptyset$ for $i \neq j$.

Put

$$\mathcal{D} := \mathcal{P}(A_1)^{\mathsf{Ind}} \times \cdots \times \mathcal{P}(A_n)^{\mathsf{Ind}}, \qquad C := \mathcal{P}(H)^{\mathsf{Ind}}.$$

An *IndetermSuperHyperSoft Set* (IndSHSS) over $U$ (with universe-part $H$) is a pair

$$(F, \mathcal{D}),$$

where $F : \mathcal{D} \to C$ is a mapping.

**Interpretation.** A point of $\mathcal{D}$ is an $n$-tuple $(S_1, \ldots, S_n)$ where each coordinate $S_i$ is either (i) a subset $S_i \subseteq A_i$ (possibly empty), or (ii) the indeterminate symbol $\mathsf{Ind}$. The output $F(S_1, \ldots, S_n)$ is either a subset of $H$, or $\mathsf{Ind}$, representing an indeterminate/unknown approximation-set.

**Nontrivial indeterminacy.** We say that $(F, \mathcal{D})$ is *properly indeterminate* if at least one of the following occurs:

(i) some attribute-side input is indeterminate, i.e., there exists $(S_1, \ldots, S_n) \in \mathcal{D}$ with $S_i = \mathsf{Ind}$ for some $i$;

(ii) the universe-side output can be indeterminate, i.e., there exists $(S_1, \ldots, S_n) \in \mathcal{D}$ with $F(S_1, \ldots, S_n) = \mathsf{Ind}$;

(iii) the description of at least one $A_i$, $H$, or $F$ is itself incomplete/ambiguous, which we represent by allowing $\mathsf{Ind}$ at the relevant places via Notation 1.

**Remark 5.4.2** (Relation to other soft-type structures)**.** If one deletes the indeterminate symbol $\mathsf{Ind}$ everywhere, one recovers the *(crisp) SuperHyperSoft Set* model, i.e., a mapping

$$F : \mathcal{P}(A_1) \times \cdots \times \mathcal{P}(A_n) \to \mathcal{P}(H)$$

. If one further restricts each input coordinate to a singleton $\{x_i\}$ with $x_i \in A_i$, one recovers a hypersoft-type parametrization $A_1 \times \cdots \times A_n \to \mathcal{P}(H)$.

**Theorem 5.4.3** (IndSHSS generalizes IndetermHyperSoft Sets and SuperHyperSoft Sets)**.** (i) *(*IndetermHyperSoft $\Rightarrow$ IndSHSS*) Let* $(G, A_1 \times \cdots \times A_n)$ *be an* IndetermHyperSoft Set *over U, modeled as a map*

$$G : \left(A_1^{\mathsf{Ind}} \times \cdots \times A_n^{\mathsf{Ind}}\right) \longrightarrow \mathcal{P}(H)^{\mathsf{Ind}}.$$

*Then there exists an IndetermSuperHyperSoft Set* $(F, \mathcal{D})$ *such that G is recovered by restricting F to singleton inputs:*

$$G(x_1, \ldots, x_n) = F(\{x_1\}, \ldots, \{x_n\}) \qquad (x_i \in A_i^{\mathsf{Ind}}).$$

(ii) *(*SuperHyperSoft $\Rightarrow$ IndSHSS*) Let* $(F_0, \mathcal{P}(A_1) \times \cdots \times \mathcal{P}(A_n))$ *be a* SuperHyperSoft Set *over U with* $F_0 : \mathcal{P}(A_1) \times \cdots \times \mathcal{P}(A_n) \to \mathcal{P}(H)$*. Then* $F_0$ *extends to an IndetermSuperHyperSoft Set* $F : \mathcal{D} \to C$ *(Definition 5.4.1) by setting* $F = F_0$ *on determinate inputs and assigning any values on inputs containing* $\mathsf{Ind}$*. In particular, every SuperHyperSoft Set is obtained as the* $\mathsf{Ind}$*-free restriction of an IndSHSS.*

*Proof.* (i) Define $F : \mathcal{D} \to C$ by the rule

$$F(S_1, \ldots, S_n) := \begin{cases} G(x_1, \ldots, x_n), & \text{if } S_i = \{x_i\} \text{ for all } i \text{ and } x_i \in A_i, \\ \mathsf{Ind}, & \text{otherwise.} \end{cases}$$

Then $F$ is well-defined because the first case uniquely determines $(x_1, \ldots, x_n)$ from the singleton tuple. By construction, for all $(x_1, \ldots, x_n) \in A_1 \times \cdots \times A_n$,

$$F(\{x_1\}, \ldots, \{x_n\}) = G(x_1, \ldots, x_n).$$

If the given IndetermHyperSoft model allows indeterminate inputs $x_i = \mathsf{Ind}$, one may either (a) treat $\{\mathsf{Ind}\}$ as a legal singleton in $\mathcal{P}(A_i)^{\mathsf{Ind}}$ and keep the same definition, or (b) map such cases to Ind; either convention yields an IndSHSS extension and preserves the claimed restriction identity on determinate points.

(ii) Given $F_0$, define $F : \mathcal{D} \to C$ by

$$F(S_1, \ldots, S_n) \;:=\; \begin{cases} F_0(S_1, \ldots, S_n), & \text{if } S_i \in \mathcal{P}(A_i) \text{ for all } i, \\ \mathsf{Ind}, & \text{otherwise.} \end{cases}$$

Then $F$ is an IndSHSS. Moreover, restricting $F$ to $\mathcal{P}(A_1) \times \cdots \times \mathcal{P}(A_n) \subseteq \mathcal{D}$ recovers $F_0$ exactly. □

## 5.5 HyperRough Sets and HyperRough Graphs

We introduce a parameterized ("hyper") rough-set framework by combining a *soft/hypersoft-style* parameter domain with Pawlak-type rough approximations [401–404]. Intuitively, each parameter tuple selects a target subset, and the rough operators are applied with respect to a fixed approximation space.

**Definition 5.5.1** (Approximation space)**.** Let $X$ be a nonempty finite universe. An *approximation space* is a pair $(X, R)$ where $R \subseteq X \times X$ is an equivalence relation. For $x \in X$, write

$$[x]_R \;:=\; \{y \in X : \; (x, y) \in R\}$$

for the (equivalence) $R$–class of $x$.

**Definition 5.5.2** (HyperRough set)**.** [405] Let $(X, R)$ be an approximation space, and let $T_1, \ldots, T_n$ be $n$ distinct attributes with (nonempty) value domains $J_1, \ldots, J_n$. Put

$$J \;:=\; J_1 \times \cdots \times J_n.$$

A *hyperrough set* (parameterized rough set) over $(X, R)$ is a pair $(F, J)$ where

$$F : J \longrightarrow \mathcal{P}(X)$$

is a mapping. For each parameter tuple $a = (a_1, \ldots, a_n) \in J$, define the *lower* and *upper* $R$–approximations of $F(a)$ by

$$\underline{F}(a) := \{\, x \in X : \; [x]_R \subseteq F(a) \,\},$$
$$\overline{F}(a) := \{\, x \in X : \; [x]_R \cap F(a) \neq \emptyset \,\}.$$

The ordered pair $\big(\underline{F}(a), \overline{F}(a)\big)$ is called the *rough view of* $F(a)$.

**Remark 5.5.3.** For every $a \in J$ one has $\underline{F}(a) \subseteq F(a) \subseteq \overline{F}(a)$, and both $\underline{F}(a)$ and $\overline{F}(a)$ are unions of $R$–equivalence classes.

**Theorem 5.5.4** (Rough sets as a special case)**.** *Let $(X, R)$ be an approximation space and let $U \subseteq X$. Define $n := 1$, $J_1 := \{a\}$ (a singleton), and $F : J_1 \to \mathcal{P}(X)$ by $F(a) := U$. Then $(F, J_1)$ is a hyperrough set and*

$$\underline{F}(a) = \underline{U} \qquad \text{and} \qquad \overline{F}(a) = \overline{U},$$

*where $(\underline{U}, \overline{U})$ is the classical Pawlak rough set of $U$.*

*Proof.* Since $J_1 = \{a\}$, the map $F$ selects exactly one target subset, namely $U$. By Definition 5.5.2,

$$\underline{F}(a) = \{x \in X : [x]_R \subseteq F(a)\} = \{x \in X : [x]_R \subseteq U\} = \underline{U},$$

and similarly

$$\overline{F}(a) = \{x \in X : [x]_R \cap F(a) \neq \emptyset\} = \{x \in X : [x]_R \cap U \neq \emptyset\} = \overline{U}.$$

□

**Definition 5.5.5** (HyperRough graph)**.** (cf. [404]) Let $G = (V, E)$ be a finite simple undirected graph, and let $(V, R)$ be an approximation space on the same vertex set (Definition 5.5.1). Let $T_1, \ldots, T_n$ be attributes with domains $J_1, \ldots, J_n$ and put $J := J_1 \times \cdots \times J_n$.

A *hyperrough graph* on $(G, R)$ is a pair $(F, J)$ with

$$F : J \longrightarrow \mathcal{P}(V).$$

For each $a \in J$, define the vertex approximations

$$\underline{V}(a) := \underline{F}(a), \qquad \overline{V}(a) := \overline{F}(a),$$

as in Definition 5.5.2. The corresponding *lower* and *upper* edge sets are defined by

$$\begin{aligned} \underline{E}(a) &:= \big\{\{u, v\} \in E : \ u, v \in \underline{V}(a)\big\}, \\ \overline{E}(a) &:= \big\{\{u, v\} \in E : \ u, v \in \overline{V}(a)\big\}. \end{aligned}$$

Thus the *lower* and *upper* induced subgraphs are

$$\underline{G}(a) := \big(\underline{V}(a), \underline{E}(a)\big), \qquad \overline{G}(a) := \big(\overline{V}(a), \overline{E}(a)\big).$$

**Remark 5.5.6.** For every $a \in J$,

$$\underline{V}(a) \subseteq \overline{V}(a) \quad \Longrightarrow \quad \underline{E}(a) \subseteq \overline{E}(a),$$

hence $\underline{G}(a)$ is a subgraph of $\overline{G}(a)$ (both as graphs on $V$).

**Theorem 5.5.7** (Rough graphs as a special case)**.** *Let $G = (V, E)$ be a finite simple graph and let $(V, R)$ be an approximation space. Fix $U \subseteq V$ and set $n := 1$, $J_1 := \{a\}$, and $F(a) := U$. Then the hyperrough graph $(F, J_1)$ satisfies*

$$\underline{G}(a) = \big(\underline{U}, \ \{\{u, v\} \in E : \ u, v \in \underline{U}\}\big), \qquad \overline{G}(a) = \big(\overline{U}, \ \{\{u, v\} \in E : \ u, v \in \overline{U}\}\big),$$

*i.e., it reduces to the usual rough approximation of the vertex-induced subgraph determined by $U$.*

*Proof.* With $J_1 = \{a\}$ and $F(a) = U$, Theorem 5.5.4 yields $\underline{V}(a) = \underline{U}$ and $\overline{V}(a) = \overline{U}$. Substituting these into Definition 5.5.5 gives exactly the displayed formulas for $\underline{E}(a)$ and $\overline{E}(a)$, hence for $\underline{G}(a)$ and $\overline{G}(a)$. □

**Definition 5.5.8** (SuperHyperRough set)**.** [406, 407] Let $(X, R)$ be an approximation space and let $T_1, \ldots, T_n$ be attributes with domains $J_1, \ldots, J_n$. Define the *super-parameter domain*

$$\mathcal{J} \ := \ \mathcal{P}(J_1) \times \cdots \times \mathcal{P}(J_n).$$

A *superhyperrough set* over $(X, R)$ is a pair $(F, \mathcal{J})$ where

$$F : \mathcal{J} \longrightarrow \mathcal{P}(X).$$

For each $A = (A_1, \ldots, A_n) \in \mathcal{J}$, define

$$\begin{aligned} \underline{F}(A) &:= \ \{ x \in X : \ [x]_R \subseteq F(A) \}, \\ \overline{F}(A) &:= \ \{ x \in X : \ [x]_R \cap F(A) \neq \emptyset \}. \end{aligned}$$

**Theorem 5.5.9** (HyperRough sets embed into SuperHyperRough sets)**.** *Let $(F, J)$ be a hyperrough set with $J = J_1 \times \cdots \times J_n$. Define $\mathcal{J} := \mathcal{P}(J_1) \times \cdots \times \mathcal{P}(J_n)$ and define*

$$F^{\uparrow} : \mathcal{J} \to \mathcal{P}(X) \quad \text{by} \quad F^{\uparrow}(\{a_1\}, \ldots, \{a_n\}) := F(a_1, \ldots, a_n),$$

*leaving $F^{\uparrow}$ arbitrary on non-singleton inputs if desired. Then $(F^{\uparrow}, \mathcal{J})$ is a superhyperrough set whose rough views coincide with those of $(F, J)$ on singleton parameter selections.*

*Proof.* By construction, $F^{\uparrow}$ is well-defined as a map $\mathcal{J} \to \mathcal{P}(X)$. For any $a = (a_1, \ldots, a_n) \in J$, let $A := (\{a_1\}, \ldots, \{a_n\}) \in \mathcal{J}$. Then $F^{\uparrow}(A) = F(a)$, so by Definitions 5.5.2 and 5.5.8,

$$\underline{F^{\uparrow}}(A) = \underline{F}(a), \qquad \overline{F^{\uparrow}}(A) = \overline{F}(a).$$

Thus the hyperrough set is recovered on singleton parameter selections. □

For reference, the relationships between the SuperhyperSoft set and the Superhyperrough set are illustrated in Figure 5.1.

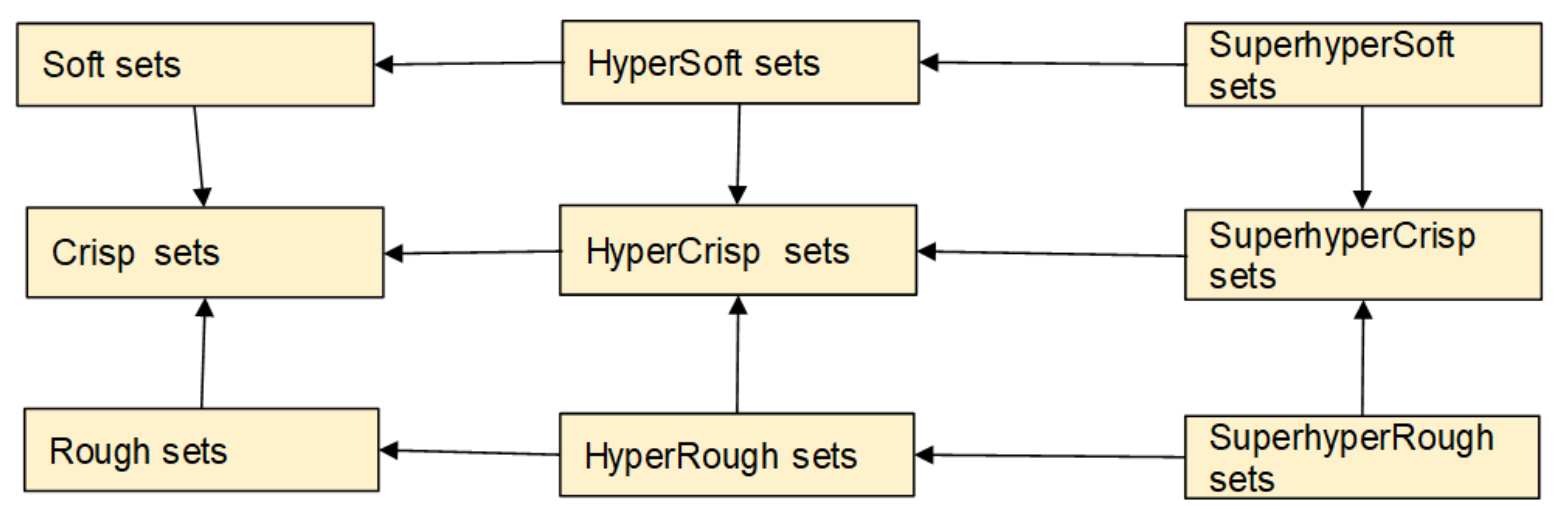

Figure 5.1: Some the SuperhyperSoft sets and the Superhyperrough sets Hierarchy. The set class at the origin of an arrow contains the set class at the destination of the arrow.

## 5.6 Single-Valued Neutrosophic Over/Under/Off Hypergraphs

We introduce three neutrosophic extensions of (crisp) hypergraphs in which the membership degrees of vertices and hyperedges are allowed to exceed 1 (overset), fall below 0 (underset), or do both (offset). Throughout, we work with *single-valued* degrees (real numbers), in contrast to interval-valued variants.

**Definition 5.6.1** (Crisp hypergraph)**.** A *(crisp) hypergraph* is a pair $H^* = (V, E)$ where $V$ is a finite nonempty set (vertices) and

$$E \subseteq \mathcal{P}(V) \setminus \{\emptyset\}$$

is a finite family of nonempty subsets of $V$ (hyperedges).

**Notation 2** (Over/Under limits)**.** *Fix real bounds*

$$\Psi < 0 < 1 < \Omega.$$

*We use* $\Omega$ *as an* overlimit *and* $\Psi$ *as an* underlimit.

**Definition 5.6.2** (Single-valued neutrosophic over/under/off triples)**.** Let $\Omega > 1$ and $\Psi < 0$.

(i) A *single-valued neutrosophic over-triple* is a triple $(T, I, F) \in [0, \Omega]^3$.

(ii) A *single-valued neutrosophic under-triple* is a triple $(T, I, F) \in [\Psi, 1]^3$.

(iii) A *single-valued neutrosophic off-triple* is a triple $(T, I, F) \in [\Psi, \Omega]^3$.

Optionally (and commonly), one may impose the sum-bound

$$0 \leq T+I+F \leq 3 \quad \text{(standard)}, \qquad 0 \leq T+I+F \leq 3\Omega \quad \text{(over/off)}, \qquad 3\Psi \leq T+I+F \leq 3 \quad \text{(under)}.$$

In what follows, we only require the coordinate-wise bounds; any additional sum-bound can be added as a modeling choice.

**Definition 5.6.3** (Single-Valued Neutrosophic Over Hypergraph)**.** Fix $\Omega > 1$. A *single-valued neutrosophic over hypergraph* is a tuple

$$H_{\text{over}} = \big(V, E;\ T_V, I_V, F_V;\ T_E, I_E, F_E\big),$$

where $(V, E)$ is a crisp hypergraph (Definition 5.6.1) and

$$T_V,\ I_V,\ F_V : V \to [0, \Omega], \qquad T_E,\ I_E,\ F_E : E \to [0, \Omega]$$

are vertex- and hyperedge-membership functions (truth/indeterminacy/falsity degrees). We call $e \in E$ *active* if $(T_E(e), I_E(e), F_E(e)) \neq (0, 0, 0)$.

Optionally, one may impose the componentwise incidence constraints

$$T_E(e) \leq \min_{v \in e} T_V(v), \qquad I_E(e) \leq \min_{v \in e} I_V(v), \qquad F_E(e) \leq \min_{v \in e} F_V(v) \qquad (\forall e \in E),$$

which mirror the usual admissibility conditions for neutrosophic hypergraph models.

**Definition 5.6.4** (Single-Valued Neutrosophic Under Hypergraph)**.** Fix $\Psi < 0$. A *single-valued neutrosophic under hypergraph* is a tuple

$$H_{\text{under}} = \big(V, E;\ T_V, I_V, F_V;\ T_E, I_E, F_E\big),$$

where $(V, E)$ is a crisp hypergraph and

$$T_V,\ I_V,\ F_V : V \to [\Psi, 1], \qquad T_E,\ I_E,\ F_E : E \to [\Psi, 1].$$

An edge $e$ is *active* if $(T_E(e), I_E(e), F_E(e)) \neq (0, 0, 0)$.

Optionally, one may impose the same componentwise incidence constraints as in Definition 5.6.3.

**Definition 5.6.5** (Single-Valued Neutrosophic Off Hypergraph)**.** Fix $\Psi < 0 < 1 < \Omega$. A *single-valued neutrosophic off hypergraph* is a tuple

$$H_{\text{off}} = \big(V, E;\ T_V, I_V, F_V;\ T_E, I_E, F_E\big),$$

where $(V, E)$ is a crisp hypergraph and

$$T_V,\ I_V,\ F_V : V \to [\Psi, \Omega], \qquad T_E,\ I_E,\ F_E : E \to [\Psi, \Omega].$$

Again, $e$ is *active* if $(T_E(e), I_E(e), F_E(e)) \neq (0, 0, 0)$.

Optionally, one may impose the componentwise incidence constraints as above.

**Theorem 5.6.6** (Reduction to the crisp hypergraph)**.** *Each of the structures in Definitions 5.6.3–5.6.5 has an underlying crisp hypergraph $(V, E)$ obtained by forgetting the membership functions.*

*Proof.* By definition, each structure explicitly contains $(V, E)$ as part of its data. □

**Theorem 5.6.7** (Graphs as the 2-uniform case)**.** *Let $H_{\text{over}}$ be a single-valued neutrosophic over hypergraph. If every hyperedge $e \in E$ satisfies $|e| = 2$, then $(V, E)$ is a simple graph, and $H_{\text{over}}$ is exactly a single-valued neutrosophic over graph on that underlying graph. The same statement holds for the under and off cases.*

*Proof.* If $|e| = 2$ for all $e \in E$, then $E \subseteq \binom{V}{2}$ and hence $(V, E)$ is a simple graph. The membership functions $(T_V, I_V, F_V)$ and $(T_E, I_E, F_E)$ assign neutrosophic degrees to vertices and edges in the same coordinate ranges as in Definitions 5.6.3–5.6.5, so the structure coincides with the corresponding graph-level notion. □

**Theorem 5.6.8** (Off includes Over and Under as special parameter regimes)**.** *Fix $\Psi < 0 < 1 < \Omega$.*

(i) *If we restrict a single-valued neutrosophic off hypergraph by additionally requiring all degrees to lie in $[0, \Omega]$ (equivalently, by setting $\Psi := 0$), then it becomes a single-valued neutrosophic over hypergraph.*

(ii) *If we restrict a single-valued neutrosophic off hypergraph by additionally requiring all degrees to lie in $[\Psi, 1]$ (equivalently, by setting $\Omega := 1$), then it becomes a single-valued neutrosophic under hypergraph.*

*Proof.* Both claims follow immediately from the interval inclusions

$$[0, \Omega] \subseteq [\Psi, \Omega] \qquad \text{and} \qquad [\Psi, 1] \subseteq [\Psi, \Omega],$$

together with the fact that the underlying crisp hypergraph $(V, E)$ is unchanged. □

## 5.7 HyperCrisp Sets, SuperHyperCrisp Sets, and SuperHyperFuzzy Sets

This section introduces three "hyper" extensions of set-valued membership models. A *hypermembership* assigns to each element a *set* of admissible membership values rather than a single number. The "superhyper" versions iterate the powerset construction once more, so that the membership information itself becomes a higher-order set object.

**Notation 3** (Nonempty powerset and iterated powersets)**.** *For any set $Y$, write*

$$\mathcal{P}(Y) := \{A \mid A \subseteq Y\}, \qquad \mathcal{P}^*(Y) := \mathcal{P}(Y) \setminus \{\emptyset\}$$

*for the powerset and the* nonempty *powerset, respectively. Define iterated powersets by*

$$\mathcal{P}^0(Y) := Y, \qquad \mathcal{P}^{k+1}(Y) := \mathcal{P}\big(\mathcal{P}^k(Y)\big) \quad (k \in \mathbb{N}_0),$$

*and similarly $\mathcal{P}^{*\,0}(Y) := Y$ and $\mathcal{P}^{*\,k+1}(Y) := \mathcal{P}^*(\mathcal{P}^k(Y))$.*

**Definition 5.7.1** (HyperCrisp set)**.** Let $X$ be a nonempty set. A *hypercrisp set* on $X$ is a map

$$\widetilde{C} : X \longrightarrow \mathcal{P}^*(\{0,1\}),$$

so that each $x \in X$ is assigned a nonempty subset $\widetilde{C}(x) \in \big\{\{0\},\{1\},\{0,1\}\big\}$. Intuitively, $\{1\}$ means "certainly in", $\{0\}$ means "certainly out", and $\{0,1\}$ means "undetermined between in/out".

**Theorem 5.7.2** (Crisp sets as a special case)**.** *Every crisp subset $A \subseteq X$ induces a hypercrisp set $\widetilde{C}_A$ by*

$$\widetilde{C}_A(x) := \begin{cases} \{1\}, & x \in A, \\ \{0\}, & x \notin A, \end{cases} \qquad (x \in X).$$

*Conversely, if a hypercrisp set $\widetilde{C}$ satisfies $\widetilde{C}(x) \in \{\{0\},\{1\}\}$ for all $x \in X$, then it uniquely determines a crisp set $A := \{x \in X : \widetilde{C}(x) = \{1\}\}$.*

*Proof.* The first claim is immediate from the definition of $\widetilde{C}_A$. For the converse, the stated $A$ is well-defined and satisfies $\widetilde{C} = \widetilde{C}_A$ by construction. □

**Definition 5.7.3** (SuperHyperCrisp set)**.** Let $X$ be a nonempty set. A *superhypercrisp set* on $X$ is a map

$$\mathbb{C} : X \longrightarrow \mathcal{P}^*\big(\mathcal{P}^*(\{0,1\})\big) = \mathcal{P}^{*\,2}(\{0,1\}),$$

so that each $x \in X$ is assigned a nonempty *family* of nonempty subsets of $\{0,1\}$. Equivalently, $\mathbb{C}(x)$ is a nonempty collection of candidates $\widetilde{C}(x) \in \{\{0\},\{1\},\{0,1\}\}$.

**Interpretation.** A hypercrisp set assigns to $x$ one admissible set of crisp-membership values; a superhypercrisp set assigns to $x$ a nonempty *set of admissible admissible-sets*, i.e., two layers of set-valued uncertainty.

**Remark 5.7.4.** Every hypercrisp set $\widetilde{C} : X \to \mathcal{P}^*(\{0,1\})$ canonically embeds into a superhypercrisp set by the singleton lift

$$\mathbb{C}_{\widetilde{C}}(x) := \{\widetilde{C}(x)\} \in \mathcal{P}^*\big(\mathcal{P}^*(\{0,1\})\big).$$

**Definition 5.7.5** (HyperFuzzy set)**.** Let $X$ be a nonempty set. A *hyperfuzzy set* on $X$ is a map

$$\widetilde{\mu} : X \longrightarrow \mathcal{P}^*([0,1]),$$

assigning to each $x \in X$ a nonempty set of admissible membership degrees in $[0,1]$.

**Definition 5.7.6** (SuperHyperFuzzy set)**.** Let $X$ be a nonempty set. A *superhyperfuzzy set* on $X$ is a map

$$\mathbb{F} : X \longrightarrow \mathcal{P}^*\big(\mathcal{P}^*([0,1])\big) = \mathcal{P}^{*\,2}([0,1]).$$

Thus, for each $x \in X$, $\mathbb{F}(x)$ is a nonempty collection of nonempty subsets of $[0,1]$.

**Interpretation.** Each inner set $A \in \mathbb{F}(x)$ can be viewed as one admissible *scenario* of possible membership degrees, while the outer set $\mathbb{F}(x)$ collects multiple such scenarios.

**Theorem 5.7.7** (SuperHyperFuzzy sets generalize HyperFuzzy sets)**.** *Every hyperfuzzy set* $\widetilde{\mu} : X \longrightarrow \mathcal{P}^*([0,1])$ *induces a superhyperfuzzy set* $\mathbb{F}_{\widetilde{\mu}} : X \longrightarrow \mathcal{P}^*(\mathcal{P}^*([0,1]))$ *by*

$$\mathbb{F}_{\widetilde{\mu}}(x) := \{\widetilde{\mu}(x)\} \qquad (x \in X).$$

*Conversely, a superhyperfuzzy set* $\mathbb{F}$ *reduces to a hyperfuzzy set precisely when* $|\mathbb{F}(x)| = 1$ *for every* $x \in X$.

*Proof.* The map $\mathbb{F}_{\widetilde{\mu}}$ is well-defined because $\widetilde{\mu}(x)$ is nonempty, so the singleton $\{\widetilde{\mu}(x)\}$ is a nonempty element of $\mathcal{P}^*(\mathcal{P}^*([0,1]))$. For the converse, if $|\mathbb{F}(x)| = 1$ then $\mathbb{F}(x) = \{A_x\}$ for a unique nonempty $A_x \subseteq [0,1]$; defining $\widetilde{\mu}(x) := A_x$ yields a hyperfuzzy set with $\mathbb{F} = \mathbb{F}_{\widetilde{\mu}}$. □

**Remark 5.7.8** (From HyperCrisp to HyperFuzzy)**.** Since $\{0,1\} \subseteq [0,1]$, every hypercrisp set $\widetilde{C} : X \rightarrow \mathcal{P}^*(\{0,1\})$ is a hyperfuzzy set by composing with the inclusion $\mathcal{P}^*(\{0,1\}) \hookrightarrow \mathcal{P}^*([0,1])$.

As discussed above, if we explicitly define the n-SuperHyperCrisp Set, it would be as follows.

**Definition 5.7.9** ($n$-SuperHyperCrisp Set)**.** Let $X$ be a non-empty set, and $n \geq 0$ be an integer. An *n-SuperHyperCrisp Set* $\tilde{C}_n$ over $X$ is a mapping:

$$\tilde{C}_n : \tilde{\mathcal{P}}_n^*(X) \rightarrow \tilde{\mathcal{P}}(\{0,1\}),$$

where $\tilde{\mathcal{P}}_n^*(X)$ denotes the family of all non-empty elements of the $n$-th PowerSet $\mathcal{P}_n^*(X)$, and $\tilde{\mathcal{P}}(\{0,1\})$ denotes the family of all non-empty subsets of $\{0,1\}$.

For each element $A \in \tilde{\mathcal{P}}_n^*(X)$, $\tilde{C}_n(A) \subseteq \{0,1\}$, representing the membership degree(s) of $A$.

**Theorem 5.7.10.** *An n-SuperHyperCrisp Set* $\tilde{C}_n : \tilde{\mathcal{P}}_n^*(X) \rightarrow \tilde{\mathcal{P}}(\{0,1\})$ *aligns with the n-th PowerSet* $\mathcal{P}_n^*(X)$ *and generalizes the SuperHyperCrisp Set to include higher-order structures.*

*Proof.* We will demonstrate that the domain of $\tilde{C}_n$ is the set of all non-empty elements of $\mathcal{P}_n^*(X)$, which inherently includes the structures of the $n$-th PowerSet.

1. **Base Case:** $n = 0$**.**
   - $\mathcal{P}_0^*(X) = X$.
   - The mapping $\tilde{C}_0 : X \rightarrow \tilde{\mathcal{P}}(\{0,1\})$.
   - This reduces to the definition of a *HyperCrisp Set*, where each element $x \in X$ is mapped to $\tilde{C}_0(x) \subseteq \{0,1\}$.
2. **First-Level PowerSet:** $n = 1$**.**
   - $\mathcal{P}_1^*(X) = \mathcal{P}(X)$.
   - The mapping $\tilde{C}_1 : \tilde{\mathcal{P}}(X) \rightarrow \tilde{\mathcal{P}}(\{0,1\})$, where $\tilde{\mathcal{P}}(X) = \mathcal{P}(X) \setminus \{\emptyset\}$.
   - This corresponds to the *SuperHyperCrisp Set* as previously defined.
3. **Recursive Case:** $n \geq 2$**.**
   - $\mathcal{P}_n^*(X) = \mathcal{P}(\mathcal{P}_{n-1}^*(X))$.
   - The mapping $\tilde{C}_n : \tilde{\mathcal{P}}_n^*(X) \rightarrow \tilde{\mathcal{P}}(\{0,1\})$.
   - Each non-empty element $A \in \mathcal{P}_n^*(X)$ is assigned a subset $\tilde{C}_n(A) \subseteq \{0,1\}$, capturing higher-order relationships.
   - This extends the SuperHyperCrisp Set to higher-order structures corresponding to the $n$-th PowerSet.

Thus, the $n$-SuperHyperCrisp Set aligns with the $n$-th PowerSet $\mathcal{P}_n^*(X)$ for all $n \geq 0$, generalizing the SuperHyperCrisp Set to include hierarchical and recursive structures. □

**Definition 5.7.11** (SuperHyperFuzzy Set)**.** Let $X$ be a non-empty set. A mapping $\tilde{\mu} : \tilde{\mathcal{P}}(X) \to \tilde{\mathcal{P}}([0,1])$ is called a *SuperHyperFuzzy Set* over $X$, where $\tilde{\mathcal{P}}(X)$ denotes the family of all non-empty subsets of $X$, and $\tilde{\mathcal{P}}([0,1])$ denotes the family of all non-empty subsets of the interval $[0,1]$.

In this structure:

- Each element $A \in \tilde{\mathcal{P}}(X)$ is a non-empty subset of $X$.
- The mapping $\tilde{\mu}$ assigns to each $A \in \tilde{\mathcal{P}}(X)$ a non-empty subset $\tilde{\mu}(A) \subseteq [0,1]$, representing the degrees of membership associated with the subset $A$.

**Example 5.7.12.** A *SuperHyperFuzzy Set* $\tilde{B}$ over $X$ is defined by a mapping $\tilde{\mu}_{\tilde{B}} : \tilde{\mathcal{P}}(X) \to \tilde{\mathcal{P}}([0,1])$, where $\tilde{\mathcal{P}}(X)$ denotes the family of all non-empty subsets of $X$.

The non-empty subsets of $X$ are:

$$\tilde{\mathcal{P}}(X) = \{\{x_1\}, \{x_2\}, \{x_3\}, \{x_1, x_2\}, \{x_1, x_3\}, \{x_2, x_3\}, \{x_1, x_2, x_3\}\}.$$

We define $\tilde{\mu}_{\tilde{B}}$ as:

$$\begin{aligned}
\tilde{\mu}_{\tilde{B}}(\{x_1\}) &= \{0.1, 0.2, 0.3\},\\
\tilde{\mu}_{\tilde{B}}(\{x_2\}) &= \{0.4, 0.5, 0.6\},\\
\tilde{\mu}_{\tilde{B}}(\{x_3\}) &= \{0.6, 0.7, 0.8\},\\
\tilde{\mu}_{\tilde{B}}(\{x_1, x_2\}) &= \{0.3, 0.5\},\\
\tilde{\mu}_{\tilde{B}}(\{x_1, x_3\}) &= \{0.4, 0.6\},\\
\tilde{\mu}_{\tilde{B}}(\{x_2, x_3\}) &= \{0.5, 0.7\},\\
\tilde{\mu}_{\tilde{B}}(\{x_1, x_2, x_3\}) &= \{0.6, 0.8\}.
\end{aligned}$$

This means:

- Each non-empty subset of $X$ is assigned a set of membership degrees in $\tilde{B}$.
- For singleton subsets, the membership degrees correspond to those in the hyperfuzzy set $\tilde{A}$.
- For subsets with more elements, the membership degrees represent the collective membership of the group.

**Theorem 5.7.13.** *A SuperHyperFuzzy Set is a generalization of a SuperHyperCrisp Set.*

*Proof.* Let $X$ be a non-empty set, and let $\tilde{C} : \tilde{\mathcal{P}}(X) \to \tilde{\mathcal{P}}(\{0,1\})$ be a SuperHyperCrisp Set over $X$, where each subset $A \in \tilde{\mathcal{P}}(X)$ is assigned a subset $\tilde{C}(A) \subseteq \{0,1\}$ as its membership degrees. Specifically, $\tilde{C}(A) = \{0\}$ if $A \notin \tilde{C}$ and $\tilde{C}(A) = \{1\}$ if $A \in \tilde{C}$.

Now, consider a SuperHyperFuzzy Set $\tilde{\mu} : \tilde{\mathcal{P}}(X) \to \tilde{\mathcal{P}}([0,1])$, where each subset $A \in \tilde{\mathcal{P}}(X)$ is assigned a non-empty subset $\tilde{\mu}(A) \subseteq [0,1]$, allowing partial membership values.

To show that a SuperHyperCrisp Set is a special case of a SuperHyperFuzzy Set, we define $\tilde{\mu}(A) = \tilde{C}(A)$, where $\tilde{C}(A) \subseteq \{0,1\} \subseteq [0,1]$. Here:

- If $\tilde{C}(A) = \{1\}$, then $\tilde{\mu}(A) = \{1\}$, indicating full membership in the SuperHyperFuzzy Set.
- If $\tilde{C}(A) = \{0\}$, then $\tilde{\mu}(A) = \{0\}$, indicating non-membership.

Thus, every SuperHyperCrisp Set can be represented as a SuperHyperFuzzy Set where membership values are restricted to the discrete set $\{0,1\}$, proving that the SuperHyperFuzzy Set generalizes the SuperHyperCrisp Set. □

**Theorem 5.7.14.** *The SuperHyperFuzzy Set generalizes both the HyperFuzzy Set and the traditional Fuzzy Set. Specifically:*

1. *Every Fuzzy Set can be represented as a SuperHyperFuzzy Set.*
2. *Every HyperFuzzy Set can be represented as a SuperHyperFuzzy Set.*

*Proof.* A Fuzzy Set is a mapping $\mu : X \to [0,1]$ that assigns to each element $x \in X$ a membership degree $\mu(x) \in [0,1]$.

To represent a Fuzzy Set as a SuperHyperFuzzy Set $\tilde{\mu} : \tilde{\mathcal{P}}(X) \to \tilde{\mathcal{P}}([0,1])$, we define:

$$\tilde{\mu}(A) = \{\mu(a) \mid a \in A\}$$

for all $A \in \tilde{\mathcal{P}}(X)$.

This means:

- For singleton subsets $\{x\}$, we have $\tilde{\mu}(\{x\}) = \{\mu(x)\}$.
- For larger subsets $A \subseteq X$, $\tilde{\mu}(A)$ collects the membership degrees of all elements in $A$.

Thus, the Fuzzy Set is embedded into the SuperHyperFuzzy Set by considering the membership degrees over subsets of $X$.

A HyperFuzzy Set is a mapping $\tilde{\mu} : X \to \tilde{\mathcal{P}}([0,1])$ that assigns to each element $x \in X$ a non-empty subset $\tilde{\mu}(x) \subseteq [0,1]$.

We construct a SuperHyperFuzzy Set $\tilde{\mu}' : \tilde{\mathcal{P}}(X) \to \tilde{\mathcal{P}}([0,1])$ by defining:

$$\tilde{\mu}'(A) = \bigcup_{a \in A} \tilde{\mu}(a)$$

for all $A \in \tilde{\mathcal{P}}(X)$.

This implies:

- For singleton subsets $\{x\}$, $\tilde{\mu}'(\{x\}) = \tilde{\mu}(x)$.
- For subsets $A \subseteq X$, $\tilde{\mu}'(A)$ is the union of the membership degree sets of all elements in $A$.

Therefore, the HyperFuzzy Set is a special case of the SuperHyperFuzzy Set when restricted to singleton subsets of $X$.

Since both Fuzzy Sets and HyperFuzzy Sets can be represented within the framework of SuperHyperFuzzy Sets, it follows that the SuperHyperFuzzy Set generalizes these concepts. □

As discussed above, if we explicitly define the n-SuperHyperFuzzy Set, it would be as follows.

**Definition 5.7.15** ($n$-SuperHyperFuzzy Set)**.** Let $X$ be a non-empty set, and $n \geq 0$ be an integer. An *$n$-SuperHyperFuzzy Set* is a mapping:

$$\tilde{\mu}_n : \tilde{\mathcal{P}}_n^*(X) \to \tilde{\mathcal{P}}([0,1]),$$

where:

- $\tilde{\mathcal{P}}_n^*(X)$ denotes the family of all non-empty elements of the $n$-th PowerSet $\mathcal{P}_n^*(X)$, defined recursively as:

  $$\mathcal{P}_0^*(X) = X, \quad \mathcal{P}_1^*(X) = \mathcal{P}(X), \quad \mathcal{P}_n^*(X) = \mathcal{P}(\mathcal{P}_{n-1}^*(X)), \text{ for } n \geq 2,$$

  with $\tilde{\mathcal{P}}_n^*(X) = \mathcal{P}_n^*(X) \setminus \{\emptyset\}$.
- $\tilde{\mathcal{P}}([0,1])$ denotes the family of all non-empty subsets of the interval $[0,1]$.

**Structure:**

1. Each element $A \in \tilde{\mathcal{P}}_n^*(X)$ is a non-empty subset within the $n$-th PowerSet hierarchy of $X$.
2. The mapping $\tilde{\mu}_n$ assigns to each $A \in \tilde{\mathcal{P}}_n^*(X)$ a non-empty subset $\tilde{\mu}_n(A) \subseteq [0,1]$, representing the degrees of membership associated with the subset $A$.

**Properties:**

- If $n = 0$, $\tilde{\mathcal{P}}_0^*(X) = X$, and the structure reduces to a standard fuzzy set:

$$\tilde{\mu}_0 : X \to [0,1].$$

- For $n = 1$, $\tilde{\mathcal{P}}_1^*(X) = \tilde{\mathcal{P}}(X)$, and the structure represents a SuperHyperFuzzy Set:

$$\tilde{\mu}_1 : \tilde{\mathcal{P}}(X) \to \tilde{\mathcal{P}}([0,1]).$$

- For $n \geq 2$, the structure recursively extends to higher-order fuzzy relationships:

$$\tilde{\mu}_n : \tilde{\mathcal{P}}_n^*(X) \to \tilde{\mathcal{P}}([0,1]).$$

The $n$-SuperHyperFuzzy Set generalizes the concept of fuzzy sets to hierarchical and recursive levels of membership, allowing for higher-order relationships and fuzzy degrees associated with subsets of subsets, and so on, up to the $n$-th PowerSet hierarchy of $X$.

**Definition 5.7.16** (SuperHyperVague Set)**.** Let $X$ be a non-empty set. A mapping $\tilde{A} : \tilde{\mathcal{P}}(X) \to \tilde{\mathcal{P}}([0,1]^2)$ is called a *SuperHyperVague Set* over $X$, where:

- $\tilde{\mathcal{P}}(X)$ denotes the family of all non-empty subsets of $X$.
- $\tilde{\mathcal{P}}([0,1]^2)$ denotes the family of all non-empty subsets of the unit square $[0,1]^2$.

For each $A \in \tilde{\mathcal{P}}(X)$, $\tilde{A}(A) \subseteq [0,1]^2$, and for each pair $(t, f) \in \tilde{A}(A)$, it satisfies:

$$0 \leq t + f \leq 1,$$

where $t$ represents the degree of truth membership and $f$ represents the degree of falsity membership for the subset $A$.

**Definition 5.7.17** (SuperHyperNeutrosophic Set)**.** Let $X$ be a non-empty set. A mapping $\tilde{A} : \tilde{\mathcal{P}}(X) \to \tilde{\mathcal{P}}([0,1]^3)$ is called a *SuperHyperNeutrosophic Set* over $X$, where:

- $\tilde{\mathcal{P}}(X)$ denotes the family of all non-empty subsets of $X$.
- $\tilde{\mathcal{P}}([0,1]^3)$ denotes the family of all non-empty subsets of the unit cube $[0,1]^3$.

For each $A \in \tilde{\mathcal{P}}(X)$, $\tilde{A}(A) \subseteq [0,1]^3$, and for each triplet $(T, I, F) \in \tilde{A}(A)$, it satisfies:

$$0 \leq T + I + F \leq 3,$$

where $T$ represents the degree of truth membership, $I$ represents the degree of indeterminacy, and $F$ represents the degree of falsity membership for the subset $A$.

**Definition 5.7.18** (SuperHyperPlithogenic Set)**.** Let $X$ be a non-empty set, and let $V = \{v_1, v_2, \ldots, v_n\}$ be a set of attributes, each with a set of possible values $P_{v_i}$. A *SuperHyperPlithogenic Set SHPS* over $X$ is defined as:

$$SHPS = (P, V, \{P_{v_i}\}_{i=1}^n, \{\tilde{pdf}_i\}_{i=1}^n, pCF),$$

where:

- $P \subseteq X$ is a subset of the universe.
- For each attribute $v_i$, $P_{v_i}$ is the set of possible values.
- For each attribute $v_i$, $\tilde{pdf}_i : P \times P_{v_i} \to \tilde{\mathcal{P}}([0,1]^s)$ is the *Hyper Degree of Appurtenance Function (HDAF)*, assigning to each element $x \in P$ and attribute value $a_i \in P_{v_i}$ a non-empty subset of $[0,1]^s$.
- $pCF : \left(\bigcup_{i=1}^n P_{v_i}\right) \times \left(\bigcup_{i=1}^n P_{v_i}\right) \to [0,1]^t$ is the *Degree of Contradiction Function (DCF).*
- $s$ and $t$ are positive integers representing the dimensions of the membership degrees and contradiction degrees, respectively.

**Theorem 5.7.19.** *Every SuperHyperVague Set can be transformed into a HyperVague Set and a SuperHyperFuzzy Set.*

*Proof.* Let $\tilde{A} : \tilde{\mathcal{P}}(X) \to \tilde{\mathcal{P}}([0,1]^2)$ be a SuperHyperVague Set over $X$.

Consider the restriction of $\tilde{A}$ to singleton subsets of $X$:

$$\tilde{A}|_X : X \to \tilde{\mathcal{P}}([0,1]^2), \quad \tilde{A}|_X(x) = \tilde{A}(\{x\}).$$

This mapping assigns to each element $x \in X$ a non-empty subset $\tilde{A}|_X(x) \subseteq [0,1]^2$, satisfying $0 \leq t + f \leq 1$ for each $(t,f) \in \tilde{A}|_X(x)$. Therefore, $\tilde{A}|_X$ is a HyperVague Set over $X$.

Define a mapping $\tilde{\mu} : \tilde{\mathcal{P}}(X) \to \tilde{\mathcal{P}}([0,1])$ by:

$$\tilde{\mu}(A) = \{t \mid (t,f) \in \tilde{A}(A),\ f \in [0,1]\}.$$

For each $A \in \tilde{\mathcal{P}}(X)$, $\tilde{\mu}(A) \subseteq [0,1]$ represents the degrees of truth membership extracted from $\tilde{A}(A)$. Thus, $\tilde{\mu}$ is a SuperHyperFuzzy Set over $X$. □

**Theorem 5.7.20.** *Every SuperHyperNeutrosophic Set can be transformed into a SuperHyperVague Set and a HyperNeutrosophic Set.*

*Proof.* Let $\tilde{A} : \tilde{\mathcal{P}}(X) \to \tilde{\mathcal{P}}([0,1]^3)$ be a SuperHyperNeutrosophic Set over $X$.

Define a mapping $\tilde{B} : \tilde{\mathcal{P}}(X) \to \tilde{\mathcal{P}}([0,1]^2)$ by projecting the neutrosophic membership degrees onto the truth and falsity components:

$$\tilde{B}(A) = \{(T,F) \mid (T,I,F) \in \tilde{A}(A),\ I \in [0,1]\}.$$

For each $A \in \tilde{\mathcal{P}}(X)$, $\tilde{B}(A) \subseteq [0,1]^2$, and $0 \leq T + F \leq 2$ (since $0 \leq T + I + F \leq 3$).

Adjust the normalization by defining:

$$\tilde{B}'(A) = \left\{ \left( \frac{T}{T+F}, \frac{F}{T+F} \right) \middle| (T,F) \in \tilde{B}(A),\ T + F > 0 \right\}.$$

Now, for each $(t,f) \in \tilde{B}'(A)$, $0 \leq t + f = 1$.

Thus, $\tilde{B}'$ is a SuperHyperVague Set over $X$.

Consider the restriction of $\tilde{A}$ to singleton subsets of $X$:

$$\tilde{A}|_X : X \to \tilde{\mathcal{P}}([0,1]^3), \quad \tilde{A}|_X(x) = \tilde{A}(\{x\}).$$

This mapping assigns to each $x \in X$ a non-empty subset $\tilde{A}|_X(x) \subseteq [0,1]^3$, satisfying $0 \leq T + I + F \leq 3$.

Therefore, $\tilde{A}|_X$ is a HyperNeutrosophic Set over $X$. □

**Theorem 5.7.21.** *Every SuperHyperPlithogenic Set can be transformed into a HyperPlithogenic Set and the following SuperHyper Sets:*

- *SuperHyperFuzzy Set with* $s = 1$ *and* $t = 1$.
- *SuperHyperVague Set with* $s = 2$ *and* $t = 1$.
- *SuperHyperNeutrosophic Set with* $s = 3$ *and* $t = 1$.

*Proof.* Let $SHPS = (P, V, \{P_{v_i}\}_{i=1}^n, \{\tilde{pdf}_i\}_{i=1}^n, pCF)$ be a SuperHyperPlithogenic Set over $X$, with $s$ and $t$ representing the dimensions of the membership degrees and contradiction degrees, respectively.

*Transformation to HyperPlithogenic Set*:

Consider the restriction of $SHPS$ to singleton subsets of $P$:

$$HPS = (P, V, \{P_{v_i}\}_{i=1}^n, \{pdf_i\}_{i=1}^n, pCF),$$

where $pdf_i : P \times P_{v_i} \to [0, 1]^s$ is defined by:

$$pdf_i(x, a_i) = \tilde{pdf}_i(\{x\}, a_i).$$

Since $\tilde{pdf}_i$ assigns a subset of $[0, 1]^s$ to each element, we can select representative values or aggregate them to obtain $pdf_i$.

Therefore, $HPS$ is a HyperPlithogenic Set over $X$.

*Transformation to SuperHyperFuzzy Set (*$s = 1, t = 1$*)*:

If $s = 1$ and $t = 1$, the membership degrees are single-dimensional, and the contradiction degrees are also single-dimensional.

Define $\tilde{\mu} : \tilde{\mathcal{P}}(X) \to \tilde{\mathcal{P}}([0, 1])$ by:

$$\tilde{\mu}(A) = \bigcup_{x \in A} \bigcup_{i=1}^{n} \bigcup_{a_i \in P_{v_i}} \tilde{pdf}_i(x, a_i).$$

This mapping assigns to each subset $A \subseteq X$ a non-empty subset of $[0, 1]$, forming a SuperHyperFuzzy Set.

*Transformation to SuperHyperVague Set (*$s = 2, t = 1$*)*:

If $s = 2$, the membership degrees are two-dimensional, corresponding to the truth and falsity components.

Define $\tilde{A} : \tilde{\mathcal{P}}(X) \to \tilde{\mathcal{P}}([0, 1]^2)$ by:

$$\tilde{A}(A) = \bigcup_{x \in A} \bigcup_{i=1}^{n} \bigcup_{a_i \in P_{v_i}} \tilde{pdf}_i(x, a_i),$$

where $\tilde{pdf}_i(x, a_i) \subseteq [0, 1]^2$.

This forms a SuperHyperVague Set over $X$.

*Transformation to SuperHyperNeutrosophic Set (*$s = 3, t = 1$*)*:

If $s = 3$, the membership degrees have three components, representing truth, indeterminacy, and falsity.

Define $\tilde{A} : \tilde{\mathcal{P}}(X) \to \tilde{\mathcal{P}}([0, 1]^3)$ similarly:

$$\tilde{A}(A) = \bigcup_{x \in A} \bigcup_{i=1}^{n} \bigcup_{a_i \in P_{v_i}} \tilde{pdf}_i(x, a_i),$$

with $\tilde{pdf}_i(x, a_i) \subseteq [0, 1]^3$.

This mapping constitutes a SuperHyperNeutrosophic Set over $X$. □

As discussed above, if we explicitly define the n-SuperHyperVague Set n-SuperHyperNeutrosophic Set, and SuperHyperPlithogenic set, it would be as follows.

**Definition 5.7.22** ($n$-SuperHyperVague Set)**.** Let $X$ be a non-empty set, and $n \geq 0$ be an integer. An *n-SuperHyperVague Set* is a mapping:

$$\tilde{V}_n : \tilde{\mathcal{P}}_n^*(X) \to \tilde{\mathcal{P}}([0,1]^2),$$

where:

- $\tilde{\mathcal{P}}_n^*(X)$ denotes the family of all non-empty elements of the $n$-th PowerSet $\mathcal{P}_n^*(X)$, defined recursively as:

  $$\mathcal{P}_0^*(X) = X, \quad \mathcal{P}_1^*(X) = \mathcal{P}(X), \quad \mathcal{P}_n^*(X) = \mathcal{P}(\mathcal{P}_{n-1}^*(X)), \text{ for } n \geq 2,$$

  with $\tilde{\mathcal{P}}_n^*(X) = \mathcal{P}_n^*(X) \setminus \{\emptyset\}$.
- $\tilde{\mathcal{P}}([0,1]^2)$ denotes the family of all non-empty subsets of the unit square $[0,1]^2$.

**Structure:**

1. Each element $A \in \tilde{\mathcal{P}}_n^*(X)$ is a non-empty subset within the $n$-th PowerSet hierarchy of $X$.
2. The mapping $\tilde{V}_n$ assigns to each $A \in \tilde{\mathcal{P}}_n^*(X)$ a non-empty subset $\tilde{V}_n(A) \subseteq [0,1]^2$, representing pairs $(t, f)$, where:

   $$0 \leq t + f \leq 1,$$

   $t$ represents the degree of truth membership, and $f$ represents the degree of falsity membership for the subset $A$.

**Properties:**

- If $n = 0$, $\tilde{\mathcal{P}}_0^*(X) = X$, and the structure reduces to a standard vague set:

  $$\tilde{V}_0 : X \to [0,1]^2.$$

- For $n = 1$, $\tilde{\mathcal{P}}_1^*(X) = \tilde{\mathcal{P}}(X)$, and the structure represents a SuperHyperVague Set:

  $$\tilde{V}_1 : \tilde{\mathcal{P}}(X) \to \tilde{\mathcal{P}}([0,1]^2).$$

- For $n \geq 2$, the structure recursively extends to higher-order vague relationships:

  $$\tilde{V}_n : \tilde{\mathcal{P}}_n^*(X) \to \tilde{\mathcal{P}}([0,1]^2).$$

The $n$-SuperHyperVague Set generalizes the concept of vague sets to hierarchical and recursive levels, allowing for higher-order relationships and vague degrees associated with subsets of subsets, and so on, up to the $n$-th PowerSet hierarchy of $X$. The use of pairs $(t, f)$ ensures the balance between truth and falsity membership for each subset at every level.

**Definition 5.7.23** ($n$-SuperHyperNeutrosophic Set)**.** Let $X$ be a non-empty set, and $n \geq 0$ be an integer. An *n-SuperHyperNeutrosophic Set* is a mapping:

$$\tilde{N}_n : \tilde{\mathcal{P}}_n^*(X) \to \tilde{\mathcal{P}}([0,1]^3),$$

where:

- $\tilde{\mathcal{P}}_n^*(X)$ denotes the family of all non-empty elements of the $n$-th PowerSet $\mathcal{P}_n^*(X)$, defined recursively as:

  $$\mathcal{P}_0^*(X) = X, \quad \mathcal{P}_1^*(X) = \mathcal{P}(X), \quad \mathcal{P}_n^*(X) = \mathcal{P}\left(\mathcal{P}_{n-1}^*(X)\right), \quad \text{for } n \geq 2,$$

  with $\tilde{\mathcal{P}}_n^*(X) = \mathcal{P}_n^*(X) \setminus \{\emptyset\}$.
- $\tilde{\mathcal{P}}([0,1]^3)$ denotes the family of all non-empty subsets of the unit cube $[0,1]^3$.

**Structure:**

1. Each element $A \in \tilde{\mathcal{P}}_n^*(X)$ is a non-empty subset within the $n$-th PowerSet hierarchy of $X$.
2. The mapping $\tilde{N}_n$ assigns to each $A \in \tilde{\mathcal{P}}_n^*(X)$ a non-empty subset $\tilde{N}_n(A) \subseteq [0,1]^3$, representing triplets $(T, I, F)$, where $T$, $I$, and $F$ satisfy:

$$T, I, F \in [0,1], \quad 0 \leq T + I + F \leq 3.$$

   $T$ represents the degree of truth membership, $I$ represents the degree of indeterminacy, and $F$ represents the degree of falsity membership for the subset $A$.

**Properties:**

- If $n = 0$, $\tilde{\mathcal{P}}_0^*(X) = X$, and the structure reduces to a standard neutrosophic set:

$$\tilde{N}_0 : X \to [0,1]^3.$$

- For $n = 1$, $\tilde{\mathcal{P}}_1^*(X) = \tilde{\mathcal{P}}(X)$, and the structure represents a SuperHyperNeutrosophic Set:

$$\tilde{N}_1 : \tilde{\mathcal{P}}(X) \to \tilde{\mathcal{P}}([0,1]^3).$$

- For $n \geq 2$, the structure recursively extends to higher-order neutrosophic relationships:

$$\tilde{N}_n : \tilde{\mathcal{P}}_n^*(X) \to \tilde{\mathcal{P}}([0,1]^3).$$

The $n$-SuperHyperNeutrosophic Set generalizes the concept of neutrosophic sets to hierarchical and recursive levels, allowing for higher-order relationships and degrees associated with subsets of subsets, and so on, up to the $n$-th PowerSet hierarchy of $X$.

**Definition 5.7.24** ($n$-SuperHyperPlithogenic Set)**.** Let $X$ be a non-empty set, and $n \geq 0$ be an integer. Let $V = \{v_1, v_2, \ldots, v_m\}$ be a set of attributes, each with a set of possible values $P_{v_i}$.

An *n-SuperHyperPlithogenic Set $SHPS_n$* over $X$ is defined as:

$$SHPS_n = (P, V, \{P_{v_i}\}_{i=1}^m, \{\tilde{pdf}_{i,n}\}_{i=1}^m, pCF_n),$$

where:

- $P \subseteq X$ is a subset of the universe.
- For each attribute $v_i$, $P_{v_i}$ is the set of possible values.
- For each attribute $v_i$, $\tilde{pdf}_{i,n} : \tilde{\mathcal{P}}_n^*(P) \times P_{v_i} \to \tilde{\mathcal{P}}([0,1]^s)$ is the *n-th order Hyper Degree of Appurtenance Function (HDAF)*, assigning to each element $A \in \tilde{\mathcal{P}}_n^*(P)$ and attribute value $a_i \in P_{v_i}$ a non-empty subset of $[0,1]^s$.
- $pCF_n : \left(\bigcup_{i=1}^m P_{v_i}\right) \times \left(\bigcup_{i=1}^m P_{v_i}\right) \to [0,1]^t$ is the *Degree of Contradiction Function (DCF)*.
- $s$ and $t$ are positive integers representing the dimensions of the membership degrees and contradiction degrees, respectively.

The $n$-SuperHyperPlithogenic Set generalizes plithogenic sets to higher-order structures, incorporating multiple attributes and their possible values, along with higher-order degrees of membership and contradiction over the $n$-th PowerSet hierarchy of $X$.

**Theorem 5.7.25.** *An n-SuperHyperPlithogenic Set generalizes the n-SuperHyperFuzzy Set, n-SuperHyperVague Set, and n-SuperHyperNeutrosophic Set.*

*Proof.* Let $SHPS_n = (P, V, \{P_{v_i}\}_{i=1}^m, \{\tilde{pdf}_{i,n}\}_{i=1}^m, pCF_n)$ be an $n$-SuperHyperPlithogenic Set over $X$, with $s$ and $t$ representing the dimensions of the membership degrees and contradiction degrees, respectively.

We will show that:

1. When $s = 1$ and $t = 1$, $SHPS_n$ reduces to an $n$-SuperHyperFuzzy Set.
2. When $s = 2$ and $t = 1$, $SHPS_n$ reduces to an $n$-SuperHyperVague Set.
3. When $s = 3$ and $t = 1$, $SHPS_n$ reduces to an $n$-SuperHyperNeutrosophic Set.

**Case 1: $s = 1, t = 1$ (Reduction to $n$-SuperHyperFuzzy Set)** In this case, the membership degrees are single-dimensional, and the contradiction degrees are also single-dimensional.

Define a mapping $\tilde{\mu}_n : \tilde{\mathcal{P}}_n^*(P) \to \tilde{\mathcal{P}}([0, 1])$ by:

$$\tilde{\mu}_n(A) = \bigcup_{i=1}^{m} \bigcup_{a_i \in P_{v_i}} \tilde{pdf}_{i,n}(A, a_i),$$

for each $A \in \tilde{\mathcal{P}}_n^*(P)$.

Since $\tilde{pdf}_{i,n}(A, a_i) \subseteq [0, 1]$, their union is a subset of $[0, 1]$.

Thus, $\tilde{\mu}_n$ is an $n$-SuperHyperFuzzy Set over $P$.

**Case 2: $s = 2, t = 1$ (Reduction to $n$-SuperHyperVague Set)** Here, the membership degrees are two-dimensional, corresponding to the truth and falsity components.

Define a mapping $\tilde{V}_n : \tilde{\mathcal{P}}_n^*(P) \to \tilde{\mathcal{P}}([0, 1]^2)$ by:

$$\tilde{V}_n(A) = \bigcup_{i=1}^{m} \bigcup_{a_i \in P_{v_i}} \tilde{pdf}_{i,n}(A, a_i),$$

for each $A \in \tilde{\mathcal{P}}_n^*(P)$.

Since $\tilde{pdf}_{i,n}(A, a_i) \subseteq [0, 1]^2$, $\tilde{V}_n$ is an $n$-SuperHyperVague Set over $P$.

**Case 3: $s = 3, t = 1$ (Reduction to $n$-SuperHyperNeutrosophic Set)** In this case, the membership degrees are three-dimensional, representing truth, indeterminacy, and falsity.

Define a mapping $\tilde{N}_n : \tilde{\mathcal{P}}_n^*(P) \to \tilde{\mathcal{P}}([0, 1]^3)$ by:

$$\tilde{N}_n(A) = \bigcup_{i=1}^{m} \bigcup_{a_i \in P_{v_i}} \tilde{pdf}_{i,n}(A, a_i),$$

for each $A \in \tilde{\mathcal{P}}_n^*(P)$.

Since $\tilde{pdf}_{i,n}(A, a_i) \subseteq [0, 1]^3$, $\tilde{N}_n$ is an $n$-SuperHyperNeutrosophic Set over $P$.

In each case, by choosing appropriate dimensions $s$ and $t$ for the membership and contradiction degrees, the $n$-SuperHyperPlithogenic Set $SHPS_n$ reduces to the corresponding $n$-SuperHyperFuzzy, $n$-SuperHyperVague, or $n$-SuperHyperNeutrosophic Set.

Therefore, the $n$-SuperHyperPlithogenic Set generalizes these sets. □

For reference, the relationships between the SuperhyperUncertain set are illustrated in Figure 5.2.

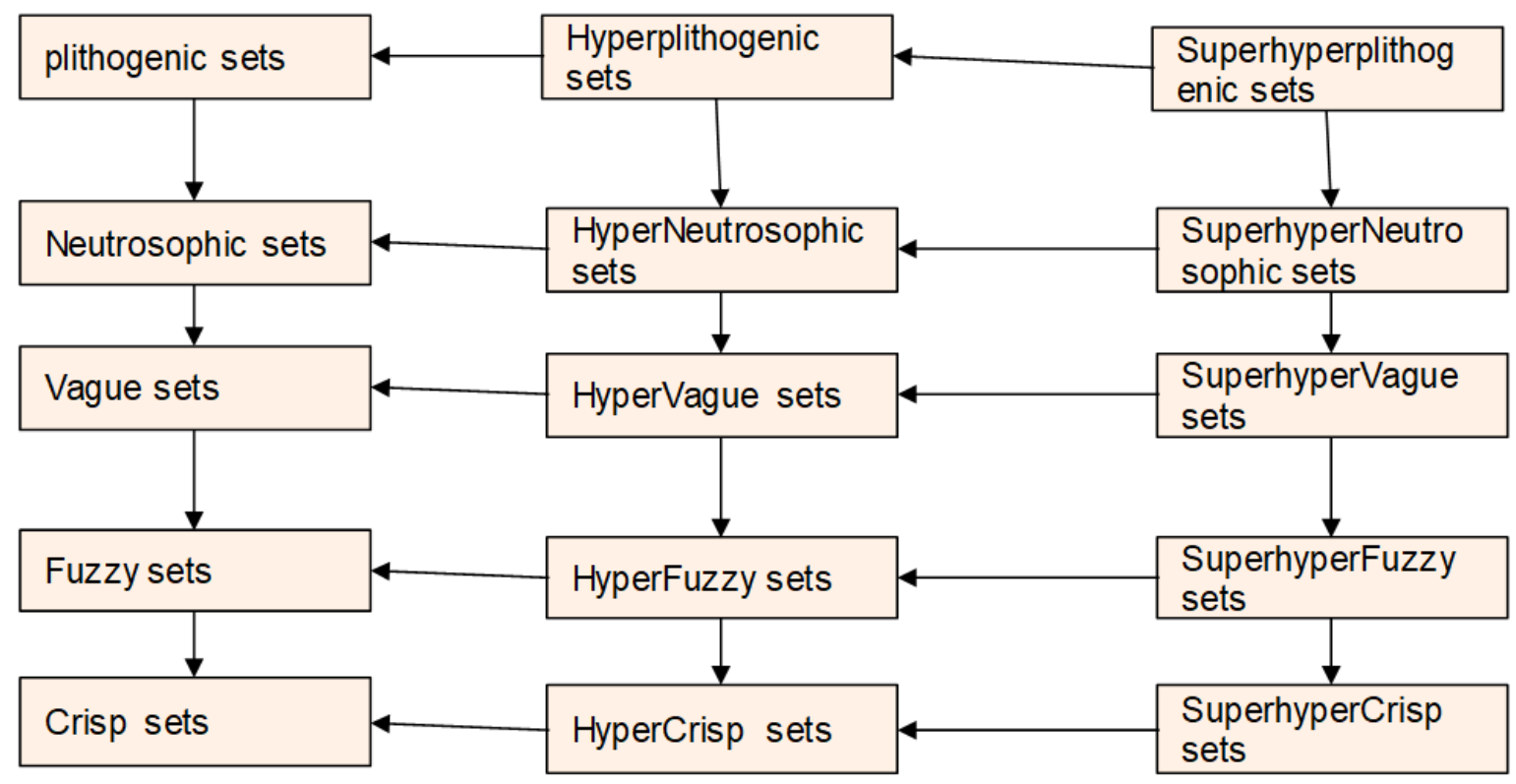

Figure 5.2: Some the SuperhyperUncertain sets Hierarchy. The set class at the origin of an arrow contains the set class at the destination of the arrow.

## 5.8 $(m, n)$-SuperHyperFuzzy, Neutrosophic, and Related Set

A $(m, n)$-Superhyperfuzzy Set maps m-level nested subsets of $X$ to n-level nested fuzzy membership-value subsets in $[0, 1]$, capturing distinct hierarchical fuzzy uncertainty patterns. The definition of the $(m, n)$-SuperhyperFuzzy Set is presented below.

**Definition 5.8.1** ($(m, n)$-SuperhyperFuzzy Set)**.** Let $U$ be a universe of discourse and let $A \subseteq U$ be nonempty. Fix integers $m, n \geq 0$. Define the $m$-th nested power-set of $A$ by

$$\mathcal{P}^0(A) = A, \qquad \mathcal{P}^k(A) = \mathcal{P}\big(\mathcal{P}^{k-1}(A)\big) \quad (k \geq 1),$$

and similarly define $\mathcal{P}^n([0, 1])$ for the unit interval. An $(m, n)$-*SuperhyperFuzzy Set* on $A$ is a function

$$\tau : \mathcal{P}^m(A) \longrightarrow \mathcal{P}^n([0, 1])$$

such that for each $X \in \mathcal{P}^m(A)$, the image $\tau(X) \subseteq [0, 1]$ is nonempty. In other words, the "membership grade" of the (possibly nested) element $X$ is not a single number but a set of values—intervals and/or discrete points—drawn from $[0, 1]$.

An $(m, n)$-SuperhyperNeutrosophic Set maps $m$-level nested subsets to nonempty families of $n$-level neutrosophic triples $(T, I, F) \in [0, 1]^3$, thereby modeling hierarchical uncertainty across levels.

**Definition 5.8.2** ($(m, n)$-SuperhyperNeutrosophic Set)**.** Let $U$ be a universe of discourse and let $A \subseteq U$ be nonempty. Fix $m, n \in \mathbb{N} \cup \{0\}$. Define recursively

$$\mathcal{P}^0(A) = A, \qquad \mathcal{P}^k(A) = \mathcal{P}\big(\mathcal{P}^{k-1}(A)\big) \quad (k \geq 1),$$

and analogously for $\mathcal{P}^n([0, 1]^3)$. An $(m, n)$-*SuperhyperNeutrosophic Set* on $A$ is a mapping

$$\mu : \mathcal{P}^m(A) \longrightarrow \mathcal{P}^n\big([0, 1]^3\big)$$

such that for each $X \in \mathcal{P}^m(A)$ the value $\mu(X)$ is nonempty and every triple $(T, I, F)$ occurring at level 1 inside $\mu(X)$ satisfies

$$0 \leq T + I + F \leq 3.$$

Equivalently, $\mathrm{U}_{n\to 1}\big(\mu(X)\big) \subseteq [0, 1]^3$ is nonempty and consists of triples obeying $T + I + F \leq 3$, where $\mathrm{U}_{n\to 1}$ denotes iterated union (flattening) from level $n$ to level 1.

A $(m, n)$-SuperhyperPlithogenic Set assigns $m$-level parameter-subsets and their attribute values to $n$-level fuzzy-contradiction degree-sets, capturing multi-faceted membership and inter-attribute conflicts and uncertainty patterns. The definition of the $(m, n)$-SuperhyperPlithogenic Set is presented below.

**Definition 5.8.3** ($(m, n)$-SuperhyperPlithogenic Set)**.** Let $X$ be a nonempty set and let $V = \{v_1, \ldots, v_k\}$ be a finite set of attributes. For each $v \in V$, let $P_v$ be the set of its possible values. Fix positive integers $m, n$ and positive dimensions $s, t$. Define the $m$-th nested powerset of $X$ by

$$\mathcal{P}^0(X) = X, \quad \mathcal{P}^r(X) = \mathcal{P}\big(\mathcal{P}^{r-1}(X)\big) \quad (r \geq 1),$$

and similarly $\mathcal{P}^n\big([0,1]^s\big)$ for the $s$-dimensional unit cube. An $(m, n)$-*SuperhyperPlithogenic Set* over $X$ is the quintuple

$$SHP^{(m,n)} = \Big(\mathcal{P}^m(X),\ V,\ \{P_v\}_{v \in V},\ \{\tilde{pdf}_v^{(m,n)}\}_{v \in V},\ pCF^{(m,n)}\Big),$$

where

(i) $\mathcal{P}^m(X)$ is the domain of "super-elements" of level $m$.

(ii) For each $v \in V$, $P_v$ is the finite set of its values.

(iii) The *Hyper Degree of Appurtenance Function*

$$\tilde{pdf}_v^{(m,n)} : \mathcal{P}^m(X) \times P_v \longrightarrow \mathcal{P}^n\big([0,1]^s\big)$$

assigns to each $(A, a)$ with $A \in \mathcal{P}^m(X)$ and $a \in P_v$ a nonempty subset $\tilde{pdf}_v^{(m,n)}(A, a) \subseteq [0,1]^s$ representing all possible membership-degree vectors of dimension $s$.

(iv) The *Degree of Contradiction Function*

$$pCF^{(m,n)} : \Big(\bigcup_{v \in V} P_v\Big) \times \Big(\bigcup_{v \in V} P_v\Big) \longrightarrow [0,1]^t$$

satisfies for all $a, b$:

(a) $pCF^{(m,n)}(a, a) = 0$ (reflexivity),

(b) $pCF^{(m,n)}(a, b) = pCF^{(m,n)}(b, a)$ (symmetry).

## 5.9 Neutrosophic Hypersoft OffGraphs / OverGraphs / UnderGraphs

We define neutrosophic hypersoft graph variants by combining: (i) a hypersoft parameter space $J = J_1 \times \cdots \times J_n$, (ii) a hypersoft vertex-selection map $F : J \to \mathcal{P}(V)$, and (iii) neutrosophic membership degrees for vertices and edges that may lie in an "under/over/off" interval. We then state precise (aggregation-based) reductions to neutrosophic soft graphs.

**Definition 5.9.1** (Hypersoft parameter space)**.** Let $T_1, \ldots, T_n$ be distinct attributes with nonempty value sets $J_1, \ldots, J_n$. The *hypersoft parameter space* is the Cartesian product

$$J := J_1 \times \cdots \times J_n.$$

Elements $a = (a_1, \ldots, a_n) \in J$ are called *attribute combinations*.

**Definition 5.9.2** (Neutrosophic hypersoft labeling domain)**.** Let $G = (V, E)$ be a finite simple undirected graph with $V \neq \emptyset$ and $E \subseteq \binom{V}{2}$. Let $J$ be as in Definition 5.9.1. A *hypersoft vertex selector* is any map

$$F :\ J \longrightarrow \mathcal{P}(V).$$

For $a \in J$, write $V_a := F(a) \subseteq V$, and define the induced edge set

$$E_a := \{\{u, v\} \in E :\ u, v \in V_a\} = E \cap \binom{V_a}{2}.$$

**Definition 5.9.3** (Off/over/under intervals)**.** Fix real bounds $\Psi < 0 < 1 < \Omega$. We use the following intervals:

$$I_{\text{off}} := [\Psi, \Omega], \qquad I_{\text{over}} := [0, \Omega], \qquad I_{\text{under}} := [\Psi, 1].$$

**Definition 5.9.4** (Neutrosophic Hypersoft OffGraph)**.** Let $G = (V, E)$ be a finite simple graph, let $J$ be a hypersoft parameter space, and let $F : J \to \mathcal{P}(V)$ be a hypersoft vertex selector with induced $(V_a, E_a)$ as in Definition 5.9.2. Fix bounds $\Psi < 0 < 1 < \Omega$ and put $I_{\text{off}} = [\Psi, \Omega]$.

A *neutrosophic hypersoft offgraph* is a tuple

$$\mathcal{G}_{\text{off}} = \big(G,\ J,\ F,\ T_V,\ I_V,\ F_V,\ T_E,\ I_E,\ F_E\big),$$

where the membership maps are

$$T_V,\ I_V,\ F_V :\ \{(v, a) \in V \times J :\ v \in V_a\} \longrightarrow I_{\text{off}},$$

$$T_E,\ I_E,\ F_E :\ \{(e, a) \in E \times J :\ e \in E_a\} \longrightarrow I_{\text{off}}.$$

(Thus membership degrees are defined precisely on those vertex/edge occurrences that are active under parameter $a$.)

Optionally, one may impose a pointwise sum constraint

$$T_\bullet(x, a) + I_\bullet(x, a) + F_\bullet(x, a)\ \leq\ 3\Omega,$$

which is automatic if each component lies in $[\Psi, \Omega]$ and $\Psi \geq 0$, but not automatic when $\Psi < 0$.

**Definition 5.9.5** (Neutrosophic Hypersoft OverGraph / UnderGraph)**.** In Definition 5.9.4, replace the codomain interval $I_{\text{off}}$ by

$$I_{\text{over}} = [0, \Omega] \quad (\Omega > 1)$$

to obtain a *neutrosophic hypersoft overgraph*, and replace it by

$$I_{\text{under}} = [\Psi, 1] \quad (\Psi < 0)$$

to obtain a *neutrosophic hypersoft undergraph*. We denote these structures by $\mathcal{G}_{\text{over}}$ and $\mathcal{G}_{\text{under}}$, respectively.

**Remark 5.9.6** (Notation)**.** We use $F_V$ for the falsity-on-vertices map and $F$ (or $F(\cdot)$) for the hypersoft selector $F : J \to \mathcal{P}(V)$. This avoids the collision present in the original draft where both were denoted by $F$.

To reduce a hypersoft (multi-attribute) model to a soft (single-parameter) model, one must fix a projection (or aggregation) from the product space $J = J_1 \times \cdots \times J_n$ to a parameter set $A$. The simplest choice is to take $A := J$ itself and treat each $a \in J$ as a single parameter.

**Definition 5.9.7** (Neutrosophic soft off/over/under graphs (minimal form))**.** Let $G = (V, E)$ be a finite simple graph and let $A \neq \emptyset$ be a parameter set. A *neutrosophic soft offgraph* over $G$ (with bounds $\Psi < 0 < 1 < \Omega$) consists of:

$$F :\ A \to \mathcal{P}(V), \qquad K :\ A \to \mathcal{P}(E),$$

and membership maps

$$T_V, I_V, F_V :\ \{(v, a) : v \in F(a)\} \to [\Psi, \Omega], \qquad T_E, I_E, F_E :\ \{(e, a) : e \in K(a)\} \to [\Psi, \Omega].$$

The over/under cases replace $[\Psi, \Omega]$ by $[0, \Omega]$ or $[\Psi, 1]$, respectively.

**Theorem 5.9.8** (Hypersoft $\Rightarrow$ soft by reindexing)**.** *Let $\mathcal{G}_{\text{off}}$ be a neutrosophic hypersoft offgraph as in Definition 5.9.4. Set $A := J$ and define*

$$F'(a) := V_a = F(a), \qquad K'(a) := E_a \qquad (a \in A).$$

*Define the soft membership maps by restriction:*

$$T'_V(v, a) := T_V(v, a),\ I'_V(v, a) := I_V(v, a),\ F'_V(v, a) := F_V(v, a),$$

*for $v \in F'(a)$, and*

$$T'_E(e, a) := T_E(e, a),\ I'_E(e, a) := I_E(e, a),\ F'_E(e, a) := F_E(e, a),$$

*for $e \in K'(a)$. Then*

$$\mathcal{G}_{\text{soft,off}} = \big(G,\ A,\ F', K',\ T'_V, I'_V, F'_V,\ T'_E, I'_E, F'_E\big)$$

*is a neutrosophic soft offgraph in the sense of Definition 5.9.7. The same construction works for overgraphs and undergraphs by changing the codomain interval accordingly.*

*Proof.* This is a pure reindexing argument. The hypersoft maps are already defined on $\{(v,a) : v \in V_a\}$ and $\{(e,a) : e \in E_a\}$. By taking $A := J$, the same domains become $\{(v,a) : v \in F'(a)\}$ and $\{(e,a) : e \in K'(a)\}$. All codomain constraints are preserved because we do not change the values. Therefore the resulting structure satisfies the definition of a neutrosophic soft offgraph. The over/under cases are identical. □

**Remark 5.9.9** (Global aggregation is optional)**.** If one wants a *single* neutrosophic labeling on $V \cup E$ (without parameters), then one must choose an aggregation operator across contexts $a \in J$ (e.g. sup, inf, averages, etc.). Such a collapse is not canonical and depends on the intended semantics; therefore it is not built into the definition.

**Theorem 5.9.10** (Forgetting neutrosophic degrees yields a hypersoft graph)**.** *Let $\mathcal{G}_{\mathrm{off}}$ (or $\mathcal{G}_{\mathrm{over}}, \mathcal{G}_{\mathrm{under}}$) be a neutrosophic hypersoft graph structure on $G$. If one forgets the membership maps $T_\bullet, I_\bullet, F_\bullet$, the remaining data $(G, J, F)$ is exactly a (hyper)soft graph skeleton given by a hypersoft vertex selector $F : J \to \mathcal{P}(V)$ (with induced edge sets $E_a$).*

*Proof.* Immediate from Definition 5.9.2. □

## 5.10 Hyperbinary Fuzzy Sets and Hyperbinary Neutrosophic Sets

The term *hyperbinary* is used in several areas with meanings different from the "hyper-" terminology in hypergraphs/hypersets. Here we adopt a set-theoretic meaning: *membership is allowed to take the values* 0, 1, 2 (or, in the fuzzy version, values in $[0,2]$). This is most naturally formalized using the language of multisets.

**Definition 5.10.1** (Hyperbinary multiset)**.** Let $X \neq \emptyset$. A *hyperbinary multiset* on $X$ is a multiset whose multiplicity function satisfies

$$m_H : X \longrightarrow \{0,1,2\}.$$

Equivalently, $H$ is completely determined by $m_H$, and we may write

$$H = \{(x, m_H(x)) : x \in X\}.$$

**Theorem 5.10.2** (Crisp sets as a special case)**.** *Every crisp (binary) subset $B \subseteq X$ is a special case of a hyperbinary multiset.*

*Proof.* Given $B \subseteq X$, define $m_B : X \to \{0,1,2\}$ by $m_B(x) = 1$ if $x \in B$ and $m_B(x) = 0$ otherwise. Then $m_B$ is a hyperbinary multiplicity function and encodes $B$. □

**Definition 5.10.3** (Hyperbinary fuzzy set)**.** Let $X \neq \emptyset$. A *hyperbinary fuzzy set* (HBFS) on $X$ is a function

$$\mu : X \longrightarrow [0,2].$$

Thus each $x \in X$ is assigned a membership degree $\mu(x)$ that may exceed 1 but is bounded above by 2.

**Remark 5.10.4.** Every hyperbinary multiset $m_H : X \to \{0,1,2\}$ is an HBFS by viewing $\{0,1,2\} \subset [0,2]$. Likewise, every ordinary fuzzy set $\mu : X \to [0,1]$ is an HBFS since $[0,1] \subset [0,2]$.

**Theorem 5.10.5** (HBFS subsumes fuzzy sets and hyperbinary multisets)**.**

(i) *If $\mu : X \to [0,1]$, then $\mu$ is an ordinary fuzzy set and hence an HBFS.*

(ii) *If $\mu : X \to \{0,1,2\}$, then $\mu$ is the multiplicity function of a hyperbinary multiset and hence an HBFS.*

*Proof.* Both statements follow from the inclusions $[0,1] \subset [0,2]$ and $\{0,1,2\} \subset [0,2]$. □

**Definition 5.10.6** (Hyperbinary neutrosophic set)**.** Let $X \neq \emptyset$. A *hyperbinary neutrosophic set* (HBNS) on $X$ is a triple of functions

$$T,\ I,\ F : X \longrightarrow [0,2].$$

Equivalently, it is a map $\mu : X \to [0,2]^3$ given by $\mu(x) = (T(x), I(x), F(x))$. Optionally, one may impose a pointwise bound such as

$$0 \leq T(x) + I(x) + F(x) \leq 6 \qquad (\forall x \in X),$$

which is automatic from $[0,2]$-valuedness.

**Remark 5.10.7** (Avoiding unnecessary two-universe notation)**.** The earlier draft used a pair $(X, Y)$ and six sets $H_{ij}$, which is not needed for the core notion. The standard neutrosophic pattern is one universe $X$ with three components $T, I, F$; Definition 5.10.6 is the direct hyperbinary analogue.

**Theorem 5.10.8** (HBNS subsumes HBFS)**.** *Every hyperbinary fuzzy set is obtained as a special case of a hyperbinary neutrosophic set.*

*Proof.* Let $\mu : X \to [0, 2]$ be an HBFS. Define $T := \mu$, $I := 0$, and $F := 0$. Then $T, I, F : X \to [0, 2]$, so $(T, I, F)$ is an HBNS and the truth component recovers $\mu$. □

**Theorem 5.10.9** (Binary neutrosophic sets as a special case)**.** *If $T, I, F : X \to \{0, 1\}$, then $(T, I, F)$ is an (ordinary) binary-valued neutrosophic set. In particular, restricting an HBNS to $\{0, 1\}$-valued components yields a binary neutrosophic set.*

*Proof.* Immediate from $\{0, 1\} \subset [0, 2]$ and the usual definition of binary-valued neutrosophic membership triples. □

**Theorem 5.10.10** (Normalization to a standard neutrosophic set)**.** *Let $(T, I, F)$ be an HBNS on $X$. Define*

$$T_0(x) := \frac{T(x)}{2}, \qquad I_0(x) := \frac{I(x)}{2}, \qquad F_0(x) := \frac{F(x)}{2} \qquad (x \in X).$$

*Then $(T_0, I_0, F_0)$ is a (single-valued) neutrosophic set on $X$, i.e., $T_0, I_0, F_0 : X \to [0, 1]$.*

*Proof.* Since $T(x), I(x), F(x) \in [0, 2]$, dividing by 2 yields values in $[0, 1]$. Thus $T_0, I_0, F_0 : X \to [0, 1]$, which is exactly the data of a single-valued neutrosophic set. □

## 5.11 Ranked Hypersoft Sets

Ranked soft sets refine soft sets by replacing the usual subset-valued output with a *ranked partition* (graded satisfaction levels). A ranked hypersoft set extends this idea to the hypersoft parameter space $J = J_1 \times \cdots \times J_n$.

**Definition 5.11.1** (Ranked partition)**.** Let $X \neq \emptyset$ and let $k \in \mathbb{N}$. A *ranked partition of height $k$* of $X$ is a $(k + 1)$-tuple

$$\mathbf{V} = (V_0, V_1, \ldots, V_k)$$

of subsets of $X$ such that:

$$V_i \cap V_j = \emptyset \quad (i \neq j), \qquad \bigcup_{i=0}^{k} V_i = X.$$

We interpret $V_0$ as the "non-satisfaction" level and $V_1, \ldots, V_k$ as increasing satisfaction (or relevance) levels. Let $\mathcal{R}_k(X)$ denote the set of all ranked partitions of height $k$, and set

$$\mathcal{R}(X) := \bigcup_{k \geq 0} \mathcal{R}_k(X).$$

**Definition 5.11.2** (Ranked Hypersoft Set)**.** Let $X \neq \emptyset$ be a universe. Let $E = \{T_1, \ldots, T_n\}$ be a finite set of attributes, and for each $i \in \{1, \ldots, n\}$ let $J_i \neq \emptyset$ be the set of admissible values of attribute $T_i$. Define the hypersoft parameter space

$$J := J_1 \times \cdots \times J_n.$$

A *ranked hypersoft set* on $X$ is a pair $(H, J)$ where

$$H : J \longrightarrow \mathcal{R}(X).$$

Equivalently, for each attribute-combination $j = (j_1, \ldots, j_n) \in J$, the value $H(j)$ is a ranked partition

$$H(j) = (V_0^{(j)}, V_1^{(j)}, \ldots, V_{k(j)}^{(j)})$$

of $X$, for some integer $k(j) \geq 0$ that may depend on $j$.

**Remark 5.11.3** (Fixed height version)**.** If one prefers a uniform number of ranks, fix a single $k \geq 0$ and require $H : J \to \mathcal{R}_k(X)$. All statements below remain valid in this fixed-height setting.

**Definition 5.11.4** (Ranked soft set and hypersoft set (for comparison))**.** A *ranked soft set* on $X$ is a map $h : A \to \mathcal{R}(X)$ from a parameter set $A \neq \emptyset$. A (crisp) *hypersoft set* on $X$ is a map $F : J \to \mathcal{P}(X)$ from a hypersoft parameter space $J$.

**Theorem 5.11.5** (Ranked hypersoft sets generalize ranked soft sets)**.** *If $n = 1$, then $J = J_1$ and any ranked hypersoft set $H : J \to \mathcal{R}(X)$ is exactly a ranked soft set on $X$ (with parameter set $J_1$).*

*Proof.* If $n = 1$, then $J = J_1$, so $H : J_1 \to \mathcal{R}(X)$ is precisely the definition of a ranked soft set (Definition 5.11.4). □

**Theorem 5.11.6** (Ranked hypersoft sets generalize hypersoft sets)**.** *Every hypersoft set $F : J \to \mathcal{P}(X)$ canonically induces a ranked hypersoft set $H : J \to \mathcal{R}_1(X)$ by setting, for each $j \in J$,*

$$H(j) := \big(X \setminus F(j),\ F(j)\big).$$

*Proof.* For each $j \in J$, the pair $\big(X \setminus F(j), F(j)\big)$ is a ranked partition of height 1: the two sets are disjoint and their union is $X$. Hence $H(j) \in \mathcal{R}_1(X)$ for all $j$, so $H : J \to \mathcal{R}_1(X)$ is a ranked hypersoft set. □

## 5.12 TreeHyperSoft Sets

A hypersoft set uses a Cartesian product of attribute-value domains, while a TreeSoft set uses a hierarchical (tree) index of attributes. A *TreeHyperSoft set* combines these two ideas by allowing *tree-structured attributes* and, simultaneously, *multiple values* for each attribute. The natural parameter domain is the set of all root-to-leaf value assignments.

**Definition 5.12.1** (Attribute tree with value sets)**.** Let $\mathrm{Tree}(A) = (\mathcal{N}, \mathcal{E})$ be a finite rooted tree with node set $\mathcal{N}$ and root $r \in \mathcal{N}$. Assume that each node $a \in \mathcal{N}$ is an attribute equipped with a nonempty set $J_a$ of admissible values.

A *root-to-leaf path* is a sequence $\pi = (r = a_0, a_1, \ldots, a_\ell)$ where $a_\ell$ is a leaf and $\{a_{i-1}, a_i\} \in \mathcal{E}$ for all $i$. For a fixed path $\pi$, define its value-assignment space by

$$J_\pi := J_{a_0} \times J_{a_1} \times \cdots \times J_{a_\ell}.$$

Finally, define the global *tree hypersoft parameter space* by the disjoint union over all root-to-leaf paths:

$$J_{\mathrm{THS}} := \bigsqcup_{\pi \in \mathsf{Path}(r \to \mathrm{leaf})} J_\pi.$$

An element $p \in J_{\mathrm{THS}}$ is thus a pair $(\pi, \mathbf{j})$ consisting of a path $\pi$ and a tuple $\mathbf{j} \in J_\pi$ of values assigned along that path.

**Remark 5.12.2.** Different paths may have different lengths, hence different Cartesian products. Using the disjoint union $\bigsqcup_\pi J_\pi$ yields a well-defined parameter space without forcing equal dimensions.

**Definition 5.12.3** (TreeHyperSoft Set)**.** Let $U \neq \emptyset$ be a universe of discourse and let $\mathrm{Tree}(A)$ be an attribute tree with value sets $\{J_a\}_{a \in \mathcal{N}}$ as in Definition 5.12.1. A *TreeHyperSoft set* over $U$ is a pair $(F, J_{\mathrm{THS}})$ where

$$F :\ J_{\mathrm{THS}} \longrightarrow \mathcal{P}(U)$$

assigns to each parameter $p = (\pi, \mathbf{j}) \in J_{\mathrm{THS}}$ a subset $F(p) \subseteq U$.

**Definition 5.12.4** (Hypersoft set and TreeSoft set (for comparison))**.** A *hypersoft set* over $U$ is a map $F : J_1 \times \cdots \times J_n \to \mathcal{P}(U)$. A *TreeSoft set* (power-set indexed form) over $U$ is a map $G : \mathcal{P}(\mathcal{N}) \to \mathcal{P}(U)$, where $\mathcal{N}$ is the node set of an attribute tree.

**Theorem 5.12.5** (Reduction to hypersoft sets)**.** *Assume* $\mathrm{Tree}(A)$ *has exactly one non-root level consisting of nodes* $a_1, \ldots, a_n$*, so every root-to-leaf path has the same length* 1*. Then*

$$J_{\mathrm{THS}} \;\cong\; J_{a_1} \times \cdots \times J_{a_n},$$

*and a TreeHyperSoft set* $F : J_{\mathrm{THS}} \to \mathcal{P}(U)$ *is equivalent (up to this identification) to a hypersoft set.*

*Proof.* If the tree has one non-root level with nodes $a_1, \ldots, a_n$, then each path consists of $(r, a_i)$ for exactly one $i$. If one identifies the root value set $J_r$ with a singleton (or omits it), the disjoint union over paths becomes the Cartesian product over the level-1 nodes (a standard hypersoft parameterization). Thus $F$ becomes a map from a Cartesian product to $\mathcal{P}(U)$, i.e. a hypersoft set. □

**Remark 5.12.6** (A cleaner hypersoft reduction)**.** To match the standard hypersoft form $J_1 \times \cdots \times J_n$, one typically assumes the root carries no value (or carries a singleton value set). Then each parameter is exactly a choice of one value for each attribute node.

**Theorem 5.12.7** (A TreeHyperSoft set induces a TreeSoft set when each $J_a$ is singleton)**.** *Assume each node* $a \in \mathcal{N}$ *has a singleton value set* $J_a = \{j_a\}$*. Then* $J_\pi$ *is a singleton for every root-to-leaf path* $\pi$*, so* $J_{\mathrm{THS}}$ *can be identified with the set of root-to-leaf paths. Consequently, a TreeHyperSoft set* $F : J_{\mathrm{THS}} \to \mathcal{P}(U)$ *is equivalent to a TreeSoft set in node-indexed form* $F' : \mathcal{N} \to \mathcal{P}(U)$ *after fixing a choice of how to represent paths by nodes (e.g. using leaves as parameters).*

*Proof.* If every $J_a$ is singleton, then for each path $\pi$ the Cartesian product $J_\pi$ is singleton, hence the only variation in $J_{\mathrm{THS}}$ comes from the choice of the path itself. Thus $J_{\mathrm{THS}}$ is in bijection with the set of root-to-leaf paths. Any such path is uniquely determined by its leaf, so one may parameterize by leaves and obtain a node-indexed TreeSoft map (taking leaves as the parameter nodes). □

**Remark 5.12.8** (Why the original "$F : \mathcal{P}(\mathrm{Tree}(A)) \to \mathcal{P}(U)$" does not follow)**.** Even when all $J_a$ are singletons, the natural parameter space becomes the set of paths (or leaves), not the full power set $\mathcal{P}(\mathcal{N})$. To obtain a power-set-indexed TreeSoft set one must add an aggregation rule, e.g. $G(S) = \bigcup_{a \in S} F'(a)$. Such an aggregation is a modeling choice and is therefore not part of the reduction.

## 5.13 Hyperweighted Graphs and Superhyperweighted Graphs

We formalize graphs whose edges (and optionally vertices) carry *sets of weights* rather than a single weight. This yields a natural "hyper" (set-valued) weighting and its iterated "superhyper" (set-of-set) analogue [408, 409].

**Definition 5.13.1** (Nonempty power set)**.** For any set $S$, write $\widetilde{\mathcal{P}}(S) := \mathcal{P}(S) \setminus \{\emptyset\}$.

**Definition 5.13.2** (Hyperweighted set)**.** [408, 409] Let $S \neq \emptyset$. A *hyperweighted set* on $S$ is a pair $(S, W)$ where

$$W :\; S \longrightarrow \widetilde{\mathcal{P}}(\mathbb{R})$$

assigns to each $s \in S$ a nonempty set $W(s) \subseteq \mathbb{R}$ of weights.

**Definition 5.13.3** (Hyperweighted graph)**.** [408, 409] Let $G = (V, E)$ be a finite simple undirected graph, where $V \neq \emptyset$ and $E \subseteq \binom{V}{2}$. A *hyperweighted graph* is a triple

$$G_H = (V, E, W),$$

where

$$W :\; E \longrightarrow \widetilde{\mathcal{P}}(\mathbb{R})$$

assigns to each edge $e \in E$ a nonempty set $W(e) \subseteq \mathbb{R}$ of edge-weights. (One may also introduce vertex hyperweights $W_V : V \to \widetilde{\mathcal{P}}(\mathbb{R})$; we restrict to edge weights here.)

**Example 5.13.4.** Let $V = \{v_1, v_2, v_3\}$ and

$$E = \big\{\{v_1, v_2\}, \{v_2, v_3\}\big\}.$$

Define $W : E \to \widetilde{\mathcal{P}}(\mathbb{R})$ by

$$W(\{v_1, v_2\}) = \{5, 7\}, \qquad W(\{v_2, v_3\}) = \{3, 4, 6\}.$$

Then $(V, E, W)$ is a hyperweighted graph. The multiple values in $W(e)$ can represent distinct metrics (distance, time, cost, risk, etc.), depending on the application.

**Theorem 5.13.5** (Weighted graphs are special cases)**.** *Every (single) weighted graph is a special case of a hyperweighted graph.*

*Proof.* Let $G = (V, E, w)$ be a weighted graph with weight function $w : E \to \mathbb{R}$. Define $W : E \to \widetilde{\mathcal{P}}(\mathbb{R})$ by $W(e) := \{w(e)\}$. Then $(V, E, W)$ is a hyperweighted graph and recovers $w$ by the singleton identification. □

**Definition 5.13.6** (Superhyperweighted graph)**.** [408,409] Let $G = (V, E)$ be a finite simple undirected graph. A *superhyperweighted graph* is a triple

$$G_{SH} = (V, E, \mathcal{W}),$$

where the *superhyperweight function* is

$$\mathcal{W} : E \longrightarrow \widetilde{\mathcal{P}}\big(\widetilde{\mathcal{P}}(\mathbb{R})\big).$$

Thus, for each edge $e \in E$, the value $\mathcal{W}(e)$ is a nonempty family of nonempty real-weight sets (i.e., a set of hyperweights).

**Theorem 5.13.7** (Hyperweighted graphs embed into superhyperweighted graphs)**.** *[408, 409] Every hyperweighted graph $G_H = (V, E, W)$ induces a superhyperweighted graph $G_{SH} = (V, E, \mathcal{W})$ by*

$$\mathcal{W}(e) := \{W(e)\} \qquad (e \in E).$$

*Conversely, if $\mathcal{W}(e)$ is a singleton for every $e$, then a superhyperweighted graph reduces to a hyperweighted graph.*

*Proof.* If $W(e) \in \widetilde{\mathcal{P}}(\mathbb{R})$, then $\{W(e)\} \in \widetilde{\mathcal{P}}(\widetilde{\mathcal{P}}(\mathbb{R}))$, so $\mathcal{W}$ is well-defined and $(V, E, \mathcal{W})$ is superhyperweighted.

Conversely, suppose $\mathcal{W}(e) = \{S_e\}$ for each $e \in E$ with $S_e \in \widetilde{\mathcal{P}}(\mathbb{R})$. Define $W(e) := S_e$. Then $W : E \to \widetilde{\mathcal{P}}(\mathbb{R})$ and $(V, E, W)$ is a hyperweighted graph. □

**Definition 5.13.8** (Hypercrisp and superhypercrisp sets)**.** Let $X \neq \emptyset$. A *hypercrisp set* on $X$ is a map $\eta : X \to \widetilde{\mathcal{P}}(\{0, 1\})$. A *superhypercrisp set* on $X$ is a map $\widetilde{\eta} : \widetilde{\mathcal{P}}(X) \to \widetilde{\mathcal{P}}(\{0, 1\})$.

**Theorem 5.13.9** (Crisp specialization of hyperweights)**.**

(i) *If a hyperweight function $W : E \to \widetilde{\mathcal{P}}(\mathbb{R})$ satisfies $W(e) \subseteq \{0, 1\}$ for all $e$, then $W$ is a hypercrisp labeling of edges.*

(ii) *If a superhyperweight function $\mathcal{W} : E \to \widetilde{\mathcal{P}}(\widetilde{\mathcal{P}}(\mathbb{R}))$ satisfies $\bigcup \mathcal{W}(e) \subseteq \{0, 1\}$ for all $e$, then it is a superhypercrisp-type edge labeling.*

*Proof.* Immediate from the codomain restrictions in Definition 5.13.8 and the stated hypotheses. □

**Remark 5.13.10.** Statements of the form "$W$ generalizes a hypercrisp set" are best understood as *restriction* statements: by restricting the codomain from $\mathbb{R}$ to $\{0, 1\}$, one recovers a crisp labeling.

## 5.14 Hyperlabeling Graphs and Superhyperlabeling Graphs

A labeling of a graph assigns to each vertex and/or edge a label from prescribed sets, often subject to constraints [410, 411]. We extend this by allowing each vertex/edge to carry *a nonempty set of labels* (hyperlabels), and then iterate once more by allowing *a nonempty set of hyperlabels* (superhyperlabels).

**Definition 5.14.1** (Nonempty power set)**.** For any set $S$, write $\widetilde{\mathcal{P}}(S) := \mathcal{P}(S) \setminus \{\emptyset\}$.

**Definition 5.14.2** (Labeling graph (minimal form))**.** Let $G = (V, E)$ be a finite simple undirected graph, where $V \neq \emptyset$ and $E \subseteq \binom{V}{2}$. Let $L_V$ and $L_E$ be nonempty label sets. A *labeling graph* on $(G, L_V, L_E)$ is a pair of functions

$$\sigma' : V \to L_V, \qquad \mu' : E \to L_E.$$

Additional constraints (graceful, edge-graceful, harmonious, lucky, etc.) are extra axioms imposed on $\sigma'$ and $\mu'$.

**Definition 5.14.3** (Hyperlabeling graph)**.** Let $G = (V, E)$ be a finite simple graph and let $L_V, L_E$ be nonempty label sets. A *hyperlabeling graph* is a triple

$$G_H = (G, \sigma, \mu),$$

where the vertex hyperlabeling and edge hyperlabeling are maps

$$\sigma : V \longrightarrow \widetilde{\mathcal{P}}(L_V), \qquad \mu : E \longrightarrow \widetilde{\mathcal{P}}(L_E).$$

Thus each vertex $v$ receives a nonempty set of labels $\sigma(v) \subseteq L_V$, and each edge $e$ receives a nonempty set of labels $\mu(e) \subseteq L_E$. Any further *rules* (e.g. relating $\mu(uv)$ to $\sigma(u), \sigma(v)$) are optional axioms depending on the application.

**Theorem 5.14.4** (Labeling graphs are special cases)**.** *Every labeling graph* $(G, \sigma', \mu')$ *is a special case of a hyperlabeling graph, via singleton hyperlabels.*

*Proof.* Given $\sigma' : V \to L_V$ and $\mu' : E \to L_E$, define

$$\sigma(v) := \{\sigma'(v)\} \in \widetilde{\mathcal{P}}(L_V), \qquad \mu(e) := \{\mu'(e)\} \in \widetilde{\mathcal{P}}(L_E).$$

Then $(G, \sigma, \mu)$ is a hyperlabeling graph and recovers $(\sigma', \mu')$ by taking the unique elements of the singletons. □

**Example 5.14.5.** Let $V = \{v_1, v_2, v_3\}$, $E = \{\{v_1, v_2\}, \{v_2, v_3\}\}$, $L_V = \{\text{red}, \text{blue}, \text{green}\}$, and $L_E = \{\text{solid}, \text{dashed}\}$. Define

$$\sigma(v_1) = \{\text{red}, \text{blue}\}, \quad \sigma(v_2) = \{\text{blue}\}, \quad \sigma(v_3) = \{\text{green}, \text{blue}\},$$

$$\mu(\{v_1, v_2\}) = \{\text{solid}, \text{dashed}\}, \qquad \mu(\{v_2, v_3\}) = \{\text{dashed}\}.$$

Then $(G, \sigma, \mu)$ is a hyperlabeling graph.

**Definition 5.14.6** (Superhyperlabeling graph)**.** Let $G = (V, E)$ be a finite simple graph and let $L_V, L_E$ be nonempty label sets. A *superhyperlabeling graph* is a triple

$$G_S = (G, \Sigma, M),$$

where

$$\Sigma : V \longrightarrow \widetilde{\mathcal{P}}\big(\widetilde{\mathcal{P}}(L_V)\big), \qquad M : E \longrightarrow \widetilde{\mathcal{P}}\big(\widetilde{\mathcal{P}}(L_E)\big).$$

Thus $\Sigma(v)$ is a nonempty set of nonempty vertex-hyperlabels, and $M(e)$ is a nonempty set of nonempty edge-hyperlabels.

**Theorem 5.14.7** (Hyperlabeling graphs embed into superhyperlabeling graphs)**.** *Every hyperlabeling graph* $(G, \sigma, \mu)$ *induces a superhyperlabeling graph* $(G, \Sigma, M)$ *by*

$$\Sigma(v) := \{\sigma(v)\}, \qquad M(e) := \{\mu(e)\}.$$

*Conversely, if each* $\Sigma(v)$ *and* $M(e)$ *is a singleton, then a superhyperlabeling graph reduces to a hyperlabeling graph.*

*Proof.* If $\sigma(v) \in \widetilde{\mathcal{P}}(L_V)$, then $\{\sigma(v)\} \in \widetilde{\mathcal{P}}(\widetilde{\mathcal{P}}(L_V))$, so $\Sigma$ is well-defined; similarly for $M$. This yields a superhyperlabeling graph.

Conversely, if $\Sigma(v) = \{S_v\}$ with $S_v \in \widetilde{\mathcal{P}}(L_V)$ and $M(e) = \{T_e\}$ with $T_e \in \widetilde{\mathcal{P}}(L_E)$, define $\sigma(v) := S_v$ and $\mu(e) := T_e$. Then $(G, \sigma, \mu)$ is a hyperlabeling graph. □

**Remark 5.14.8** (Type-correct comparison)**.** The maps $\Sigma : V \to \widetilde{\mathcal{P}}(\widetilde{\mathcal{P}}(L_V))$ and $M : E \to \widetilde{\mathcal{P}}(\widetilde{\mathcal{P}}(L_E))$ are *set-valued labelings* and can be compared to "superhypercrisp" maps only after fixing a common codomain. For instance, if one sets $L_V = L_E = \{0, 1\}$, then $\Sigma$ and $M$ become superhypercrisp labelings. Without such a restriction, claiming a literal isomorphism to a binary-valued SuperHyperCrisp set is not type-correct.

## 5.15 MultiHypersoft Graphs

MultiHypersoft graphs are intended as multi-parameter (hypersoft) analogues of multisoft graphs: a family of subgraphs is indexed not by a single parameter set, but by (possibly many) *combinations* of values drawn from several attribute domains. We give a type-correct definition and record basic reduction relations.

**Definition 5.15.1** (Hypersoft parameter space)**.** Let $T_1, \dots, T_n$ be distinct attributes and let $E_1, \dots, E_n$ be their nonempty value sets. Define the hypersoft parameter space

$$J := E_1 \times \cdots \times E_n.$$

**Remark 5.15.2** (Why $\mathcal{P}(J)$ appears)**.** Elements of $J$ are single attribute-combinations. To model *multiple* combinations simultaneously, it is natural to consider a family $A \subseteq \mathcal{P}(J)$, so that a parameter $a \in A$ is itself a set of combinations.

**Definition 5.15.3** (MultiHypersoft Graph)**.** Let $G^* = (V, E)$ be a finite simple undirected graph, where $V \neq \emptyset$ and $E \subseteq \binom{V}{2}$. Let $J = E_1 \times \cdots \times E_n$ be a hypersoft parameter space (Definition 5.15.1). Let $A \subseteq \mathcal{P}(J) \setminus \{\emptyset\}$ be a nonempty family of parameter-sets.

A *multihypersoft graph* over $G^*$ is a quadruple

$$G_{MH} = (G^*, F, K, A),$$

where

$$F : A \longrightarrow \mathcal{P}(V), \qquad K : A \longrightarrow \mathcal{P}(E),$$

and for each $a \in A$ the pair $(F(a), K(a))$ defines a (not necessarily induced) subgraph of $G^*$, i.e.,

$$K(a) \subseteq E \cap \binom{F(a)}{2}.$$

We write

$$H(a) := \big(F(a), K(a)\big)$$

for the subgraph indexed by $a$, and we may view $G_{MH}$ as the $A$-indexed family $\{H(a)\}_{a \in A}$.

**Remark 5.15.4** (Induced-subgraph option)**.** If one prefers $H(a)$ to be induced, replace the condition $K(a) \subseteq E \cap \binom{F(a)}{2}$ by $K(a) = E \cap \binom{F(a)}{2}$. The general definition above is more flexible.

**Definition 5.15.5** (Hypersoft graph skeleton (single-combination index))**.** With $G^* = (V, E)$ and $J$ as above, a *hypersoft graph skeleton* is a pair of maps

$$F_0 : J \to \mathcal{P}(V), \qquad K_0 : J \to \mathcal{P}(E),$$

such that for each $j \in J$, $K_0(j) \subseteq E \cap \binom{F_0(j)}{2}$.

**Theorem 5.15.6** (Reduction to a hypersoft graph)**.** *Let $G_{MH} = (G^*, F, K, A)$ be a multihypersoft graph. Assume $A \subseteq \{\{j\} : j \in J\}$, i.e., every parameter-set $a \in A$ is a singleton $\{j\}$. Define $J_A := \{j \in J : \{j\} \in A\} \subseteq J$ and define*

$$F_0(j) := F(\{j\}), \qquad K_0(j) := K(\{j\}) \qquad (j \in J_A).$$

*Then $(F_0, K_0)$ is a hypersoft graph skeleton on $J_A$.*

*Proof.* If $a = \{j\} \in A$, then $H(a) = (F(a), K(a))$ is a subgraph of $G^*$, so $K(\{j\}) \subseteq E \cap \binom{F(\{j\})}{2}$. Hence $K_0(j) \subseteq E \cap \binom{F_0(j)}{2}$ for all $j \in J_A$, proving that $(F_0, K_0)$ is a hypersoft graph skeleton. □

**Definition 5.15.7** (Multisoft graph (minimal form))**.** Let $G^* = (V, E)$ be a finite simple graph and let $P \neq \emptyset$ be a parameter set. A *multisoft graph* over $G^*$ is a triple $(F, K, P)$ with

$$F : P \to \mathcal{P}(V), \qquad K : P \to \mathcal{P}(E), \qquad K(p) \subseteq E \cap \binom{F(p)}{2} \ (\forall p \in P).$$

**Theorem 5.15.8** (MultiHypersoft graphs generalize multisoft graphs)**.** *Every multisoft graph is a special case of a multihypersoft graph.*

*Proof.* Let $(F, K, P)$ be a multisoft graph over $G^*$. Take $n = 1$ and set $E_1 := P$, so $J = E_1 = P$. Let

$$A := \{\{p\} : p \in P\} \subseteq \mathcal{P}(J) \setminus \{\emptyset\}.$$

Define $F^*, K^* : A \to \mathcal{P}(V), \mathcal{P}(E)$ by

$$F^*(\{p\}) := F(p), \qquad K^*(\{p\}) := K(p).$$

Then $(G^*, F^*, K^*, A)$ satisfies Definition 5.15.3 and recovers the original multisoft graph under the singleton identification $p \leftrightarrow \{p\}$. □

# Chapter 6

# Other Concepts

In this chapter, we examine several concepts that are not easily classified as graph-theoretic notions.

## 6.1 Filters and Matroids on Offsets, Oversets, and Undersets

This section recalls standard notions of filters and ultrafilters and then records consistent "hyper" and "superhyper" variants by changing the ambient carrier from $X$ to $\mathcal{P}(X)$ or $\mathcal{P}(\mathcal{P}(X))$. We emphasize type-correct definitions: a hyperfilter on $\mathcal{P}(X)$ is simply a filter on the set $\mathcal{P}(X)$, and similarly for the superhyper level. We also state natural "pullback" constructions along canonical embeddings. (Generalizations to offset/over/under membership frameworks can be introduced by replacing crisp subsets with the corresponding offset-valued memberships; we do not formalize those here.)

**Definition 6.1.1** (Filter)**.** Let $X \neq \emptyset$. A *filter* on $X$ is a nonempty family $\mathcal{F} \subseteq \mathcal{P}(X)$ such that

(i) $X \in \mathcal{F}$ and $\emptyset \notin \mathcal{F}$,

(ii) if $A, B \in \mathcal{F}$ then $A \cap B \in \mathcal{F}$,

(iii) if $A \in \mathcal{F}$ and $A \subseteq B \subseteq X$, then $B \in \mathcal{F}$.

**Definition 6.1.2** (Ultrafilter)**.** (cf. [412]) Let $X \neq \emptyset$. An *ultrafilter* on $X$ is a filter $\mathcal{U}$ on $X$ such that for every $A \subseteq X$,

$$A \in \mathcal{U} \quad \text{or} \quad X \setminus A \in \mathcal{U}.$$

Equivalently, $\mathcal{U}$ is a maximal proper filter under inclusion.

**Definition 6.1.3** (Hyperfilter and hyperultrafilter)**.** Let $X \neq \emptyset$ and view $\mathcal{P}(X)$ as a set. A *hyperfilter* on $X$ is a filter on $\mathcal{P}(X)$, i.e., a nonempty family

$$\mathcal{HF} \subseteq \mathcal{P}(\mathcal{P}(X))$$

satisfying:

(i) $\mathcal{P}(X) \in \mathcal{HF}$ and $\emptyset \notin \mathcal{HF}$,

(ii) if $A, B \in \mathcal{HF}$ then $A \cap B \in \mathcal{HF}$,

(iii) if $A \in \mathcal{HF}$ and $A \subseteq B \subseteq \mathcal{P}(X)$, then $B \in \mathcal{HF}$.

A *hyperultrafilter* on $X$ is an ultrafilter on $\mathcal{P}(X)$, i.e., a hyperfilter $\mathcal{HUF}$ such that for every $A \subseteq \mathcal{P}(X)$,

$$A \in \mathcal{HUF} \quad \text{or} \quad \mathcal{P}(X) \setminus A \in \mathcal{HUF}.$$

**Definition 6.1.4** (Superhyperfilter and superhyperultrafilter)**.** Let $X \neq \emptyset$ and view $\mathcal{P}(\mathcal{P}(X))$ as a set. A *superhyperfilter* on $X$ is a filter on $\mathcal{P}(\mathcal{P}(X))$, i.e., a nonempty family

$$\mathcal{SHF} \subseteq \mathcal{P}\big(\mathcal{P}(\mathcal{P}(X))\big)$$

satisfying the filter axioms with carrier $\mathcal{P}(\mathcal{P}(X))$. A *superhyperultrafilter* on $X$ is an ultrafilter on $\mathcal{P}(\mathcal{P}(X))$, i.e., a superhyperfilter $\mathcal{SHUF}$ such that for every $B \subseteq \mathcal{P}(\mathcal{P}(X))$,

$$B \in \mathcal{SHUF} \quad \text{or} \quad \mathcal{P}(\mathcal{P}(X)) \setminus B \in \mathcal{SHUF}.$$

**Remark 6.1.5** (On the earlier "$Z \subseteq X \cap Y$" clause)**.** The condition "for all $A, B \in \mathcal{HF}$ there exists $Z \in \mathcal{HF}$ with $Z \subseteq A \cap B$" is the defining axiom of a *filter base*, not of a filter. Definitions 6.1.3–6.1.4 use the standard filter axioms.

**Definition 6.1.6** (Singleton embedding)**.** Let $X \neq \emptyset$. Define $\iota : X \to \mathcal{P}(X)$ by $\iota(x) = \{x\}$, and define $\iota_{\mathcal{P}} : \mathcal{P}(X) \to \mathcal{P}(\mathcal{P}(X))$ by $\iota_{\mathcal{P}}(A) = \{A\}$.

**Theorem 6.1.7** (From a hyperultrafilter on $\mathcal{P}(X)$ to an ultrafilter on $X$)**.** *Let $\mathcal{HUF}$ be a hyperultrafilter on $X$, i.e., an ultrafilter on the set $\mathcal{P}(X)$. Define*

$$\mathcal{U} := \{\, A \subseteq X : \ \iota_{\mathcal{P}}(A) = \{A\} \in \mathcal{HUF} \,\} \ \subseteq \ \mathcal{P}(X).$$

*Then $\mathcal{U}$ is an ultrafilter on $X$.*

*Proof. (Nonempty and proper.)* Since $\mathcal{P}(X) \in \mathcal{HUF}$ and $\{X\} \subseteq \mathcal{P}(X)$, upward closure yields $\{X\} \in \mathcal{HUF}$, hence $X \in \mathcal{U}$. Also $\emptyset \notin \mathcal{U}$ because $\{\emptyset\} = \emptyset$ in $\mathcal{P}(\mathcal{P}(X))$ is false, but more directly $\emptyset \notin \mathcal{HUF}$.

*(Upward closure.)* If $A \in \mathcal{U}$ and $A \subseteq B \subseteq X$, then $\{A\} \in \mathcal{HUF}$. Since $\{A\} \subseteq \{B\} \subseteq \mathcal{P}(X)$ implies $\{B\}$ is a superset of $\{A\}$ inside $\mathcal{P}(X)$, upward closure in $\mathcal{HUF}$ gives $\{B\} \in \mathcal{HUF}$, hence $B \in \mathcal{U}$.

*(Finite intersection.)* If $A, B \in \mathcal{U}$, then $\{A\}, \{B\} \in \mathcal{HUF}$, hence $\{A\} \cap \{B\} \in \mathcal{HUF}$. If $A \neq B$ this intersection is empty, which cannot lie in $\mathcal{HUF}$, so necessarily $A = B$ whenever both $\{A\}$ and $\{B\}$ belong to an ultrafilter. In that case $A \cap B = A \in \mathcal{U}$, so closure holds.

*(Ultrafilter property.)* For any $A \subseteq X$, consider the subset $\{A\} \subseteq \mathcal{P}(X)$. Since $\mathcal{HUF}$ is an ultrafilter on $\mathcal{P}(X)$, either $\{A\} \in \mathcal{HUF}$ or $\mathcal{P}(X) \setminus \{A\} \in \mathcal{HUF}$. In the first case $A \in \mathcal{U}$. In the second case, $\{X \setminus A\} \subseteq \mathcal{P}(X) \setminus \{A\}$, so upward closure yields $\{X \setminus A\} \in \mathcal{HUF}$, hence $X \setminus A \in \mathcal{U}$. Thus $\mathcal{U}$ is an ultrafilter on $X$. □

**Remark 6.1.8** (Caveat: principal behavior)**.** The proof above uses a strong fact: if $\{A\} \in \mathcal{HUF}$ and $\{B\} \in \mathcal{HUF}$, then $A = B$, since otherwise $\{A\} \cap \{B\} = \emptyset$ would belong to $\mathcal{HUF}$, impossible. Hence any $\mathcal{U}$ obtained in Theorem 6.1.7 is necessarily *principal* (generated by that $A$). So Theorem 6.1.7 is a correct construction but yields only principal ultrafilters.

**Theorem 6.1.9** (From a superhyperultrafilter to a hyperultrafilter)**.** *Let $\mathcal{SHUF}$ be a superhyperultrafilter on $X$, i.e., an ultrafilter on $\mathcal{P}(\mathcal{P}(X))$. Define*

$$\mathcal{HUF} := \{\, A \subseteq \mathcal{P}(X) : \ \{A\} \in \mathcal{SHUF} \,\}.$$

*Then $\mathcal{HUF}$ is a hyperultrafilter on $X$, i.e., an ultrafilter on $\mathcal{P}(X)$.*

*Proof.* This is the same pullback argument as Theorem 6.1.7, with $X$ replaced by $\mathcal{P}(X)$. □

## 6.2 SuperHypercubes and SuperHyperspheres

We propose set-valued analogues of the unit hypercube and the $n$-sphere. The guiding idea is that a "super" point carries *sets of possible coordinates* (hypercube case), or a "super" sphere element is a *nonempty subset* of the classical sphere (hypersphere case).

**Definition 6.2.1** (SuperHypercube)**.** Let $n \in \mathbb{N}^+$. Define the *superhypercube* of dimension $n$ by

$$C_n := \big(\widetilde{\mathcal{P}}([0,1])\big)^n = \{(X_1,\dots,X_n) : \ X_i \subseteq [0,1],\ X_i \neq \emptyset \text{ for } i = 1,\dots,n\},$$

where $\widetilde{\mathcal{P}}([0,1]) = \mathcal{P}([0,1]) \setminus \{\emptyset\}$. An element $(X_1,\dots,X_n) \in C_n$ is interpreted as a set-valued coordinate vector.

**Definition 6.2.2** (Classical unit hypercube)**.** The (closed) unit hypercube in $\mathbb{R}^n$ is

$$C_n := [0,1]^n = \{(x_1,\dots,x_n) \in \mathbb{R}^n : \ x_i \in [0,1] \ \forall i\}.$$

**Theorem 6.2.3** (Selections from a superhypercube yield cube points)**.** *Let $(X_1,\dots,X_n) \in C_n$. Choose $x_i \in X_i$ for each $i = 1,\dots,n$. Then $(x_1,\dots,x_n) \in C_n$.*

*Proof.* Since $X_i \subseteq [0,1]$ and $x_i \in X_i$, we have $x_i \in [0,1]$ for each $i$. Hence $(x_1,\dots,x_n) \in [0,1]^n = C_n$. □

**Remark 6.2.4** (No canonical single point)**.** Theorem 6.2.3 uses an arbitrary choice $x_i \in X_i$. There is generally no canonical selection unless additional structure is imposed (e.g. choosing $\min X_i$, $\max X_i$, or a representative such as the midpoint when $X_i$ is an interval).

**Definition 6.2.5** ($n$-sphere)**.** Let $n \in \mathbb{N}$, let $c \in \mathbb{R}^{n+1}$, and let $r > 0$. The *$n$-sphere* of radius $r$ centered at $c$ is

$$S^n(c,r) := \{x \in \mathbb{R}^{n+1} : \ \|x - c\|_2 = r\}.$$

**Definition 6.2.6** (SuperHypersphere)**.** Let $n \in \mathbb{N}$, let $c \in \mathbb{R}^{n+1}$, and let $r > 0$. Define the *superhypersphere* of dimension $n$ (radius $r$, center $c$) by

$$\mathcal{S}_n(c,r) := \widetilde{\mathcal{P}}\big(S^n(c,r)\big) = \{\, X \subseteq \mathbb{R}^{n+1} : \ X \neq \emptyset,\ X \subseteq S^n(c,r) \,\}.$$

Thus, an element of $\mathcal{S}_n(c,r)$ is a nonempty subset of the classical $n$-sphere.

**Theorem 6.2.7** (Union of all superhypersphere elements equals the sphere)**.** *For any $n \in \mathbb{N}$, $c \in \mathbb{R}^{n+1}$, and $r > 0$,*

$$\bigcup_{X \in \mathcal{S}_n(c,r)} X \ = \ S^n(c,r).$$

*Proof.* ($\subseteq$) If $y \in \bigcup_{X \in \mathcal{S}_n(c,r)} X$, then $y \in X$ for some nonempty $X \subseteq S^n(c,r)$, hence $y \in S^n(c,r)$.

($\supseteq$) If $y \in S^n(c,r)$, then $\{y\} \in \widetilde{\mathcal{P}}(S^n(c,r)) = \mathcal{S}_n(c,r)$, and $y \in \{y\} \subseteq \bigcup_{X \in \mathcal{S}_n(c,r)} X$. □

**Remark 6.2.8** (Interpretation)**.** Definition 6.2.6 is the simplest "super" analogue: it replaces points on the sphere by nonempty subsets of points. Other variants are possible (e.g. allowing sets that intersect the sphere, or using set-valued coordinates with a set-valued norm constraint), but those require additional modeling choices.

It is anticipated that set-valued geometric objects (such as the superhypercube and superhypersphere introduced above) can be studied further using the theory of hypervector spaces and related hyper-structures [413–416]. Below we record type-correct definitions that align with standard hyperstructure terminology: scalar multiplication is set-valued (a hyperoperation), so results are subsets of the ambient vector space.

**Definition 6.2.9** (Hypervector space (one common axiom scheme))**.** Let $K$ be a field and let $(V,+)$ be an abelian group. A *hypervector space* over $K$ is a triple $(V,+,\circ)$ where

$$\circ : \ K \times V \longrightarrow \mathcal{P}(V) \setminus \{\emptyset\}, \qquad (a,x) \longmapsto a \circ x,$$

is a set-valued scalar multiplication (a hyperoperation) satisfying, for all $a, b \in K$ and $x, y \in V$, the (weak) distributive inclusions

$$a \circ (x+y) \ \subseteq \ (a \circ x) + (a \circ y), \qquad (a+b) \circ x \ \subseteq \ (a \circ x) + (b \circ x),$$

where for subsets $A, B \subseteq V$ the Minkowski sum is

$$A + B := \{u + v : \ u \in A,\ v \in B\}.$$

One typically also assumes additional axioms (depending on the chosen convention), such as

$$1 \circ x \ni x, \qquad 0 \circ x = \{0\}, \qquad a \circ 0 = \{0\}, \qquad (ab) \circ x \subseteq a \circ (b \circ x),$$

interpreting $a \circ (b \circ x) := \bigcup_{z \in b \circ x} a \circ z$.

**Remark 6.2.10.** The literature contains multiple (non-equivalent) axiom systems for hypervector spaces. Definition 6.2.9 records the core set-valued scalar multiplication together with the distributive inclusions needed to make the expressions well-typed.

**Definition 6.2.11** (Subhypervector space)**.** Let $(V, +, \circ)$ be a hypervector space over $K$. A nonempty subset $W \subseteq V$ is a *subhypervector space* (subhyperspace) if

(i) $W$ is a subgroup of $(V, +)$, and

(ii) for all $a \in K$ and $w \in W$, one has $a \circ w \subseteq W$.

**Definition 6.2.12** (Hyperplane (codimension-one subspace))**.** Let $V$ be a classical $n$-dimensional vector space over $K$ (so scalar multiplication is single-valued). A *hyperplane* in $V$ is a linear subspace $W \leq V$ with $\dim(W) = n - 1$. Equivalently, $W = \ker(\varphi)$ for some nonzero linear functional $\varphi : V \to K$.

**Remark 6.2.13** (On "superhyperspace")**.** The statement "a superhyperspace is an $(n-1)$-dimensional subhyperspace" mixes hypervector-space terminology with classical dimension theory. A mathematically standard formulation is:

- In the *classical* setting, codimension-one subspaces are hyperplanes (Definition 6.2.12).
- In the *hypervector* setting, one first defines an appropriate notion of dimension for hypervector spaces (e.g. via a basis theory if available in the chosen axiom system) and then defines a "hyper-hyperplane" accordingly.

If one adopts a hypervector-space dimension theory and can define $\dim(W)$, then one may call a subhyperspace $W$ a *superhyperspace* when $\dim(W) = \dim(V) - 1$; however, the kernel-of-functional characterization then requires additional hypotheses and is not automatic from Definition 6.2.9.

# Chapter 7

# Discussion: Procedures for Graphization, Hyperization, and Uncertainization

This chapter records illustrative procedures for (i) *graphization* (turning set-based data into a graph-based model), (ii) *hyperization/superhyperization* (replacing single-valued attributes by set-valued, then set-of-set-valued attributes), and (iii) *uncertainization* (replacing crisp membership by fuzzy/neutrosophic-type uncertainty). These are not universal recipes: the meanings of "hyper" and "superhyper" vary substantially across fields, and any practical construction must be tailored to the target application.

**Note 1** (Illustrative procedure of graphization)**.** *Let a* set-based structure *be given by data $\mathcal{S}$ on a universe $U$ (e.g. a set system, a soft set, a fuzzy set, a neutrosophic set, etc.). A generic graphization pipeline can be stated as follows.*

***1. Identify the base universe and semantics.*** *Specify the universe $U$, the objects of interest, and the meaning of the available structure (membership, labels, parameters, relations, etc.).*

***2. Choose a vertex carrier.*** *Select a vertex set $V$ that represents the entities to be connected. Common choices include $V = U$, a subset of $U$, or a derived set such as parameters, attributes, or equivalence classes.*

***3. Choose an edge carrier (a relation).*** *Specify an edge set $E \subseteq \binom{V}{2}$ (undirected) or $E \subseteq V \times V \setminus \{(v, v)\}$ (directed), based on a relation extracted from $\mathcal{S}$ (intersection, adjacency, similarity threshold, co-occurrence, etc.).*

***4. Transport set structure into graph attributes.*** *Encode the set-based structure as vertex/edge attributes, such as weights, labels, fuzzy degrees, neutrosophic triples, or set-valued labels. Formally this produces maps like*

$$\sigma : V \to \mathcal{A}_V, \qquad \mu : E \to \mathcal{A}_E$$

*for appropriate attribute codomains $\mathcal{A}_V, \mathcal{A}_E$.*

***5. State coherence axioms (if needed).*** *If the application requires consistency between vertex and edge attributes (e.g. edge degrees bounded by endpoint degrees), state these as explicit axioms.*

***6. Validate well-posedness.*** *Check that the resulting object satisfies the relevant graph-theoretic type constraints (finite vs. infinite, simple vs. multigraph, symmetry, etc.) and that each attribute map is well-defined on its domain.*

**Example 7.0.1** (Fuzzy set $\to$ fuzzy graph (one standard route))**.** Let $U \neq \emptyset$ and let $\mu : U \to [0,1]$ be a fuzzy set. Choose $V := U$ and choose a crisp relation $R \subseteq U \times U$ (e.g. a similarity relation). Define $E := \{\{u,v\} : (u,v) \in R,\ u \neq v\}$. A common fuzzy-graph construction assigns the vertex membership $\sigma(u) := \mu(u)$ and an edge membership such as

$$\mu_E(\{u,v\}) := \min\{\mu(u),\mu(v)\} \quad \text{or} \quad \mu_E(\{u,v\}) := \mu(u)\mu(v),$$

yielding a fuzzy graph $(V,E,\sigma,\mu_E)$.

**Example 7.0.2** (Neutrosophic set $\to$ neutrosophic graph (one standard route))**.** Let $U \neq \emptyset$ and let $T,I,F : U \to [0,1]$ define a neutrosophic set. Choose $V := U$ and an edge relation $E \subseteq \binom{V}{2}$. Assign vertex triples $(T(u),I(u),F(u))$ and define edge triples by a chosen rule, e.g.

$$T_E(\{u,v\}) := \min\{T(u),T(v)\}, \quad I_E(\{u,v\}) := \max\{I(u),I(v)\}, \quad F_E(\{u,v\}) := \max\{F(u),F(v)\},$$

obtaining a neutrosophic graph $(V,E,T,I,F,T_E,I_E,F_E)$ (or the appropriate subcollection).

**Note 2** (Illustrative procedure of hyperization/superhyperization)**.** *Let a structure be given by a set $S$ and an attribute map*

$$f : S \longrightarrow T,$$

*where $T$ is an attribute codomain (weights, labels, membership degrees, etc.).*

***1. Hyperization (first level).*** *Replace $f$ by a set-valued map*

$$f^{(1)} : S \longrightarrow \widetilde{\mathcal{P}}(T), \qquad f^{(1)}(s) \subseteq T,\ f^{(1)}(s) \neq \emptyset.$$

*This yields a* hyper-*attribute structure (multiple admissible attributes per element).*

***2. Superhyperization (second level).*** *Iterate once more to obtain*

$$f^{(2)} : S \longrightarrow \widetilde{\mathcal{P}}\big(\widetilde{\mathcal{P}}(T)\big),$$

*so that each $s \in S$ carries a nonempty set of hyperattributes, i.e. a family of nonempty subsets of $T$.*

***3. Specify reduction conditions.*** *To ensure these constructions genuinely extend the original one, record the canonical embeddings*

$$t \in T \ \mapsto\ \{t\} \in \widetilde{\mathcal{P}}(T), \qquad A \in \widetilde{\mathcal{P}}(T) \ \mapsto\ \{A\} \in \widetilde{\mathcal{P}}\big(\widetilde{\mathcal{P}}(T)\big),$$

*and observe that restricting to singleton images recovers the previous level.*

***4. Extend operations if present.*** *If the original model carries operations on $T$ (addition, order, aggregation), specify how these lift to $\widetilde{\mathcal{P}}(T)$ and $\widetilde{\mathcal{P}}(\widetilde{\mathcal{P}}(T))$ (e.g. via Minkowski sum, union, supremum/infimum, etc.). This step is not canonical and must be chosen explicitly.*

**Note 3** (Illustrative procedure of uncertainization)**.** *A generic uncertainization pipeline replaces crisp membership* $0/1$ *information by graded information.*

***1. Crisp $\to$ fuzzy.*** *Replace an indicator $\chi_A : U \to \{0,1\}$ by a fuzzy membership $\mu : U \to [0,1]$.*

***2. Fuzzy $\to$ neutrosophic.*** *Replace a single degree $\mu(u)$ by a triple $(T(u),I(u),F(u)) \in [0,1]^3$, optionally with a normalization constraint.*

***3. Offset/over/under variants.*** *Enlarge the codomain interval to* $[\Psi, \Omega]$ *with* $\Psi < 0 < 1 < \Omega$ *when under/over behavior is needed.*

***4. Subset-valued and probabilistic enrichments.*** *Replace scalar degrees by subsets of degrees, probability measures on degree spaces, or other uncertainty models, each requiring explicit choices of codomain and aggregation rules.*

**Note 4** (Illustrative procedure of uncertainization)**.** *The following is an illustrative (and non-canonical) procedure for extending a mathematical concept to fuzzy, intuitionistic fuzzy, neutrosophic, and plithogenic forms. The precise generality relations depend on the chosen axioms and codomains; in particular, neutrosophic models often* contain *intuitionistic fuzzy models as special cases under additional constraints, and plithogenic models typically require an explicit attribute/contradiction apparatus.*

***1. Identify the core concept.*** *Specify the original structure* $\mathcal{S}$ *(e.g., a set system, function, relation, graph, hypergraph) and the data it carries (membership, weights, labels, parameters, etc.).*

***2. Select the carriers to uncertainize.*** *Choose which components of* $\mathcal{S}$ *will become graded (e.g., vertex membership, edge membership, relation strength, parameter relevance). Formally, identify the underlying carrier* $U$ *and a crisp predicate* $\chi : U \to \{0, 1\}$ *or crisp label* $\ell : U \to L$ *that will be replaced by uncertainty data.*

***3. Fuzzy extension.*** *Replace crisp membership* $\chi$ *by a fuzzy membership function*

$$\mu : U \longrightarrow [0, 1].$$

*Update definitions/operations of* $\mathcal{S}$ *so that occurrences of "*$x \in A$*" or "x satisfies property" are interpreted through* $\mu(x)$ *(e.g., via thresholds, t-norms, or degree-based constraints).*

***4. Intuitionistic fuzzy extension (Atanassov).*** *Refine the fuzzy model by assigning both membership and non-membership degrees*

$$\mu,\ \nu : U \longrightarrow [0, 1] \quad \text{with} \quad 0 \le \mu(x) + \nu(x) \le 1 \ \ (\forall x \in U).$$

*Define the hesitation (indeterminacy) degree by*

$$\pi(x) := 1 - \mu(x) - \nu(x) \in [0, 1].$$

*Replace fuzzy constraints by their intuitionistic counterparts using* $(\mu, \nu, \pi)$*.*

***5. Neutrosophic extension (single-valued core form).*** *Assign three degrees*

$$T,\ I,\ F : U \longrightarrow [0, 1],$$

*interpreted as truth-, indeterminacy-, and falsity-membership, respectively. Unlike the intuitionistic case,* $T, I, F$ *are typically treated as* independent *subject only to the codomain bounds; optionally one may impose a normalization such as* $T(x) + I(x) + F(x) \le 1$ *or* $\le 3$ *depending on the adopted convention. Observe that an intuitionistic fuzzy datum* $(\mu, \nu)$ *embeds into the neutrosophic form by*

$$T := \mu, \qquad F := \nu, \qquad I := 1 - \mu - \nu.$$

***6. Plithogenic extension (attribute-based).*** *Introduce an attribute system:*

$$PS = (P,\ v,\ P_v,\ pdf,\ pCF),$$

*where* $P$ *is the object set,* $v$ *is an attribute with value set* $P_v$*,* $pdf : P \times P_v \to [0, 1]^s$ *is a degree-of-appurtenance function, and* $pCF : P_v \times P_v \to [0, 1]^t$ *is a contradiction function. Replace single-valued degrees by attribute-conditioned degrees and use* $pCF$ *to model interactions among (possibly contradictory) attribute values.*

***7. Validation.*** *For each extension, verify: (i) type correctness (domains/codomains, nonemptiness conditions), (ii) consistency with the original structure when degrees collapse to crisp values, and (iii) closure of any defined operations (e.g., unions, intersections, graph products, etc.) within the chosen model.*

# Chapter 8

# Conclusion

This chapter outlines the future prospects of this research.

## 8.1 Conclusion of this book

In this book, we have examined a variety of graph concepts, uncertainty-aware concepts, and hierarchical concepts (including HyperStructures and SuperHyperStructures). We sincerely hope that these frameworks will enable clearer and more intuitive representations of the highly complex concepts that arise in the modern world.

## 8.2 Future Works: Other Graph Class Extension (Revisited)

The author is interested in exploring extensions of graph classes and investigating the mathematical structures within specific graph classes. Our aim is to identify suitable graph classes by integrating these various perspectives. Although theoretical generalization is fundamental to mathematics, it does not always lead directly to practical applications. From an applied mathematics perspective, it is equally important to evaluate the practical relevance of such concepts. To support theoretical advancements, experimental approaches and algorithmic explorations are essential. Research in this area often involves expanding or refining graph classes based on the following factors:

- *Classic Graph Properties:* Regular [417,418], Irregular [419,420], Complete [421,422], Perfect [423, 424], claw-free [425], Tree, Path [426, 427], Planar [428], Linear [429, 430], OuterPlanar [431, 432], Median [433–435], Multigraph [436,437].
- *Graph Using Operations:* Intersection Graph [181], Product Graph [438,439], Union Graph [440,441].
- *Subgraph/Hypergraph Properties:* Supergraph [72], Hypergraph [94, 95, 442], Superhypergraph [292], n-Superhypergraph [238,443], Subgraph, Induced Subgraph [444,445], Induced Supergraph [181].
- *Graph Directionality:* Undirected, Directed, Mixed [446,447], Bidirected [448,449], Bunch [450,451].
- *Graph Partition:* Bipartite [452–454], Tripartite [455,456], n-partite [457,458].
- *Uncertain Properties:* Fuzzy, Neutrosophic, Plithogenic, Rough [459, 460], Vague [43, 461, 462], Soft [463, 464],Hypersoft [465, 466], Weighted [232, 467], Picture Fuzzy [468–470], Grey [471], Triangular Fuzzy [472], Z-Number [473, 474], Refined Plithogenic [91], q-Rung Orthopair Fuzzy [475, 476], Quadripartitioned Neutrosophic [121, 122], Pentapartitioned Neutrosophic, Entropy [477], HyperFuzzy [70, 261], Type-2 Fuzzy [478–480], Hesitant [155, 481], spherical [482], Bipolar [483, 484], Tripolar [485–487] etc.
- *Graph Dimensionality:* 2-dimensional, 3-dimensional [488–490], 4-dimensional [491–493], Multidimensional [494, 495] etc.

- *Themes:* Graph Classes Hierarchy [86], Mathematical Structure of Graph Classes [86], Graph Parameters [496, 497], Algorithms [22], Computational Complexity [20, 21], Real-world Applications, Combinatorics [498, 499].

The above is merely one example, and the author believes there are numerous concepts and perspectives that remain unrecognized. It is sincerely hoped that further research in this field will continue to advance.

# Disclaimer

## Funding

This study did not receive any financial or external support from organizations or individuals.

## Acknowledgments

We extend our sincere gratitude to everyone who provided insights, inspiration, and assistance throughout this research. We particularly thank our readers for their interest and acknowledge the authors of the cited works for laying the foundation that made our study possible. We also appreciate the support from individuals and institutions that provided the resources and infrastructure needed to produce and share this book. Finally, we are grateful to all those who supported us in various ways during this project.

## Data Availability

This research is purely theoretical, involving no data collection or analysis. We encourage future researchers to pursue empirical investigations to further develop and validate the concepts introduced here.

## Ethical Approval

As this research is entirely theoretical in nature and does not involve human participants or animal subjects, no ethical approval is required.

## Use of Generative AI and AI-Assisted Tools

I use generative AI and AI-assisted tools for tasks such as English grammar checking, and I do not employ them in any way that violates ethical standards.

## Conflicts of Interest

The authors confirm that there are no conflicts of interest related to the research or its publication.

## Disclaimer

This work presents theoretical concepts that have not yet undergone practical testing or validation. Future researchers are encouraged to apply and assess these ideas in empirical contexts. While every effort has been made to ensure accuracy and appropriate referencing, unintentional errors or omissions may still exist. Readers are advised to verify referenced materials on their own. The views and conclusions expressed here are the authors' own and do not necessarily reflect those of their affiliated organizations.

# Appendix (List of Tables)

*

# Appendix (List of Figures)

*

*Combinatorics is a branch of mathematics focused on counting, arranging, and combining elements within a set under specific rules and constraints. This field is particularly fascinating due to its ability to yield novel results through the integration of concepts from various mathematical domains. Its significance remains unchanged in areas that address uncertainty in the real world. Set theory, another foundational area of mathematics, explores "sets," which are collections of objects that can be finite or infinite. Recent years have seen growing interest in "non-standard set theory" and "non-standard analysis." To better handle real-world uncertainty, concepts such as fuzzy sets, neutrosophic sets, rough sets, and soft sets have been introduced. For example, neutrosophic sets, which simultaneously represent truth, indeterminacy, and falsehood, have proven to be valuable tools for modeling uncertainty in complex systems. These set concepts are increasingly studied in graphized forms, and generalized graph concepts now encompass well-known structures such as hypergraphs and superhypergraphs. Furthermore, hyperconcepts and superhyperconcepts are being actively researched in areas beyond graph theory. Combinatorics, uncertain sets (including fuzzy sets, neutrosophic sets, rough sets, soft sets, and plithogenic sets), uncertain graphs, and hyper and superhyper concepts are active areas of research with significant mathematical and practical implications. Recognizing their importance, this book explores new graph and set concepts, as well as hyper and superhyper concepts. Additionally, this work aims to consolidate recent findings, providing as survey-like resource to inform and engage readers. For instance, we extend several graph concepts by introducing Neutrosophic Oversets, Neutrosophic Undersets, Neutrosophic Offsets, and the Nonstandard Real Set. This book defines a variety of concepts with the goal of inspiring new ideas and serving as a valuable resource for researchers in their academic pursuits. In this second edition, we add several recent concepts to the first edition and also revise typographical errors and re-examine mathematical correctness.*

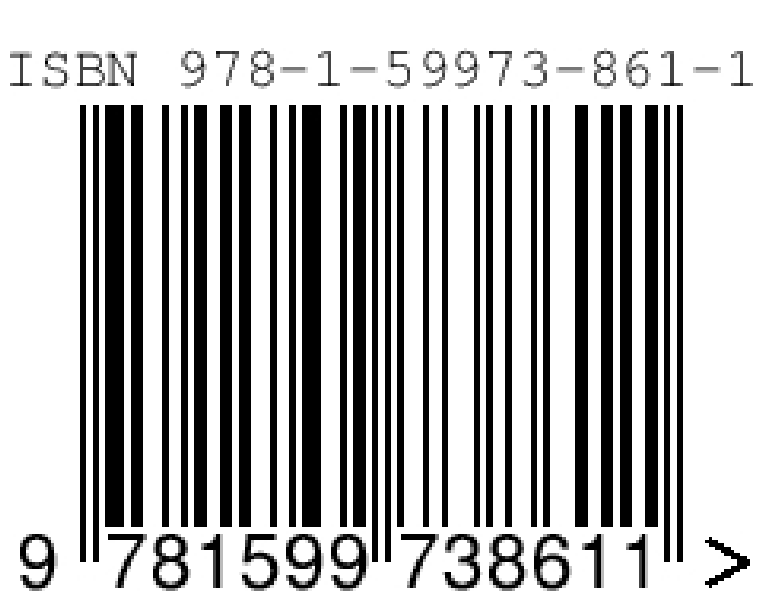